\documentclass[12pt]{report} 

\usepackage[
a4paper,
vmargin=2.5cm,
hmargin=3cm
]{geometry}

\usepackage[a-1b]{pdfx}
\usepackage{algorithm}
\usepackage{algorithmicx}
\usepackage{booktabs}
\usepackage{hyperref}
\usepackage{float}
\usepackage{verbatim} 
\usepackage{algpseudocode}
\usepackage{multirow}
\usepackage{multicol}
\usepackage{xcolor}
\usepackage{graphicx}
\usepackage{graphicx}[realboxes]
\usepackage{lscape}
\usepackage{afterpage}
\usepackage{adjustbox}
\usepackage{rotating}
\usepackage{amsmath, amssymb, amsfonts}
\usepackage{comment}
\usepackage{amsthm}
\usepackage{mathrsfs}
\usepackage[title]{appendix}
\usepackage{textcomp}
\usepackage{manyfoot}
\usepackage{listings}
\usepackage{subcaption}
\usepackage{changepage}

\AtBeginDocument{}%
\AtBeginDocument{}%
\AtBeginDocument{}%

\newtheorem{definition}{Definition}

\usepackage{hyperref}
\hypersetup{colorlinks=true,
	linkcolor=black, 
	urlcolor=blue} 

\usepackage{amsmath,amssymb,amsfonts,amsthm}

\usepackage{txfonts} 
\usepackage[T1]{fontenc}
\usepackage[utf8]{inputenc}

\usepackage[english]{babel} 
\usepackage[babel, english=american]{csquotes}
\AtBeginEnvironment{quote}{\small}

\usepackage{fancyhdr}
\fancypagestyle{plain}{\pagestyle{fancy}}

\usepackage{titlesec}
\usepackage{titletoc}
\titleformat{\chapter}[block]
{\large\bfseries\filcenter}
{\thechapter.}
{5pt}
{\MakeUppercase}
{}
\titlespacing{\chapter}{0pt}{0pt}{*3}
\titlecontents{chapter}
[0pt]                                               
{}
{\contentsmargin{0pt}\thecontentslabel.\enspace\uppercase}
{\contentsmargin{0pt}\uppercase}                        
{\titlerule*[.7pc]{.}\contentspage}                 

\titleformat{\section}
{\bfseries}
{\thesection.}
{5pt}
{}
\titlecontents{section}
[5pt]                                               
{}
{\contentsmargin{0pt}\thecontentslabel.\enspace}
{\contentsmargin{0pt}}
{\titlerule*[.7pc]{.}\contentspage}

\titleformat{\subsection}
{\normalsize\bfseries}
{\thesubsection.}
{5pt}
{}
\titlecontents{subsection}
[10pt]                                               
{}
{\contentsmargin{0pt}                          
	\thecontentslabel.\enspace}
{\contentsmargin{0pt}}                        
{\titlerule*[.7pc]{.}\contentspage}  

\usepackage{chngcntr}
\counterwithout{footnote}{chapter}

\usepackage{graphicx}
\graphicspath{{imagenes/}} 

\usepackage[backend=biber, style=ieee, isbn=false,sortcites, maxbibnames=6, minbibnames=1]{biblatex} 

\begin{document}
\pagenumbering{roman} 
\begin{titlepage}
	\begin{center}
		\begin{Huge}
			Market Making Strategies with Reinforcement Learning\\			
			\bigskip
			by\\
			\bigskip
			Óscar Fernández Vicente\\
		\end{Huge}
	    \vspace{4cm}	
	    \begin{Large}
            A dissertation submitted in partial fulfillment of the requirements for the degree of Doctor of Philosophy in\\
            \bigskip
            COMPUTER SCIENCE AND TECHNOLOGY\\
		    \vspace{2cm}
		    Universidad Carlos III de Madrid\\
		    \vspace{3cm}
		    Advisors:\\
            Fernando Fernández Rebollo\\
		    \bigskip
		    Francisco Javier García Polo\\

		    \bigskip
		    Tutor:\\
		    \bigskip
		    Fernando Fernández Rebollo\\
		    \bigskip
		    <January 2025>\
		 \end{Large}
		 
	\end{center}
	\end{titlepage}
	
	\thispagestyle{empty}
	\vspace*{\fill}
	\begin{center}
	This thesis is distributed under license ``Creative Commons \textbf{Atributtion - Non Commercial - Non Derivatives}''.\\
	\medskip
	\includegraphics[width=4.2cm]{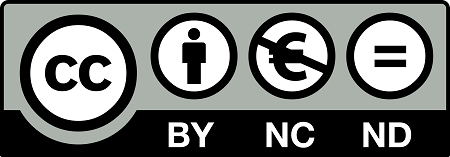} 
	\end{center}



    \chapter*{Acknowledgements}
\thispagestyle{empty}

Quiero dar las gracias en especial a mis directores, Fernando y Javier por guiarme en todo este largo e intenso camino. Sin su conocimiento, dedicación y apoyo no hubiera sido posible conseguir este gran hito. Sin duda es una suerte haber caído de su lado. De igual forma agradecer a los miembros del Tribunal, Daniel, Inés y Juan Ramón, por su profesionalidad y sus más que relevantes aportaciones en la mejora de la Tesis.

Mención especial, por supuesto, a mi familia por el cariño y la paciencia, especialmente a Lina, durante todo el desarrollo de la investigación. A veces se hace cuesta arriba y es un lujo poder contar con personas así. Sin olvidarme tampoco de los \textit{Warriors}, que siempre están ahí cuando se les necesita.

Y por último, y como no puede ser de otra manera, a toda la gente, cercana o lejana, apasionados por este complejo y emocionante campo que aportan cada día su granito de arena para conseguir grandes avances.

\newpage
\section*{Funding Acknowledgements}
This research was funded in part by JPMorgan Chase \& Co. Any views or opinions expressed herein are solely those of the authors listed, and may differ from the views and opinions expressed by JPMorgan Chase \& Co. or its affiliates. This material is not a product of the Research Department of J.P. Morgan Securities LLC. This material should not be construed as an individual recommendation for any particular client and is not intended as a recommendation of particular securities, financial instruments or strategies for a particular client. This material does not constitute a solicitation or offer in any jurisdiction.

\chapter*{Published and submitted content}
\thispagestyle{empty}

The compilation of research articles published in peer-reviewed journals and conference presentations conducted throughout this dissertation is as follows:

\noindent\textbf{Journal Papers}

\begin{itemize}
    \item Vicente, O.F., Fernández, F., García, J.: Automated market maker inventory management with deep reinforcement learning. Applied Intelligence (2023) 
    \url{https://doi.org/10.1007/s10489-023-04647-9}. Contributor author. 
    Included in \autoref{amm_sec_appliedintelligence}.
\end{itemize}

\noindent\textbf{International Conference Papers}
\begin{itemize}
    \item Vicente, O.F., Fernández, F., García, J. Deep q-learning market makers in a multi-agent simulated stock market. In: Proceedings of the Second ACM International Conference on AI in Finance. Association for Computing Machinery, New York, NY, USA, ICAIF ’21 (2022) 
    \url{https://doi.org/10.1145/3490354.3494448} .Contributor author. 
    Included in \autoref{chap_icaif}.
\end{itemize}

\noindent\textbf{Unplublished Papers}

\begin{itemize}
\item Vicente, O.F., García, J., Fernández, F.: Optimizing market making Strategies: A Multi-Objective Reinforcement Learning Approach with Pareto Fronts. Submitted to Expert Systems with Applications (2023). Contributor author. (Under second review\footnote{As of October 2024}).
Included in \autoref{morl_sec-morlpareto}.
\item Vicente, O.F., García, J., Fernández, F.: Policy Weighting via Discounted Thompson Sampling for Non-Stationary market making. Submitted to Artificial Intelligence Review (2024). Contributor author. (Under review$^1$).
Included in \autoref{chap_nonstationarity}.
\end{itemize}

 The material from all the mentioned sources included in this thesis is not singled out with typographic means and references.




\tableofcontents
\thispagestyle{fancy}

\listoffigures
\thispagestyle{fancy}

\listoftables
\thispagestyle{fancy}

\clearpage
\pagenumbering{arabic} 

\chapter{Introduction}

This chapter introduces the doctoral thesis by describing the context in which the work is situated. It presents the motivation behind exploring market making in financial markets using Reinforcement Learning (RL) and outlines the challenges and opportunities that arise from this integration (\autoref{sec_motivation}). The chapter further elaborates on the central hypothesis that RL-based market making agents can outperform traditional strategies by autonomously adapting to dynamic market conditions (\autoref{sec_hypothesis}). Additionally, the chapter briefly describes the objectives that the research aims to achieve, including the development of RL agents capable of efficient market making, the evaluation of their performance in various scenarios, and the exploration of advanced inventory management and multi-objective optimization techniques  (\autoref{sec-goals}). Finally, the chapter concludes with an overview of the thesis structure, summarizing how each subsequent chapter contributes to the comprehensive investigation of RL applications in market making (\autoref{sec_thesisoverview}).

\section{Motivation}\label{sec_motivation}

\begin{figure}[ht]
        \centering
        \includegraphics[width=1\textwidth]{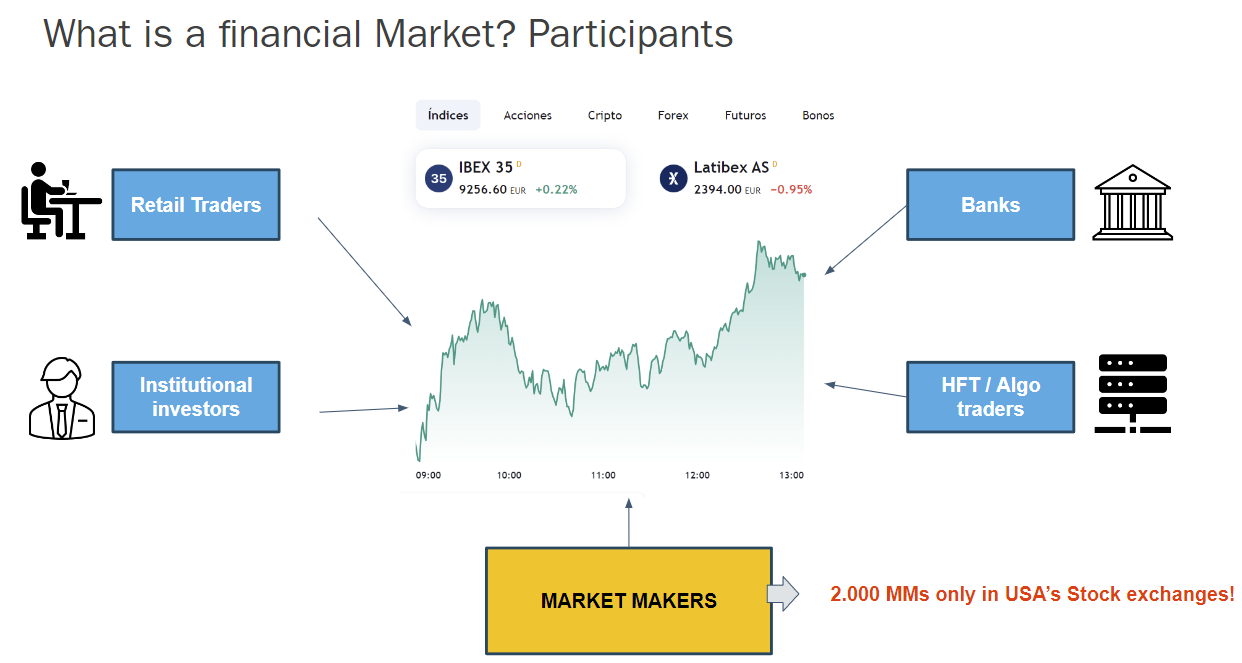}
        \caption{Composition of a Financial Market. Example of participants.}
        \label{fig:participants}
\end{figure}

Financial markets are complex systems characterized by the dynamic interactions of diverse participants, each pursuing distinct objectives. This ecosystem comprises various actors including retail traders, large institutions, arbitrageurs, market makers, and high-frequency traders, all engaging in transactions across different types of markets such as stock markets, bond markets, foreign exchange (Forex), commodities markets, and cryptocurrencies~\cite{hendershott2003electronic, HUBERMAN2005, Chordia2013}. These interactions shape the continual flux of asset prices, as represented in \autoref{fig:participants}. Each participant in the financial markets plays a unique role, contributing to the overall liquidity and depth of the markets. Besides these traditional participants, the financial markets are also influenced by regulatory bodies and technology platforms. Regulatory bodies oversee the functioning of markets, setting rules that aim to ensure fairness and transparency while minimizing systemic risk. Technology platforms, including electronic trading platforms and blockchain-based systems, facilitate the rapid execution of trades and are pivotal in the dissemination of market information.

In this financial ecosystem, \textbf{market makers} (MM) play a crucial role. The primary function of classical market making is to enhance market liquidity by issuing multiple buy and sell orders at various price levels within the order book (OB). This activity not only provides traders with a broader range of options for executing their trades but is also vital for maintaining \textbf{liquidity}, especially in markets where it is typically low~\cite{Chakraborty2011}\footnote{All these concepts are broadly detailed at \autoref{sec_finmarkets}}.

MMs are essential in ensuring that financial markets operate efficiently. By continuously offering to buy and sell securities, they help fill the gap between supply and demand, which can prevent large price jumps and ensure that trades can be executed quickly. Their presence reduces transaction costs for other market participants and increases market transparency, thereby attracting more traders and investors to the market.  To gain a better understanding of the significance of these agents, as of 2008, there were over 2,000 MMs on the US Stock Exchange \cite{us_market_makers} and more than 100 in Canada \cite{canadian_market_makers}, numbers that have likely increased since then.

\begin{figure}[ht]
        \centering
        \includegraphics[width=1\textwidth]{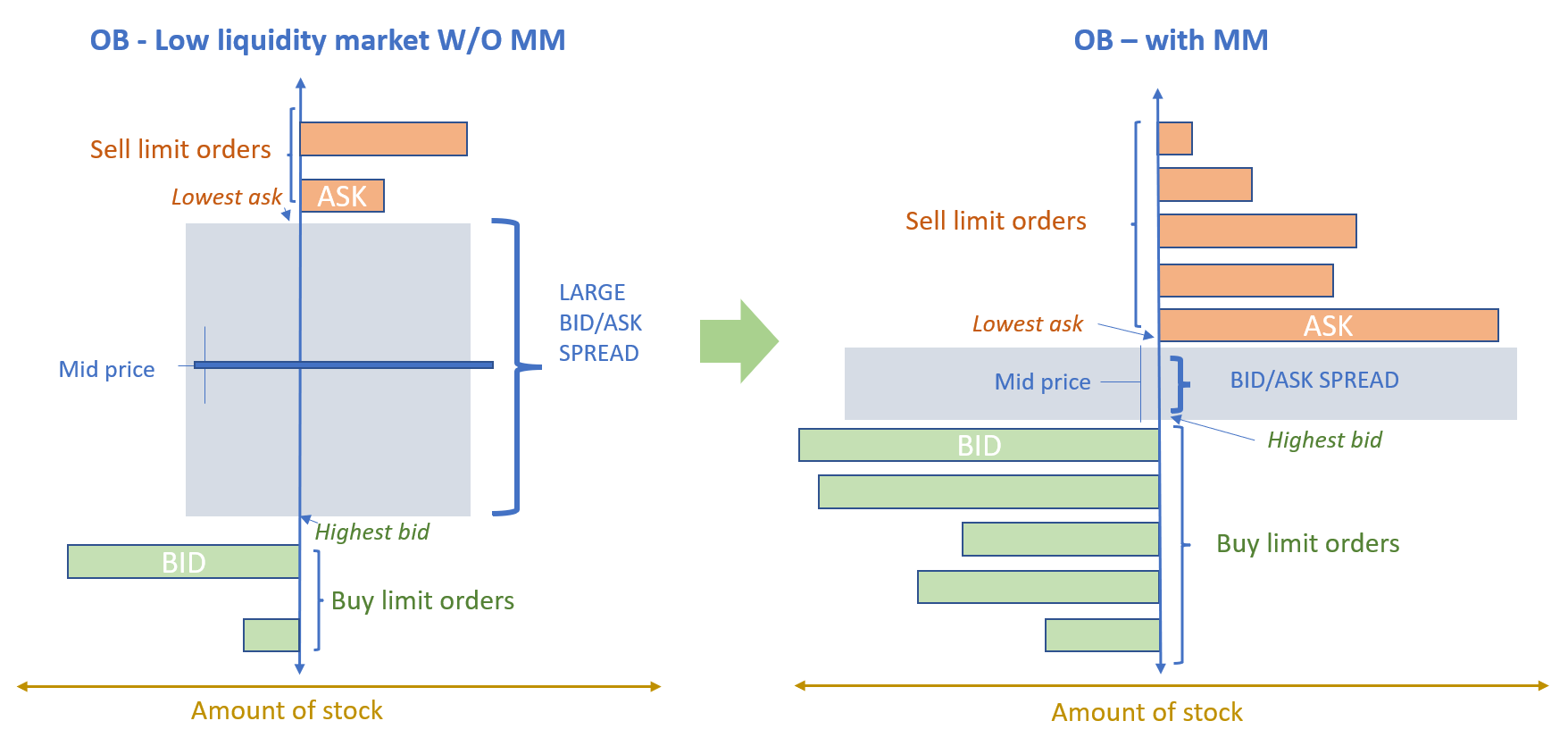}
        \caption[Example of the impact of a MM in a low liquidity asset order book]{Example of the impact of a MM in a low liquidity asset order book. A MM usually populates the order book allowing the rest of the market's participants to have more prices to trade with.}
        \label{fig:OBMMExample}
        \end{figure}

In their routine operations, MMs typically generate income by matching trades against other market participants. This income primarily arises from the bid-ask spread, which is the difference between the lowest price a seller is willing to accept (ask) and the highest price a buyer is willing to pay (bid)~\cite{https://doi.org/10.1111/j.1540-6261.1983.tb03834.x}, as depicted in \autoref{fig:OBMMExample}. For MMs, a wider bid-ask spread means higher profit margins per transaction. The spread not only reflects the liquidity of an asset but also serves as an indicator of the general market conditions. Typically, markets with lower liquidity exhibit wider spreads, suggesting a higher compensation for MMs who provide liquidity under these conditions.

Furthermore, MMs are engaged in sophisticated inventory management control. As they execute trades, their inventory levels fluctuate; they may increase if buying activity outweighs selling, and decrease in the opposite scenario. The main challenge with holding inventory is the risk associated with market price fluctuations, which can significantly affect the value of held assets. Such price volatility can lead to substantial financial gains or losses in their Mark-to-Market (MtM) value, depending on the nature of the inventory and market price movements at the time of liquidation. MtM is a financial accounting technique used to value assets and liabilities at their current market prices. MtM is widely used in the trading of securities and derivatives, where market prices can fluctuate significantly in short periods, making timely valuation crucial. By updating the valuation of these financial instruments regularly, MtM accounting allows for a transparent view of investment performance and risk exposure. 

Managing inventory is a critical aspect that necessitates meticulous risk assessment and management strategies. The optimization of this balance, aimed at maximizing profitability while minimizing risks, revolves around strategic decisions regarding the setting of bid and ask prices—the spread. Effective spread management not only ensures profitability but also reduces market risk exposure by maintaining an appropriate level of inventory that aligns with current market dynamics and future market expectations.

Additionally, the non-stationary aspect of financial markets demands continuous monitoring and adaptive strategies from those involved. Traders, investors, and MMs must constantly update their models and approaches to align with the current market state, ensuring their methods are responsive to the latest market conditions. This dynamic environment makes financial markets complex and unpredictable, requiring sophisticated tools and analytical techniques to navigate effectively.

Due to the complexities and challenges outlined, the operation of every MM presents a particularly challenging problem for several reasons:

\begin{enumerate}
    \item \textbf{Competitive environment}: MMs operate in a highly competitive environment where interactions are not limited to independent agents; they also face competition from other MMs and participants. This competitive landscape requires that they continuously adapt their strategies to stay ahead or simply keep pace with others in the market.
    
    \item \textbf{Inventory risk management}: A significant part of a MM's role involves managing inventory risk effectively. They must balance the inventory of financial instruments they hold to avoid substantial losses due to adverse price movements. Proper inventory management is crucial to mitigate risks associated with holding too much of a depreciating asset.

    \item \textbf{High dimensionality of information}: The information that MMs must process is typical of high dimensionality, involving continuous variables such as price, volume, and inventory levels. The complexity of this data requires sophisticated models and algorithms to make informed trading decisions quickly and accurately.
    \item \textbf{Dynamic and evolving environment}: The trading environment is not static but evolves continuously and unpredictably. Changes can occur without prior warning, altering the dynamics of the market and potentially rendering previous strategies ineffective. MMs must be agile, with the ability to quickly adapt their strategies in response to these changes.
\end{enumerate}

Each of these factors contributes to the intricate nature of a MMs' role within financial markets. They must navigate through a labyrinth of competitive pressures, risk management, complex data, and an ever-changing environment, which demands a high level of expertise, sophisticated analytical tools, and adaptive strategies. This dynamic setting not only challenges MMs but also underscores the importance of continuous learning and evolution in their methodologies.

\section{Hypothesis}\label{sec_hypothesis}

Market making presents a significant control challenge. Managing this complexity is difficult, which is where RL comes into play. RL offers a robust framework for decision-making in complex environments. RL, as described by Sutton and Barto \cite{Sutton1998}, represents a paradigm within machine learning (ML) where agents learn to determine optimal behaviors through trial and error, interacting with their environment. These agents are designed to optimize their decision-making processes, aiming to maximize cumulative rewards over time. This technique is initially well suited to address the challenges presented in market making due to several reasons:

\begin{itemize}
    \item \textbf{High performance in complex scenarios}: RL agents are capable of achieving high performance in scenarios with single or multiple objectives, making them versatile in handling the diverse goals inherent in financial markets. This adaptability is crucial in market making where agents must simultaneously optimize for various factors such as cost, speed, and risk.
    
    \item \textbf{Handling of continuous and high-dimensional data}: They can process continuous and high-dimensional information effectively. Many RL algorithms, particularly those involving deep RL (DRL), employ neural networks (NN), which are powerful universal function approximators. This capability allows them to digest vast amounts of market data and extract actionable insights, which is essential in the fast-paced and complex world of financial trading.
    
    \item \textbf{Adaptability to market conditions}: RL is adaptable to both stochastic and deterministic environments, employing various algorithms and approaches to suit the specific context, as detailed throughout this thesis. This flexibility enables RL agents to perform robustly across a spectrum of market dynamics, adjusting their strategies in real-time to deal with uncertainties and volatilities.
    
    \item \textbf{Continuous learning and adaptation}: Beyond their initial training, RL agents can continuously learn from their interactions with the market, improving their decisions over time based on new data and experiences. This feature is particularly beneficial in market making, where market conditions can change rapidly and unpredictably.
    
    \item \textbf{Optimization of trade execution}: RL agents can also optimize the execution of trades by determining the optimal times and prices for order placement, thereby minimizing the market impact and the cost of trading. This aspect of RL is critical for maintaining competitive edges in market making, where profit margins can be tightly bound to execution efficiency.
\end{itemize}

More precisely, RL agents develop optimal policies based solely on the feedback (rewards or penalties) received from their interactions with the environment. This reward-based learning guides their actions towards the most beneficial outcomes without predefined strategies for goal attainment, relying instead on discerning beneficial from detrimental actions. Thus, RL is particularly effective at uncovering novel and surprisingly efficient strategies. A prominent example of RL's innovative capability is the ``move 37'' performed by AlphaGo agent against Go master Lee Sedol \cite{10.1145/3600211.3604730}. This move, unforeseen by most human experts, was instrumental in securing a victory and is acclaimed as a display of ``creativity''.

This thesis is founded on the premise that the dynamic and complex nature of stock markets, characterized by the interplay of a diverse range of participants, presents a unique set of challenges and opportunities for MMs. It hypothesizes that RL, with its inherent capability to learn optimal behaviors through trial and error by interacting with a complex environment, is particularly suited to address these challenges. This suitability comes from RL's ability to adapt and optimize decision-making processes in environments that are as unpredictable and multifaceted as financial markets.

The first layer of the hypothesis proposes that an RL-based approach can significantly enhance the MM's ability to inject liquidity into the market while ensuring profitability. This assertion takes into account the critical role of MMs in providing depth and stability to the market through the issuance of multiple buy and sell orders. It further acknowledges the nuanced balance MMs must strike between facilitating liquidity and managing the inherent risks of holding inventory, especially in markets characterized by low liquidity and high volatility.

Expanding upon this foundation, the hypothesis delves into the more complex scenario of inventory risk management. It suggests that RL can effectively manage the dual objectives of liquidity provision and inventory risk control, transforming the MM problem into a multi-objective optimization challenge. 

Furthermore, the hypothesis anticipates that an RL-based MM agent can achieve autonomous adaptation to the evolving dynamics of financial markets. This aspect considers the non-stationary nature of financial markets, influenced by numerous factors including economic events, market participant behavior, and trading volume fluctuations. It suggests that through continuous learning and adaptation, an RL agent can dynamically adjust its strategy in response to these changes, maintaining effectiveness and resilience in its market making activities. Such autonomous adaptation is expected to not only sustain the profitability and liquidity provision efficacy of MMs but also to potentially discover innovative strategies that enhance market efficiency and stability.

\section{Objectives}\label{sec-goals}

This dissertation aims to comprehensively explore and evaluate the application of RL in market making strategies within competitive financial markets. The research focuses on the profitability, adaptability, and strategic inventory management of RL-based MMs under dynamic market conditions. 

This research is driven by several key questions:
\begin{itemize}
    \item How can RL improve the decision-making processes in market making in addition to traditional quantitative methods?
    \item What are the optimal strategies for managing inventory and risks in market making using RL techniques?
    \item Can these strategies effectively handle the multi-objective nature of market making, balancing profit maximization with other critical operational goals?
    \item How do market making strategies adapt to and perform in non-stationary market environments?
\end{itemize}

Exploring these questions further, the specific objectives to be pursued include:

\begin{enumerate}
    \item \textbf{Evaluate RL for profitable market making in competitive scenarios:} \label{goal_1}
    \begin{itemize}
        \item Develop and test RL models that can operate effectively in competitive trading environments, emphasizing the creation of profitable market making strategies.
        \item Analyze the performance of RL-driven MMs against traditional and other algorithmic MMs to assess relative profitability and efficiency.
        \item Conduct simulations in a simulated market to understand how RL MMs interact with different types of trading agents and the resultant effects on market dynamics and liquidity.
    \end{itemize}
    
    \item \textbf{Advanced management of inventory using RL:} \label{goal_2}
    \begin{itemize}
        \item Investigate various RL approaches for inventory management, focusing on multi-objective strategies that minimize the risk associated with holding inventory while maximizing profitability. 
        \item Explore the impact of inventory levels on the decision-making processes of RL agents, incorporating models that dynamically adjust strategies based on current inventory states.
        \item Evaluate the effectiveness of these inventory management strategies through rigorous testing in simulated environments, measuring performance through metrics such as cost minimization and avoidance of inventory imbalances.
    \end{itemize}
    
    \item \textbf{Adaptability in market  making strategies to address non-stationarity:} \label{goal_3}
    \begin{itemize}
        \item Develop adaptive RL models capable of responding to non-stationary market conditions, ensuring sustained performance and relevance despite changes in market dynamics.
        \item Implement and assess mechanisms to enable RL models to shift strategies based on evolving market conditions effectively.
        \item Analyze how these adaptive strategies perform in long-term simulations, focusing on their ability to maintain competitiveness and profitability over time.
        \item Analyze and compare different existing learning strategies performed to adapt to this non-stationarity.
    \end{itemize}
    
\end{enumerate}

In summary, this thesis encapsulates a comprehensive exploration of the application of RL to the domain of market making. It examines the potential of RL to surpass the traditional approaches to market making by equipping MMs with strategies that are adaptive, robust, and optimized for the complexities of modern financial markets. Through this exploration, the thesis aims to contribute to the advancement of both the theoretical understanding and practical implementation of RL in financial market applications, ultimately promoting more resilient and efficient markets.

\section{Thesis overview}\label{sec_thesisoverview}

This dissertation is organized into six chapters, each serving a distinct purpose within the overall research narrative. The journey begins with this introductory segment, which sets the stage for the comprehensive exploration to follow. The following \autoref{chap_sota} takes a deep dive into the current state of the art and the main background in several interconnected fields: market making, RL, multi-objective reinforcement learning (MORL), and the nuanced application of RL within the context of market making. Additionally, it introduces the main concepts of financial markets and describes the simulated environment used throughout this work. This entire analysis is pivotal in understanding the research.

The heart of the dissertation resides in \autoref{chap_icaif}, \ref{chap_inventory}, and \ref{chap_nonstationarity}, where the bulk of the research unfolds. In \autoref{chap_icaif}, the narrative begins with the development of a RL market making agent capable of operating within a simulated stock market environment. This agent is designed to enhance profitability while providing liquidity, even in the presence of competitors. Through rigorous policy analysis and competitive testing against other RL-based MM agents, the chapter describes the agent's strategic efficacy.

\autoref{chap_inventory} shifts the focus to the critical aspect of inventory management, approached from two distinct angles. The initial perspective, presented in \autoref{amm_sec_appliedintelligence}, treats inventory management as a problem of reward engineering. Here, a novel reward function is proposed, leveraging dynamic factors that adjust rewards based on the cash-to-inventory ratio in real-time. This model is shown to outperform existing strategies, marking a significant advancement in the field. However, a more robust way to solve this multi-objective problem was sought. Reward engineering, although the main approach used in RL to deal with these kinds of challenges, suffers from many problems. This is why, in \autoref{morl_sec-morlpareto}, a more sophisticated multi-objective approach is explored, allowing the agent to focus on both objectives independently through the use of dual NNs and Pareto front (PF) optimization. This approach is demonstrated to not only enhance stability but also significantly improve performance compared to the reward-engineered model.

With these achievements, this research has not yet addressed the problem of the non-stationarity of financial markets. To this point,  a policy able to perform well under some predefined circumstances has been learned. However, the performance is not guaranteed if the market evolves due to any of the aforementioned factors. For this reason, in \autoref{chap_nonstationarity}, a non-stationary algorithm based on Thompson sampling (TS), \textit{Policy Weighting through Discounted Thompson sampling} (POW-dTS), tailored for RL market making agents is introduced. This algorithm cleverly balances a suite of pre-trained policies, derived from the multi-objective RL agent, adjusting dynamically to the ever-changing market conditions. With a set of pre-trained policies, the algorithm can adapt autonomously to the best combination of known policies in terms of rewards.

In the concluding \autoref{chap_conclusion}, summarizes the key findings of this study, discussing their impact on market making and RL. It highlights the contributions to academic knowledge and suggests directions for future research. This final chapter aims to encapsulate the essence of the dissertation, offering both a summation of the research conducted and a visionary outlook toward the future of market making strategies empowered by RL.

\begin{figure}[H]
\centering
\includegraphics[width=1\textwidth]{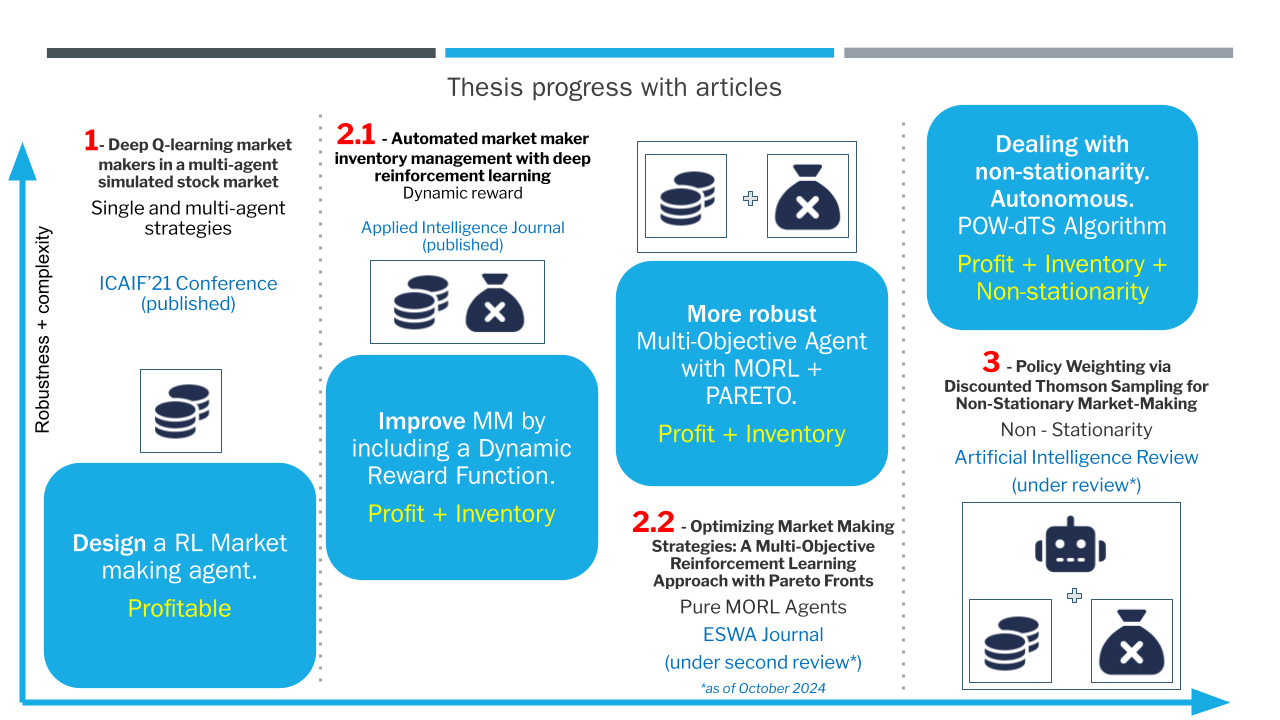}
\caption{A detailed depiction of the research's evolution, including the rationale and associated scholarly articles at each developmental phase.}
\label{fig:thesis_evol}
\end{figure}

The evolution of this research is visually represented in \autoref{fig:thesis_evol}, which not only chronicles the progression of the study through its main milestones but also lists the articles published or undergoing review at each pivotal moment. This graphical timeline reveals the dissertation's logical and methodical advancement.

\chapter{Background}\label{chap_sota}

The landscape of financial markets is continuously reshaped by advancements in technology and quantitative analysis, influencing strategies in trading and particularly in market making. This chapter provides a detailed review of the state of the art in market making, emphasizing the evolution from traditional models to sophisticated computational approaches that leverage RL and its derivatives. The purpose is to frame the current research within the broader context of recent technological advancements and to highlight the progress that has been made as well as the gaps that this thesis aims to address.

This chapter begins in \autoref{sota_rldrl} by focusing on RL and deep reinforcement learning (DRL). The foundational concepts, including Markov Decision Processes (MDP), value functions in RL, and various RL algorithms, are introduced. The discussion then delves into MORL, exploring both reward engineering approaches and broader MORL frameworks. Additionally, the complexities of non-stationary RL are addressed, highlighting strategies for adapting to changing environments. Next, the discussion shifts to \autoref{sec_finmarkets}. This section provides an introduction to financial markets, explaining what they are and how they work. Some relevant concepts such as the OB are presented. To understand market making, it is crucial to first understand the nature of these environments. Following this, the analysis of the market making problem is provided in \autoref{sec_dealingwithmarketmaking}. Both classical and modern (supervised learning and RL) market making approaches are detailed. It is important to recognize how different techniques have been employed to address this challenge. The pros and cons of each approach are also discussed. Attention is then given to simulated environments in RL, specifically focusing on the ABIDES framework. In \autoref{sec_abides}, the role of simulation in developing and testing market making strategies is explored, illustrating how simulated environments provide a robust and controlled setting for experimenting with and refining RL algorithms. Finally, \autoref{sec_discussion_background} details the advantages and disadvantages of current market making algorithms and techniques. The reference papers in the literature are described, along with the goals aimed to be achieved at every stage. 

By reviewing these diverse aspects, this chapter sets a solid foundation for understanding the innovative approaches explored in this thesis, highlighting the shift towards more dynamic, responsive, and efficient market making mechanisms. This comprehensive overview serves to position the current research within the continuum of financial technology development, emphasizing both the achievements and the ongoing challenges in the field.

\section{RL and DRL}\label{sota_rldrl}

In this section, the main concepts of RL and DRL are described in detail. Various approaches currently applied to address the challenges associated with multi-objective optimization and non-stationarity in dynamic environments are also explored. Specifically, the foundational principles of RL and DRL, their key components, and the algorithms that underpin these techniques are discussed. Additionally, how these methods are tailored and enhanced to handle the complexities of managing multiple objectives simultaneously and adapting to changes in the environment over time is examined. This comprehensive overview provides the necessary theoretical background for understanding the advanced methodologies employed in this dissertation.

\subsection{Markov decission process}

\begin{figure}[H]
\centering
\includegraphics[width=0.8\textwidth]{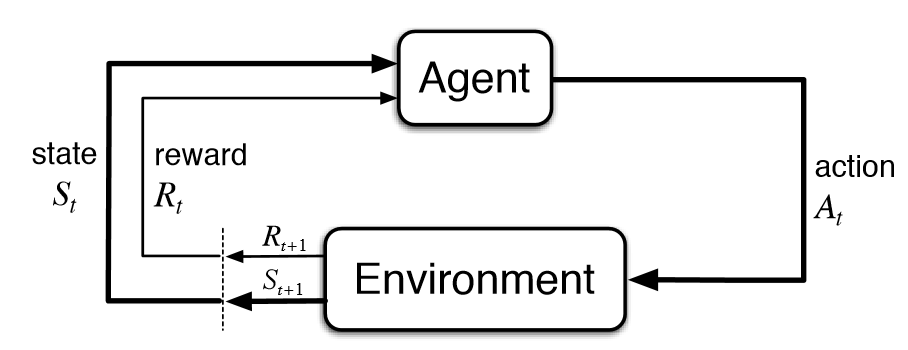}
\caption[RL agent and environment interactions]{RL Agent and environment interactions. (Sutton \& Barto - Reinforcement learning an Introduction).}
\label{fig:rlbasic}
\end{figure}

As introduced, RL~\cite{Sutton1998} is the pivot technique used around the market making problem. The foundational framework for RL tasks is the MDP. The MDP is formally defined by a tuple $\mathcal M=\langle S, A, T, R \rangle$, where $S$ represents the set of all possible states in the environment, $A$ denotes the set of actions available to the agent, $T$, the transition function, defines the probability of transitioning from one state to another given a specific action, and $R$ is the reward function, providing immediate feedback based on the current state and action taken, as depicted in \autoref{fig:rlbasic}. This framework allows for the modeling of both deterministic environments, where actions lead to predictable outcomes, and stochastic environments, where outcomes may vary even with the same action in the same state.

The final objective in RL is to discover a policy $\pi$, a strategy that dictates the action the agent should choose when in a given state, that maximizes the expected return $J(\pi)$ from any initial state, as in \autoref{eq_mdp_reward_1}. 

\begin{equation}\label{eq_mdp_reward_1}
       J(\pi) = \mathbb{E}_{\pi} \left[\sum_{t=0}^{\infty} \gamma^t R_t \right]
\end{equation}

Where \(\mathbb{E}_{\pi}[\cdot]\) represents the expectation over the distribution of states and actions by following policy \(\pi\), \(R_t\) is the reward received at time step \(t\), and \(\gamma \in [0, 1)\) is the discount factor, indicating the present value of future rewards.

The objective is to find the optimal policy \(\pi^*\) that maximizes \(J(\pi)\) as in \autoref{eq_mdp_argmax}:

\begin{equation}\label{eq_mdp_argmax}
     \pi^* = \arg\max_{\pi} J(\pi) 
\end{equation}

\subsection{Value functions in RL}\label{sec_valuefunctions}

In RL, value functions are essential for evaluating the potential outcomes of states and actions. These functions help agents decide by estimating the expected rewards. The two primary types of value functions are the state value function (V) and the action-value function (Q). The state-value function, denoted as \( V(s) \), estimates the expected return (present and future rewards) starting from a state \( s \) while following a given policy \( \pi \). It is formally defined as in \autoref{eq_valuef}:

\begin{equation} \label{eq_valuef}
 V^\pi(s) = \mathbf{E}^\pi \left[ \sum_{t=0}^{\infty} \gamma^t r_{t+1} \mid s_t = s \right]     
\end{equation}

In this equation, \( \mathbf{E}^\pi \) represents the expected value following policy \( \pi \), \( \gamma \) is the discount factor, which discounts future rewards,    \( r_{t+1} \) is the reward received at time step \( t+1 \); and  \( s_t \) is the state at time step \( t \). The state-value function indicates how favorable it is for an agent to be in a specific state, assuming it follows the policy \( \pi \). The main difference between \autoref{eq_mdp_reward_1} and this function is that the former represents the expectation over the distribution of states and actions encountered while following policy $\pi$, whereas \autoref{eq_valuef} focuses on the expected return when following policy $\pi$ starting from a specific state $s$. In summary, $J(\pi)$ provides a global view of the policy performance, whereas $V^\pi(s)$ offers a local view from each state.

The action-value function, or Q-function, denoted as \( Q(s, a) \), estimates the expected return for taking an action \( a \) in state \( s \) and then continuing to follow policy \( \pi \). It is defined as in \autoref{eq_qvaluef}:

\begin{equation} \label{eq_qvaluef}
 Q^\pi(s, a) = \mathbf{E}^\pi \left[ \sum_{t=0}^{\infty} \gamma^t r_{t+1} \mid s_t = s, a_t = a \right] 
\end{equation}

Here, \( \mathbf{E}^\pi \) represents the expected value following policy \( \pi \), \( \gamma \) is the discount factor, \( r_{t+1} \) is the reward received at time step \( t+1 \), and \( s_t \) and \( a_t \) are the state and action at time step \( t \). The Q-function indicates how favorable it is to take a specific action in a given state, assuming the agent continues following policy \( \pi \).

The state-value function \( V(s) \) and the action-value function \( Q(s, a) \) are very related. Indeed, the value of a state \( V(s) \) can be derived from  \( Q(s, a) \) by averaging over all possible actions, weighted by the policy \( \pi \), as shown in \autoref{eq_qvrelation} and \autoref{eq_qvrelation_2}, depending on whether stochastic or deterministic policies are applied:

\begin{equation}\label{eq_qvrelation}
 V^\pi(s) = \sum_{a \in A} \pi(a \mid s) \: Q^\pi(s, a) ,  \quad \forall s \in \mathcal{S} \quad \text{(when stochastic policy)}
\end{equation}

\begin{equation}\label{eq_qvrelation_2}
    V^\pi(s) = Q^\pi(s, \pi(s)),  \quad \forall s \in \mathcal{S} \quad \text{ (when deterministic policy)}
\end{equation}

This means the value of a state is the expected value of the Q-function for all actions, considering the probability of each action under the policy \( \pi \).

In RL, the main goal is to find the optimal policy \( \pi^* \) that maximizes the expected return. The optimal state-value function \( V^*(s) \) and the optimal action-value function \( Q^*(s, a) \) are defined as in \autoref{eq_vmax_wmax} and \autoref{eq_vmax_qmax}:

\begin{equation}\label{eq_vmax_wmax}
 V^*(s) = \max_\pi V^\pi(s), \quad \forall s \in \mathcal{S}
\end{equation}
\begin{equation}\label{eq_vmax_qmax}
 Q^*(s, a) = \max_\pi Q^\pi(s, a), \quad  \forall  s \in \mathcal{S} \quad \text{and} \quad \forall a \in A
\end{equation}

For the optimal policy, the relationship between \( V^* \) and \( Q^* \) is as in \autoref{eq_opvalf}:

\begin{equation}\label{eq_opvalf}
    V^*(s) = \max_{a \in A} \space Q^*(s, a) 
\end{equation}

This indicates that the optimal value of a state is the highest expected return achievable by any action in that state, followed by the optimal policy. This value function tends to work very well when dealing with discrete and not very large state and action spaces. In such scenarios, algorithms can be computed iteratively in tabular modes, such as Q-tables, where each state-action pair is explicitly stored and updated. The most representative value-function tabular algorithm is called Q-learning\cite{watkins1989}. Q-learning uses a Q-table to store and update the Q-values for each state-action pair. This algorithm updates the q-values of each state-action pair according to the \autoref{eq_qlearning}, where $\max_{a'} Q(s', a')$ represents the maximum estimated Q-value for the next state s' over all possible actions a':

\begin{equation}\label{eq_qlearning}
Q(s, a) \leftarrow Q(s, a) + \alpha \left[ r + \gamma \max_{a'} Q(s', a') - Q(s, a) \right]
\end{equation}

However, when the state or action space becomes very large or even continuous, the ``curse of dimensionality'' is encountered. This term refers to the exponential growth in computational complexity and resource requirements as the dimensionality of the space increases. In these situations, using a tabular approach becomes impractical due to the immense number of entries required and the infeasibility of updating each one through direct experience.

To address the curse of dimensionality, it is mandatory to employ some form of function approximator that can generalize across the vast or infinite number of states and actions. This is where neural networks (NNs) come into play, transforming traditional RL into deep reinforcement learning (DRL). NNs, with their ability to learn complex patterns and representations, can approximate value functions effectively by mapping high-dimensional input spaces (states) to output spaces (value estimates or actions). Hence, NNs enable RL to scale to problems with continuous or very large discrete spaces by learning a compact, parametric representation of the value function. Instead of storing a value for every possible state-action pair, the network learns to predict these values based on the features of the state and action. This not only makes the approach more memory efficient but also allows the agent to generalize learned knowledge to unseen states, improving its ability to make informed decisions in new or previously unexplored parts of the state space. 

However, the use of NNs in RL introduces new complexities and challenges. Training deep NNs is computationally intensive and requires careful tuning of hyperparameters. Additionally, NNs can be unstable and prone to divergence during training, especially in the context of bootstrapped updates like those used in Q-learning. Techniques such as \textit{experience replay} \cite{pmlr-v119-fedus20a}, where past experiences are stored and sampled randomly to break the correlation between consecutive updates, and target networks, which stabilize learning by slowly updating the network parameters, are essential to mitigate these issues. 

Finally, there are some other issues regarding the use of NNs in this type of problem. The first is the loss of explainability \cite{8466590}, as the many parameters a NN can have may obscure the relationship between inputs and outputs. Additionally, NNs must interpolate or extrapolate between continuous inputs, making it difficult to ensure whether the state-action space has been adequately explored, how frequently it has been visited, and whether the interpolations are performing as expected across the entire dimensional space. Nevertheless, DRL has proven to be an effective solution for high-dimensional problems. The following subsection takes a deeper look into various RL and DRL algorithms.

\subsection{RL and DRL Algorithms}

There exists a rich spectrum of RL algorithms. These methodologies fall into two primary classifications: value-based or policy-based; and model-based or model-free algorithms. Each one is tailored for specific scenarios and challenges inherent to RL tasks. \textbf{Value-based algorithms}, as introduced in the previous section, prioritize the evaluation of action-values within states. Central to these methods is the concept of a value function, which predicts the cumulative rewards expected from each state or state-action combination, thereby guiding the agent towards actions that maximize future returns. Q-learning, for example, leverages a tabular approach while, in Deep Q-Networks' case (DQN), a NN to approximate these values across potentially vast or continuous state spaces. These algorithms are particularly effective in settings with discrete, clearly defined action spaces, offering versatile solutions across a multitude of applications. The return is calculated as the discounted sum of rewards, as in \autoref{eq_valuef} and \autoref{eq_qvaluef}, where $r_{k}$ represents the reward received at step $k$, $\gamma$ is the discount factor determining the present value of future rewards (where $0 \leq \gamma \leq 1$), and $\mathbf{E}$ denotes the expectation over the distribution of paths generated by policy $\pi$. The discount factor $\gamma$ plays a crucial role in balancing the importance of immediate vs. future rewards, enabling the agent to make strategic decisions that consider both short-term gains and long-term objectives. 

Conversely, \textbf{policy-based algorithms} such as REINFORCE\cite{Williams92}, Soft Actor Critic (SAC)\cite{haarnoja2018soft}, and Proximal Policy Optimization (PPO)\cite{SchulmanWDRK17} focus on directly learning the policy that determines the agent's actions in various states. This class of algorithms eschews indirect value estimation in favor of optimizing the policy itself, often through techniques like gradient ascent, to maximize expected returns. Their direct approach to learning and optimizing policies renders them well-suited to environments with continuous states and action spaces.

An additional classification for RL algorithms depends on whether they have a model of the environment or not. \textbf{Model-based RL} strategies endeavor to understand and replicate the environment's dynamics to inform their decision-making process. Techniques in this regard, like Dyna-Q\cite{Sutton1998} and Monte Carlo Tree Search (MCTS)\cite{browne2012survey}, merge the insights of value-based and policy-based methods, utilizing models of the environment—whether learned or predefined—to simulate potential future states and rewards. This predictive capability allows for enhanced strategic planning and decision-making, reducing the necessity for exhaustive interaction with the actual environment. Conversely, \textbf{model-free algorithms} learn without knowledge of the transition probabilities and reward functions that describe the dynamics of the environment.

This research primarily focuses on DQN with various adaptations. The main reason for choosing DQN is its suitability for handling continuous state spaces while allowing for the discretization of action spaces. DQN excels in scenarios where defining a precise model of the environment is challenging or impractical. Additionally, as a model-free approach, DQN does not require a predefined model of the underlying environment, making it particularly relevant for applications in financial markets. As a value function method, it computes the optimal policy according to the \autoref{eq_qvaluef} detailed at \autoref{sec_valuefunctions}. Financial markets are inherently stochastic and complex, characterized by unpredictable fluctuations and a multitude of interacting variables. Traditional model-based methods struggle to capture this complexity accurately. In contrast, DQN can learn effective policies directly from interaction with the environment, making it a robust choice for developing adaptive and resilient market making strategies in such a volatile context. Additionally, experience replay has also been applied in every experiment. Instead of processing all the transitions sequentially, experience replay gets random samples from all the stored transitions. This helps achieve more stable and reliable learning, preventing the algorithm from overfitting to recent experiences, among other benefits. Algorithm \ref{algo_dqnbasic} details the standard DQN with experience replay algorithm, as described in \cite{Mnih2015}. All the details can be found in the mentioned work.

\begin{algorithm}
\caption{Deep Q-learning with Experience Replay Algorithm (From \cite{Mnih2015})\label{algo_dqnbasic}}
\begin{algorithmic}[1]
\State Initialize replay memory $\mathcal{D}$ to capacity $N$
\State Initialize action-value function $Q$ with random weights
\For{episode $= 1, M$}
    \State Initialize sequence $s_1 = \{x_1\}$ and preprocessed sequence $\phi_1 = \phi(s_1)$
    \For{$t = 1, T$}
        \State With probability $\epsilon$ select a random action $a_t$
        \State otherwise select $a_t = \max_a Q^*(\phi(s_t), a; \theta)$
        \State Execute action $a_t$ in emulator and observe reward $r_t$ and image $x_{t+1}$
        \State Set $s_{t+1} = s_t, a_t, x_{t+1}$ and preprocess $\phi_{t+1} = \phi(s_{t+1})$
        \State Store transition $(\phi_t, a_t, r_t, \phi_{t+1})$ in $\mathcal{D}$
        \State Sample random minibatch of transitions $(\phi_j, a_j, r_j, \phi_{j+1})$ from $\mathcal{D}$
        \State Set $y_j = 
        \begin{cases} 
            r_j & \text{for terminal } \phi_{j+1} \\
            r_j + \gamma \max_{a'} Q(\phi_{j+1}, a'; \theta) & \text{for non-terminal } \phi_{j+1} 
        \end{cases}$
        \State Perform a gradient descent step on $(y_j - Q(\phi_j, a_j; \theta))^2$ according to equation 3
    \EndFor
\EndFor
\end{algorithmic}
\end{algorithm}

\subsection{Multi-objective RL}\label{sec-multiobjectivestrategies}

To this point, the focus has primarily been on solving single-objective tasks. However, many real-world problems involve addressing multiple objectives simultaneously, which are commonly referred to as multi-objective problems, as discussed in the literature \cite{roijers2013survey,jin2006multi,deb2016multi}. Over time, various techniques have been developed to address this optimization problem.


More precisely, in the domain of RL, addressing multi-objective challenges necessitates strategies that differ notably from those used for single-objective optimization \cite{nguyen2020multi,van2014multi,Hayes2022}. With multiple, potentially competing objectives to consider, the agent must learn to find a balance, often optimizing one aspect while compromising on another to some degree. The common method to navigate this complexity is through the integration of all objectives into a single reward function, which is accomplished by assigning weights or parameters that reflect the importance of each subgoal. The result is a composite objective that guides the agent's behavior.

Yet, the field of multi-objective RL offers an array of alternative strategies that may prove more adequate for certain applications. Multi-objective RL frameworks (MORL) are designed to pursue multiple objectives in parallel, capturing the essence of the trade-offs between them. Instead of forcing a compromise through a single reward signal, these frameworks maintain a spectrum of policies, each attuned to different objective balances, thus offering a suite of solutions among which decision-makers can choose based on their preferences or changing situational demands.

To gain a better understanding of both techniques, the following subsection provides detailed explanations of the two multi-objective approaches: reward engineering and MORL.

\subsubsection{Multi-objective RL via reward engineering}\label{sec-rewardeng-back}

Reward function engineering stands as a cornerstone in the domain of multi-objective RL, offering a methodical way to direct an agent towards achieving multiple objectives \cite{dewey2014reinforcement,Eschmann2021}. The process involves creating a reward function that includes all the goals the agent needs to achieve. However, engineering such functions is complex. It often leads to elaborate formulas that complicate the learning process \cite{vamplew2022scalar}. The task becomes even more challenging when the agent must balance many objectives within a single scalar reward signal. This consolidation necessitates careful calibration, as the weights or parameters involved present a challenge in their specification without prior empirical knowledge \cite{garcia2017incremental}. 

The art of reward engineering extends to more sophisticated techniques than mere weight adjustments. These advanced methods craft complex reward functions that may implement penalty terms. Such terms are designed to modulate one objective in the context of another's performance, thus creating a deliberate interdependence between goals \cite{YOO2021487, Zakaria_2021}. The inclusion of penalty terms represents an effort to balance competing objectives, where success in one aspect may necessitate a controlled concession in another. This nuanced approach allows for a more refined guidance of the agent's behavior, allowing a better alignment of the agent's actions with the multifaceted nature of real-world tasks.

By shaping the decision-making processes of RL agents, these engineered rewards guide them toward achieving a balance of objectives that mirrors the complexities and trade-offs inherent in their tasks. These approaches, therefore, have multiple weaknesses, including the need to define one reward function per a priori goal, difficulty in handling nonlinear relationships between sub-goals, and a lack of explainability regarding the reward, among others.

\subsubsection{Multi-objective  RL via MORL}\label{sec-morlpareto}


Considering the mentioned weaknesses of reward-engineered approaches, another alternative for addressing multi-objective problems with RL involves using MORL frameworks.

MORL differs from single-objective RL in that it requires optimization across multiple dimensions, effectively transforming the classic MDP into a Multi-Objective MDP (MOMDP). This transformation leads to the simultaneous optimization of multiple goals, where rewards and Q-values are treated as vectors to accommodate the complexity and multi-dimensionality of the problem. Consequently, the Q-value function is defined to reflect this vector nature, as in \autoref{qfunction-morl}:

\begin{equation}
\vec Q(s,a) = (Q_1(s,a), Q_2(s,a), ..., Q_n(s,a))  \label{qfunction-morl}  
\end{equation}

Where every $Q_x$ corresponds to one sub-goal. In the same way, the reward is represented by a vector instead of a scalar, as in \autoref{rewards-morl}

\begin{equation}
\vec R(s,a) = (R_1(s,a), R_2(s,a), ..., R_n(s,a))
\label{rewards-morl}  
\end{equation}

Q-functions can be managed using Q-tables when dealing with discrete state spaces or with NNs when handling continuous space states, as in their case. Some MORL algorithms use weights to balance all the goals and, in this sense, obtain the policies that define the PF. However, these weights differ from their reward-engineered alternative, as they \textbf{do not collapse the multiple objectives into a single scalar reward}. Instead, they influence the exploration and exploitation of the solution space to find a balance between the objectives. The main reason for not scalarizing the rewards in MORL is \textbf{to preserve the diversity of the solution set}. By maintaining multiple objectives and representing trade-offs, MORL allows for a richer and more flexible set of solutions. Scalarization can sometimes oversimplify the problem by aggregating the objectives into a single value, potentially losing important information about the relationships between the objectives.

Additionally, MO and MORL solutions are typically supported by Pareto fronts (PFs) \cite{Hayes2022, GARCIA2020104021}. PFs have been extensively utilized in solving various multi-objective problems across different domains, such as engineering, finance, and healthcare \cite{ngatchou2005pareto}. Within the market making domain, for instance, PFs have proven effective in optimizing trading strategies to balance conflicting objectives, such as maximizing profits while minimizing risk \cite{10.1007/978-3-642-15381-5_10}.

\begin{figure}
\centering
\includegraphics[width=0.6\textwidth]{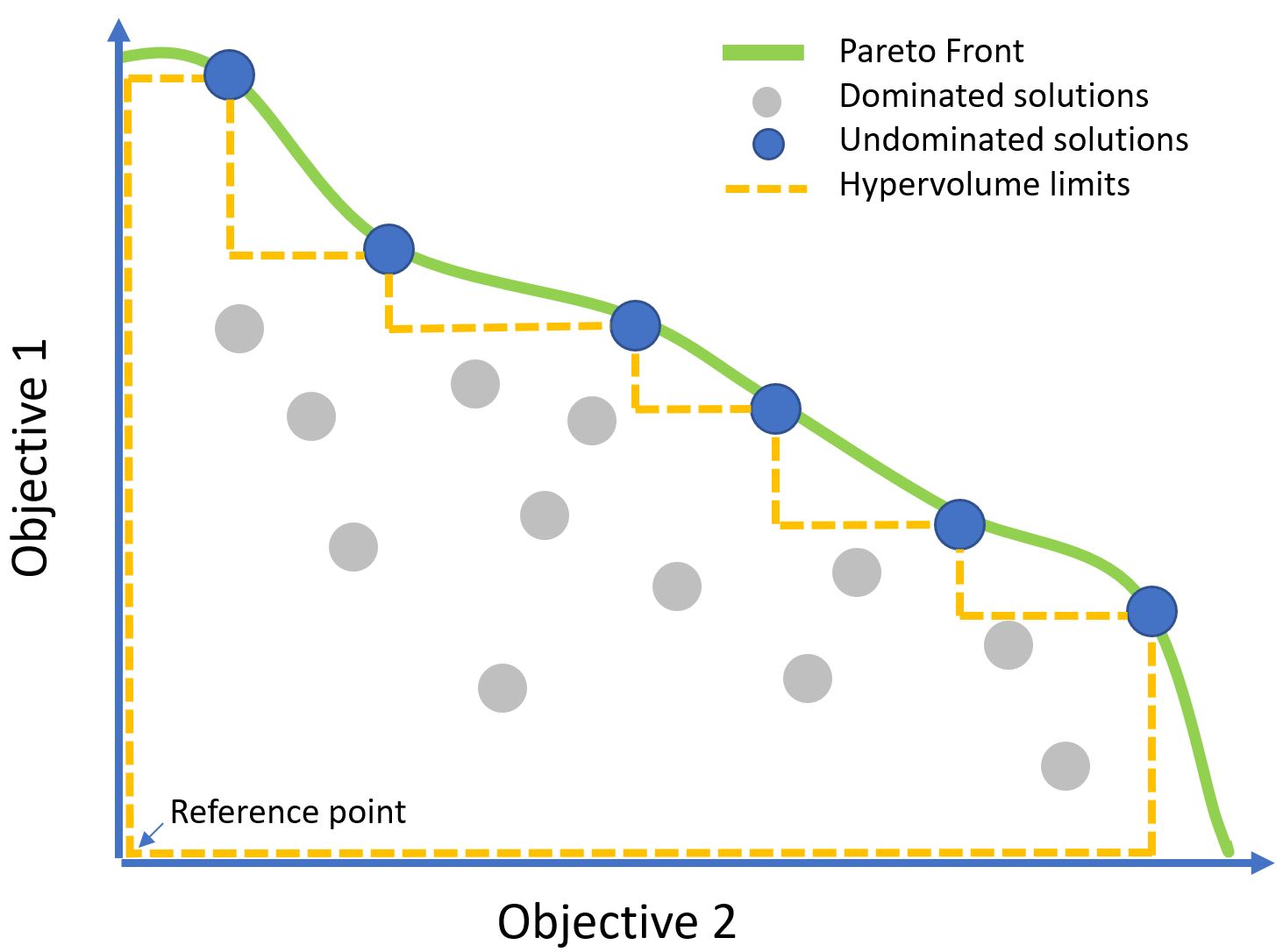}
\caption[Overview of multi-objective artifacts and concepts]{Overview of multi-objective artifacts and concepts.
}
\label{fig:pareto-explain}
\end{figure}

PFs depict the non-dominated policies, which represent the optimal solutions where enhancing one objective would necessitate deteriorating at least one other objective (\autoref{fig:pareto-explain}). By offering decision-makers a spectrum of optimal solutions, PFs enable them to navigate the trade-offs between competing objectives effectively. Decision-makers can analyze a PF to identify the optimal trade-off based on their preferences and priorities. This analytical process facilitates informed decision-making and objective exploration, allowing stakeholders to make decisions aligned with their strategic goals and objectives. Moreover, PFs are not limited to RL applications but are broadly applied in different fields, including engineering design, resource allocation, and portfolio optimization \cite{1599245, 8477730, TANABE2020106078}. Across these diverse domains, PFs serve as a valuable tool for decision-makers to evaluate alternative solutions and make decisions that strike a balance between conflicting objectives.

In summary, MORL frameworks supported by PFs offer a powerful approach to addressing multi-objective RL problems. By providing decision-makers with a comprehensive view of optimal solutions, PFs enable them to navigate complex decision spaces and identify trade-offs that align with their strategic objectives. This approach enhances decision-making processes across various domains, leading to more effective and informed outcomes.

\subsection{Non-stationary RL}\label{sota_nonstation}

In the previous sections, it was assumed that the underlying MDP  remains constant throughout the tasks. However, another issue that may arise is that the MDP changes over time. In this case, it is said that the environment is non-stationary. Non-stationary environments present a significant challenge for RL algorithms, as the policies learned may become suboptimal as the environment evolves. This requires the development of adaptive methods capable of responding to changes and maintaining optimal performance.

In the domain of RL, the quest to tackle non-stationary challenges present in many research areas remains a significant focus \cite{10.1145/3459991}. The strategies used are customized to each problem's unique features, including factors like how long decisions need to be planned for (finite or infinite), the availability of a context library, and the degree of understanding regarding the MDPs and their contexts. Particularly in non-episodic tasks, such as our approach in market making, two predominant strategies for navigating the dynamic environment are performed. Firstly, there is the strategy of adapting the learned policy to reflect the ongoing changes in the environment (single-policy). This adaptation involves continuously updating the policy based on the latest observations and feedback from the environment. By doing so, the agent can better respond to shifts in market conditions, ensuring its actions remain effective and relevant over time. Secondly, there is the approach of maintaining a collection of policies and selecting the optimal one at each decision point (multi-policy). This strategy involves training multiple policies or models to cover a range of possible scenarios or market conditions. Then, at runtime, the agent selects the most suitable policy from this collection based on the current state of the environment. By maintaining a diverse set of policies and dynamically choosing the best one, the agent can effectively handle the variability and unpredictability inherent in non-episodic tasks like market making. 

In addition to the two primary strategies utilized to address non-stationarity in RL, there exists a plethora of other techniques within classical ML aimed at managing this challenge. Continual learning (CL), for instance, represents a subset of methods designed to enable models or agents to seamlessly incorporate new incoming data while maintaining or even improving their performance over time. Unlike traditional batch learning paradigms, CL methods focus on the incremental acquisition of knowledge, allowing models to adapt and evolve as they encounter new information. This iterative learning process enables agents to stay relevant and effective in dynamic environments where the underlying data distribution may change over time. 

Regarding this, the following details various algorithms and techniques from the literature used to address the problem of non-stationarity:

\noindent\textbf{Single-policy algorithms}: Within this domain, strategies often revolve around a single policy that evolves in response to environmental changes. Infinite-horizon scenarios have seen limited focus, with RUQL (Referenced Update Q-learning) by \cite{JMLR:v17:14-037} standing out. RUQL adjusts Q-values in reaction to new experiences or changes in transition probabilities, applying uniform updates based on new data. 
This method is based on having a full understanding of MDP dynamics and is designed for discrete state spaces. 
The Q-learning framework has seen various modifications to better suit non-stationary environments through learning rate adjustments \cite{noda2010recursive,george2006adaptive}. For example, \cite{noda2010recursive} enhanced Q-learning by dynamically altering the learning rate through gradient descent, showing superiority over traditional methods in shifting environments. 
Similarly, Optimal Step Size Algorithm~\cite{george2006adaptive} (OSA)  fine-tunes the Q-learning rate to counteract noise in non-static environments, focusing on minimizing estimation errors rather than directly improving learning outcomes. OSA presupposes prior knowledge of transition and reward structures. 
Regarding financial market non-stationarity, studies like \cite{WU2020142} leverage prior deep learning models to glean insights from technical indicators for RL policy training, though they lack adaptability to newly emerging changes.

\noindent\textbf{Multi-policy algorithms}: Contrarily, some solutions involve maintaining and selectively applying multiple policies or contexts, especially in non-episodic settings. The RL-CD model (RL with context detection)  \cite{10.1145/1160633.1160779} exemplifies this, utilizing a series of models chosen based on performance signals and generating new ones as needed. It requires meticulous parameter adjustments specific to each challenge and is limited to discrete decision-making environments. While other studies like \cite{hadoux:hal-01200817} aim to simplify these aspects, they remain largely conceptual. Alternatives such as HM-MDP~\cite{Choi2001} and ContextQL~\cite{Padakandla2019} explore the use of model-based and policy-variance strategies. ContextQL is also based on change point detection strategies, specifically the Online Parametric Dirichlet Changepoint detector~\cite{K.J.2022} (ODCP), a multivariate detector that identifies whether the dynamics of the environment have changed.

\noindent\textbf{Continual Learning (CL)}: Given the unique challenges of the multi-objective continuous MDP, exploration into CL \cite{Lange2022, Chen2018}, a subset of ML focused on adapting to new data without succumbing to Catastrophic Forgetting (CF) \cite{french1999catastrophic}, is undertaken. CF, the performance degradation following new data integration, is mitigated through strategies such as freezing layers (FL) \cite{Sorrenti2023}, data rehearsal (DR) \cite{ROBINS1995,9412614, Atkinson2021}, and elastic weight consolidation (EWC) \cite{doi:10.1073/pnas.1611835114}, targeting parameter isolation, data replay, and regularization, respectively. While FL, a tactic of adjusting weights in certain layers while preserving others, shows efficacy in supervised learning (SL) and transfer scenarios, its RL application, especially in non-stationary contexts, is less examined. EWC, distinguishing crucial NN weights via a Fisher information matrix, discourages adjustments to key weights, thereby safeguarding vital learned information. DR, alternatively, incorporates a mix of old and new data during training, maintaining a broad experiential spectrum for the agent, a practice particularly valued in RL for enhancing performance and stability.


\section{Financial markets}\label{sec_finmarkets}

To this point, the context of RL in which this thesis is framed has been introduced in detail. The following section will describe the financial context, thus completing the general background of the thesis. Key concepts of financial markets are discussed to provide a better understanding of their movements, dynamics, and characteristics, and where the MM agent will have to perform.

As introduced in \autoref{sec_motivation}, financial markets are characterized by the interaction of various participants, each with different goals, strategies, operations, and impacts on the market itself. For instance,
\textbf{retail traders}, often individuals, engage in trading on a smaller scale, primarily influenced by personal goals and risk tolerance. They add significant volume to trades, especially in highly liquid markets like Forex and stocks. However, the influence of individual retail traders is typically smaller compared to institutional traders. \textbf{Large institutions}, such as mutual funds, pension funds, and insurance companies, trade in large volumes and can significantly impact market prices with their large-scale transactions. Their strategies are typically more sophisticated, involving long-term positions and hedging techniques. \textbf{Arbitrageurs} seek to exploit price discrepancies between markets or securities, adding to market efficiency by ensuring that prices do not deviate substantially from their fair values for long. Their strategies are crucial in maintaining the interconnectedness of global markets, ensuring that anomalies between similar financial instruments are minimized. \textbf{High-frequency traders (HFTs)} use advanced algorithms to move in and out of positions in seconds or milliseconds. HFTs thrive in markets where speed is a critical factor and can capitalize on very small price differences. Their activities contribute to market liquidity and can also lead to increased volatility when large volumes of trades are executed quickly. The interaction between these diverse market participants and technologies not only drives the day-to-day movements in asset prices but also shapes the strategic evolution of financial markets. Innovations such as automated trading and artificial intelligence are reshaping strategies, while global economic events and policy changes continue to test the resilience and adaptability of market structures.

The mentioned participants take part in a perpetual auction, submitting buy and sell orders through a centralized public limit order book (OB), which is crucial in determining asset price movements~\cite{Lu2018}. The OB serves as an electronic catalog that lists all pending (unmatched) orders. It is continuously updated, with new limit orders being added and executed orders being removed in a first-in-first-out (FIFO) sequence. This dynamic process offers a real-time representation of the market's supply and demand dynamics in terms of price and volume, showcasing the vibrant and fluid nature of financial markets.

\begin{figure}[H]
    \centering
  \includegraphics[width=0.5\linewidth]{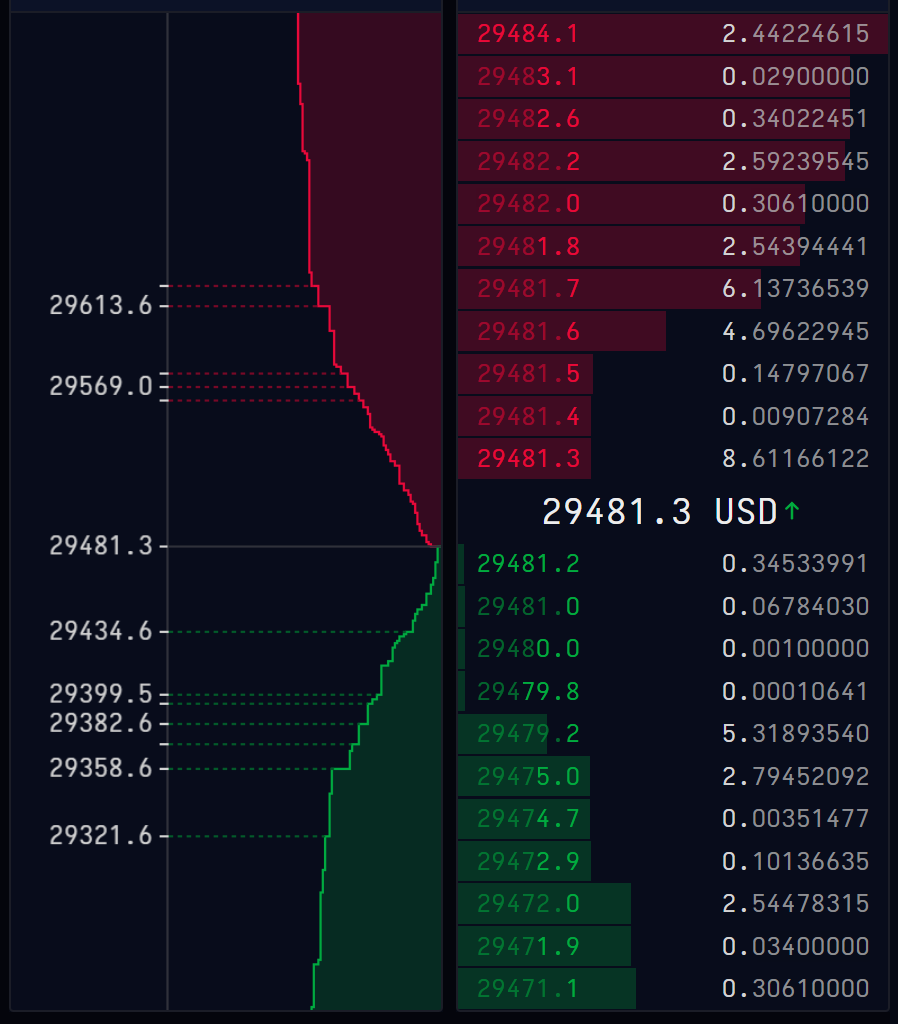}
  \caption[Running OB example.]{Running OB example. Source: \url{https://www.simtrade.fr/blog_simtrade/understanding-order-book-how-impacts-trading/}. (last accessed: October 24, 2024)}
  \label{fig:ob_general}
\end{figure}

Each level within the OB displays the total volume of orders at specific price points, providing valuable insights into the market depth and the potential for upcoming price movements. Such information is vital for market participants in strategizing their entries and exits. Additionally, the transparency of the OB allows for an understanding of market sentiment and potential price support or resistance levels. A graphical depiction of the OB's key elements, illustrating its structure and function, can be found in \autoref{fig:ob_general}. This visualization aids in comprehending how orders are organized and executed, influencing overall market liquidity and price stability.

Liquidity is critical because it reflects the ease with which assets can be bought or sold at stable prices. The level of liquidity in a market is heavily influenced by the number of active participants and the volume of transactions taking place. In conditions where liquidity is lacking, even large orders can lead to severe price fluctuations. Such orders quickly fill up the OB, causing significant changes in the asset's price. These volatile conditions can lead to substantial financial losses for investors in a short period, which can diminish confidence in the asset or even the entire market.

MMs counteract these risks by ensuring a steady stream of orders, which helps to stabilize the market. They act as a buffer against sharp price movements by balancing the buying and selling pressures through their continuous order flows. The vital role of MMs in enhancing the stability and liquidity of financial markets is illustrated in \autoref{fig:OBMMExample}, which shows the impact of MMs on the OB and, consequently, on market liquidity.

One significant characteristic of every financial market is the non-stationarity \cite{MANUCA1996134,MCCAULEY2008820,procacci2023non,Schmitt_2013}. 
Financial markets are not static; they evolve continuously, influenced by a variety of factors. Elements such as trading session schedules \cite{MCINISH2002287}, 
news \cite{OBERLECHNER2004407}
, and economic expectations \cite{doi:10.1177/0093650217705528} 
serve as triggers for these changes, generating non-stationarity in the market itself. These factors collectively impact the inherent dynamism of price movements, complicating the task of defining stable and robust strategies. Furthermore, the introduction of new financial products and the entrance of new technology in trading systems, such as algorithmic and high-frequency trading, also contribute to the non-stationary nature of markets. These technological advancements can drastically change trading patterns and volumes within very short periods, often leading to unexpected volatility or new market trends.

\section{Dealing with the market making problem}\label{sec_dealingwithmarketmaking}

Once introduced to how the financial markets work and provided the main background on RL and ML, it is time to dive into the market making problem. The literature presents many types of strategies within the realm of market making. Before advanced ML techniques revolutionized trading systems, more traditional mathematical and algorithmic approaches were predominant. These classical methods relied on statistical models, stochastic processes, and other quantitative techniques to make trading decisions. However, with the advent of sophisticated ML and RL techniques, the landscape of market making has significantly evolved.

In the following section, both classical and modern approaches to market making are explored. The traditional mathematical and algorithmic strategies that laid the groundwork for automated trading systems are introduced first. These foundational methods have been extensively used in the past.

The focus then shifts to the more recent advancements in ML and RL technologies. This section provides a detailed exploration of how these cutting-edge techniques are applied to market making. This includes an overview of various ML models, mainly SL algorithms, and a deep dive into RL methods that enable agents to learn optimal trading strategies through interaction with the market environment.

By comparing and contrasting these approaches, the strengths and limitations of both traditional and modern strategies are highlighted. This comprehensive review sets the stage for understanding the advanced methodologies discussed later in this dissertation.

\subsection{Market making as a classic financial problem}\label{MM_as_classical}

From a financial point of view, the market making problem has been addressed for many years and even decades. One of the most important contributions and a reference was made by Avellaneda and Stoikov \cite{Avellaneda2008}, where they combine the Ho and Stoll \cite{HO198147} framework with the microstructure of the OB, based on the current inventory and a calculated reservation price. In a nutshell, they try to find a symmetrical optimal spread based on a reference price and a pre-defined inventory strategy. The reference price is conditioned by the imbalance between buy and sell orders placed in the OB. This reservation price is calculated with the following terms: (i) the current mid-price of the stock, (ii) the inventory they want to hold under the strategy, (iii) a risk aversion coefficient, and (iv) the volatility of the traded asset. With this reservation price, and including another additional term that represents the liquidity (density) of the OB, they calculate an optimal spread that fits the initial strategy. 

This work by Avellaneda and Stoikov inspired other authors, such as Gueant et al.~\cite{GUEANT}, who introduced an evolution with a new change of variables based on Hamilton-Jacobi-Bellman equations and a limited inventory to improve the model. More recently, the same authors introduced another solution also based on Avellaneda's framework where they face the problem of multi-asset market making, addressing the challenge of managing correlated assets~\cite{gueant:hal-02862554}. All these works at the end of the day propose an evolution of this first work, by focusing on certain points.  
\newline Excluding Avellaneda's approaches, other works deal with this problem. For example, regarding HFT and its underlying challenges, some authors \cite{Ait_Sahalia_2017} propose specific solutions based on a previous liquidity analysis of these environments. The analysis of liquidity as a predictor of future movements is not only used in market making tasks but also gives us insights into the underlying short-term price intentions. 

As presented, many different paths have been followed to address the market making challenge. Some of them are based on the OB, while others are more focused on liquidity, or other strategies.

\subsection{New techniques to deal with the market making problem}\label{sota_newtech}

Apart from the traditional market making strategies, which are predominantly based on mathematical techniques, the adoption of ML-based strategies has become increasingly common in recent years. The integration of ML techniques into the financial sector has revolutionized many aspects of trading, including market making. This subsection examines how ML, diverging from classical and purely mathematical approaches, offers new dimensions of analysis and strategy development in market making. With the advent of sophisticated algorithms, the traditional landscape of market making has been transformed, allowing for more dynamic and responsive trading mechanisms.

ML in market making can be broadly categorized into two streams: supervised learning (SL) and RL. Each stream provides a unique approach to tackling the complex challenges faced by MMs, from predicting market movements to developing autonomous trading agents that interact and learn from the market environment in real-time.

The following subsections analyze both types of techniques applied to address the market making challenge.

\subsubsection{Market making as a machine learning problem}
Beyond classical and pure mathematical approaches presented in \autoref{MM_as_classical}, there are also some interesting works in the context of ML. The application of ML in trading is primarily categorized into two streams. The first stream focuses on SL, which aims to model or forecast market dynamics directly. The second stream revolves around RL, adopting a distinctly different methodology by developing agents that engage with the market based on a policy refined through learning experiences.

Within the realm of SL, and in the context of market making, researchers leverage historical market data to predict future market trends or to generate trading signals. This predictive capability is fundamental to the design of trading strategies, where signals serve as indicators for executing trades. These predictive signals are typically derived from a variety of sources, including OBs, financial news, and other alternative data sets. The essence of SL in this context lies in its ability to provide foresight into imminent market movements, thereby empowering MMs with the ability to devise more effective strategies. Notably, some studies \cite{10.1007/s11704-014-3312-6} have focused on utilizing SL for the generation of trading signals, while others, like Dixon \cite{Dixon_2017}, have explored its utility in predicting short-term stock price movements, offering valuable insights for MMs in their quest for optimal trading positions. However, all these methods are not easily adaptable to changes in the underlying environment.

\subsubsection{Reinforcement learning in market making}\label{related-work-rlmarketmaker}

In the exploration of RL within the market making domain, a variety of strategies and focuses have emerged, each contributing uniquely to the understanding and advancement of RL applications in financial markets. This section outlines the progression and differentiation in research efforts, highlighting both the evolution of RL strategies and the specialization towards market making in both single-agent and multi-agent environments.

\textbf{Development of RL strategies in market making:} Early works by Chan and Shelton~\cite{Chan2001} introduced the use of different RL strategies, such as SARSA and actor-critic, laying the groundwork for further exploration in this field. They use a very simple simulated market, with discrete state variables. They demonstrate that RL can be further explored to solve MM problems in more complex scenarios. Following this, Yagna Patel~\cite{Patel2018} innovated by combining micro and macro agents, each receiving different inputs, to craft a profitable strategy. Spooner et al.~\cite{Spooner2018} then shifted focus towards creating a realistic simulation of an OB using Q-learning agents, advancing the realism and applicability of RL models in market making scenarios. Continuing with Q-learning, Lokhacheva et al.~\cite{Lokhacheva2020/01} developed an agent influenced by trading signals such as Exponential Moving Average (EMA) and Relative Strength Index (RSI) to define the state space.

\textbf{Multi-agent environments and market making:} Research expanded to include multi-agent environments, recognizing the complexity and dynamism of real-world markets. Wang, Zhou, and Zeng~\cite{Wang2011} conducted multiple experiments in simulated markets with both informed and uninformed traders, with a particular focus on inventory optimization. More recently, Ganesh et al.~\cite{Ganesh2019} delved into competitive multi-agent scenarios using ABIDES~\cite{Byrd2019} as a simulation environment, where RL agents aimed to outperform other MM agents through the discovery of profitable policies.

\textbf{Consideration of inventory risk:} A notable gap in many market making RL strategies is the lack of attention to inventory risk, with a predominant focus on maximizing profits and losses (PnL). Approaches that do not consider inventory management in their solutions present a significant limitation. To bridge this gap, research has evolved to incorporate inventory management and associated risks within RL strategies. The methods applied to address this vary: some works apply penalties to changes in inventory held during trading sessions~\cite{ijcai2020p615, Spooner2018}, while others penalize the entire inventory held to discourage asset accumulation~\cite{Selser_2021, Gasperov2021}. An alternative method penalizes the inventory held only at the end of episodes, often utilizing \textit{Constant Absolute Risk Aversion} (CARA) utility to gauge risk aversion~\cite{symposium, Zhang_2020, Mani2019}.


\section{Simulated environments in RL. ABIDES}\label{sec_abides}

RL utilizes a variety of training methodologies, primarily categorized into offline and online learning strategies. Depending on the learning strategy employed, they can be classified into offline RL algorithms (data-based), and online RL algorithms (interacting with the environment). Furthermore, online learning strategies can be implemented in two distinct ways: one involves operating in real environments, and the other employs simulators. Each approach addresses different research needs and presents its own set of advantages and challenges.

\textbf{Online RL performed in real environments} allows the agent to interact directly with the actual system, enabling real-time learning and adaptation. This method ensures that strategies are developed and refined in the exact environment where they will be implemented, potentially enhancing their effectiveness. However, direct interaction with the live environment means that errors can have immediate and significant financial consequences. Additionally, the complexity and variability of real-world environments can complicate the learning process and affect the stability of the model.

\textbf{Online RL performed in simulators} provides a controlled yet dynamic setting for training and testing RL policies. Using simulators for online RL allows for extensive testing of strategies without the risk of real-world repercussions, provides the ability to experiment with a wide range of scenarios, and generates diverse data scenarios to enrich learning opportunities. However, ensuring that the simulated environment accurately reflects real-world conditions poses significant challenges, and discrepancies in simulation accuracy can lead to strategies that perform well in the simulator but less so in reality. Sim2Real strategies \cite{10.1109/ICRA.2018.8460528,hofer2021sim2real} have to be applied to reduce that gap. Additionally, developing and maintaining a high-fidelity simulator can be resource-intensive.

For this thesis, the decision to employ a simulator-based RL approach is driven by several key considerations, primarily the necessity for a method that offers both adaptability and safety. Utilizing a simulator allows for dynamic interaction and strategic exploration across a diverse range of market conditions, without incurring the financial risks and potential losses associated with real-world market engagement. This controlled environment enables the detailed testing and iteration of RL strategies under simulated market conditions that can be adjusted to reflect various levels of complexity and unpredictability.

The use of a simulator also provides the opportunity to model and analyze how these strategies would perform under extreme market conditions that are rare in the real world but critical for testing the robustness of financial strategies. Furthermore, simulators can accelerate the learning process, as they allow for ``faster than real-time'' simulations, enabling the collection of vast amounts of interaction data in a relatively short period. This capability is invaluable for deep learning approaches that require extensive data to refine and validate their algorithms.

Additionally, the simulated environment supports a level of granularity in control and monitoring that is often unfeasible in real markets. Researchers can easily adjust variables, introduce new elements, or isolate specific factors to observe how these changes affect the performance of RL agents. This flexibility is crucial for dissecting the mechanics of complex strategies and for understanding the nuanced behaviors of RL agents under different market dynamics.

\begin{figure}
    \centering
    \includegraphics[width=\textwidth]{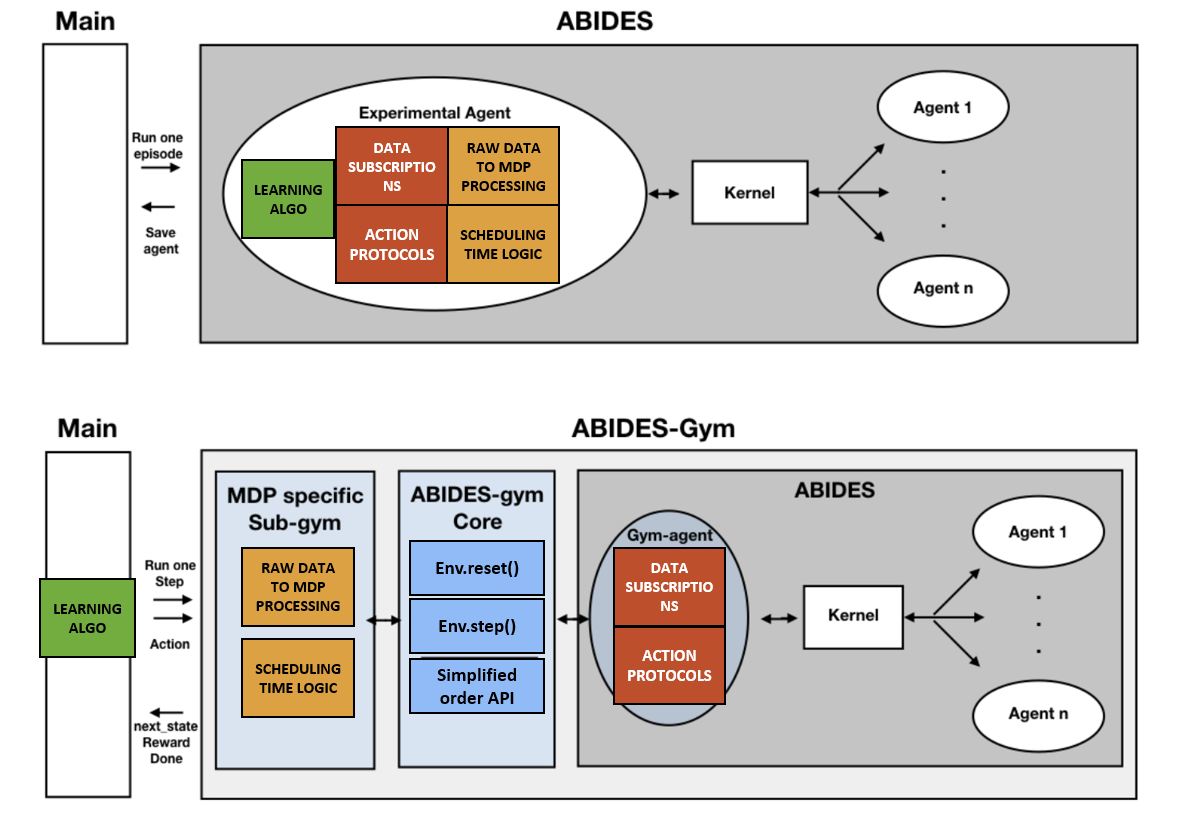}
    \caption[Reinforcement learning framework in ABIDES-Gym vs. regular ABIDES]{Reinforcement learning framework in ABIDES-Gym vs. regular ABIDES. Picture from \cite{10.1145/3490354.3494433}. (adapted fonts)}.
    \label{fig_abidesgym}
\end{figure}

Overall, the choice of a simulated environment for online RL in this thesis provides a practical and effective framework for conducting rigorous academic research. In this regard, all experiments have been conducted in a simulated event-based trading environment called ABIDES\cite{Byrd2019}. ABIDES (Agent-Based Interactive Discrete Event Simulation) is a market simulator that enables the running of multiple setups, which increases the variance of the data and reduces over-fitting, among other benefits. In its more recent versions, ABIDES is being integrated as an OpenAI gym environment \cite{10.1145/3490354.3494433}. Essentially, it is an agent-based simulation tool where defined agents interact with each other in a trading session, much like a real market. Each agent in the market has its strategy (non-informed traders, MMs, momentum agents, etc.), and the natural interaction among them defines the price movement of the negotiated assets. The main architecture of the ABIDES system, both the standard ABIDES and the gym version, is depicted in \autoref{fig_abidesgym}.

The ABIDES simulation framework is centered around a discrete event-based kernel, found in the Kernel class, and is required for all simulations. All agent messages pass through the kernel’s event queue. The kernel supports geography simulation, computation time, and network latency while enforcing simulation `physics' by maintaining the current simulation time and ensuring no inappropriate time travel. Key features of the ABIDES kernel, as described by authors in \cite{Byrd2019}, include:

\begin{itemize}
    \item \textbf{Historical date simulation:} Simulations occur on configurable historical dates, allowing real historical data to be seamlessly injected.
    \item \textbf{Nanosecond resolution:} Time is simulated at nanosecond resolution using Pandas Timestamp objects, allowing agents to operate on different time scales with minimal overhead.
    \item \textbf{Global virtual time (GVT):} Tracks the latest simulated time for which all messages are processed. GVT typically advances faster than wall clock time.
    \item \textbf{Current time per agent:} Maintains a “current time” per agent, incremented after Agent.receiveMessage() or Agent.wakeup(), with message delivery delayed if the agent's current time is in the future.
    \item \textbf{Computation delay:} Stores a computation delay per agent, added to the agent’s current time and outbound message times, adjustable for different computation events.
    \item \textbf{Configurable network latency:} Maintains a pairwise agent latency matrix and a latency noise model, simulating network conditions and agent location, including co-location.
    \item \textbf{Deterministic but random execution:} Uses a single pseudo-random number generation (PRNG) seed for initialization, generating individual PRNGs per agent to ensure identical simulations for the same seed. This setup supports A/B testing without affecting other agents' stochastic sources.
\end{itemize}

Additionally, ABIDES works by using an internal OB, which tracks all open orders and the last trade price for a single stock symbol. All the activity is logged through the exchange agent. The OB implements the following functionality:

\begin{itemize}
    \item \textbf{Order matching:} Matches incoming orders against the appropriate side of the OB, selecting the best price match. If multiple orders have the same price, the oldest order is selected.
    \item \textbf{Partial execution:} Incoming or matched limit orders may be partially executed. The remaining quantity is reduced, and the matching continues. Participants receive an ``order executed'' message for each partial execution, noting the fill price. The average price per share is recorded as the last trade price when the incoming order is executed in multiple parts.
    \item \textbf{Order acceptance:} If the incoming limit order has a remaining quantity after all possible matches, it is added to the OB for later fulfillment, and an ``order accepted'' message is sent.
    \item \textbf{Order cancellation:} Locates the requested order by unique order id, removes any remaining unfilled quantity and sends an ``order canceled'' message.
\end{itemize}

The platform also produces realistic slippage naturally due to dynamic computation and network delays, without the need for an explicit slippage model.

This tool is used in many trading studies \cite{10.1145/3383455.3422554,9383217,doi:10.1080/1350486X.2021.1967767} as it can simulate very realistic trading markets. It allows for the generation of diverse market configurations according to the needs of a specific study.

\section{Discussion}\label{sec_discussion_background}

As stated in \autoref{sec_dealingwithmarketmaking}, the market making problem has been approached from many different perspectives, ranging from more mathematical and quantitative methods to more recent ML techniques. Some works focus solely on profitability \cite{Lokhacheva2020/01,Dixon_2017}, while many others also attempt to manage the inventory problem \cite{ijcai2020p615,Spooner2018,Selser_2021,Gasperov2021,symposium,Zhang_2020,Mani2019,10.1007/s11704-014-3312-6,Chan2001}. Regarding those works that include inventory in their strategies, the main strategy is to include a static penalty term in the reward function. Some apply this penalty at the end of the episode \cite{symposium,Zhang_2020,Mani2019}, and others along the trading session \cite{ijcai2020p615,Spooner2018}. However, in both alternatives, \textbf{the reward function is static and does not adapt to the changing market conditions}, as proposed in this dissertation. Additionally, there are different approaches to training. Some works use historical data to train the agent \cite{Spooner2018,Lokhacheva2020/01,Gasperov2021}, while others generate some kind of simulation based on the OB \cite{symposium}, fitted on historical data \cite{Spooner2018,Selser_2021}, or as the result of agent-based interactions \cite{Chan2001,Ganesh2019,Mani2019}. In this dissertation, the use of an agent-based market simulator is proposed because of the many advantages it has, as broadly discussed in \autoref{sec_abides}. 

Additionally, unlike prior works that primarily targeted profitability and risk management from an RL perspective, \textbf{this thesis expands the focus to include the interaction among MM agents}. The analysis delves into how these interactions affect profitability and explores the diversity of strategies adopted by competing MM agents in both single-agent and multi-agent setups. Direct transfer learning is employed to assess the performance and policies of pre-trained RL MMs in these experiments. This dissertation focuses on applying RL to this problem, using various methods. While there exists some ML approaches in the state of the art, we manage the problem as a non-episodic method that does not rely on a ``world model'', or just a model, of the financial market or some of its dynamics. Attempting to model a financial market given its heavy stochasticity and incomplete information is not an appropriate alternative, from our point of view. The proposed MM must be able to operate continuously, adapting to the non-stationarity of a changing market. 

Last but not least, it is crucial to mention that \textbf{none of the aforementioned strategies consider non-stationarity in their approaches}, which is a substantial weakness. There are RL algorithms that attempt to manage the non-stationarity of the environments, as stated in \autoref{sota_nonstation}. Within the domain of single-policy algorithms, various strategies revolve around a single policy that evolves in response to environmental changes. One notable algorithm is RUQL (Referenced Update Q-learning) by \cite{JMLR:v17:14-037}, which adjusts Q-values based on new experiences or changes in transition probabilities, applying uniform updates based on new data. Despite its innovation, RUQL faces several challenges, such as the diminishing relevance of prior experiences amidst new explorations and a full understanding of MDP dynamics. Additionally, RUQL's design, aimed at discrete state spaces, is not fully compatible with dynamic spaces. Q-learning frameworks have seen various modifications to better suit non-stationary environments through learning rate adjustments. For instance, \cite{noda2010recursive} enhanced Q-learning by dynamically altering the learning rate through gradient descent, demonstrating superiority over traditional methods in shifting environments. However, this model is limited to scenarios with state dynamics. Similarly, \cite{george2006adaptive}'s OSA fine-tunes the Q-learning rate to counteract noise in non-static environments, focusing on minimizing estimation errors rather than directly improving learning outcomes. OSA, however, presupposes prior knowledge of transition and reward structures, a premise not assumed in our work.

In contrast, multi-policy algorithms involve maintaining and selectively applying multiple policies or contexts, especially in non-episodic settings. The RL-CD model \cite{10.1145/1160633.1160779} exemplifies this approach, utilizing a series of models chosen based on performance signals and generating new ones as needed. Despite its potential for adaptability, RL-CD requires meticulous parameter adjustments specific to each challenge and is limited to discrete decision-making environments. While other studies like \cite{hadoux:hal-01200817} aim to simplify these aspects, they remain largely conceptual. Alternatives such as HM-MDP by \cite{Choi2001} and ContextQL by \cite{Padakandla2019} explore the use of model-based and policy-variance strategies. However, they often assume known environmental patterns, which is a risky assumption in trading environments. Notably, change point detection strategies, despite their innovation, may suffer under gradual environmental shifts or false positives, affecting agent performance. For these reasons, CL techniques have been chosen instead of directly applying the aforementioned non-stationary RL algorithms.

In this thorough pursuit, there is a clear line of works and publications that must be considered. In the  \autoref{table_papers}, the evolution of this research is shown. Every objective presented in \autoref{sec-goals} is accompanied by a group of relevant works and techniques. Both works that serve as a comparison with the proposed methods and/or techniques and frameworks that are the basis of the research are clearly presented in the table.

The \textbf{first objective}, ``Evaluate RL for profitable market making in competitive scenarios'', is addressed by building on the previous work of Ganesh et al.\cite{ganesh2019reinforcement}, who explored the use of RL in market making and provided a basis for further research in this area. This first study establishes the foundations by validating RL as an appropriate technique for addressing the market making problem.

The \textbf{second objective}, ``Advanced management of inventory using RL'', focuses on inventory management and is driven by two different approaches. The first, inventory management via reward engineering, is a common method addressed in the literature. The main difference between the works lies in how they define the reward function and the penalty term for inventory holding. This is evident in the works by \cite{Spooner2018} and \cite{Gasperov2021}, which serve as comparison points. However, these approaches share a common limitation: they do not consider the adapting situation of the MM in terms of liquidity. A fixed penalty is established throughout the experiment, affecting it equally at all times. Such restrictive reward functions can negatively impact total returns, as they prompt the agent to act more cautiously. Conversely, the reward function presented in this thesis goes a step further by considering the liquidity ratio (cash/inventory value) of the MM throughout the experiment. This allows the agent to assume more risks dynamically if it gathers more cash from its operations. Risk is, therefore, adapted to the changing situation of the agent without increasing risk aversion. This enables the agent to obtain more profits when the situation is favorable.

Following this approach, and considering the multi-objective nature of the market making problem, a strict MORL alternative is proposed. Reward engineering has many weaknesses and challenges, such as the trial-and-error process of debugging, the need to balance sub-goals, and other issues presented in \autoref{sec-rewardeng-back}. To find a more robust approach that combines RL and multi-objective problems, the work evolves into a MORL framework. Based on previous works by \cite{Vamplew2011}, a MORL MM is proposed. This empowered agent can specialize in different sub-goals separately, avoiding the complex and sometimes biased reward engineering process. This alternative is compared to the previous reward-engineered approach \cite{Vicente2023} to demonstrate the power of this novel application.

Finally, \textbf{the third objective} ``Adaptability in
market making strategies to address non-stationarity'', addresses non-stationarity in market making strategies. To the best of our knowledge, no work addresses the market making problem in non-stationary environments using RL. All the literature revolves around training an agent in a specific market situation. Regarding dealing with non-stationary environments, different strategies have been introduced in \autoref{sota_nonstation}. None of the current non-stationary algorithms, in our opinion, fits well with the current problem. Some rely on having a model of the environment, some are adapted to discrete state and action spaces, and others use changepoint detectors. Apart from RL, non-stationarity has also been addressed using different ML techniques. Most of them, under the concept of CL, allow the model or the agent to learn from new data without losing past knowledge. This dissertation proposes a novel way to deal with non-stationarity from a multi-policy perspective. Additionally, the proposed technique is compared with existing CL techniques, such as EWC \cite{doi:10.1073/pnas.1611835114}, DR \cite{ROBINS1995}, and FL \cite{Sorrenti2023}. Combining a robust multi-objective RL approach with non-stationarity adaptation based on RL is a powerful proposition that can serve as the foundation for many MMs. There are no other works that address this problem in such a holistic way as presented here.

In summary, \autoref{table_papers} presents a comprehensive overview of the research goals and the relevant reference papers that support each objective. This table not only illustrates the evolution of the research but also demonstrates how each objective is grounded in and supported by previous works and techniques. 

All the experiments in this research have been conducted in Python 3.11. All the simulations have been run on a PC Intel i7-9700 Linux (WSL2) with an ASUS 1070GTX GPU, and 32GB RAM. Torch library has been used to train the models in GPU with CUDA.

\begin{table}[H]
\centering
\resizebox{\textwidth}{!}{
\begin{tabular}{|p{3cm}|p{3cm}|p{13cm}|}
\hline
\textbf{Title} & \textbf{Objective} & \textbf{Reference Papers.} \\ \hline
\textbf{RL in marker making operative} & \textbf{Objective \ref{goal_1}}: \newline Evaluate RL for Profitable market making in Competitive Scenarios. & Evolution of previous work:
\begin{itemize}
  
    \item S. Ganesh, N. Vadori, M. Xu, H. Zheng, P. Reddy, and M. Veloso, “Reinforcement learning for market making in a multi-agent dealer market,” Nov. 2019.
\end{itemize} \\ \hline

\textbf{Inventory management via Reward Engineering} & \textbf{Objective \ref{goal_2} (a)}: \newline Advanced Management of Inventory Using RL:  \newline Reward engineered approach & Comparison to other similar RL solutions:
\begin{itemize}

    \item B. Gasperov and Z. Kostanjcar, “Market making with signals through deep reinforcement learning,” IEEE Access, vol. 9, pp. 61611–61622, 2021.
    \item T. Spooner, J. Fearnley, R. Savani, and A. Koukorinis, “Market making via reinforcement learning,” AAMAS ’18, pp. 434–442, Apr. 2018.
\end{itemize} 

Solution based on previous work:

\begin{itemize}
    \item O. Fernández, F Fernández, and F. García. 2022. Deep Q-learning market makers in a multi-agent simulated stock market. In Proceedings of the Second ACM International Conference on AI in Finance (ICAIF '21). Association for Computing Machinery, New York, NY, USA, Article 19, 1–9.
\end{itemize}

\\ \hline

\textbf{Inventory management via MORL} & \textbf{Objective \ref{goal_2} (b)}: \newline Advanced Management of Inventory Using RL:  \newline Multi-objective RL (MORL) Approach & MORL solution based on framework :
\begin{itemize}
    \item P. Vamplew, R. Dazeley, A. Berry, R. Issabekov, and E. Dekker, “Empirical evaluation methods for multiobjective reinforcement learning algorithms,” Machine Learning, vol. 84, no. 1, pp. 51–80, Jul. 2011.
\end{itemize} 

Comparison with the former reward-engineered approach:
\begin{itemize}
    \item O. Fernández, F. Fernández, and J. García. 2023. Automated market maker inventory management with deep reinforcement learning. Applied Intelligence 53, 19 (Oct 2023), 22249–22266
\end{itemize}\\

\hline

\textbf{Non-Stationarity: POW-dTS} & \textbf{Objective \ref{goal_3}}: \newline Adaptability in market making Strategies to Address Non-Stationarity. & Comparison of the proposed method to other CL ML techniques:
\begin{itemize}
    \item Elastic Weight Consolidation:
    \begin{itemize}
        \item J. Kirkpatrick et al., “Overcoming catastrophic forgetting in neural networks,” Proceedings of the National Academy of Sciences, vol. 114, no. 13, pp. 3521–3526, 2017.
    \end{itemize}
    \item Data Rehearsal:
    \begin{itemize}
        \item “Catastrophic forgetting, rehearsal, and pseudorehearsal,” Connection Science, vol. 7, pp. 123–146, 1995.
    \end{itemize}
    \item Freezing Layers:
    \begin{itemize}
        \item A. Sorrenti, G. Bellitto, F. Salanitri, M. Pennisi, C. Spampinato, and S. Palazzo, “Selective freezing for efficient continual learning,” IEEE Computer Society, Oct. 2023, pp. 3542–3551.
    \end{itemize}
\end{itemize} \\ \hline

\end{tabular}
}
\caption{Summary of research goals and reference papers.}
\label{table_papers}
\end{table}


\chapter{RL in marker making operative}\label{chap_icaif}

The initial objective of this thesis is to affirm the efficacy of RL in addressing the challenges associated with market making and to deepen our knowledge and application of RL in the realm of market making in financial markets. The first aim is to develop an agent proficient in executing the MM task of injecting liquidity into a particular asset, ensuring profitability. In this phase, inventory management is not considered, although is included as part of the reward.

Bearing all these aspects in mind, the goal of this chapter is to study the behavior of the MMs from an agent-based perspective. Here, the proposal is the application of RL for the creation of intelligent MMs. In particular, it aims to define an environment where RL MMs compete with other predefined MM agents, and also among themselves, to achieve the most profitable strategy. According to this, the intention is to analyze the behavior of MMs from a competitive and non-competitive point of view. This analysis allows for a better understanding of the trading strategies they employ when trading alone and when they compete against other RL MMs.

The experiments are conducted using ABIDES, introduced in \autoref{sec_abides}, which simulates a regular financial market. It is inside this environment where experimental market makers interact with investor agents that always pick up the cheapest MM in terms of spread.

The chapter is organized as follows: \autoref{icaif_sec::mapping} introduces the task of applying RL on financial markets, defining states, actions, and rewards. It also represents the architecture of RL agents.
\autoref{icaif_sec:singlemulti} introduces the rationale behind the experiments. 
\autoref{icaif_experiments} describes all the experiments launched, with the results obtained.
Next in \autoref{icaif_conclusions}, the main conclusions obtained from the research are discussed.
At last \autoref{contrib_icaif} details the main contributions of this study.

\section{Mapping a market making task onto a RL task}
\label{icaif_sec::mapping}

This section describes the process of mapping the market making problem onto a RL framework. Initially, it delineates the state space $S$, the action space $A$, and the reward function $R$ to provide a comprehensive understanding of the system. The transition function $T$, which models the dynamics of the environment, is assumed to be unknown to the learning agent. Following this, the architecture of the DQN agent, including its NN, and the algorithm that underpins the learning process are presented.

\subsection{State and action space}

The proposed RL MM agent is defined by a state space comprising 10 features that capture the essence of market dynamics: (i) the count of buy transactions completed in the last time step, (ii) the volume of stocks acquired, (iii) the count of sell transactions completed, (iv) the volume of stocks sold, (v) the inventory level (which can be positive for a surplus or negative for a deficit) from the last time step, (vi) the current inventory level, (vii) the fluctuation in the stock's mid-price ($P_{var} = \text{current mid price} - \text{last mid price}$), (viii) the current reference spread, (ix) the previous time step's spread, and (x) the total traded stock volume, typically referred to as $volume$. This selection of state space variables is deemed optimal for capturing crucial market elements, detailing not only the current market conditions like spread and inventory but also the transition from the prior time step, including price evolution—an essential factor for valuing inventory. While other methods, such as employing rolling averages or relative differences of specific inputs, could also be viable, they have not been considered. However, in later chapters, an experimental analysis has been performed evaluating the impact of including trend information.

\textbf{Actions:} The action space for the MMs consists of three primary decision variables at each time step: (i) the buy spread, (ii) the sell spread, and (iii) the inventory volume to hedge. These variables have been discretized, forming an action space spanning these three dimensions. This discretization aims to enhance the learning agent's convergence speed by prioritizing rapid decision-making over pinpoint accuracy in value selection. It is worth mentioning that other approaches based on continuous actions have also been tested previously, but the discretized version has achieved better results. Actions entail choosing a value, referred to as \textit{epsilon} (buy, sell or hedge) throughout this dissertation, from a predefined list of values $\eta_{(b,s)i}\  \in \{-1, -0.8, -0.6, ..., 0.6, 0.8, 1\}$ to adjust buy and sell spreads as $Spread_{mm} = Spread_{mkt} \cdot (1+\eta_{(b,s)i})$, where $Spread_{mkt}$ represents the market's current spread and $Spread_{mm}$ the spread offered by the MM. Hedging actions similarly involve selecting an epsilon to proportionally reduce inventory by a percentage ranging from 0\% to 100\%, as $\eta_{hi}  \in \{ 0, 0.25, 0.5, 0.75, 1\}$. This strategy simplifies the action space to 605 discrete actions, derived from all possible combinations of buy, sell, and hedge epsilons.

\textbf{Rewards:} The reward function is designed to optimize the MtM value, reflecting not only the profits from individual transactions but also the valuation of the held inventory. Although inventory value is included in the MtM calculation, no inventory control measures are applied. Consequently, a single-objective approach is pursued. The reward $R$ at each time step $i$ is calculated as:

\begin{equation}
    R_i =  E_i + PnL_i - HgC_i
\end{equation}

Where $E_i$ represents the earnings or profit from trading activity, calculated as the product of the number of traded stocks and the market spread; $PnL_i$ reflects the inventory's profit or loss based on mid-price changes; and $HgC_i$ accounts for the costs incurred from hedging inventory. 

Through the delineation of state and action spaces, alongside a carefully constructed reward function, the RL MM agent is designed to learn profitable trading strategies within the simulated market environment.

\subsection{Deep Q-learning MM agents (DQL-MM)}
\label{icaif_sec:dqlmm} 

\begin{figure}[H]
        \centering
        \includegraphics[width=0.6\textwidth]{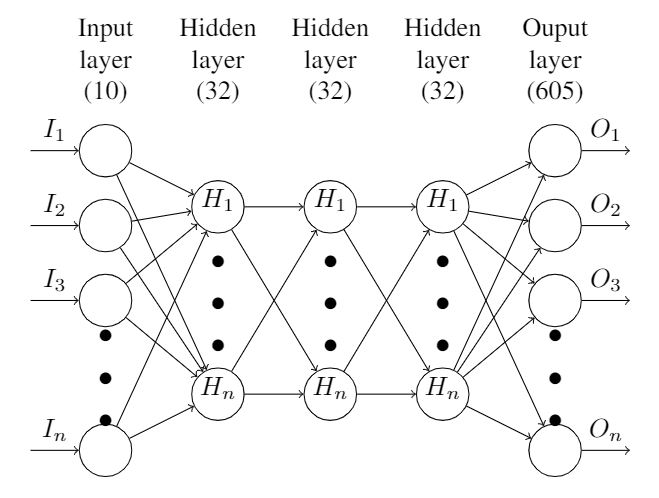}
        \caption{Deep NN architecture used for the DQL-MM agent.}
        \label{pic:nn}
        \end{figure}

Due to the continuous and high-dimensional nature of the proposed RL task, this first approach opts for using deep learning strategies to learn the behavior policy of the RL MMs. Each RL MM is based on a DQN architecture \cite{Mnih2015}. These DQN MM agents (DQL-MM) consist of a fully connected NN that predicts the expected reward of taking an action from a specific state at each time step. The NN is retrained, with its weights adjusted every 200 time steps. The defined architecture for the experiments includes an input layer with 10 nodes, three hidden layers with 32 nodes each, and one output layer with 605 nodes, representing one output node for every possible action. \autoref{pic:nn} provides a graphical representation of the proposed DQN architecture.

All three hidden layers use the ReLU (rectified linear unit) activation function, while the output layer employs a linear activation function. ReLU is an activation function well-known for mitigating the vanishing gradient problem, a common issue in neural networks. Additionally, it allows for faster convergence. For the output layer, since this is a regression problem where expected returns are predicted for each action, the linear activation function is a good option. The Adam optimizer has been chosen because it handles noisy data effectively, and Mean Absolute Error (MAE) is used as the loss metric. Input data has been previously standardized using StandardScaler from Scikit-learn library \cite{Scikit-Learn}.  Each output represents a combination of one buy's epsilon, one sell's epsilon, and one hedge's epsilon. Keras for Tensorflow has been used for the NN model. This architecture was ultimately chosen after several trials. Additional layers (4 and 5) and more neurons (64, 128, and 512) were tested, but no performance improvement was observed. Furthermore, smaller networks typically require fewer gradient steps to converge.

Algorithm \ref{alg:example_DQN} shows the learning process of DQL-MM agents, and it is based on experience replay \cite{Mnih2015}. For every simulation considered, DQL-MM imports NN from the last simulation (line 5 in Algorithm \ref{alg:example_DQN}). Once a simulation starts, along all-time steps DQL-MM queries the current reference market spread (line 8) $Spread_{mkt}$ that comes from the interaction of the rest of agents (\autoref{icaif_subsec:experimentalsetting}). After that, it is time to pick an action. According to the $\epsilon$-greedy strategy and the value of the parameter $\epsilon$, the selected action $a_{t}$ will be the best possible (greedy) (line 11) or a random one (line 13). The best possible action will be returned by the NN that computes the value function of every possible action for that state. DQL-MM calculates rewards earned the previous time step according to taken actions, and the tuple $\langle s_{t},a_{t}, s_{t+1},r_{t}\rangle$ is stored to a buffer $\mathcal D$ (line 17). Every 200 time steps, NN is retrained with stored tuples in $\mathcal D$ (line 20).

\begin{algorithm}[H]
   \caption{DQL-MM algorithm}
   \label{alg:example_DQN}
    \begin{algorithmic}[1]
    \State Initialize memory $\mathcal D \leftarrow \emptyset$, $t \leftarrow 0$
     \For{each simulation}
   \State Init simulation
   \If{Saved agent state}
   \State Get last NN weights
   \EndIf

   \For{$every \ time \ step \ in \ simulation$}
   \State Get  $Spread_{mkt}$ from market data.
   \State Initialise state, $s_{t}$
   
   \If{$rand < \epsilon$}
   \State $a_{t}$=best\_action($buy$, $sell$ and $hedge$ $epsilons$)
   \Else
   \State $a_{t}$=random\_action($buy$, $sell$ and $hedge$ $epsilons$)
   \EndIf
   \State Execute action $a_{t}$, observe $r_t$ and $s_{t+1}$
  \State Apply $\epsilon$ decay
   \State Store transition $\langle s_{t}, a_{t}, s_{t+1}, r_{t} \rangle$ buffer in $\mathcal D$
   \State $t \leftarrow t + 1$
   \If{$t \bmod 200 = 0$}
   \State Retrain NN with transitions sampled from  $\mathcal D$ 
   \EndIf
   \State $s_{t} \leftarrow s_{t+1}$
   \EndFor
   \EndFor
\end{algorithmic}
\end{algorithm}

\section{Single-agent and multi-agent scenarios}
\label{icaif_sec:singlemulti}

Two main cases are studied in this work. Initially, the aim is to evaluate how a single DQL-MM agent behaves in a stationary competitive trading domain, alongside two other non-RL MM agents (\autoref{icaif_subsec:learningsingle}). This evaluation focuses on the DQL-MM agent's performance in terms of profitability and stability compared to non-RL MM agents, as well as on the significance of the variables involved in the decision-making process. Once this scenario is examined, it becomes especially relevant to assess how DQL-MM agents behave when competing against each other in an environment perceived as non-stationary by every agent (\autoref{icaif_subsec:learningmultiple}). In this more complex scenario, MMs must adapt their strategies dynamically in response to the actions of their competitors to maximize rewards. Furthermore, in this multi-agent environment, the effectiveness of policy transfer is evaluated to determine the performance of winning pre-trained policies in new scenarios (\autoref{icaif_subsec:transfer}). This evaluation is crucial for understanding whether successful strategies from simulated environments, such as ABIDES, can be effectively transferred to real markets. The subsequent experiments analyze both the single-agent and multi-agent approaches, providing valuable insights into MM behaviors, strategies, and profitability.

\section{Experiments}
\label{icaif_experiments}

Three distinct configurations are explored in this experimental segment. The first two configurations aim to identify the best strategies in two different contexts: (i) a scenario featuring a single DQL-MM agent in competition with two predefined non-RL MMs (\autoref{icaif_subsec:learningsingle}), and (ii) a scenario where three DQL-MM agents compete against each other, alongside two additional non-RL MM agents (\autoref{icaif_subsec:learningmultiple}). The third experiment (iii) focuses on assessing the effectiveness of transferring pre-trained MM policies into a competitive setting (\autoref{icaif_subsec:transfer}).

The setup for these experiments is detailed in the following \autoref{icaif_subsec:experimentalsetting}.

\subsection{Experimental setting}
\label{icaif_subsec:experimentalsetting}

Each simulation conducted in ABIDES was configured with the following lineup of agents:

\begin{itemize}
\item \textbf{100 Noise agents}: These agents execute orders of a fixed size in random directions.
\item \textbf{10 Value agents}: Equipped with fundamental time series data, these agents make trades based on the deviation of the mid-price from their forecasted mid-price.
\item \textbf{10 Momentum agents}: These agents act upon the crossover of 50-step and 20-step moving averages.
\item \textbf{1 Adaptive POV (percentage of volume) MM agent}: By placing orders at predetermined intervals around mid-price, this agent injects a certain level of liquidity into the synthetic market. POV, or percentage of volume, refers to the proportion of the market maker's order size relative to the total transacted volume observed during the previous lookback window.
\item \textbf{1 Exchange agent}: This agent facilitates the interactions among all participating agents and manages the OB.
\end{itemize}

Each simulation represents a 2-hour market session, limited by computational constraints. However, the proposed strategy performs numerous operations during these sessions, initially providing sufficient data to validate its potential. During these two-hour market sessions, the agents' interactions establish a reference spread ($Spread_{mkt}$) and induce specific price volatility patterns based on their trading behaviors. Within this dynamic environment, the activities of experimental MMs are scrutinized. Each experiment features three distinct types of MMs:

\begin{itemize}
\item \textbf{DQL market maker (DQL-MM)}: The RL MM. It utilizes deep Q-learning (DQL) to make market making decisions.
\item \textbf{Random Market Maker (Random-MM)}: Employs random spreads and hedges at each time step.
\item \textbf{Persistent market maker (Persistent-MM)}: Maintains a constant, randomly selected spread and hedging strategy throughout the simulation. More precisely, it randomly selects one buy, one sell, and one hedging strategy at the beginning of every round and keeps them until that trading round is over. It repeats this process for all rounds.
\end{itemize}

The market making agents in this study are supplemented by interactions with 50 additional investor agents. These investor agents issue buy or sell orders in a uniformly random manner, consistently opting to match their orders with the MM offering \textbf{the narrowest spread—thus, the most cost-effective option for either buying or selling}. Therefore, as it is a zero-sum game, all existing market makers (RL, random, and persistent) \textbf{compete with each other to match the 50 investor orders.} In scenarios where MMs present identical spreads, investors select one at random to prevent any bias that could arise from the order in which MM orders are placed. It is important to note that transaction fees were not considered. This means that all the buys and sells in the simulations were performed at 0 cost.

The DQL agents are configured with specific initial parameters: starting $\epsilon$ is set at 0.99, with an $\epsilon$ decay rate tailored for 250 simulations and a minimum floor of 0.01. The discount factor, $\gamma$, is chosen to be 0.6.

To ensure the robustness and reliability of the experimental findings, a total of 250 simulations were conducted for each experimental setup. Moreover, to mitigate the impact of stochastic variability in the results, 5 rounds of these simulations were executed in parallel using different seeds.

\subsection{Single-agent market maker}
\label{icaif_subsec:learningsingle}

\begin{figure}[H]
        \centering
        \includegraphics[width=0.7\textwidth]{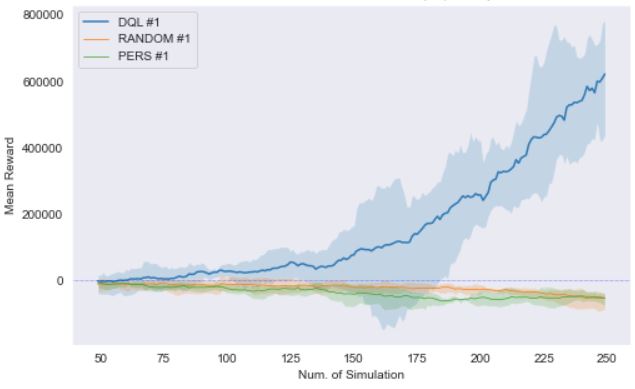}
        \caption{Single-agent training experiment. 50-simulation rolling average window.}
        \label{fig:exp1}
        \end{figure}

The first experiment focuses on training a single DQL-MM agent in a competitive setting against two non-RL MMs: a Random-MM and a Persistent-MM. \autoref{fig:exp1} illustrates the comparative performance evolution of the three market making agents throughout the training period. Notably, the DQL-MM agent begins to secure positive rewards from the early stages of the experiment, outperforming the other two agents, as expected. However, a marked performance improvement is observed around simulation 175, indicating a significant enhancement in the agent's market-making strategy due to increased exploitation from that point onward.

\autoref{table-1} presents the average rewards from 250 simulations for each experimental MM. It is observed that both the Random-MM and Persistent-MM agents conclude the experiment with negative rewards on average. Conversely, the DQL-MM agent successfully learns an effective policy that distinctly surpasses the performance of the former two agents in terms of earnings.

\begin{figure}[H]
    \centering
    \includegraphics[width=0.7\textwidth]{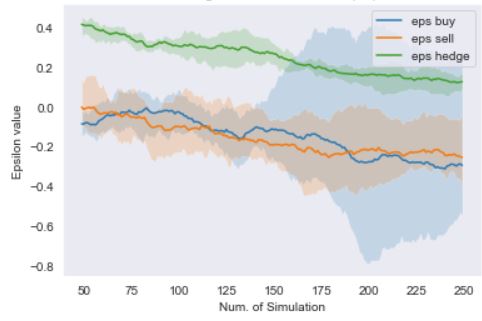}
    \caption{Evolution of epsilons throughout the single-agent experiment 50-simulation rolling average window.}
    \label{fig:eps1}
\end{figure}

\begin{table}[H]
\caption{Single-agent experiment average results (250 exp) (USDx10$^3$).}
\label{table-1}
\vskip 0.15in
\begin{center}
\begin{small}
\begin{sc}
\begin{tabular}{lcccr}
\toprule
MM Agent & MEAN & TOP & BOTTOM & STD \\
\midrule
\bfseries DQL-MM    & \bfseries159   & \bfseries 782 & \bfseries -148& \bfseries 180\\
RANDOM-MM    & -23& 13  & -94 & 11  \\
PERS-MM    & -37& 0 & -91& 15 \\
\bottomrule
\end{tabular}
\end{sc}
\end{small}
\end{center}
\vskip -0.1in
\end{table}
\autoref{fig:eps1} illustrates the progression of buy, sell, and hedge values throughout the experiment, reflecting the actions executed by the market maker. The values represented are the moving averages across 50 simulations for each of these metrics, showcasing the adaptive strategy of the DQL-MM over time. Notably, both buy and sell epsilons tend to converge around a value of $-0.2$, as depicted in \autoref{fig:eps1}. However, the volatility of the buy epsilon increases after the 150th simulation, with noisy and variable buy prices, suggesting that an optimal strategy was not found. The adoption of negative values for buy/sell epsilons implies offering prices lower than the $Spread_{mkt}$, enhancing the likelihood of matching with an investor but yielding lower rewards per trade. Meanwhile, the hedge epsilon value converges near $0.2$, indicating the DQL-MM's strategy to liquidate 20\% of its inventory at each time step, thereby incurring hedging costs at the $Spread_{mkt}$.

\begin{figure}[H]
        \centering
        \includegraphics[width=0.8\textwidth]{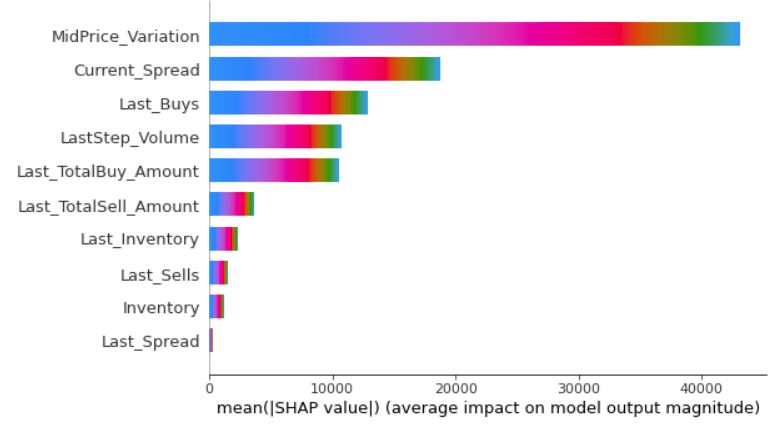}
        \caption{SHAP variable importance summary plot.}
        \label{fig:shap}
        \end{figure}

\textbf{Input variable analysis}:      
A closer examination of the DQL-MM single-agent, which exhibits superior performance despite volatility, allows for an in-depth analysis of the input variables influencing its NN. Utilizing tools like SHAP \cite{Shap}, the significance of various variables in shaping the agent's final policy and strategy can be determined. This analysis not only aids in optimizing variables but also enhances comprehension of the MM policy and its inputs. Correspondingly, \autoref{fig:shap} presents the impact of different state space variables on the agent's action selection process. The length of each bar signifies the variable's importance, with the bar color denoting the specific action taken, as each color represents a distinct action. The graph underscores the special relevance of $P_{var}$ (MidPrice\_Variation) in action determination. The second most influential variable is $Spread_{mkt}$ (Current\_spread), while variables related to inventory $inv$ (Last\_Inventory and Inventory) and the previous time step's $Spread_{mkt}$ (Last\_Spread) exhibit minimal effect on the model.

\subsection{Multi-agent RL market makers}
\label{icaif_subsec:learningmultiple}

\begin{figure}[H]
        \centering
        \includegraphics[width=0.8\textwidth]{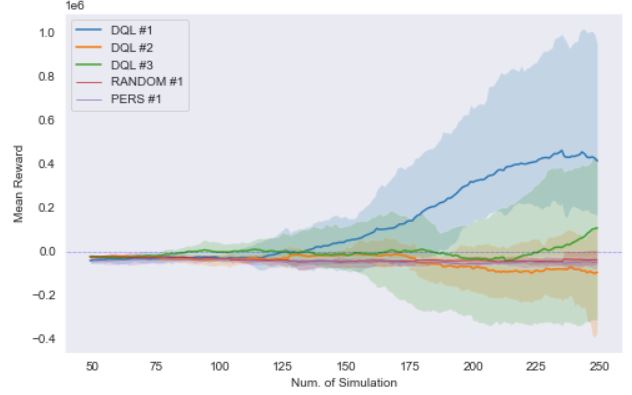}
        \caption{Multi-agent experiment. 50-simulation rolling average window.}
        \label{fig:exp2}
        \end{figure}

In the second experiment, the competitive landscape is expanded to include two additional RL MM agents, resulting in a total of three DQL-MM agents contending against each other within the same trading environment. It is important to note that, to date, no transfer learning has been applied. All the RL agents are being trained in the competitive environment from scratch. This scenario also features the participation of one Random-MM and one Persistent-MM agent. \autoref{fig:exp2} showcases that only the first DQL-MM agent (\#1) manages to secure notable returns, a trend that becomes particularly evident starting from simulation number 175. Interestingly, towards the final phases of the experiment, DQL-MM agent \#3 begins to demonstrate a profitable strategy, which in turn slightly diminishes the performance of DQL-MM \#1. On the other hand, DQL-MM agent \#2 struggles to identify a viable competitive strategy throughout the experiment, ultimately concluding with negative returns.

\begin{figure}[H]
        \centering
        \includegraphics[width=0.7\textwidth]{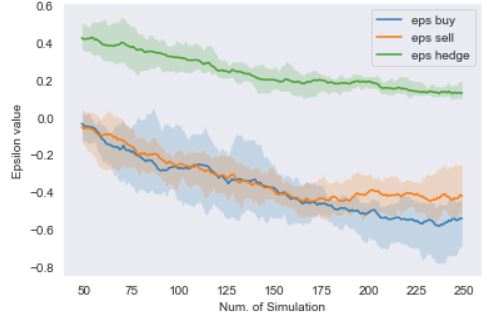}
        \caption{DQL-MM \#1 Epsilons buy, sell and hedge ($\eta$) -50-simulation rolling average window.}
        \label{fig:eps2a}
        \end{figure}

\begin{figure}[H]
        \centering
        \includegraphics[width=0.7\textwidth]{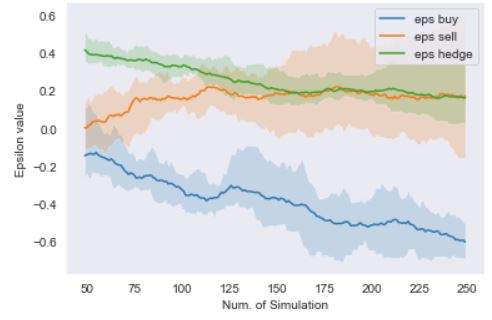}
        \caption{DQL-MM \#2 Epsilons buy, sell and hedge ($\eta$). 50-simulation rolling average window.}
        \label{fig:eps2b}
        \end{figure}

\begin{figure}[H]
        \centering
        \includegraphics[width=0.7\textwidth]{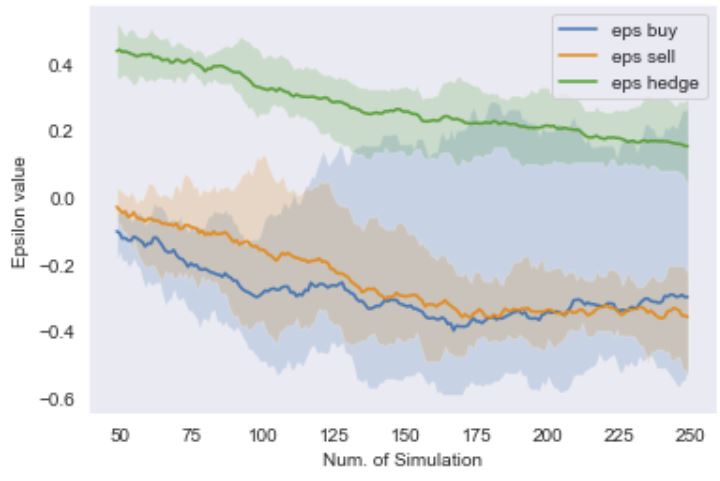}
        \caption{DQL-MM \#3 Epsilons buy, sell and hedge ($\eta$). 50-simulation rolling average window.}
        \label{fig:eps2c}
        \end{figure}

As depicted in \autoref{table-2}, only the first DQL-MM agent (\#1) manages to conclude the experiment with positive average returns. Throughout the experiment, each agent embarks on a journey to discover and adapt to the most beneficial strategy, evidenced by their evolving tactics. This evolutionary process of strategy adaptation is detailed in figures \autoref{fig:eps2a} (DQL-MM \#1), \autoref{fig:eps2b} (DQL-MM \#2), and \autoref{fig:eps2c} (DQL-MM \#3). These figures specifically illustrate the progression of buy, sell, and hedge epsilons for each of the RL agents, showcasing the individual strategic adjustments made throughout the experiment.

\begin{table}[H]
\caption{Multi-agent experiment average results (250 exp) (USDx10$^3$)}
\label{table-2}
\vskip 0.15in
\begin{center}
\begin{small}
\begin{sc}
\begin{tabular}{lcccr}
\toprule
MM Agent & MEAN & TOP & BOTTOM & STD \\
\midrule
\textbf{DQL-MM \#1}   & \textbf{129}  & \textbf{1.018}  & \textbf{-94} & \textbf{179} \\
DQL-MM \#2   & -45 & 111  & -393  & 27 \\
DQL-MM \#3   & -5 & 434  & -343  &  24 \\
RANDOM-MM    & -38  & 9& -87  &  7 \\
PERSISTENT-MM    & -45  & 2  & -85 & 6  \\
\bottomrule
\end{tabular}
\end{sc}
\end{small}
\end{center}
\vskip -0.1in
\end{table}

As observed in the mentioned figures, the buy and sell epsilons among the three DQL-MM agents vary as they dynamically adjust their strategies to respond to the evolving environment. It is important to highlight that, in this zero-sum scenario, any strategy adopted by one agent has a direct impact on the performance of the others; an improvement in one agent's performance typically results in a decrease for the others. Based on this dynamic, the agents develop their strategies in distinct manners. 

DQL-MM \#1 (\autoref{fig:eps2a}) adopts a strategy for epsilons that resembles the approach taken by the sole DQL-MM agent in the first experiment, exhibiting less variance throughout the experiment. The strategy of DQL-MM \#3 (\autoref{fig:eps2c}) closely mirrors that of DQL-MM \#1, albeit with a noticeable increase in the volatility of the buy epsilon as it explores new policies. For both, the average buy and sell epsilons tend towards approximately $-0.4$. Conversely, the strategy of DQL-MM \#2 (\autoref{fig:eps2b}) deviates from the other two agents by shifting its sell epsilons towards positive values.

Hedging epsilons are very similar in all three DQL-MM, converging smoothly to $0.2$. This hedging epsilon seems to be the most favorable in this environment too. \newline It can be concluded that when multiple RL agents participate in the same experiment, they face greater challenges due to the competitive environment, although they strive to adapt their policies to seek profitability.

The hedging epsilons for all three DQL-MM agents display considerable similarity, steadily converging towards a value of $0.2$. This convergence suggests that a hedging epsilon of $0.2$ may be the most advantageous strategy within this specific environment. It can be deduced that the presence of multiple RL agents in a singular experiment intensifies the competitive dynamics, significantly challenging each agent. Despite these challenges, the agents endeavor to adjust their policies in pursuit of profitability.

\subsection{Direct transfer learning in multi-agent RL market makers}
\label{icaif_subsec:transfer}

\begin{figure}[H]
        \centering
        \includegraphics[width=0.8\textwidth]{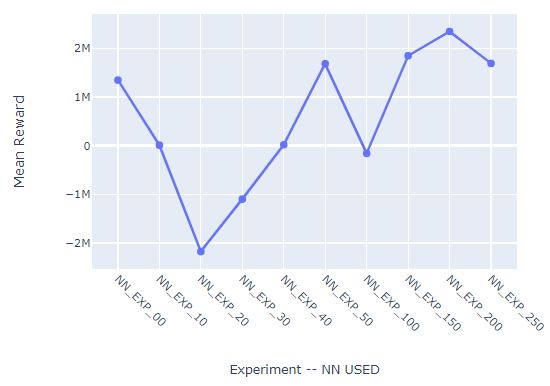}
        \caption{NN Performance evaluation. Mean rewards obtained by every NN tested.}
        \label{fig:nnexp}
        \end{figure}

After completing the experiments in sections~\ref{icaif_subsec:learningsingle} and~\ref{icaif_subsec:learningmultiple}, an interesting area to explore is the performance of the top DQL-MM agents when competing against each other in the same scenario. This exploration aims to verify whether a pre-trained policy exists that outshines the others, thereby having the potential for direct application in subsequent experiments. This concept has been analyzed through direct transfer learning, whereby NNs from the ``best performance'' DQL-MM agents were applied. Specifically, concerning the initial experiment in \autoref{icaif_subsec:learningsingle}, to determine the most effective learned policy, the performance of NNs from 10 distinct simulations was evaluated at intervals: $[0, 10, 20, 30, 40, 50, 100, 150, 200, 250]$. The outcomes of this performance assessment are depicted in figure \autoref{fig:nnexp}. According to the findings illustrated in this figure, the NN that achieved the best performance was from simulation 200; hence, this particular NN was chosen for transfer.

\begin{figure}[H]
    \centering
    \includegraphics[width=0.8\textwidth]{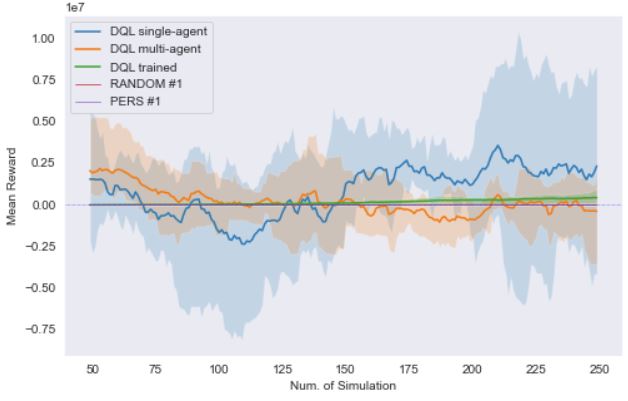}
    \caption{Results from the direct transfer learning experiment. 50-simulation rolling average window.}
    \label{fig:tlexp}
\end{figure}

Regarding the second experiment detailed in Sections~\ref{icaif_subsec:learningmultiple}, the same NN, notably from simulation 200, was employed. This pre-trained NN was introduced into a novel environment where it competed alongside another learning DQL-MM agent, in addition to facing off against a Random-MM and a Persistent-MM agent. \autoref{fig:tlexp}, consistent with previous experiments, illustrates the rewards achieved by each MM  throughout the experiment. It is important to note that pre-trained agents were not continuously trained.

The findings offer several interesting observations. Firstly, the pre-trained DQL-MM agent from the first experiment (depicted in blue) outperforms the other DQL-MM in terms of average earnings (780 compared to 165 for the second one, as shown in \autoref{table-3}). Secondly, the pre-trained DQL agent from the multi-agent scenario of the second experiment, shown in orange, also yields positive returns. While not as high as the first, these returns come with reduced volatility, indicating a potentially less risky approach than that of the single-agent policy. Both pre-trained DQL-MM agents exhibit strong performance up until the 175-200 simulation mark. It is at this juncture that the learning DQL-MM agent, illustrated in green, begins to show an improvement in results, akin to observations from previous experiments. This enhancement in performance results in a decrease in rewards for the pre-trained multi-agent DQL-MM (orange).

\begin{figure}[H]
        \centering
        \includegraphics[width=0.8\textwidth]{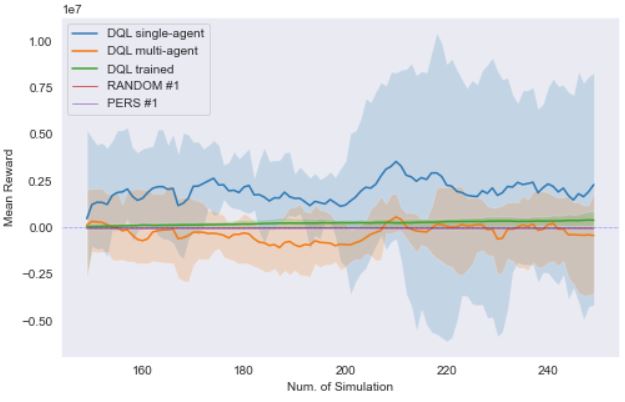}
        \caption{Last simulations transfer learning results. 50-simulation rolling average window.}
        \label{fig:nnexp_zoom}
        \end{figure}

Additionally, post-simulation 200, the pre-trained DQL-MM single-agent (blue) displays a notable increase in the variance of its results, as highlighted in \autoref{fig:nnexp_zoom}. This figure zooms in on the last 100 simulations of \autoref{fig:tlexp}, providing a detailed view of this phenomenon. The performances of the Random-MM and Persistent-MM agents remain negative, consistent with prior experiments. 

It is crucial to point out that the learning DQL-MM did not achieve profitability on par with its performance in the first experiment, where it was the sole RL MM. Moreover, the learning DQL-MM (green) exhibits significantly lower volatility compared to the pre-trained DQL-MM agents, suggesting that learning agents, despite not reaching the highest profitability, offer a less risky and more consistent strategy than relying solely on fixed pre-trained policies, regardless of their initial effectiveness. \autoref{table-3} compiles the average final rewards achieved by each agent.        

\begin{figure}[H]
    \centering
    \includegraphics[width=0.7\textwidth]{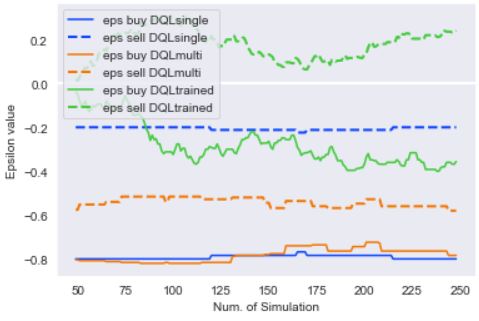}
    \caption{Evolution of epsilon values in a single experiment. 50-simulation rolling average window.}
    \label{fig:single4}
\end{figure}

\begin{table}[H]
\caption{Direct transfer learning experiment average results (250 exp) (USDx10$^3$)}
\label{table-3}
\vskip 0.15in
\begin{center}
\begin{small}
\begin{sc}
\begin{tabular}{lcccr}
\toprule
MM Agent & MEAN & TOP & BOTTOM & STD \\
\midrule
\textbf{DQL-MM single}   & \textbf{780}  & \textbf{10.447} & \textbf{-8.187}  & \textbf{1.442} \\
DQL-MM multi   &165  & 5.268  & -3.622 & 742 \\
DQL-MM trained   & 121  & 877  & -76 & 138 \\
RANDOM-MM    & -37 & 15  & -64   & 5  \\
PERSISTENT-MM    & -50  & -15  & -90  & 5 \\
\bottomrule
\end{tabular}
\end{sc}
\end{small}
\end{center}
\vskip -0.1in
\end{table}
In examining strategies and epsilon values, particularly during one of the five rounds in the current experiment, the evolution of epsilons among the three competing agents throughout the simulations is observed. \autoref{fig:single4} distinctly showcases the emergence of three divergent strategies, each unique to its respective agent. Both the DQL-MM single-agent and DQL-MM multi-agent exhibit similar buy epsilon values, but their sell epsilon strategies diverge significantly. Specifically, the DQL-MM single-agent represented in blue, opts for a less negative epsilon, indicating a preference for fewer but more profitable sell trades. Conversely, the DQL-MM multi-agent (depicted in orange) aims for a higher volume of sell trades, albeit with a lower reward per transaction. This reveals that both pre-trained policies maintain a considerable degree of stability.

On the other hand, the training DQL-MM agent illustrated in green, demonstrates a significant shift in strategy from the outset, aiming to adapt to the competitive environment. It learns a distinctly different policy, even adopting positive sell epsilons, markedly diverging from the strategies of the pre-trained agents.

This analysis leads to the conclusion that pre-trained policies exhibit greater volatility in reward outcomes compared to those still in the training phase. Moreover, the introduction of new RL agents significantly influences the effectiveness of pre-trained policies. Consequently, the pursuit of profitability should not focus solely on fixed, predefined strategies but rather on the development of adaptable, responsive strategies capable of adjusting to evolving market conditions and competitive dynamics.

\section{Discussion}
\label{icaif_conclusions}

In exploring the realm of research, both experimental scenarios—operating as a singular DQL-MM agent and contending amongst multiple DQL-MM agents—demonstrate commendable performance concerning rewards and consistency. RL MM agents exhibit superior behavior compared to random agents, distinctly improving their profit margins through strategic adaptations after a series of learning simulations. This adaptation becomes particularly pronounced in scenarios involving competition with other RL agents, underscoring the dynamic nature of strategy adjustment in pursuit of maximal rewards based on the information at hand.

The concept of direct transfer learning emerges as a noteworthy strategy, facilitating the migration of successful policies into new competitive environments with promising outcomes. This approach underscores the adaptability required in the ever-evolving landscape of RL market making, where pre-trained agents must continuously adjust to the influx of new competitors, rendering the environment increasingly non-stationary. Despite the initial success of pre-trained policies, their effectiveness is invariably tested against the dynamism introduced by new entrants, regardless of the policy's prior achievements.

From the perspective of risk management, pre-trained policies exhibit a higher degree of volatility in average returns compared to those still undergoing learning. This observation suggests a preference for agents that engage in ongoing learning from their environment, rather than relying on static, pre-defined strategies. Such adaptability not only mitigates risk but also enhances the agent's responsiveness to market changes.

From a financial point of view, these experiments lay the groundwork for further exploration into profitable market making strategies within competitive scenarios leveraged on RL. They call for deeper investigation, not just from the MM's viewpoint but also from broader market perspectives. This opens avenues for understanding how RL can be used to optimize trading strategies, improve market liquidity, and navigate the complexities of financial markets with a higher degree of sophistication and efficiency.  Additionally, these experiments open a door to deep dive into profitable MM strategies in competitive scenarios based on RL, not only from MM's perspective but also from other market approaches.

\section{Contribution}\label{contrib_icaif}

The primary contributions of this foundational work include:

\begin{itemize}

\item {\textbf{Application of RL to market making:}} The study demonstrates the innovative application of RL, specifically deep Q-learning (DQL), to create intelligent MM agents in simulated stock markets. This approach allows for the exploration of how RL can be utilized to develop strategies for providing liquidity.

\item {\textbf{Analysis of competitive and non-competitive scenarios}}: One of the novel contributions is the detailed analysis of RL MM agents' behavior in both non-competitive (single-agent) and competitive (multi-agent) scenarios. This differentiation provides insights into how these agents adapt their strategies based on the presence or absence of competition, offering a comprehensive understanding of their performance under varying market conditions.

\item{\textbf{Direct transfer learning evaluation}}: The study contributes to the field by investigating the efficacy of direct transfer learning in market making. It examines how strategies learned by RL agents in one scenario (either single-agent or multi-agent) can be transferred to a new, competitive environment, assessing the performance implications of such transfers.

\item{\textbf{Framework for future research}}: Lastly, the research provides a robust framework for future exploration in the application of RL to financial markets. By highlighting the capabilities and adaptability of RL MM agents, the work paves the way for further studies on the application of ML techniques to trading, investment strategies, and beyond.

\end{itemize}

From a financial perspective, the experiments show that, when RL MMs interact against simple agents, they are much more profitable than when other RL MMs participate simultaneously. It is remarkable how RL MMs adapt their strategies dynamically to adapt to more challenging scenarios, where multiple RL agents take place. These insights can be useful to define further profitable strategies based on ML, not only from MM perspective but also from a stock trading point of view. It is worth mentioning that all these experiments are being performed in simulators. A necessary line of work is required for these strategies to be applied in the real world. This involves first finding a way to accurately simulate the asset of interest and then addressing the gap between the simulator and the real asset. This point will also be discussed in further sections.


\chapter{Inventory management}\label{chap_inventory}

Upon evaluating the profitability of a market making RL agent, the progression moves to the subsequent challenge of managing inventory risk. As stated previously, inventory management is crucial in market making. This evolution of the problem introduces a multi-objective dimension to the task. In the field of RL, there are several strategies to address such multi-objective problems, as presented in \autoref{sec-multiobjectivestrategies}. These mainly include reward engineering and the adoption of a genuine MORL framework.

In this thesis, both two multi-objective approaches are evaluated. Initially, a reward engineering strategy that considers the changing situation of the MM along the trading session is explored in \autoref{amm_sec_appliedintelligence}, demonstrating superior performance compared to other documented methods. Subsequently, a comprehensive MORL framework that leverages PF optimization is delved into in \autoref{morl_sec-morlpareto}. This framework not only addresses the inventory risk management challenges more effectively but also surpasses the performance of the reward engineering approach, offering a more sophisticated solution to the multi-faceted problem of market making.

\section{Reward engineering: Automated market maker inventory management with DRL}\label{amm_sec_appliedintelligence}

Market making inventory management through RL is a topic that has been addressed by some authors, as it is broadly described in \autoref{related-work-rlmarketmaker}. However, most of those solutions rely on adding some sort of penalty term to the reward function, an action that reduces the reward obtained at every time step according to the full inventory value or its value variation. In some cases, this penalty is computed at the end of the trading sessions by removing the full value of this inventory. Nevertheless, both kinds of approaches do not take into consideration the changing conditions of the MM during the trading session, penalizing linearly the accumulated inventory. In this regard, it can be acceptable to have higher levels of inventory if the cash-inventory value ratio is very positive. Being more flexible in this way allows the MM to adopt more aggressive strategies in terms of profitability. 

In this section, it is described how to design an intelligent RL MM agent that can be profitable while managing inventory risk dynamically and adaptively, through a novel reward function. This agent takes into account its changing situation, in terms of liquidity (cash), adapting its policy through the trading sessions. Additionally, an in-depth analysis of the underlying policies is conducted to understand the behavior of different risk-averse agents. A comparison is also run against other existing reward functions to evaluate the performance of the proposed solution.

\begin{figure}[H]

  \includegraphics[width=\linewidth]{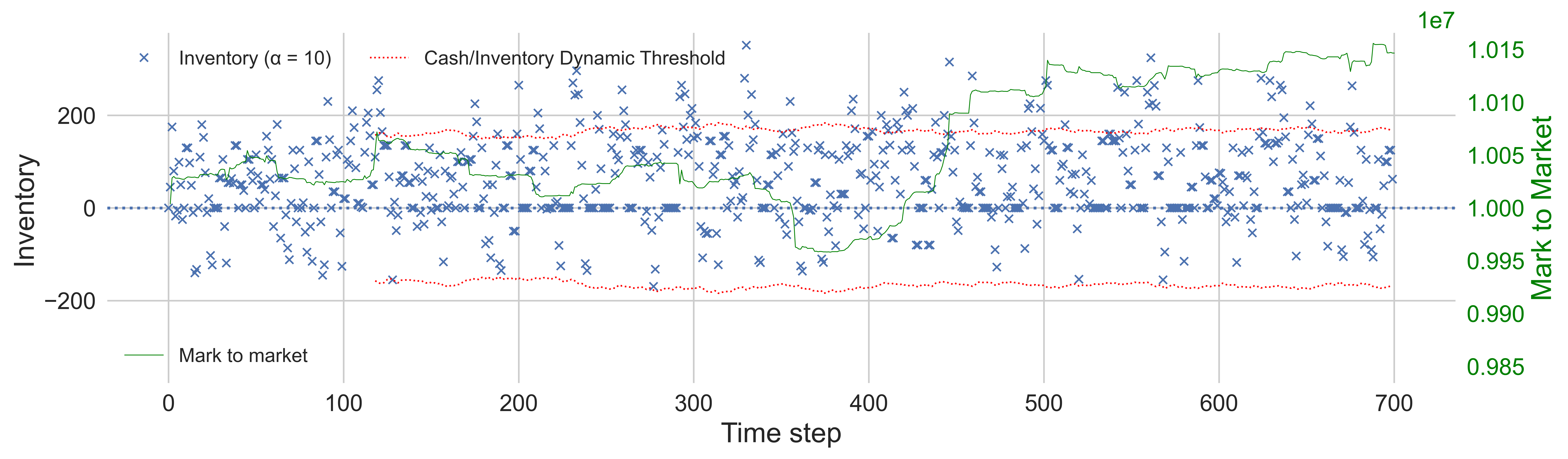}
  \caption[Example of a trading session performed by a RL MM that applies the proposed reward function with both AIIF and DITF coefficients]{Example of a trading session performed by a RL MM that applies the proposed reward function with both AIIF and DITF coefficients. In green it is shown the performance in terms of MtM of the agent, while in blue dots it is shown how the inventory is controlled with the red threshold line. 
  }\label{fig:aaifditfexample}

\end{figure}

To address this issue, two coefficients in the proposed reward function are defined to modulate the inventory exposure: \textit{Alpha Inventory Impact Factor} (AIIF) and \textit{Dynamic Inventory Threshold Factor} (DITF). These two coefficients, widely described in \autoref{amm_sec:newrlmodel}, determine the behavior of the MM through the definition of a dynamic inventory threshold, managing the inventory dynamically at every time step through its contribution to the reward function. \autoref{fig:aaifditfexample} depicts an example of the coefficients in action. 

Regarding the organization of the remainder of the section, \autoref{amm_sec:newrlmodel} details the RL MM agent, including the proposed dynamic reward function. ~\autoref{amm_methods} describes the environment and all the details of the experimental setup. ~\autoref{amm_conclusions} analyses the results of the proposed MM, including a comparison with other existing approaches. Finally, ~\autoref{contrib_aaif} presents all the conclusions derived from this second study.

\subsection{New RL model for inventory management}
\label{amm_sec:newrlmodel}

In order to control the MM inventory, an RL agent has two main levers to interact with. On one hand, it can increase buys or sells if it wants to increase or reduce inventory respectively. Both actions can be achieved by populating more competitive buy over sell prices, or vice versa, depending on the desired goal. With this `skewness', the MM inventory will be increased or decreased accordingly along the trading session due to that imbalance. On the other hand, every agent has the option of selling stocks (or buying) in the market at every moment. This action, usually known as hedging, allows the agent to reduce almost instantly the desired level of inventory held. The main problem with this alternative is that every hedging action is executed by matching a market-order. This means that it has a cost, as every market-order has to pay the current market spread.
Restrictive penalty terms or risk-averse policies usually impact negatively on final returns, as agents must operate more cautiously. Generally, less restrictive penalty terms will translate into higher and more volatile returns. Finding an equilibrium between both goals, to earn profits and to manage risks, should be the main objective of every operator. According to this, non-adapting penalty terms as stated in previous works cannot be optimal by definition. The idea behind the reward function is that MMs should be able to hold, and thereby, manage varying amounts of inventory in response to changing conditions during the trading session. Therefore, if a MM is operating throughout a trading session, its situation in terms of MtM and cash/inventory value evolves as well. This changing situation should be considered to adapt the inventory holding strategy to improve profits. For instance, the trading strategy should not be the same when having a 50\%-50\% cash/inventory value ratio, 80\%-20\%, or 20\%-80\%. In the first case, with an 80\% cash and 20\% inventory ratio, allowing the inventory to increase could be beneficial if it also increases profits, resulting from a less restrictive policy. The opposite approach should be taken in a 20\%-80\% scenario, where inventory risk should be managed more cautiously due to the imbalance. These proportions may also change during the trading session, compelling the MM to adapt its policy accordingly, aiming to find the best balance. This is why the MM operating strategy should not only be a matter of keeping inventory always close to zero units or consistently applying a linear penalty term, but rather finding a balance that enhances profits while managing inventory risk.

In the following subsection, the incorporation of these concepts into the reward function is described. Additionally, the state and action spaces used in this work are detailed. Finally, the training flow performed by the MM is outlined.

\subsubsection{Reward function}
\label{amm_reward-function}

Learning good policies in RL is usually tied to the design of appropriate reward functions. Reward function engineering is, in fact, a matter of research by itself \cite{8329917,7965896,Adams2022,Behboudian2022}. The proposed reward function presented here aims to find the highest trading profitability in terms of returns while controlling inventory dynamically using a specific penalty factor, as stated previously. It is important to remark on again the dynamic nature of this approach, as the penalty factor will adapt to available cash and inventory value at every moment. 
To learn this adapting policy, a proper reward function must be designed that is capable of being profitable while controlling risks. This involves adjusting the operations to the operator's risk aversion and also adapting to the changing situation of the MM.

To achieve that, the following key concepts are introduced:

\begin{definition}[Dynamic Inventory Threshold Factor (DITF)] factor that determines the reference ratio between the value of cash and inventory, which must be respected by the agent throughout the trading session. (\autoref{eq:thr})
 \end{definition}

According to this term, established at the beginning of the training stage, if the goal is for the MM  agent to follow a 1:1 ratio between cash value and inventory value as a reference, then \( DITF = 1 \) should be assigned. This factor defines a dynamic threshold (\( thr \)) throughout the trading session, a threshold that determines the maximum inventory limit that the MM must adhere to at any given time.

\begin{definition}[Alfa Inventory Impact Factor (AIIF)] factor that modulates the agent's risk aversion by including a coefficient in the penalty term of the reward function (as in \autoref{eq-pny})
 \end{definition}

Regarding this factor, the higher this coefficient is, the stronger the penalty the agent will receive from holding inventories out of the allowed threshold, previously defined. 

These two concepts will aid the agent in controlling risks while seeking profitability and achieving it intelligently. It is recognized that the objectives of earning profits and controlling inventory are often at odds, necessitating a strategy that optimizes both. While a static reward function that simply `scalarizes' both goals according to some fixed weight could have been chosen, as is done in other works, it is crucial to note that as the MM operates within a trading session, the proportion between cash and inventory value changes according to the results of the strategy performed. Consequently, the MM might admit more inventory in some cases than others, while maintaining an overall risk aversion ratio. The two previously introduced factors assist in this task. The AIIF factor establishes a general risk aversion threshold that is respected throughout the trading session; while the DITF modulates the overall risks depending on the evolving cash/inventory value ratio. These concepts are incorporated into the penalty term of the reward function.

The reward function is defined as follows:

\begin{definition}[The Reward function for Inventory Management (RIM)]
 is a reward function $R:E \times PnL \times HgC \times Pny \to \mathbb{R} $,  computed as defined in \autoref{eq:reward},

\begin{equation}
\label{eq:reward}
    R_i = \ E_i + PnL_i - HgC_i - Pny_i
\end{equation}
\end{definition}
The different terms of the reward function are described as follows:

\begin{itemize}
\item (i) $E_i$: The trading profits obtained by the MM at time step $t_i$ result from buys or sells. More precisely, as shown in \autoref{eq:profits}:

\begin{equation}
\label{eq:profits}
    E_i = \sum_{x=0}^{n} (B_x \cdot Spr_{Bi}) + \sum_{x=0}^{n} (S_x \cdot Spr_{Si})
\end{equation} 
where $B_x$ and $S_x$ refer to the number of stocks traded (buys or sells) by investors with the MM, and $Spr_{(b,s)i}$ is the buy or sell spread quoted by the MM respectively at that time step $t_i$. Thereby, the higher the spread quoted by the MM, the higher the profits earned, as previously mentioned.
\item (ii) $PnL_i$: The inventory held by the MM will gain or lose value according to mid-price variation in every time step $t_i$. Therefore, if the price rises the inventory value will rise as well, and vice versa. This is what $PnL_i$ reward term refers to, computed as shown in \autoref{eq:pnl}:
\begin{equation}
\label{eq:pnl}
    PnL_i = inv_i \cdot {\Delta} Pr_i
\end{equation}
where $inv_i$ is the quantity of stock held by the MM (positive or negative) at $t_i$, and ${\Delta} Pr_i$ the difference between current and previous stock prices: ${\Delta} Pr_i$ = $Pr_i - Pr_{i-1}$.

\item (iii) $HgC_i$: The MM has the option to reduce its inventory instantly by buying or selling it at every time step. When doing this, it pays a hedging cost as a penalty, as it is executed through a market-order. Hedging costs $HgC_i$ are equal to the amount of inventory hedged multiplied by the market spread $Spread_{mkt}$. It is computed as shown in \autoref{eq:hci}:

\begin{equation}
\label{eq:hci}
    HgC_i =  Hg_i \cdot Spread_{mkt}
\end{equation}
where $Hg_i$ is the amount of inventory hedged, and $Spread_{mkt}$ is the current market spread, at $t_i$.

\item (iv) $Pny_i$: The last term concerns the penalty for holding inventory. This term is especially relevant in this approach as it drives the policy according to the inventory risk. As introduced previously, MMs should carry the least amount of inventory possible, in accordance with their liquidity situation, to mitigate risks caused by price volatility. To achieve this, it is necessary to establish some restriction or penalty that discourages the agent from accumulating inventory during the trading session. Particularly, a dynamic term has been opted for, which can adapt to the MM inventory and cash values at every moment. This term, $Pny_i$, impacts the reward as it penalizes the inventory that the MM holds outside of a specific range. This range, or threshold, is not a fixed value but is dynamically adjusted taking into account the desired proportion between the MM's cash value (liquidity) and the average value of the inventory held in the last $n$ steps.

Hence, the penalty term is defined as in \autoref{eq-pny}: 
\begin{equation}
\label{eq-pny}
    Pny_i = \ AIIF \cdot min\{ \lvert R_i \rvert, \lvert  R_i \cdot \frac{\overline{inv_i}}{\overline{thr_i}} \rvert  \} 
\end{equation}
where:
\begin{equation}
    thr_i =  DITF \cdot \big \lvert  \frac{cash_i}{ \frac{1}{n}\sum_{x=t-n}^{t}  mid_i } \big \rvert   
\label{eq:thr}
\end{equation}
\begin{equation}
\overline{thr_i} = \ \frac{1}{n} \sum_{t-n}^{t} thr_x  
\end{equation}
\begin{equation}
\overline{inv_i} = \ \frac{1}{n} \sum_{t-n}^{t} inv_x  
\end{equation}

The dynamic threshold $thr_i$ (\autoref{eq:thr}) is calculated by dividing the MM cash value at $t_i$ by the average of the last $n$ stock mid-prices $mid_i$. The resultant value is multiplied by $DITF$, the  \textit{Dynamic Inventory Threshold Factor}. This factor defines the proportion between cash and inventory value mentioned above. This proportion is between the cash value at $t_i$ and the average inventory held in the last $n$ time steps. An average value is used instead of the spot value to reduce the impact of price volatility and to stabilize convergence.
 The whole penalty is modulated using the first coefficient $AIIF$, our Alpha Inventory Impact Factor.

 According to \autoref{eq-pny}, it is important to remark that this penalty term behaves as a piecewise function, where this penalty increases linearly from $inv_i=0$ until the dynamic threshold $thr_i$ is reached. The idea behind this is to have a smooth penalty term that discourages proportionally the accumulation of inventory.

In the experimental \autoref{amm_evaluation-exp} the impact of the $AIIF$ factor on profitability, inventory management, and policies is described.

\end{itemize}

\subsubsection{State and action spaces}\label{sec_amm_states}

As with every RL problem definition, some key elements have to be defined apart from the reward function. On one hand, the agent must have a good understanding of the environment. This is achieved by defining a proper state space that contains enough valuable information to select the best possible actions at every time step. On the other hand, it is necessary to determine the action space, hence, all the actions that can be performed by the agent. State and action spaces have been designed based on the previous work of Ganesh et al.~\cite{ganesh2019reinforcement}, with some adaptations such as the action space's discretization.
All these elements are detailed as follows.

\textbf{States}:
The observation state of our agent consists of 8 features: (i) the number of stocks bought in the previous time step, (ii) the number of stocks sold in the previous time step, (iii) the current inventory size (positive or negative), (iv) the inventory size in the previous step, (v) the stock mid-price variation between current and previous time step ($ {\Delta} Pr_i = Pr_i - Pr_{i-1}$), (vi) the current reference bid-ask spread, (vii) the bid-ask spread in previous time step, and (viii) the total amount of stock traded ($V{_i}$), known as $volume$, by the MM in previous time step. The state space was reduced compared to the previous section because it was observed that the agent's performance improved with this reduction. Other space states have been also tested, such as more reduced space states, with worse results.

\textbf{Actions}\label{sec:actions}:
Following the approach outlined in the previous section, the proposed MM agent has three possible actions to execute. First, it can quote a buy price and/or a sell price. These prices, related to the current market spread, will determine how many operations are matched and how much trading profits the agent will earn. In addition, it can decide how many inventory units are hedged through a market-order. To perform these actions there are three variables that MMs can interact with in every time step $t_i$: (i) buy spread $B^s_i$, (ii) sell spread $S^s_i$, and (iii) amount of inventory hedged $H^s_i$. With these three variables, a discrete action space has been defined.  

Hence, buy and sell actions will consist of picking one $\eta$ in the following list of evenly distributed values respectively: $\eta_{(b,s)i}\  \in \{-1, -0.8, -0.6, ..., 0.6, 0.8, 1\}$. 

Therefore, streamed buy/sell spreads (prices) correspond to \autoref{eq:buysellspreads}: 
\begin{equation}
\label{eq:buysellspreads}
    B_i\textsuperscript{Sp}, S_i\textsuperscript{Sp} = Spread_{mkt} \cdot (1+\eta_{(b,s)i})
\end{equation}
where $Spread_{mkt}$ is the current price spread streamed by the market, and  $B_i\textsuperscript{Sp}$, $S_i\textsuperscript{Sp}$ are the spread that MM quotes at time $t_i$.  Both buy and sell $\eta_{(b,s)i}$ can be equal or different, incurring in asymmetry in the latter case. This asymmetry, also known as skewness, allows MM to balance between buys and sells if needed. 

In addition, to buy or sell actions, hedging action will consist of picking another $\eta$ from the list $\eta_{hi}  \in \{ 0, 0.25, 0.5, 0.75, 1\}$. Depending on the specific $\eta_{hi}$ selected by the agent, its inventory will be reduced proportionally by 0\% to 100\% in the following time step $t_{i+1}$. Thereby, as shown in \autoref{eq:inv-i}:
\begin{equation}
\label{eq:inv-i}
    inv_{i+1} = inv_i \cdot \eta_{hi}
\end{equation}
 Where $inv_i$ is the inventory held by the MM at $t_i$. Hedging action is not free, as it will incur specific costs as described in \autoref{amm_reward-function}. 

In conclusion, the action space consists of $11$ buy $\times$ 11 sell $\times$ 5 hedge $\eta$, represented by the tuple $a_i = (\eta_{bi}, \eta_{si}, \eta_{hi})$. This solution was chosen over a continuous action space to increase the convergence rate while sacrificing a more concrete value, which is not considered necessary for this exercise. Additionally, using relative values instead of absolute figures helps the agent to generalize better if the domain values change. For example, having at $t_i$ a market spread of $Spi_t = 7$, an inventory of $inv_i = 150$ units, and the following $\eta$ returned by the agent $a_i = (0.8, -0.2, 0.5)$, the MM would stream the following values: $B_i^{\text{Sp}}=7 \times 0.8 = 5.6$, and $S_i^{\text{Sp}} = 7 \times -0.2 = 1.4$. Additionally, inventory would be reduced to $inv_{i+1} = inv_i - 75 = 75$ units, regardless of inventory increases or decreases due to the rest of trading operations.

\subsubsection{Market maker's training algorithm}\label{sec_amm_algo}

The training flow follows the same process as presented in Algorithm~\ref{alg:example_DQN}, in the previous section. The only difference is that, in this work, two neural networks are used instead of one to enhance stability during the learning phase. In this setup, after completing the training stage (line 20), the neural network used for training is copied to the target network. This target network is then utilized by the algorithm to select actions.

\subsection{Experimental setting and methods} \label{amm_methods}

Due to the continuous state space, and the non-episodic nature of our approach, a Deep Q-Network for the MM agent has been used. As introduced in previous sections, the Q-function Q(s,a) is approximated by a DQN, by using a fully connected network. In fact two, instead of one, identical NNs are used: a training NN and a target NN. This improves the stabilization in the learning stage~\cite{Mnih2015}. These NNs have been defined with a total of 8 state space variables in the input layer and 605 possible actions in the output layer. Furthermore, and similarly to the previous section, the agent's network has 3 hidden layers with 32 neurons in each one. The input and the three hidden layers have ReLu as the activation function. The output layer uses a linear activation function. For every given state the NN approximates the reward of every action, selecting the action with the best-expected reward. Epsilon-greedy is, again, the exploration strategy. Other network architectures have also been tested, such as $3\times32$, $3\times64$, and $5\times64$ (layers $\times$ neurons).

Regarding the training stage, the NN is fitted every 200 time steps. Experience replay is used to improve the training. According to this, all the experiences tuples $e_i = (s_i,a_i,r_i, s_{i+1})$  are stored in a replay-buffer of 1M size.
Every epoch, the training NN takes mini-batches of 1.024 random experiences from this replay-buffer fitting the weights using back-propagation. Adam has been chosen as the learning optimizer and MAE as the loss metric. The learning rate is set to $lr = 0.01$. The discount factor has been set to $\gamma = 0.6$. These hyperparameters have been selected after testing other alternative values.

Regarding the trading environment, there are three types of experimental MMs competing with each other in every simulation, as follows:

\begin{itemize}
\item \textbf{1 DQN market maker} (DQN MM): The deep Q-learning intelligent MM proposed in this approach. 
\item \textbf{1 Random market maker} (Random MM), a MM with uniform random spreads and hedges on every time step. Same action space as DQN. 

\item \textbf{1 Persistent market maker} (Persistent MM), a MM with fixed uniform random spread and hedges over every simulation. The spreads and hedges are selected at the beginning of each simulation and kept until the simulation is finished. Same action space as DQN.

\end{itemize}

\begin{figure}[H]
\centering
  \includegraphics[width=0.8\linewidth]{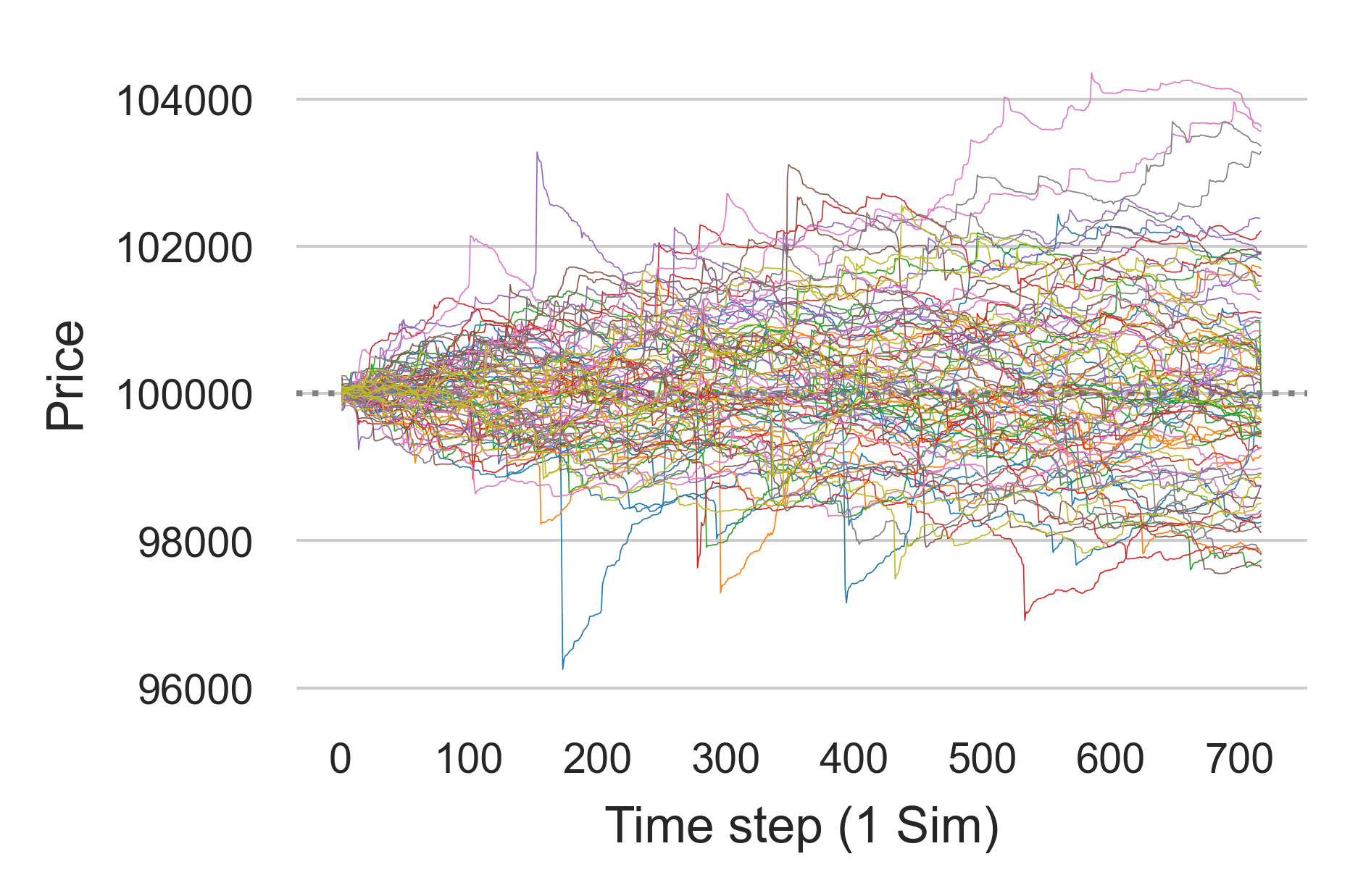}
  \caption[Mid-prices evolution sample.]{Mid-prices evolution sample. This figure illustrates how the price changes along a random sample of different simulations, showing the stochastic nature of ABIDES simulator.}\label{fig:prices_evolution}
\end{figure}

\begin{figure}[H]
\centering
  \includegraphics[width=0.8\linewidth]{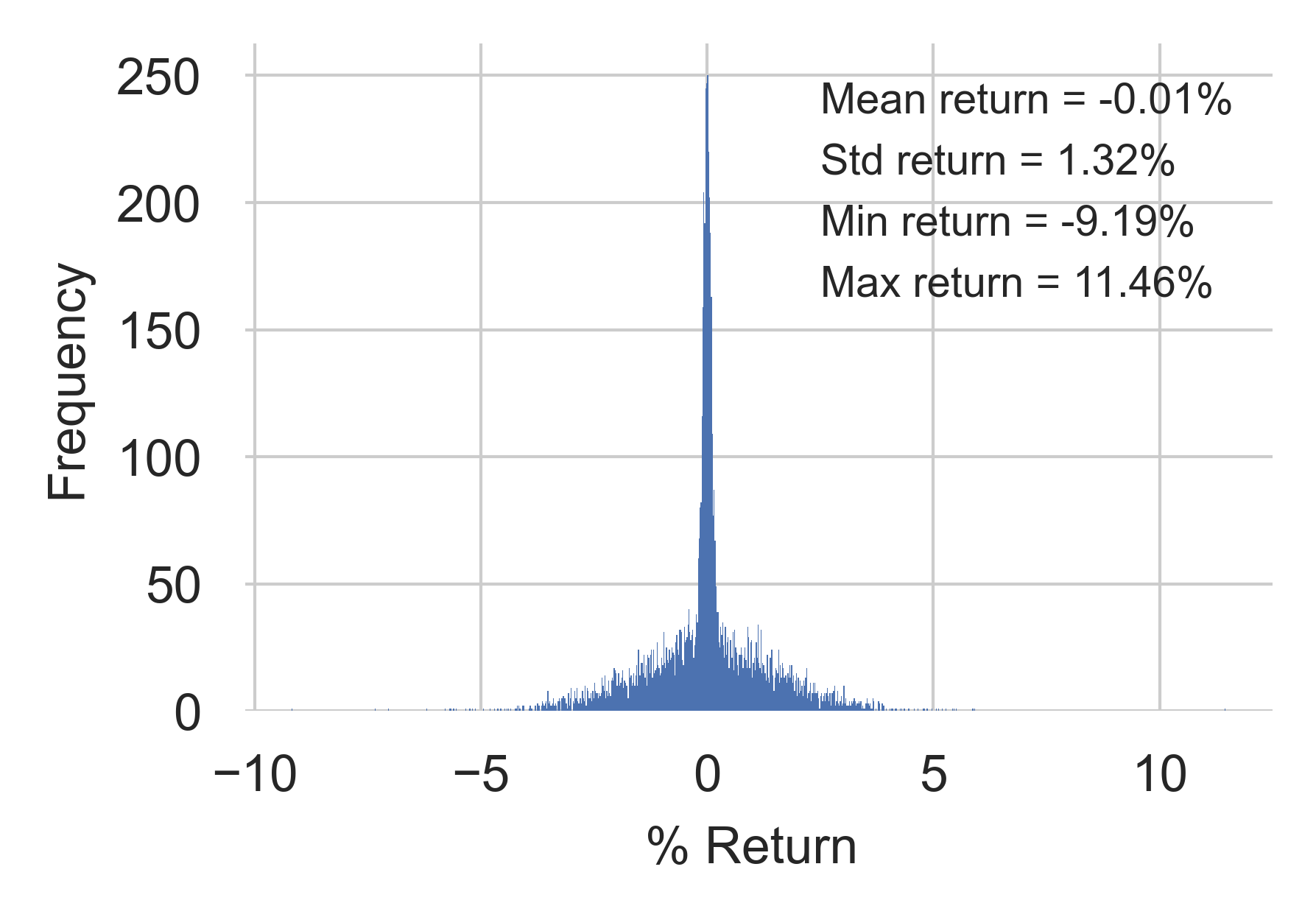}
  \caption[Histogram of returns.]{Histogram of returns.  The plot shows the returns (price variations) distribution at the end of all simulations. $Ret_t = (Price_t/Price_0) - 1 $. }\label{fig:price_returns}

\end{figure}

In this work, an attempt was made to generate a `standard' market in ABIDES in terms of price volatility. With this setup, the goal was to achieve credible price movements, while maintaining the stochastic essence of every trading environment. According to this, the market simulation configuration in which the experiments were conducted included the following participants/setup: 100 noise agents, 10 value agents, 10 Momentum agents, 1 adaptive POV agent, and 1 exchange agent. The market generated from the interaction of these agents has enough liquidity to allow for different spreads along the simulations. Additionally, the returns are reasonable and the price movements are suitable for a standard financial market. To illustrate these points, Figs.\ref{fig:prices_evolution} and \ref{fig:price_returns} show different price trajectories and returns respectively across full experiments.

All these agents interact for two hours, determining a trading session. This interaction defines the movement of the stock price, its volatility, and also the reference market spread $Sp$. In addition to these agents, 50 investor agents interact solely with the MMs. These investor agents place random buy or sell orders with a fixed size. As greedy investors, they always choose the MM with the narrowest spread available in the OB, hence the cheapest one. If there were many MM with the same cheapest $B_i\textsuperscript{Sp}$ or $S_i\textsuperscript{Sp}$, the investor would pick one of them randomly.

As mentioned above, every training/testing simulation is based on a 2-hour market session (9:30h to 11:30h). One DQN MM agent, one Random MM, and one Persistent MM are included. Full experiments have a total of 150 independent simulations (trading sessions), with 5 total experiments per setup. No trading costs have been considered. Asset prices evolve stochastically according to agents' natural interaction. All simulations begin with an opening price of \$1.000,00. The experiment results are evaluated in terms of the total MtM and cash/inventory value ratios at the end of the trading sessions.

\subsubsection{Experimental evaluation}\label{amm_evaluation-exp}

In this subsection, all the experiments are described, from the agent training to the comparison with other existing reward functions. Initially, to evaluate the impact of different penalty factors, the DQN MM is trained with 10 different setups. This phase provides an initial idea of the learning and performance of the different MM setups. Once these DQN MM are trained, they are tested in new fresh environments. The test results are then analyzed in terms of profitability, inventory management, and cash/inventory value ratio as the main metrics. Furthermore, the policies performed by the DQN in all the testing environments are inspected. Finally, to compare the proposed solution against other existing approaches, additional experiments are set up in which these reward functions are evaluated, demonstrating how the proposed solution is capable of generating better MM policies.

\subsubsection{Training the market maker}\label{amm_training-section}
Our DQN MM competes against two additional MMs in the trading sessions: one random agent, and one persistent agent. To evaluate the DQN MM performance in terms of profitability and inventory risk management, multiple training setups have been launched applying different \textit{Alpha Inventory Impact Factors} (AIIF). A total of 10 different AIIFs have been tested. Starting from  $AIIF = 0$, where no inventory penalty is applied to the reward function, up to $AIIF = 100$ where inventory impact is severe. Therefore, $AIIF = \{0, 0.2, 0.5, 0.8, 1, 1.5, 2, 5, 10, 100\} $. In addition, the \textit{Dynamic Inventory Threshold Factor} DITF, has been established in this experimentation with a fixed value of $DITF = 0.5$. This value was chosen because it is considered feasible to keep the inventory value below 25\% of the total MtM at any time. All MMs start with cash of \$100,000.00 and 0 units of asset inventory. A total of approximately 612K gradient steps have been performed per single experiment.

\begin{figure}[H]

  \includegraphics[width=\linewidth]{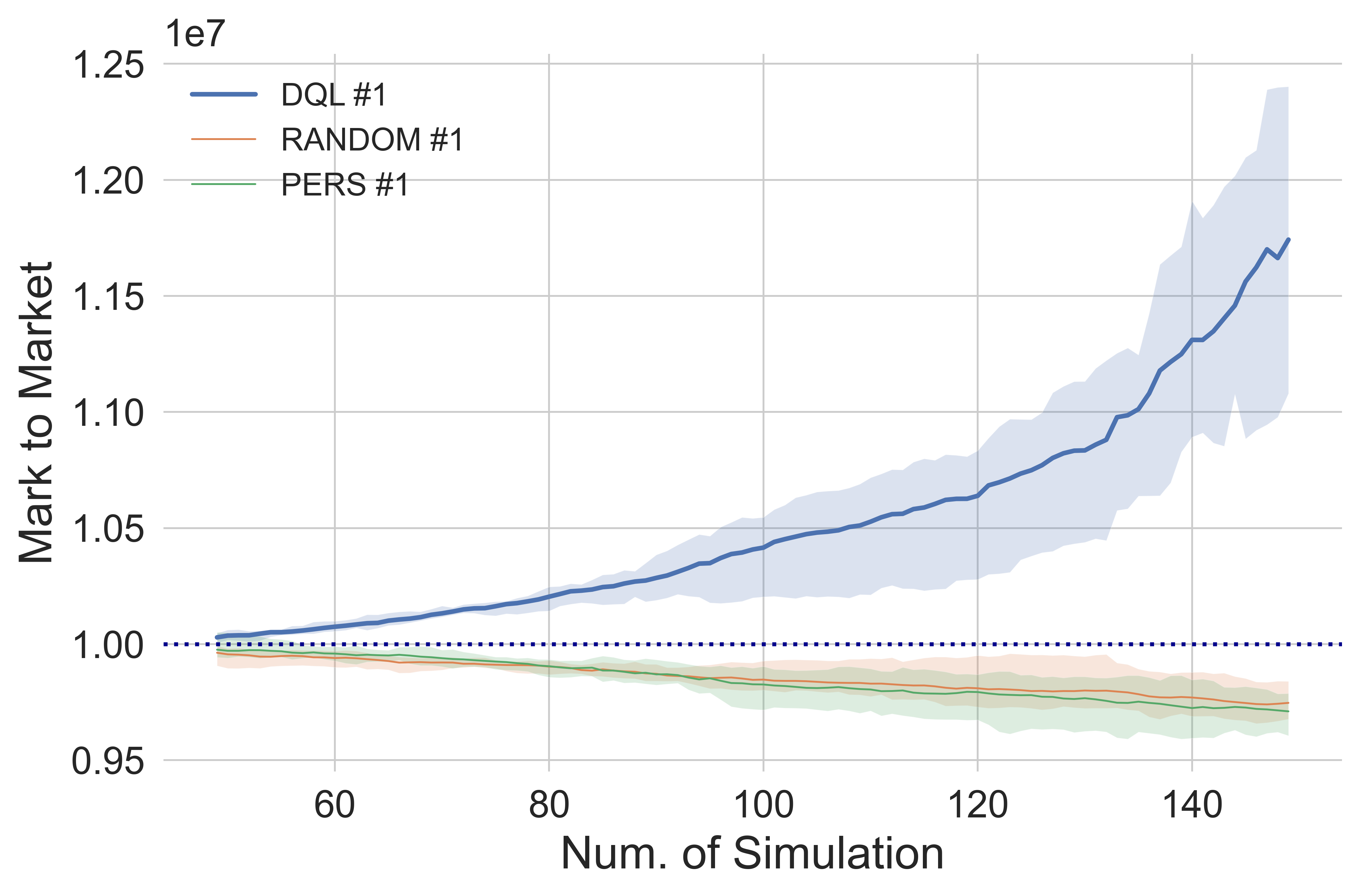}
  \caption[Single training experiment, applying an AIIF = 0 in the reward function]{Single training experiment, applying an AIIF = 0 in the reward function. The three competing agents are shown. It is noted how the DQN MM agent can learn along simulations how to be profitable, compared to the other two random agents.}\label{fig:training_single_exp}

\end{figure}

\begin{figure}[H]

  \includegraphics[width=\linewidth]{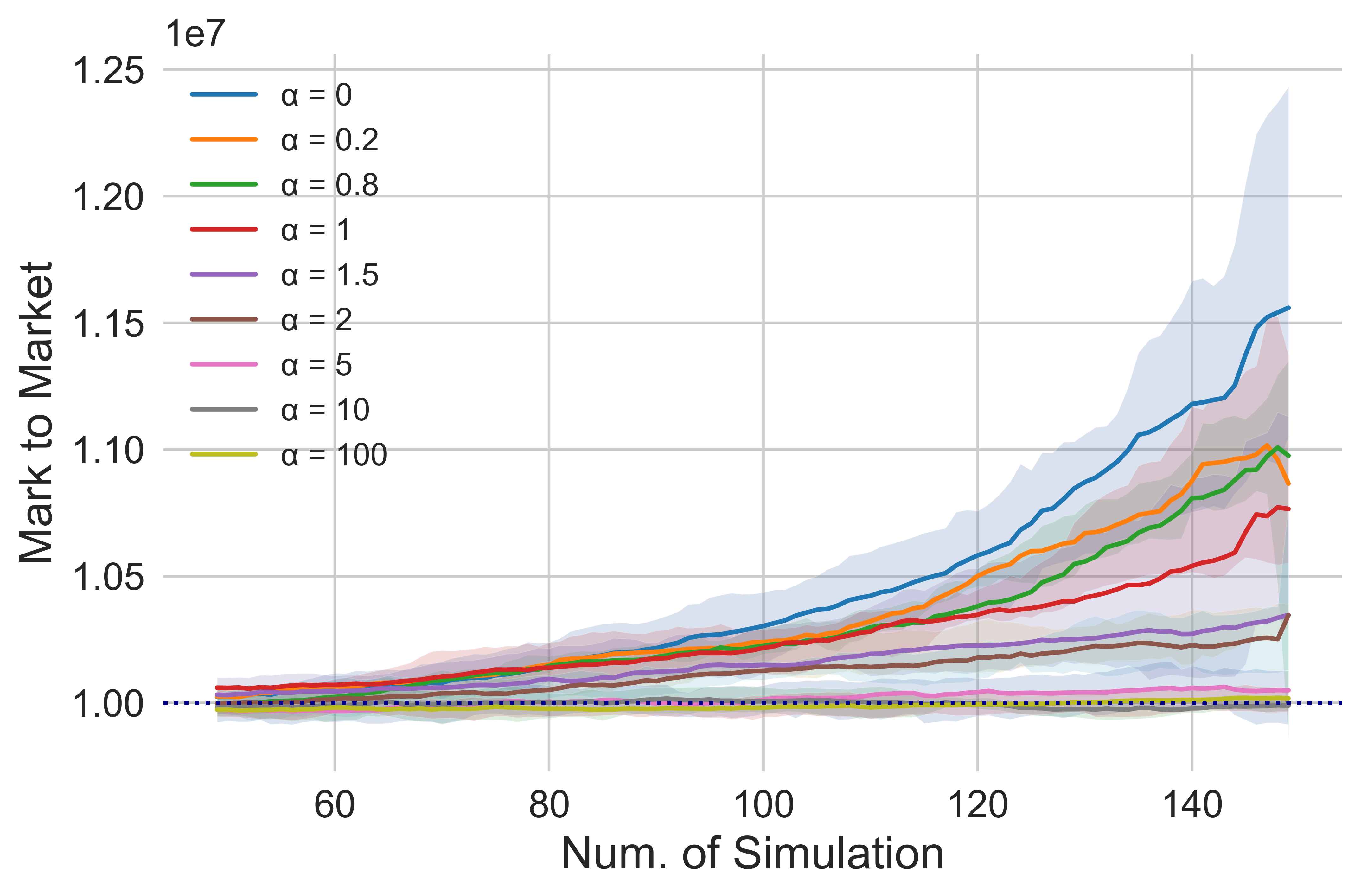}
  \caption[DQN MM training results applying different $AIIF$.]{DQN MM training results applying different $AIIF$. It is noticeable how MtM return reduces as long as the $AIIF$ factor is increased. This effect has to do with the control of inventory. (Random and Persistent agents are not shown).}\label{fig:training_alphas_mtm}
\end{figure}

The first training results (with $AIIF = 0$) show very good performance in terms of MtM, finding a more profitable policy than random and persistent MMs (\autoref{fig:training_single_exp}). These two non-intelligent MM, in fact, end with lower MtM even than at the starting point. Regarding alpha coefficients (AIIF), and taking a closer look at \autoref{fig:training_alphas_mtm} where the different values of $AIIF$ are represented, it is remarkable to see how the slopes of the MtM curves moderate as long as the AIIF factor increases.


\subsubsection{Testing the policies}\label{amm_test-section}

\begin{figure}[H]  \includegraphics[width=\linewidth]{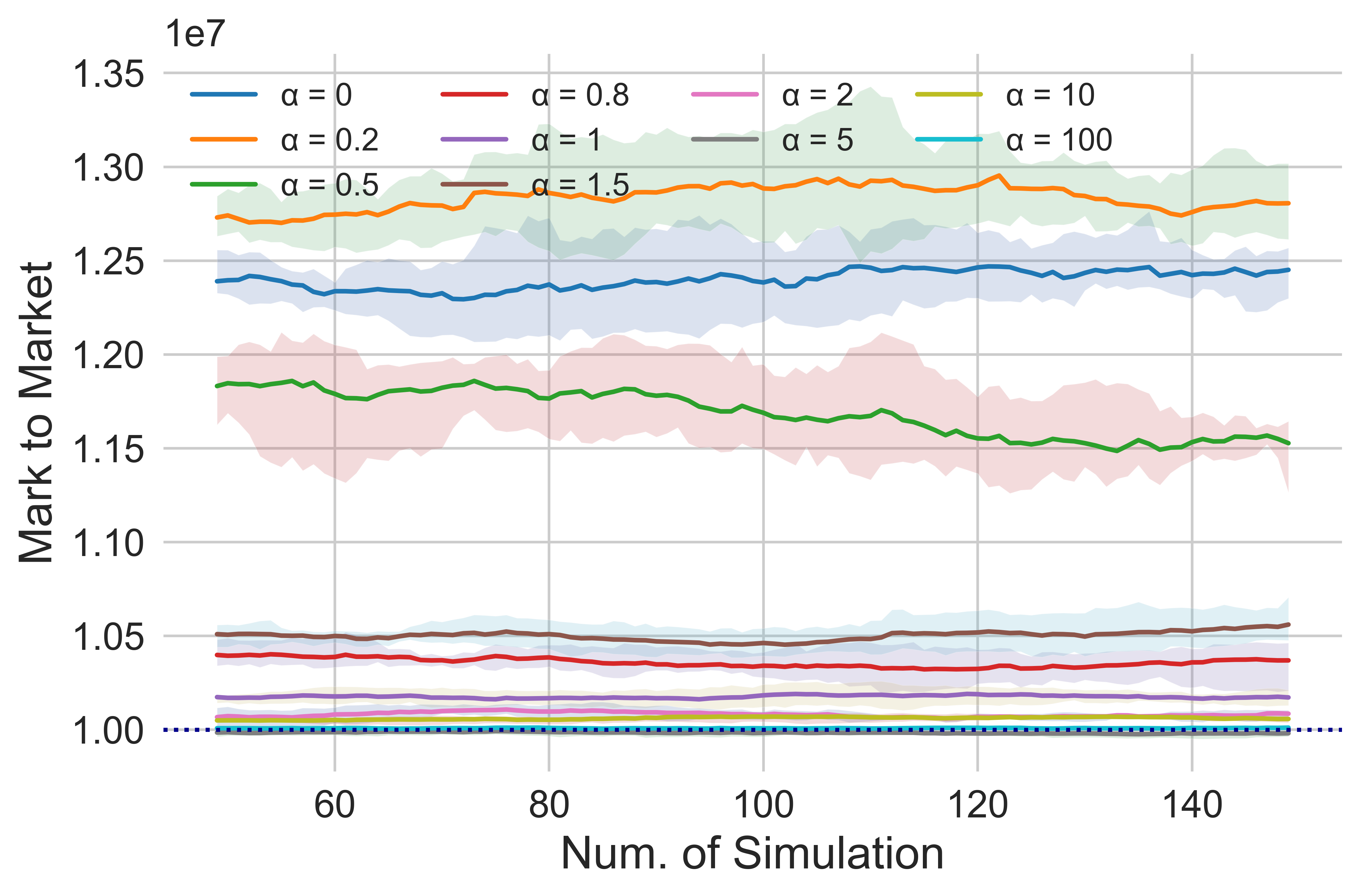}
  \caption[DQN MM testing rounds, with different AIIF factors.]{DQN MM testing rounds, with different AIIF factors. Higher profits are linked to lower $AIIF$ factors, as inventory control is more relaxed.}\label{fig:testing_alphas_mtm}
\end{figure}

\begin{figure}[H]  \includegraphics[width=\linewidth]{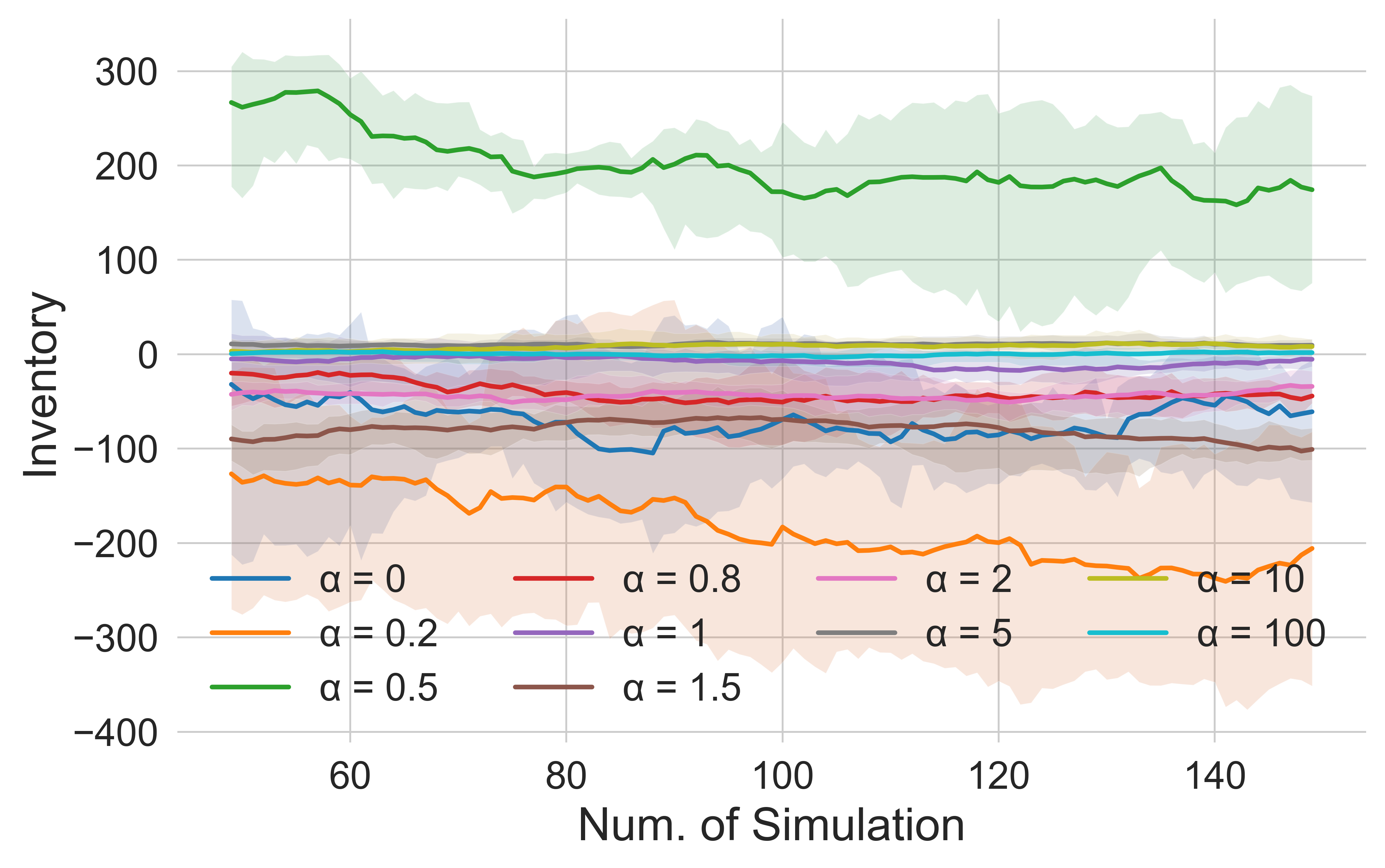}
  \caption[DQN MM testing's rounds inventories, applying different $AIIF$ factors]{DQN MM testing's rounds inventories, applying different $AIIF$ factors. In general terms, as long as the AIIF factors are increased, inventories are reduced as well, due to the controlling effect that this factor has on them.}\label{fig:testing_alphas_inventories}
\end{figure}

Once the agents have been trained using different AIIF factors, they are all tested on new experiments. At first sight, they perform well in terms of profitability with almost all small AIIFs (\autoref{fig:testing_alphas_mtm}). As long as alpha is increased, profitability is impacted. However, the inventory is controlled as well, as expected (\autoref{fig:testing_alphas_inventories}).


\begin{figure}[H]
        \centering        \includegraphics[width=1\textwidth]{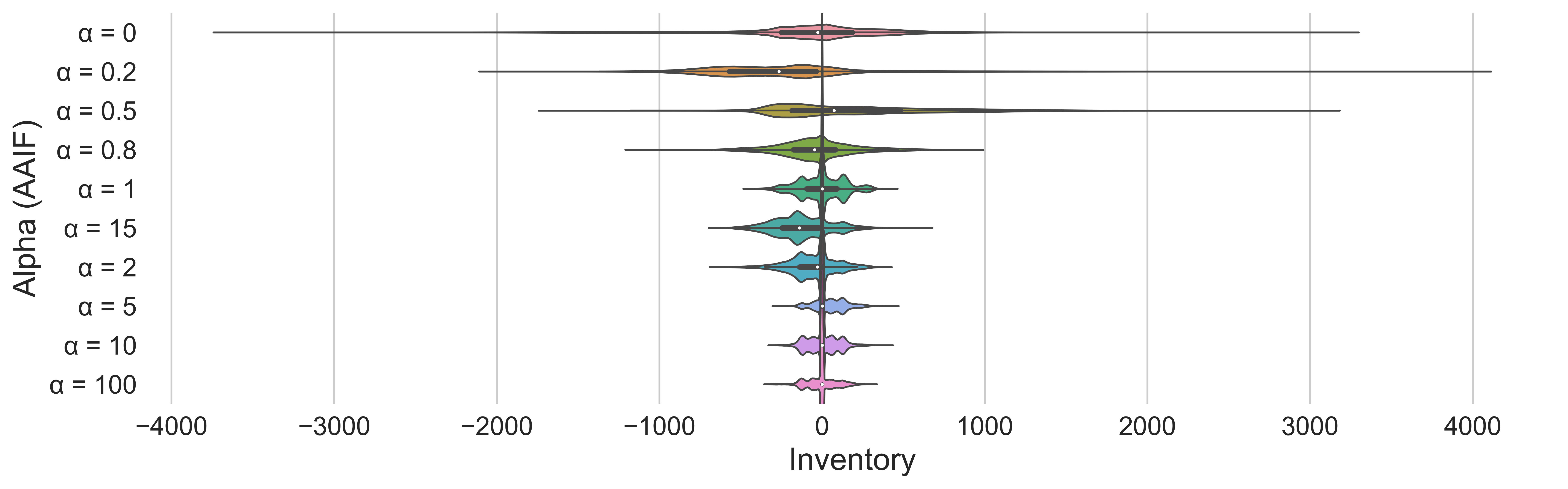}
        \caption[DQN MM inventory distributions along different experiments.]{DQN MM inventory distributions along different experiments. This Figure illustrates the different inventories held by every agent during the testing stage, according to their different AIIF factors. 
        }        \label{fig:violin_inventories}
        \end{figure}

Analyzing inventory distributions (\autoref{fig:violin_inventories}), it is clear that as soon as the AIIF penalty factor increases the inventory also shrinks. Therefore, the distributions get narrower according to this coefficient as expected.

\begin{figure}[H]
\minipage{1\textwidth}
  \includegraphics[width=\linewidth]{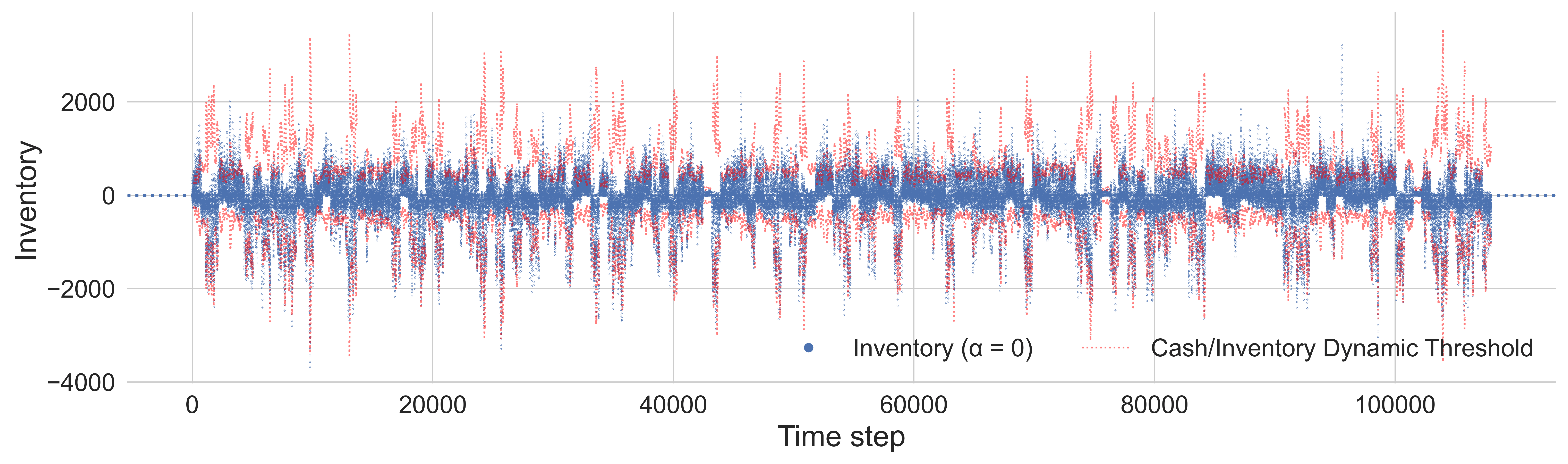}
  \caption[DQN MM instant inventories held and thresholds along the testing experiment every time step.]{DQN MM instant inventories held and thresholds along the testing experiment every time step. Due to the lowest restrictive $alpha$ factor applied ($AIIF = 0$),  noisy inventory thresholds and big variances in inventories are noted. 
  }
  \label{fig:testing_inventory_thresholds_alpha0}
\endminipage\hfill
\minipage{1\textwidth}
  \includegraphics[width=\linewidth]{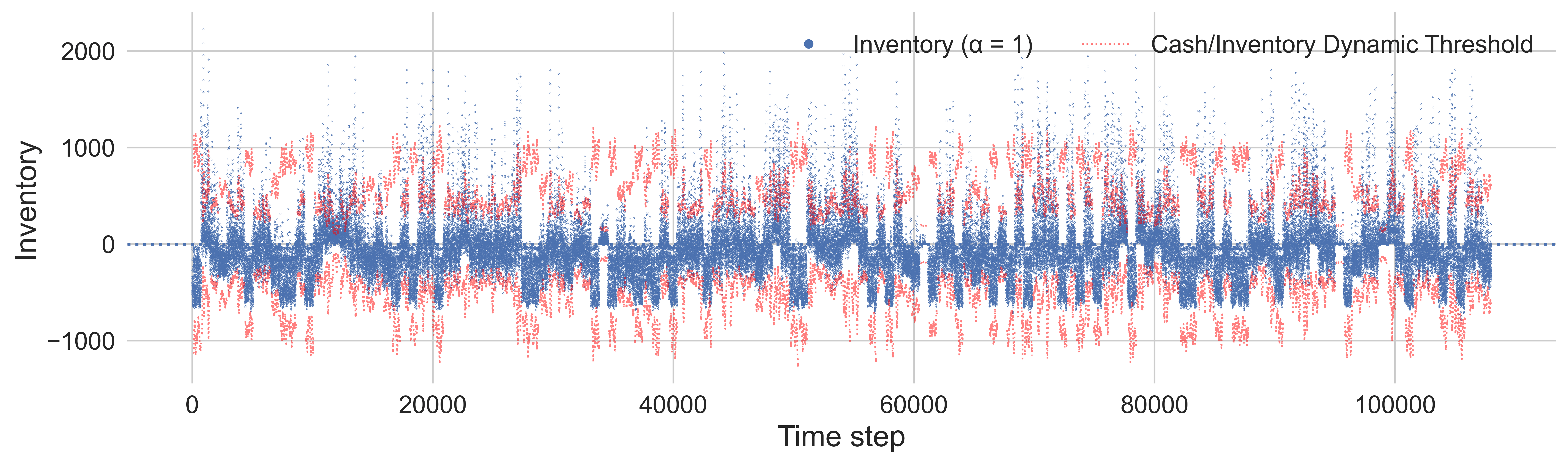}
  \caption[DQN MM inventories and thresholds along the experiment with AIIF 
 = 1.]{DQN MM inventories and thresholds along the experiment with AIIF 
 = 1. More stable thresholds are shown compared to the lower AIIFs of the previous \autoref{fig:testing_inventory_thresholds_alpha0}. In addition, the inventories are also more constrained.}\label{fig:testing_inventory_thresholds_alpha1}
\endminipage\hfill
\minipage{1\textwidth}
  \includegraphics[width=\linewidth]{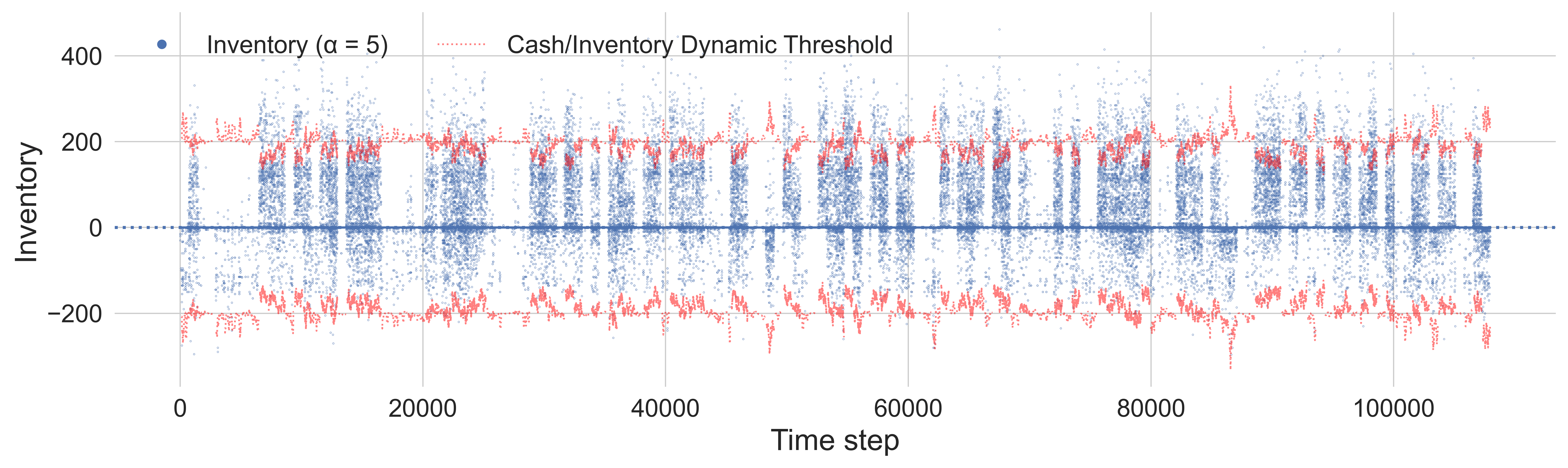}
  \caption[DQN MM inventories and thresholds along the experiment with AIIF 
 = 5.]{DQN MM inventories and thresholds along the experiment with AIIF 
 = 5. Here, extremely stable and low thresholds with very constrained inventories inside these thresholds. For the sake of legibility, y-axis ranges in Figures \ref{fig:testing_inventory_thresholds_alpha0}, \ref{fig:testing_inventory_thresholds_alpha1} and \ref{fig:testing_inventory_thresholds_alpha5} have been adjusted.}\label{fig:testing_inventory_thresholds_alpha5}
\endminipage\hfill
\end{figure}

But not only the AIIF factor plays a key role in managing the inventory risk. This management is also related to \textit{Dynamic Inventory Threshold Factor} (DITF). This factor, which is established in this research to a value of 0.5, defines the proportion between cash and inventory value along the experiments. In Figures \ref{fig:testing_inventory_thresholds_alpha0},  \ref{fig:testing_inventory_thresholds_alpha1}, and \ref{fig:testing_inventory_thresholds_alpha5} can be noticed how this threshold (in red) adapts dynamically along them according to the different AIIF factors, based only on the cash held by the MM at every moment, and the inventory value. Here three examples are shown, $AIIF=0$ (no penalty), $AIIF=1$, and $AIIF=5$. As long as the AIIF factor is increased, inventories are kept more strictly inside the thresholds. Note that y-axis ranges of Figures \ref{fig:testing_inventory_thresholds_alpha0}, \ref{fig:testing_inventory_thresholds_alpha1} and \ref{fig:testing_inventory_thresholds_alpha5} have been adjusted independently for the sake of legibility. Additional AIIF factors tested are shown in Appendix \ref{appendix_aiifplots}.

Finally, it has been explained that the dynamic threshold is calculated with the value of the inventory at every time. In this sense, the discussion revolves around cash/inventory value ratios, as it is the main driver of the penalty term. Regarding this point, Table \ref{table-ratios} presents average ratios per experiment. Once again, it can be observed how increasing the AIIF factor directly impacts this ratio, thereby improving this value. This is very reasonable, as the agent is constantly managing the trade-off between profitability and inventory value. Hence, the higher the AIIF factor, the higher the cash/inventory value ratio returned.

\begin{table}[H]
\begin{center}
\begin{minipage}{\textwidth}
\caption{DQN cash/inventory average ratio and benchmarks.}\label{table-ratios}
\begin{tabular*}{\textwidth}{@{\extracolsep{\fill}}lccccccc@{\extracolsep{\fill}}}
\toprule%
& \multicolumn{2}{@{}c@{}}{\textbf{DQN MM}}
& \multicolumn{2}{@{}c@{}}{Random MM}
& \multicolumn{2}{@{}c@{}}{Persistent MM} 
\\\cmidrule{2-3}\cmidrule{4-5} \cmidrule{6-7}%
Experiment &  Avg\footnotemark[1] & $\sigma$\footnotemark[2] & Avg\footnotemark[1] &  $\sigma$\footnotemark[2] & Avg\footnotemark[1] & $\sigma$\footnotemark[2]  \\
\midrule
$AIIF = 0$  &$1,08$ & $0,12 $  &  $1,31 $ & $ 0,31 $ & $ 1,44 $ & $ 1,18$
 \\ 
$AIIF = 0,2$  & $1,18$ & $0,06 $  &  $1 $ & $ 0,3 $ & $ 1,24 $ & $ 0,96$
  \\
  $AIIF = 0,5$  & $1,01$ & $0,15 $  &  $1,34 $ & $ 0,23 $ & $ 1,59 $ & $ 0,95$
  \\

$AIIF = 0,8$  & $1,32$ & $0,38 $  &  $1,33 $ & $ 0,29 $ & $ 2,22 $ & $ 2,36$
 \\
$AIIF = 1$  & $1,48$ & $0,65 $  &  $1,22 $ & $ 0,31 $ & $ 1,29 $ & $ 0,71$
 \\
$AIIF = 1,5$ & $1,39$ & $0,21 $  &  $1,05 $ & $ 0,34 $ & $ 1,31 $ & $ 1,32$
\\
$AIIF = 2$  & $3,06$ & $16,55 $  &  $1,2 $ & $ 0,28 $ & $ 1,3 $ & $ 0,69$
 \\
$AIIF = 5 $ & $11,62$ & $90,3 $  &  $1,23 $ & $ 0,29 $ & $ 1,28 $ & $ 0,43$
 \\
$AIIF = 10 $ & $7,96$ & $24,42 $  &  $1,27 $ & $ 0,29 $ & $ 1,18 $ & $ 0,45$
 \\
\boldmath{$AIIF = 100 $}  & \boldmath{$59,34$} & \boldmath{$413,12 $}  &  $1,21 $ & $ 0,28 $ & $ 1,22 $ & $ 0,44$
 \\
\end{tabular*}
\footnotetext{Note: Table with the average ratio between the cash of the MM and the value of the inventory held at every experiment. It is noticeable that as long as the AIIF factor is increased, so is the cash over the inventory value. Every experiment corresponds to different Alpha factors (AIIF).}
\footnotetext[1]{Average cash/inventory ratio per experiment.}
\footnotetext[2]{Cash/inventory ratio standard deviation per experiment.}
\end{minipage}
\end{center}
\end{table}

\subsubsection{Qualitative analysis of the generated policies}
\label{amm_qualitative-section}

As formerly detailed, DQN agent behavior relies on a NN that returns the best action for each state. On many occasions, NNs are considered ``black boxes'', as they suffer from a certain lack of interpretability. To have a better understanding of the experimental results and the policies followed by the different MMs, the analysis delves into the different buy, sell, and hedging spreads performed in the various experiments.

\begin{figure}[H]
\centering  \includegraphics[width=0.8\linewidth]{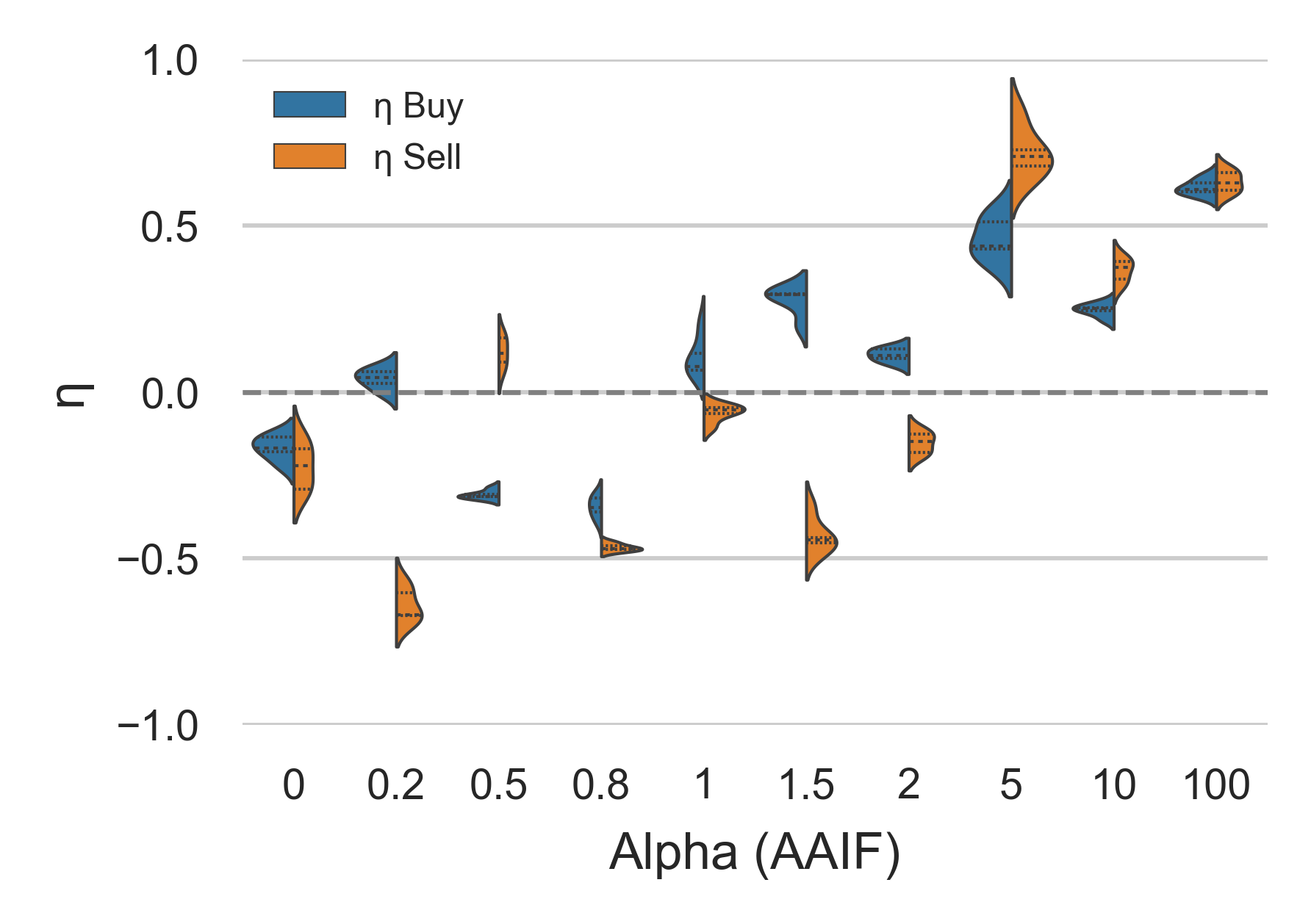}
  \caption[DQN MM buy and sell strategy distributions, in terms of prices, according to different AIIF.]{DQN MM buy and sell strategy distributions, in terms of prices, according to different AIIF. The higher $\eta$s, the more expensive the price streamed by the MM. $\eta_h$=0 represents the same price as the last market spread. With high AIIF factors, higher spreads are streamed by the MM. } \label{fig:testing_buysell_epsilons}
\end{figure}

Regarding buy and sell strategies, as shown in \autoref{fig:testing_buysell_epsilons}, it is noticeable how both $\eta_{(b,s)}$ move from negative to positive values as long as the AIIF rises. This suggests that with smaller or negative $\eta_{(b,s)}$ the MM is more aggressive in terms of prices (narrower spread), being able to capture a higher number of trades. Furthermore, it is also remarkable how strategies evolve not only in terms of negative or positive values but increasing the distance between buy and sell $\eta_{(b,s)}$ (as it happens around $AIIF=0.5$ and $AIIF=1.5$). Thus, different approaches and behaviors are found by the MM. 

\begin{figure}[H]
\centering  \includegraphics[width=0.8\linewidth]{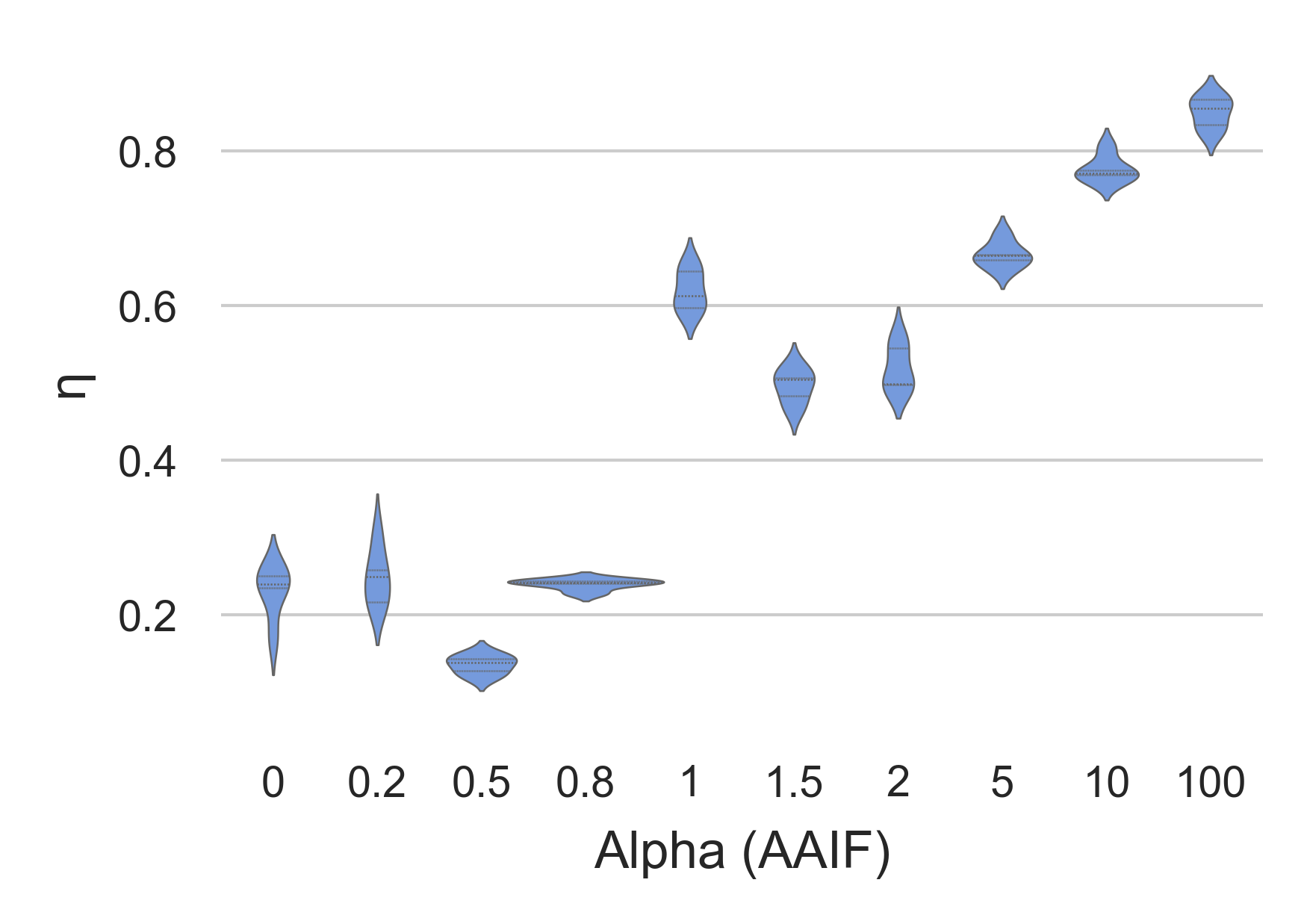}
  \caption[DQN MM hedge strategy distributions according to different AIIF.]{DQN MM hedge strategy distributions according to different AIIF. $\eta_h$ represents the amount of inventory the MM is reducing instantly by buying or selling at market price in a specific time step. This `urgency' has an extra cost. It is remarkable how higher hedges $\eta_h$ are performed as long as the alfa factor is increased. }\label{fig:testing_hedge_epsilons}
\end{figure}

Regarding the hedging strategies (\autoref{fig:testing_hedge_epsilons}), it is noticed that as long as the inventory penalty increases so does hedging $\eta_h$. This increase makes MM keep low inventory despite losing profitability due to hedging costs. Thereby, buy/sell $\eta_{(b,s)}$ adaptive strategies combined with different hedging actions $\eta_h$ compose the unique learned policies.

\begin{figure}[H]
  \centering  \includegraphics[width=0.8\linewidth]{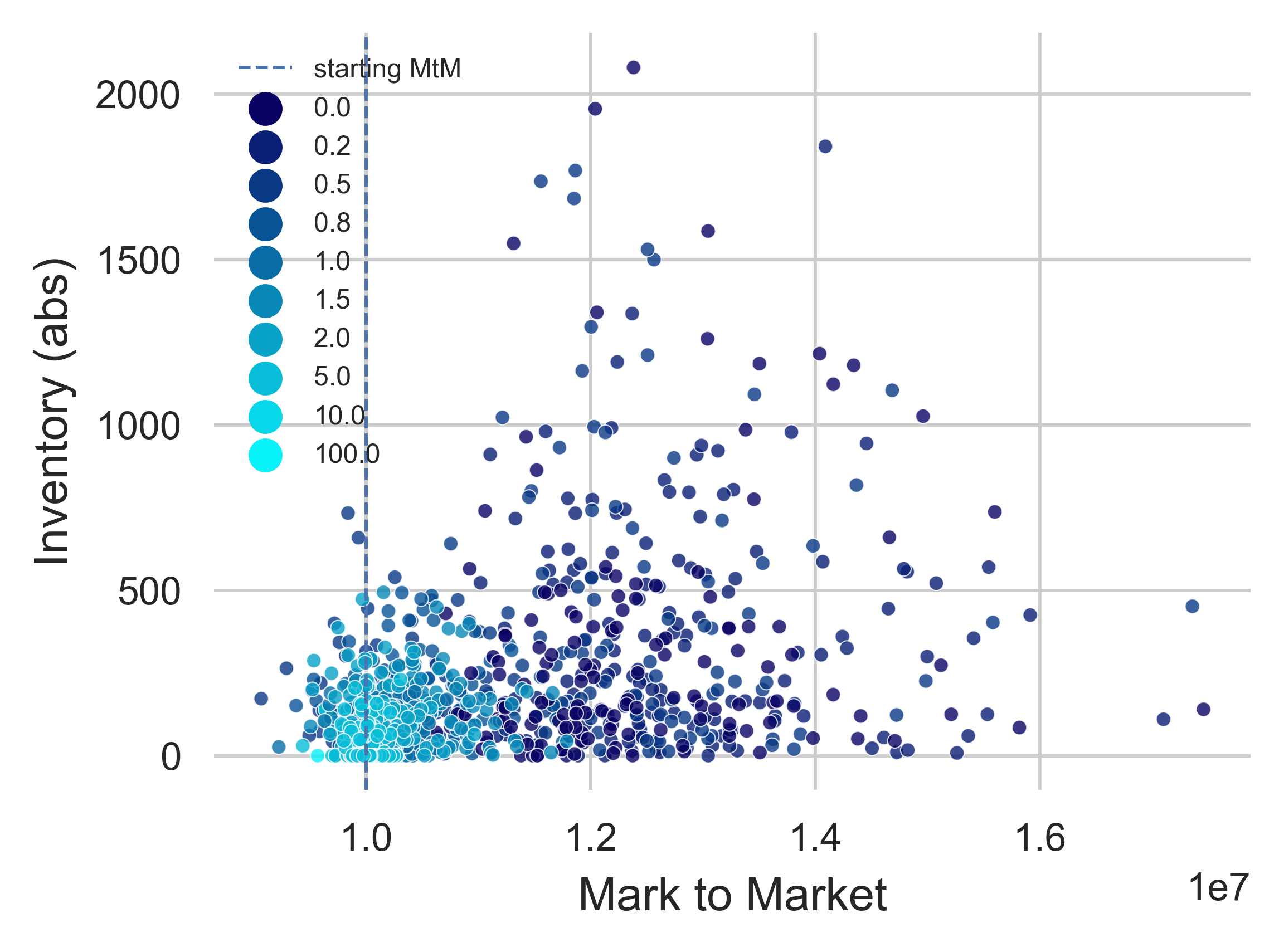}
  \caption[DQN single inventory and MtM per AIIF results are obtained from every test experiment.]{DQN single inventory and MtM per AIIF results are obtained from every test experiment. Each point represents the average MtM and Inventory of the MM of every testing simulation (150 simulations x 10 $alphas$). As noticed, the dispersion is higher with lower AIIF factors.}  \label{fig:mtm_inventory_individual_zoomed}
\end{figure}

\begin{figure}[H]
\centering  \includegraphics[width=0.8\linewidth]{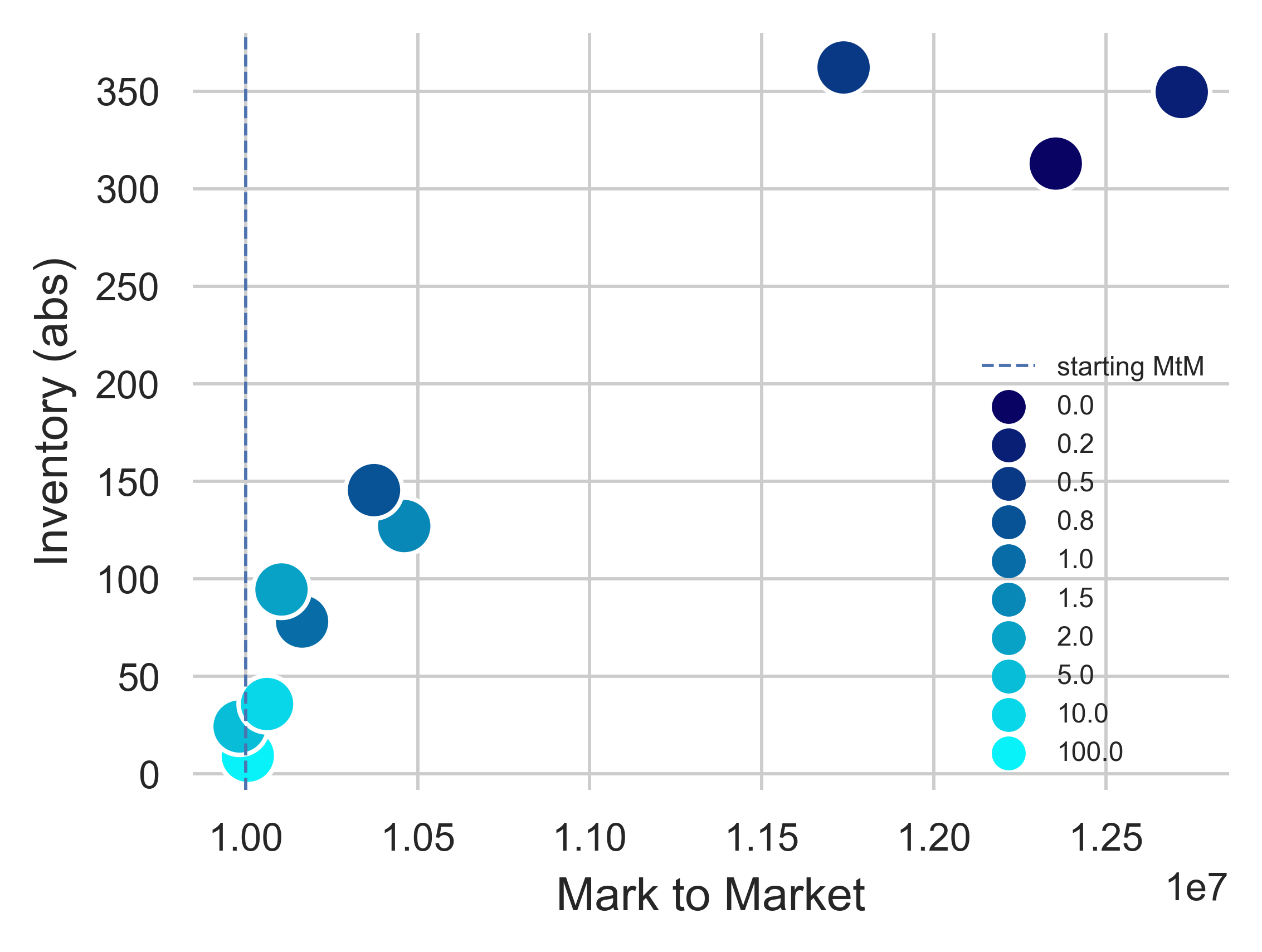}
  \caption[DQN inventory and MtM in average per AIIF.]{DQN inventory and MtM in average per AIIF are obtained from every test experiment(grouped by $alpha$). Here it can be noticed the inverse correlation between MtM+inventory and AIIF factors. In addition, it clearly shows the competing nature of both goals: to reduce the inventory while increasing the MtM.}\label{fig:mtm_inventory_avg}
\end{figure}

Taking a look at the average status of the MMs (Figures \ref{fig:mtm_inventory_individual_zoomed} and \ref{fig:mtm_inventory_avg}), it is remarkable how higher AIIFs can even finish with less value in terms of MtM than the starting value. Focusing on $AIIF=5$, although the inventory position ends in a very constrained situation, it can be seen how the experiment finishes below the starting MtM reference line. So, according to this, it is very important to be aware of the impact of increasing AIIF factor on the overall performance. All the details regarding the MtM can be found in Table \ref{table-mtm}.

\begin{table}[H]
\begin{center}
\begin{minipage}{\textwidth}
\caption{DQN Mark-to-Market experiment results (x$10^3$) and baseline agents.}\label{table-mtm}
\begin{tabular*}{\textwidth}{@{\extracolsep{\fill}}lcccccccccc@{\extracolsep{\fill}}}
\toprule%
& \multicolumn{3}{@{}c@{}}{\textbf{DQN MM}}
& \multicolumn{3}{@{}c@{}}{Random MM}
& \multicolumn{3}{@{}c@{}}{Persistent MM} 
\\\cmidrule{2-4}\cmidrule{5-7} \cmidrule{8-10}%
Experiment & Avg\footnotemark[1]  & $\sigma$\footnotemark[2] & Var\footnotemark[3] & Avg\footnotemark[1] & $\sigma$\footnotemark[2] & Var\footnotemark[3] & Avg\footnotemark[1] & $\sigma$\footnotemark[2] & Var\footnotemark[3] \\
\midrule
$AIIF = 0$  & $12.399$ & $48 $ &24,50\% & $9.711$ & $10 $ &-2,98\% & $9.648$ & $13 $ &-3,35\%  \\ 
\boldmath{$AIIF = 0.2$}  & \boldmath{$12.833$} & \boldmath{$66 $} & \textbf{28,05\%} & $9.559$ & $7 $ &-4,30\% & $9.575$ & $14 $ &-4,45\%   \\
$AIIF = 0.5$  & $11.688$ & $122 $ &15,27\% & $9.957$ & $39 $ &-0,33\% & $9.889$ & $17 $ &-0,71\%    \\
$AIIF = 0.8$  & $10.358$ & $23 $ &3,69\% & $9.823$ & $9 $ &-1,96\% & $9.884$ & $14 $ &-1,26\%  \\
$AIIF = 1$  & $10176$ & $7 $ &1,72\% & $9.889$ & $9 $ & -0,94\% & $9.900$ & $9 $ &-0,88\%  \\
$AIIF = 1.5$  & $10.501$ & $24 $ &5,60\% & $9.775$ & $12 $ & -2,71\% & $9.811$ & $13 $ &-1,76\%  \\
$AIIF = 2$  & $10.081$ & $12 $ &0,85\% & $9.944$ & $13 $ & -0,47\% & $9.922$ & $13 $ &   -0,81\%  \\
$AIIF = 5 $ & $9.984$ & $3 $ & -0,18\% & $10.064$ & $16 $ &0,69\% & $10.038$ & $13 $ & 0,37\% \\
$AIIF = 10$ & $10.061$ & $6 $ &0,57\% & $9.969$ & $7 $ & -0,36\% & $9.998$ & $9 $ & 0,15\%   \\
$AIIF = 100 $ & $10.008$ & $1 $ &0,11\% & $10.019$ & $15 $ & 0,04\% & $10.048$ & $11$ & 0,41\%  \\
\end{tabular*}
\footnotetext{Note: Table with all the MtM values along different testing experiments. Every experiment corresponds to different Alpha factors (AIIF). Starting MtM = 10.000 (x$10^3$)}
\footnotetext[1]{Average MtM along the experiment.}
\footnotetext[2]{MtM standard deviation along the experiment.}
\footnotetext[3]{MtM increase/decrease at end of the experiment (return).}
\end{minipage}
\end{center}
\end{table}

\begin{figure}[H]
        \centering        \includegraphics[width=0.8\textwidth]{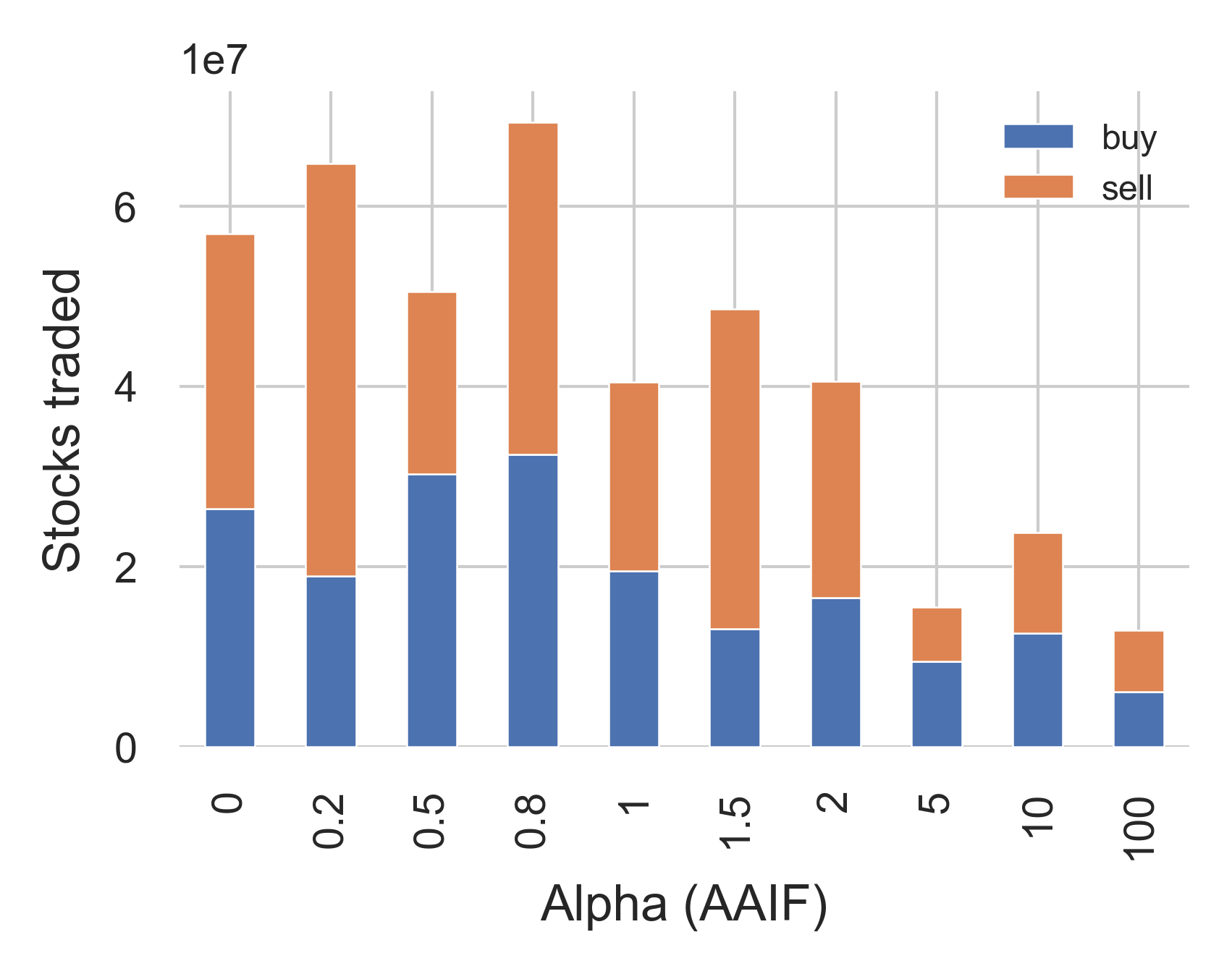}
        \caption[Buy and sell amount of stock traded according to different AIIF]{Buy and sell amount of stock traded according to different AIIF. As long as the AIIF factor is increased, the number of executed operations is reduced. This has to do with the increase of buy and sell $\eta_{(b,s)}$ shown in \autoref{fig:testing_buysell_epsilons}, a point that makes the MM to be less competitive in terms of price, capturing fewer orders from the investors.}
        \label{fig:stocks_traded}
\end{figure}

Finally, aiming to fully understand the policies behind actions, it is relevant to analyze how the AIIF factor impacts the number of stocks traded by the DQN MM. Taking into account that higher AIIFs usually rely on higher buy and sell $\eta_{(b,s)}$ as stated before, it may be expected that the number of operations performed by the MM could decrease as long as the AIIF factor increases. In \autoref{fig:stocks_traded}, it can be appreciated that this intuition is accurate and especially relevant with higher AIIFs. 
In summary, and with all these insights raised, it can be said that, as long as the AIIFs increase, the MM becomes more selective with actions performed, preferring higher spreads and increasing the use of hedging.

\subsubsection{Comparison with existing reward functions}
\label{amm_sec:benchmark}

To test the robustness of the approach, it is important to perform a benchmark with other reward functions. Accordingly, different and recent works ~\cite{Gasperov2021,Spooner2018}, already introduced in \autoref{related-work-rlmarketmaker}, have been selected. These three different works represent common MM RL approaches, as they focus on different metrics to define the reward of their agents. To do a `fair' comparison among methods, only their respective reward functions have been used, while preserving the rest of the parameters as similar as possible. Therefore, every agent shares the same state and action space, the same DQN architecture, and the same hedging policy, as presented before in \autoref{amm_methods}. The rest of the experimental settings (environment agents, episode lengths, random competing agents, etc.) have been preserved as well.

Hence, a total of 4 different reward functions have been compared:

\begin{itemize}
    
    \item{\textbf{Full inventory penalty}}: Inspired in \cite{Gasperov2021}\footnote{$R_{t+1} = (Q^{ask}_t$ - $M_{t+1}) 1 {Q^{ask}_t exe} +(M_{t+1}$ - $Q^{bid}_t ) 1 \{ Q^{bid}_t exe \}$  $-\lambda \lvert I_{t+1} \rvert $}, this reward function includes a penalty term related to the absolute amount of inventory held by the MM at every time step, multiplied by a risk aversion coefficient,  as shown in \autoref{bench-fully}:

    \begin{equation}
    \label{bench-fully}
        R_i = \ E_i - \lambda *\lvert inv_i \rvert - HgC_i
    \end{equation}
    $\lambda$ has been set to $\lambda = 0.15$, following the experimental setup of the mentioned work.
    
    \item{\textbf{Asymmetrically dampened PnL}}: Based on \cite{Spooner2018}\footnote{$r_i =  \psi (t_i)$ - $\max (0, \eta \cdot Inv(t_i)\triangle m(t_i))$}, this reward function includes a penalty term that penalizes the incomes coming from Inventory value increases. It is modulated by a scale factor $\eta$, as shown in \autoref{bench-asym}:
    \begin{equation}
    \label{bench-asym}
         R_i = \ E_i + PnL_i  - max(0, \eta \cdot PnL_i) - HgC_i
    \end{equation}
    $\eta$ has been initially set to $\eta = 0.1$ as shown in the respective work.
    
    \item{\textbf{PnL}}: No inventory penalty term is included. This is a common approach in many works as introduced in \autoref{related-work-rlmarketmaker}. Here, only the profits derived from trades are considered, apart from hedging costs, as shown in \autoref{bench-pnl}:
    \begin{equation}
    \label{bench-pnl}
        R_i = \ E_i - HgC_i
    \end{equation}
    \item{\textbf{RIM} (Our reward function, \autoref{eq:reward}): For the sake of comparison, different AIIFs have been included: $AIIF \in \{0.2, 0.5, 0.8, 1\}$}.

\end{itemize}

\begin{figure}[H]        \includegraphics[width=1\textwidth]{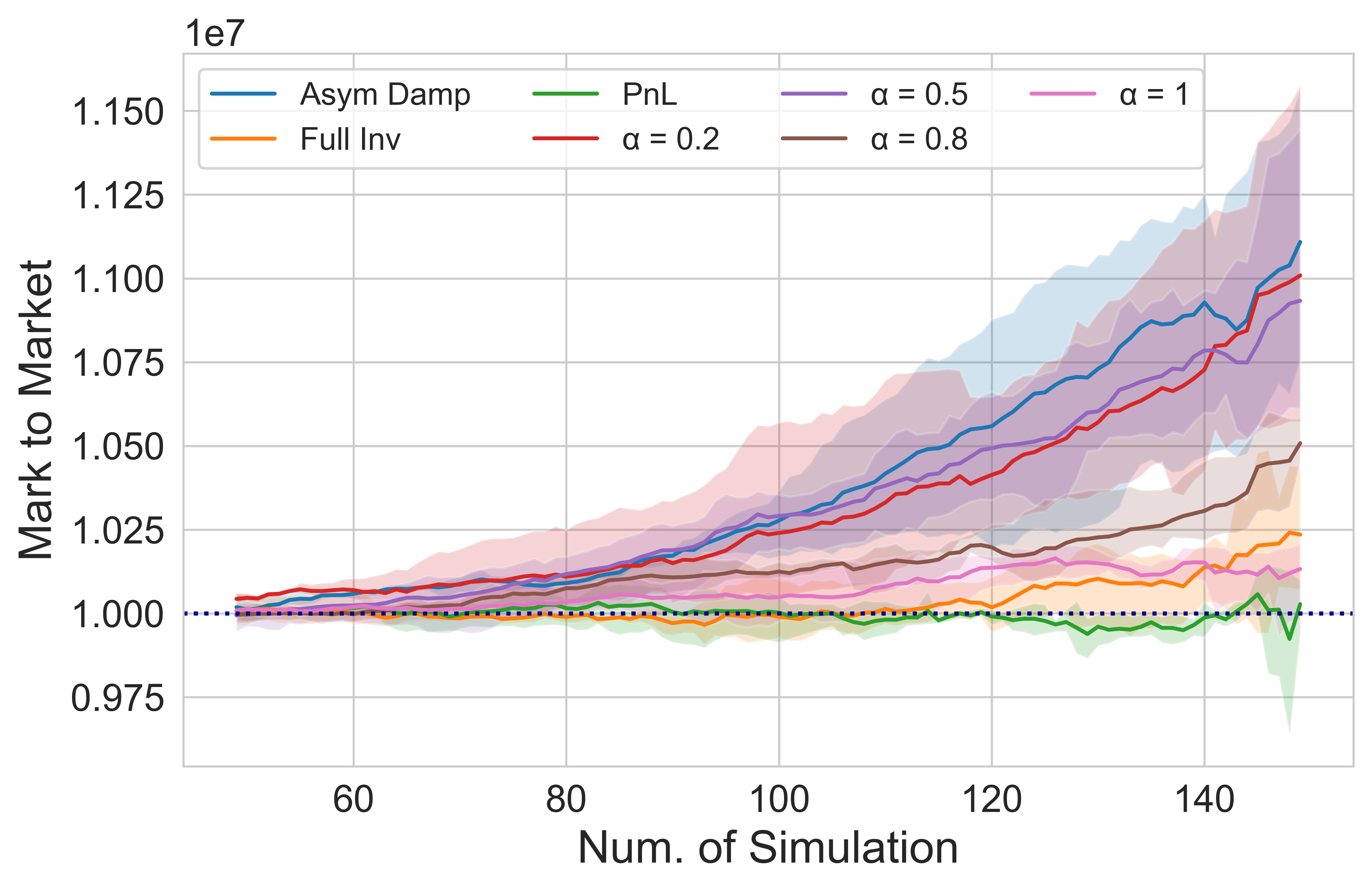}
        \caption[Benchmark results in terms of MtM (cash + inventory value). Training stage]{Benchmark results in terms of MtM(cash + inventory value). Training stage.  Four different reward functions are compared: Full Inv~\ref{bench-fully}, Asym Damp~\ref{bench-asym}, PnL~\ref{bench-pnl}, and our reward function RIM~\ref{eq:reward}($alpha$). Different AIIF factors ($\alpha$) have been included for our reward function. 
        }
        \label{fig:aiif_trainig}

\end{figure}

\begin{figure}[H]
        \includegraphics[width=1\textwidth]{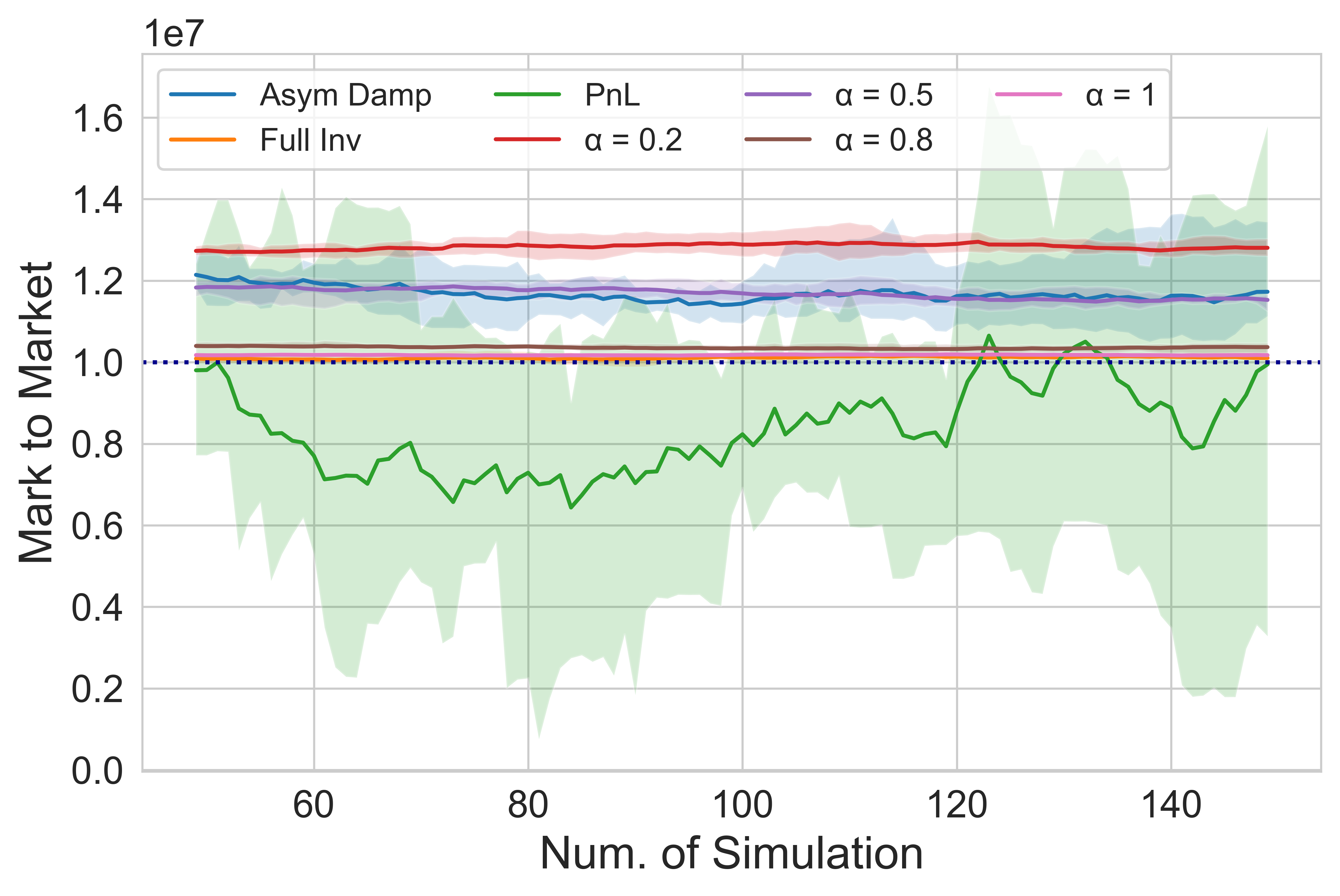}        
        \caption[Benchmark results in terms of MtM(cash + inventory value). Testing stage.]{Benchmark results in terms of MtM(cash + inventory value). Testing stage. Four different reward functions are compared: Full Inv~\ref{bench-fully}, Asym Damp~\ref{bench-asym}, PnL~\ref{bench-pnl}, and our reward function RIM~\ref{eq:reward}($alpha$). Different AIIF factors ($\alpha$) have been included for our reward function. 
        }
        \label{fig:benchmarks}
\end{figure}

Seven different DQN agents were trained independently, based on the described reward functions. The training process is shown in \autoref{fig:aiif_trainig}. Examining the results of the testing stage in \autoref{fig:benchmarks} and Table~\ref{table-benchmark}, it can be seen that the reward function with AIIF = 0.2 was the most profitable during the 150 tested trading sessions, followed by the \textit{Asymmetrically dampened} reward function and the reward function with AIIF = 0.5. These two last cases obtain almost the same profitability. With higher AIIF values ($\alpha = 0.8$ and $\alpha = 1$), MtM decreases, as stated earlier, due to inventory control.

These agents had stable returns during the test experiments, as shown in \autoref{fig:benchmarks}. However, the agent with the \textit{PnL} reward function performed much worse in both MtM and inventory management, indicating that inventory metrics should be included in the reward function to design better market making agents.

\begin{table}[H]
\begin{center}
\begin{minipage}{\textwidth}
\caption{Benchmark results. MtM, inventory, and cash/inventory ratio are presented.}\label{table-benchmark}
\begin{tabular*}{\textwidth}{@{\extracolsep{\fill}}lcccccccc@{\extracolsep{\fill}}}
\toprule%
& \multicolumn{3}{@{}c@{}}{Mark-To-Market}
& \multicolumn{2}{@{}c@{}}{Inventory}
& \multicolumn{2}{@{}c@{}}{Cash/Inv.Value ratio} 
\\\cmidrule{2-4}\cmidrule{5-6} \cmidrule{7-8}%
Experiment &  Avg\footnotemark[1] & $\sigma$\footnotemark[2] &  $Var$\footnotemark[3] & Avg\footnotemark[1] &  $\sigma$\footnotemark[2] & Avg\footnotemark[1] & $\sigma$\footnotemark[2]  \\
\midrule
$Asym Damp$ & $11.672$ & $162$ & $17,27\%$ & $2.975$ & $288$ & $1,08$ & $0,25 $\\
$Full Inv$ & $10.107$ & $27$ & $1,02\%$ & $-32$ & $14$ & $1,15$ & $0,19 $\\
$PnL$ & $8.338$ & $1.048$ & $-0,53\%$ & $16.569$ & $1.882$ & $1,03$ & $0,14 $\\
\boldmath{$AIIF = 0.2$} & \boldmath{$12.833$} & \boldmath{$66$} & \boldmath{$28,05\%$} & $-183$ & $35$ & $1,18$ & $0,06 $\\
$AIIF = 0.5$ & $11.688$ & $122$ & $15,27\%$ & $199$ & $31$ & $1,01$ & $0,15 $\\
$AIIF = 0.8$ & $10.358$ & $23$ & $3,69\%$ & $-40$ & $9$ & $1,32$ & $0,38 $\\
\boldmath{$AIIF = 1$} & $10.176$ & $7$ & $1,72\%$ & \boldmath{$-8$} & \boldmath{$4$} & \boldmath{$1,47$} & \boldmath{$0,64 $}\\
\end{tabular*}
\footnotetext{Note: Table with the results of the different reward functions tested, in terms of MtM, inventory, and cash/inventory value ratio.}
\footnotetext[1]{Average value along the experiment.}
\footnotetext[2]{Standard deviation along the experiment.}
\footnotetext[3]{MtM increase/decrease at end of the experiment (return).}
\end{minipage}
\end{center}
\end{table}

\begin{figure}[H]
        \centering
        \includegraphics[width=1\textwidth]{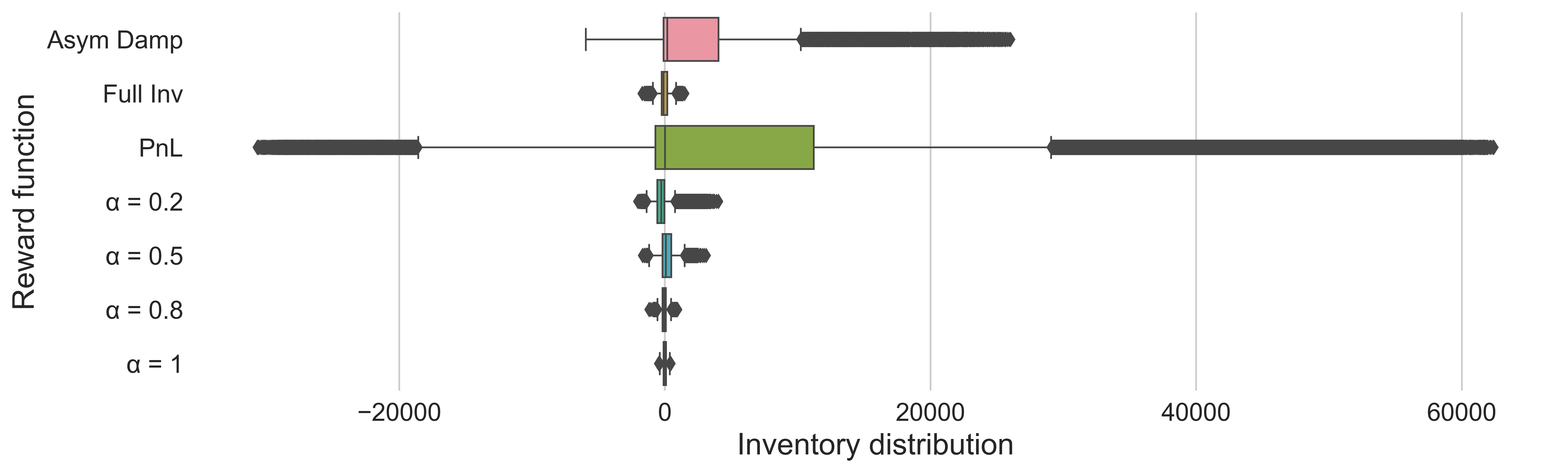}        
        \caption[Benchmark results in terms of Inventory distribution.]{Benchmark results in terms of Inventory distribution. This Figure illustrates the different inventories held by every agent during the testing stage, according to their different reward functions. Four different reward functions are compared: Full Inv~\ref{bench-fully}, Asym Damp~\ref{bench-asym}, PnL~\ref{bench-pnl} and our reward function RIM~\ref{eq:reward} (\textit{alphas}). 
        }
        \label{fig:benchmarks_inventory}
\end{figure}

Regarding inventory management, and observing \autoref{fig:benchmarks_inventory}, it is noted how our reward function performed well in all cases. The \textit{Full inventory penalty} reward function also performed well, despite lower returns. However, the \textit{Asymmetrically dampened PnL} and \textit{PnL} reward functions perform poorly in managing inventory, as shown in the distribution dispersion.

As an overall conclusion of this benchmark, it can be observed that our reward function allows the MM to perform better in terms of MtM, and also, in terms of inventory management than the rest of the tested functions. The average ratio figures between the cash and the inventory value presented in Table~\ref{table-benchmark} also illustrate how our MMs can manage these proportions properly. Furthermore, from the obtained results, it can be affirmed that including some kind of penalty term in the reward function helps to improve the profitability of the agent while constraining the inventory risks. In fact, those reward functions that are based only on PnL, ignoring inventory, can lead to the design of ineffective and also unstable agents.

\subsection{Discussion}\label{amm_conclusions}

Achieving different goals in a multi-objective environment is not always easy. Market making can be clearly considered a multi-objective task. The presented results (Table \ref{table-benchmark}) demonstrate that the proposed agent and reward function can deal with this, adapting properly to the changing environment while seeking the two main objectives: 
\begin{itemize}
    \item {To be profitable, as shown in the results. In this sense, the MtM average results show that our agent obtains good returns and performs better than most of the previous ones, except for AsymDamp.}
    \item {To keep inventory controlled, as evidenced by comparing the average inventory returns with those of PnL and AsymDamp alternatives.}
\end{itemize}

As a limitation of this method, it can be remarked that one pre-trained agent per configuration is required before going live. Defining a scalarized reward function, as done in this work, requires prior knowledge of the future utility function. If the utility function changes \textit{a posteriori}, pre-trained agents that can adapt to it are needed.

Conversely, RL enables the definition of agents by specifying the reward function. This means it always seeks the best policy, regardless of system complexity. This is a noteworthy advantage in stochastic and dynamic environments such as stock markets, where rule-based agents would require continuous calibrations and adaptations. 

As the main conclusions of this first way of controlling inventory, a robust approach for learning a market making agent based on RL has been presented, where the agent can modulate its inventory automatically according to a desired liquidity ratio. A reward function has been introduced that, when combined with the DQN algorithm, can modulate the inventory while seeking profitability and adapting to the evolving liquidity of the agent. This reward function has two key factors that allow the agent to control the inventory risk dynamically while looking for profitability. First, the AIIF factor modulates the risk aversion related to inventory risk by serving as a penalty term. Additionally, the DITF factor can dynamically adjust the behavior of the agent as it adapts to the cash/inventory value held at every time. These two terms combined enable the MM to control its risks dynamically while adhering to the initial risk aversion strategy. Therefore, the MM will not follow a static policy driven by a standard scalarized multi-objective reward function. Rather, it will adapt to the market based on its holdings of cash and inventory.

The MM agent has been tested with different penalty factors to evaluate the performance of these parameters in a simulated but realistic environment and has also been compared to random baseline agents. Experimental results show that the presented approach allows for the design of an RL MM that, with only two liquidity parameters, efficiently modulates its operations. 

Additionally, not only has this intelligent agent been designed, but the underlying strategies followed by the MM in terms of price and hedge spreads (actions) have also been understood and evaluated. This analysis has been conducted taking into account different penalty coefficients, obtaining valuable insights about the trading operative in terms of pricing, inventory hedging, and inventory dynamic thresholds. This analysis allows delving into the reasoning behind the learned strategies, without leaving the agent to operate as a ``black box''.

Last but not least, the proposed solution has been compared with other existing reward functions, concluding that our function can achieve good returns while managing inventory risk in a more robust way than the alternative tested approaches.

Regarding future work, some aspects can also be explored and even improved. The main aspect found relevant is related to the selection of a proper AIIF factor. With the current proposal, this coefficient is predetermined and does not consider the changing market conditions over time. It is well known that market volatility evolves, even for the same asset, behaving differently depending on the hour of the day, the arrival of new economic data, or some other events. Including a volatility parameter that modulates this AIIF coefficient according to the market `texture' at every time could be a good upgrade of the reward function.

\subsection{Contribution}\label{contrib_aaif}

This work makes several important contributions to the field of automated market making strategies using DRL. There is a special focus on inventory control through a dynamic reward function. 

These contributions are outlined as follows:

\begin{enumerate}

    \item \textbf{Adaptive reward function:} Development of an innovative adaptive reward function that adjusts to changes in market conditions and the agent's inventory status, enhancing the traditional static reward functions typically used in market making.

    \item \textbf{Control coefficients for dynamic inventory management:} Introduction of two novel control coefficients, Alpha Inventory Impact Factor (AIIF) and Dynamic Inventory Threshold Factor (DITF), which enable dynamic adjustments to the MM's behavior based on real-time liquidity and cash-inventory ratio conditions.

    \item \textbf{In-depth strategy analysis:} Comprehensive analysis of the strategies employed by RL agents, focusing on their impact on profitability and inventory management under various market conditions.

    \item \textbf{Comparative evaluation of reward functions:} Detailed comparative study of the new reward function against existing methodologies, demonstrating the advantages of the proposed approach in real trading environments.

\end{enumerate}

These contributions collectively advance the field of market making, providing significant theoretical and practical insights that help bridge the gap between academic research and industry application.


\section{MORL with Pareto fronts}\label{morl_sec-morlpareto}

Once the first multi-objective reward-engineered strategy was developed and its pros and cons analyzed, an evolution of the method is presented. In this regard, a strict MORL alternative is developed. Aiming to surpass the main problems of the reward-engineered approach, such as the trial-and-error iterative design process and the need to engineer both subgoals into a single complex function, a MORL alternative comes into play. As already introduced, the market making problem involves addressing two main objectives: (i) increasing MM profitability and (ii) maintaining low inventory levels to mitigate devaluation risks. To tackle these challenges, the integration of a multi-objective framework with PFs and DRL for addressing the MM problem is proposed. This approach is capable of specializing in both subgoals without introducing design biases, keeping the process more straightforward. To demonstrate the robustness of the MORL solution, a comparison with other classical reward engineering market making alternatives is presented, illustrating the strengths and weaknesses of all of them. 

The following subsections describe the entire process: First, in \autoref{morl_sec-morl}, the foundation of the agent is defined, starting with a brief review of the technique, the agent architecture, the algorithm used by the agent, and the metrics that will be used to determine the performance of the agent. After that, the setup of the experiments is detailed in \autoref{morl_sec-method}. Then, the results of the experiments are presented in \autoref{morl_sec-expsetup}. \autoref{morl:Trends} provides an analysis of the impact of including trend information in the presented agent. This is followed by the discussion in \autoref{morl_sec-conclusion} and the contributions in \autoref{morl_contribs}.

\subsection{Multi-objetive RL MM: M3ORL}\label{morl_sec-morl}

 The contribution of this section is twofold: i) the novel modeling of MMs as RL multi-objective agents with two distinct goals: maximizing profitability while minimizing inventory, and ii) the comparison of these agents with other MMs that utilize traditional reward engineering techniques.  Consequently, a MM agent based on MORL called M$^3$ORL is developed, leveraging the advantages of this approach. PFs have been widely employed in solving diverse multi-objective problems \cite{ngatchou2005pareto}, including those in the market making domain \cite{10.1007/978-3-642-15381-5_10}. The utilization of PFs offers numerous benefits in handling multi-objective problems, which have been previously discussed in detail in \autoref{sec-morlpareto}.

 \subsubsection{Market maker as an MOMDP}\label{sec_momdp_mm}

As part of the RL framework, the main characteristics of the RL agent as an MOMDP need to be defined. This includes specifying the state and action spaces, designing the reward function, and determining the agent's architecture. The design of the state and action spaces has been based on the previous section.

\textbf{Space state}: The agent's observation state comprises eight distinct features, the same as in the previous \autoref{amm_sec_appliedintelligence}  due to the good performance achieved by this configuration. These include (i) the number of stocks purchased in the previous time step, (ii) the number of stocks sold in the previous time step, (iii) the current inventory size (which may be either positive or negative), (iv) the inventory size from the previous step, (v) the difference in the stock mid-price between the current and previous time steps (${\Delta} Price_i = Price_i - Price_{i-1}$), (vi) the current reference bid-ask spread, (vii) the bid-ask spread from the previous time step, and (viii) the total volume of stock traded ($Vol{_i}$) by the MM in the previous time step. Other state representations have also been experimented with, including more simplified versions, but these have yielded worse results. 

One relevant element not considered in this dissertation is trend information. Upward and downward trends are situations where markets stabilize over certain, yet undetermined, periods. Indeed, many strategies rely on this aspect  \cite{SZAKMARY2010409,rohrbach2017momentum,5331484}. Deep analysis has been performed regarding this topic in \autoref{morl:Trends}, where trend information is incorporated into the agent's state space to evaluate the impact of including this type of information.

\begin{figure}[H]
\centering
\includegraphics[width=\textwidth]{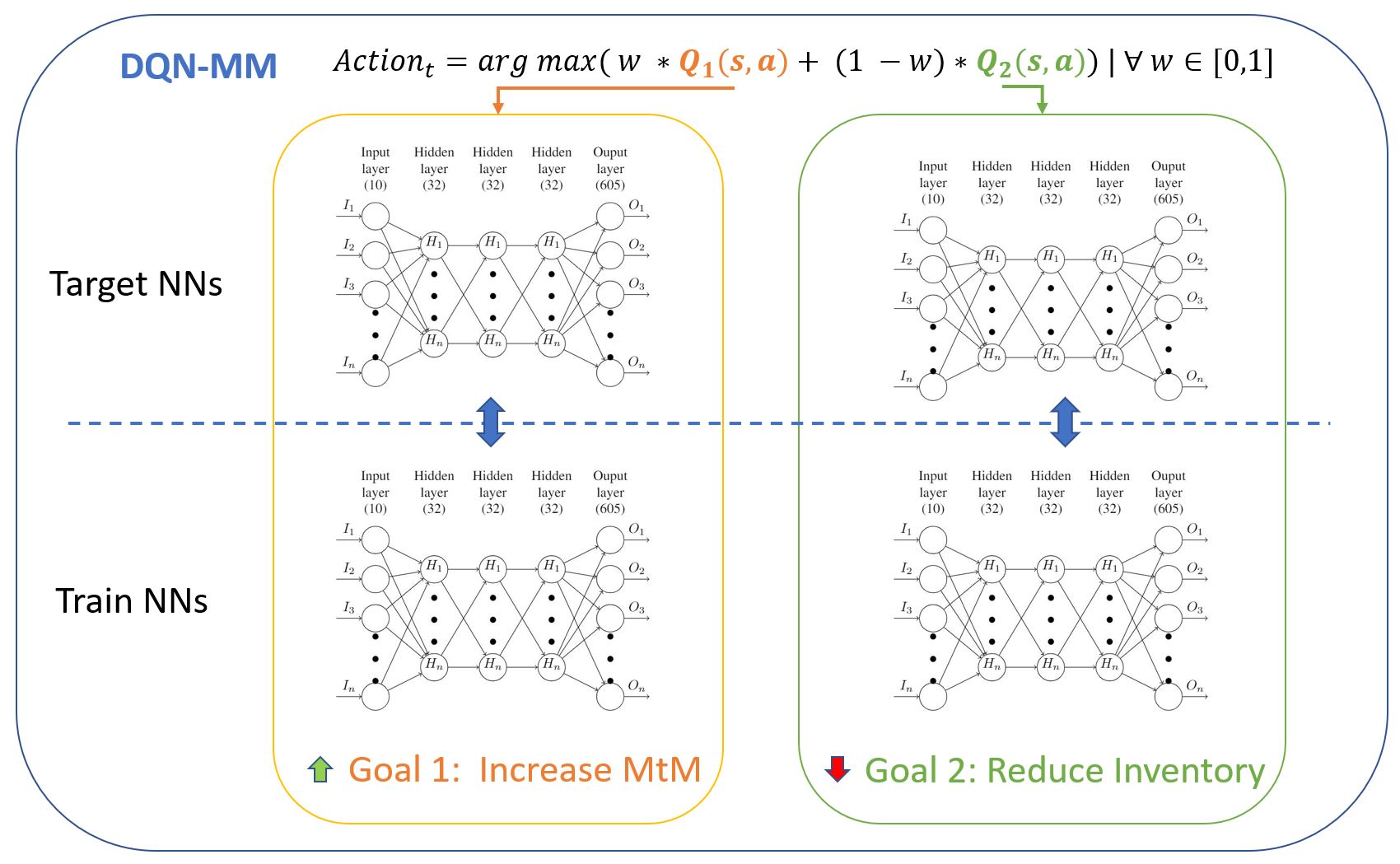}
\caption[MM DQN architecture and reward function.]{MM DQN architecture and reward function. The MM DQN architecture includes separate blocks of NNs (train+target) to handle each subgoal independently. These networks compute the Q-function for both the MtM and Inventory subgoals. The weight parameter, denoted as 'w', is used to balance the importance of MtM and Inventory in the overall reward function.}
\label{fig:dqn_mo_rewards}
\end{figure}

\textbf{Action space:} As previously, the MM agent has three possible actions: quote a buy price, quote a sell price, and decide how many inventory units to hedge through a market order. The agent can interact with three variables in every time step $t_i$: (i) buy spread $Spr^{Buy}_i$, (ii) sell spread $Spr^{Sell}_i$, and (iii) amount of inventory hedged $Hg_i$, to perform these actions. A discrete action space is defined, where buy and sell actions consist of selecting one $\eta$ from a list of evenly distributed values $\eta_{(b,s)i}\  \in \{-1, -0.8, -0.6, ..., 0.6, 0.8, 1\}$, and hedging action consists of selecting another $\eta$ from $\eta_{hi}  \in \{ 0, 0.25, 0.5, 0.75, 1\}$. Therefore, the complete action space consists of 11 buy $\times$ 11 sell $\times$ 5 hedge $\eta$, represented by the tuple $a_i = (\eta_{bi}, \eta_{si}, \eta_{hi})$. When selecting an action, and due to the multi-objective nature of the solution, the best action will be selected based on a utility coefficient $w$. This coefficient, presented in \autoref{fig:dqn_mo_rewards} and detailed in the following subsection, balances the action that aims to optimize the first objective (MtM) with the second objective (inventory control). Its value, therefore, depends on the a priori utility function imposed by the operator. It should be noted that no trading costs were considered in our study, and the asset prices evolved stochastically based on agents' natural interaction.

\textbf{Reward functions}:  Considering two different objectives, and following the MORL framework introduced previously, two reward functions are taken into account. Therefore, the agent will receive a vector as shown in \autoref{reward_eq_morl} at each learning step.

\begin{equation}
    \vec R(s,a) = (R_{1}(s,a), R_{2}(s,a))
\label{reward_eq_morl}
\end{equation}

In this context, the reward obtained from increasing the agent's profitability is denoted as $R_1$ and it represents the MtM of the agent. The MtM includes not only the profits earned or lost by the agent but also its value based on the inventory at each time step. Therefore, it represents the overall value of the agent at every time step. $R_1$ (MtM) is defined as in \autoref{eq:reward_mtm_morl}:

\begin{equation}
\label{eq:reward_mtm_morl}
    R_1(s,a)_i =  E_i + PnL_i - HgC_i
\end{equation}

Where:

\begin{itemize}
\item (i) $E_i$: The trading profits or earnings obtained by the MM at time step $t_i$ result from buying or selling stocks. The profits are given by Equation \ref{eq:profits_morl}:

\begin{equation}
\label{eq:profits_morl}
E_i = \sum_{x=0}^{n} (B_x \cdot Spr_{Bi}) + \sum_{x=0}^{n} (S_x \cdot Spr_{Si})
\end{equation}

Where $B_x$ and $S_x$ refer to the number of stocks traded (buys or sells respectively) by investors with the MM, and $Spr_{(B,S)i}$ is the buy or sell spread quoted by the MM respectively at time step $t_i$.

\item (ii) $PnL_i$: The inventory held by the MM gains or loses value according to mid-price variation at every time step $t_i$. $PnL_i$ is computed as shown in Equation \ref{eq:pnl_morl}:

\begin{equation}
\label{eq:pnl_morl}
PnL_i = I_i \cdot {\Delta} Price_i
\end{equation}

where $I_i$ is the quantity of stock held by the MM at $t_i$, and ${\Delta} Price_i$ is the difference between current and previous stock prices.

\item (iii) $HgC_i$: The MM can reduce its inventory instantly by buying or selling it at every time step, but it incurs a hedging cost as a penalty, as it is executed through a market-order. The hedging costs $HgC_i$ are given by Equation \ref{eq:hci_morl}:

\begin{equation}
\label{eq:hci_morl}
HgC_i = Hg_i \cdot Spread_{mkt}
\end{equation}

where $Hg_i$ is the amount of inventory hedged, and $Spread_{mkt}$ is the current market spread, at $t_i$.
\end{itemize}

On the other hand, $R_2$ is associated with the inventory level maintained by the agent. $R_2$ (Inventory control) is defined as in \autoref{eq:reward_inventory_morl}:

\begin{equation}
\label{eq:reward_inventory_morl}
    R_2(s,a)_i =  - \alpha \cdot | I_{i} |  - HgC_i
\end{equation}

where,

\begin{itemize}

\item(i): $ |I_{i} |$: is the absolute amount of stock held by the MM as inventory in time step $i$.

\item (ii) $HgC_i$: Same definition as described in $R_1(s,a)$.

\item(iii) : $\alpha$ is a correction term that normalizes  $R_2(s,a)$ to be in similar scales as $R_1(s,a)$. In our case, and according to a first study of the preliminary results, it has a value of $\alpha$ = 5. Other normalization techniques can be applied as well. It is important to remark that this coefficient \textbf{is only used to scale the parameter $w$ of the action selection between 0 and 1}. This coefficient \textbf{does not affect the learning process}, as both networks aim to independently optimize their sub-objectives due to the use of a reward vector instead of a scalar. Therefore, any other value could have been used without impacting the agent's performance.

\end{itemize}

Note that \textit{hedging} is an action that impacts both goals independently. It can be performed at any moment, but the agent is discouraged from doing so, as it has a cost. This is why the $HgC_i$ term is included in both goals' rewards.

\subsubsection{Agent's architecture} \label{sec_morl_arquitecture} Each independent MM objective is supported by two independent DQN blocks conformed by 3 hidden layers with 32 neurons each, which compute the Q-value actions for each objective, thus composing the vector $\vec Q(s,a) = (Q_{1}(s,a), Q_{2}(s,a))$. There are two identical NNs per goal/block (training and target NN) for the sake of learning stability, as detailed in \autoref{morl_sec-algo}.
\autoref{fig:dqn_mo_rewards} illustrates the complete $M^{3}ORL$ agent architecture, including the reward blocks. Both groups of NN are also included, one per goal.

\subsubsection{M3ORL algorithm}\label{morl_sec-algo}

The training flow of the market making agent is defined in Algorithm \ref{alg:alg-flow_morl}. The MORL algorithm follows the guidelines outlined by Vamplew et al. \cite{10.1007/978-3-540-89378-3_37, Vamplew2011}, incorporating them into the training flow of the MM. First, the buffer and parameters are initialized (line 0). At each time step $t_i$, the MM agent selects three actions: the buy spread, the sell spread, and the hedge percentage. This selection can be either random or greedy, based on the epsilon-greedy strategy (lines 7-13). To pick the action greedily, both value functions are weighted according to the predefined $w$ value. This $w$ guides the exploration according to the a priori utility function (i.e. to emphasize MtM over inventory control). Nevertheless, it is important to remark that, in the MORL framework, both Q-value functions are trained individually, as depicted in lines 22 and 23. This enriches the agent's knowledge about each sub-goal. Going back to the flow, and after selecting the proper actions, the MM agent quotes the prices to the market (line 14), and the hedging is performed (line 15). Once all the agents have spread their respective prices, the investor agents choose the agent with the lower spreads to match buy and sell random orders (line 16). The MM agent then computes the rewards $r_{1i}$ and $r_{2i}$ for each objective and stores the transition in the replay buffer (lines 17 and 18). Every 200-time steps, both Q-networks are updated using random transitions (1.024) from the replay buffer $\mathcal D$ (line 21). Importantly, the algorithm updates each NN separately based on the specific sub-goals (lines 22 and 23). Finally, the Train and Target networks are synchronized on both objectives (line 24).

\begin{algorithm}[H]
   \caption{M$^3$ORL MM training flow details}
   \label{alg:alg-flow_morl}
    \begin{algorithmic}[1]
        \State Initialize $\mathcal D \leftarrow \emptyset$, $t \leftarrow 0$, $w \leftarrow x$, $n \leftarrow 200$, $z \leftarrow 1.024$
        \For{$simulation$}
           \State Init ABIDES simulation
           \For{every time step $t_i$  in the simulation}
               \State get current market spread $Spread_{mkt}$  from market data.
               \State initialize state, $s_{i}$               
               \If{$rand < \epsilon$} \Comment{pick the best action}        
                    \State $Qval1_{i} \leftarrow Q_1(s_i)$
                    \State $Qval2_{i} \leftarrow Q_2(s_i)$
                    \State $a_{i} \leftarrow arg max(w \cdot Qval1_{i} + (1-w) \cdot Qval2_{i})$ \Comment{$buy$, $sell$ and $hedge$ $\eta$}
               \Else
                    \State $a_{i} \leftarrow$ random\_action() \Comment{$buy$, $sell$ and $hedge$ random $\eta$}
                \EndIf
                \State execute action $a_{i}$ (place buy and sell orders)
                \State execute hedging from the previous time step $Hg_{i}$  
                \State investor agents launch buy/sell (random) operations on the cheapest MM
                \State MM computes rewards $r_i$, gathers $s_{t+1}$ and applies $\epsilon$ decay
                \State MM stores transition $\langle s_{i}, a_{i}, s_{i+1}, (r_{1i}, r_{2_i}) \rangle$ in buffer  $\mathcal D$
                \State $t \Leftarrow t + 1$
                \If{$t \mod n = 0$}
                    \State Train NN with z transitions from $\mathcal D$:
                    \State \hskip1.5em  $Q_1(s,a) \leftarrow Q_1(s,a) + \alpha (R_1(s,a) + \gamma Q_1(s' , a' ) - Q_1(s, a))$
                    \State \hskip1.5em  $Q_2(s,a) \leftarrow Q_2(s,a) + \alpha (R_2(s,a) + \gamma Q_2(s' , a' ) - Q_2(s, a))$
                    \State Target NN $\Leftarrow$  Training NN
                \EndIf
                \State $s_{i} \Leftarrow s_{i+1}$
            \EndFor
        \EndFor    
    \end{algorithmic}
\end{algorithm}

\subsubsection{Multi-objective performance metrics} \label{sec_morl_metrics}

To evaluate the performance of the solution and to make the results comparable between algorithms (MORL and reward-engineered algorithms), the following well-known multi-objective metrics are used \cite{7360024, doi:10.1142/S0219622019500093, 8756240}:

\begin{itemize} 
    \item {\textbf{Hypervolume metric}}: It is a metric used to measure the extent of dominance in the objective space by a set of solutions on the PF, relative to a reference point. Hypervolume provides a quantitative measure of the quality of a PF approximation, with a larger hypervolume indicating a better approximation. To ensure comparability, it is often recommended to normalize hypervolumes when they are on different scales. 
    \item { \textbf{Sparsity}}: The sparsity measures the average distance among undominated solutions produced by the algorithm. Specifically, the sparsity is calculated as the sum of the Euclidean distances between each undominated point in the undominated set and its closest point, divided by the number of points in the reference set. A smaller sparsity value indicates better coverage of the PF.
    \item{\textbf{Number of undominated solutions}}: This point refers to the amount of undominated solutions in the non-dominated set. In this sense, the algorithm is better if it has a higher number of undominated policies.
\end{itemize}

By using these metrics, different algorithms, including single-objective and MORL ones, can be objectively compared. It is important to consider all of the metrics in the comparison, rather than relying solely on one of them, to avoid inaccurate results. This comprehensive approach is necessary to make an informed final decision.

\subsection{Methodology and experimental setup}\label{morl_sec-method}

In the study, all experiments were conducted within the simulated trading environment ABIDES~\cite{Byrd2019}, which is an event-based market simulator. This type of environment allows for testing multiple setups, thus increasing the variance of the data and reducing the risk of overfitting, among other advantages. Moreover, having a simulator provides the opportunity to train agents in an online manner, enabling RL algorithms such as DQN to explore various state spaces, as discussed previously.

In the study, inspired by our previous work (\autoref{amm_sec_appliedintelligence}), the aim was to generate a `standard' market in terms of price volatility. The market simulation configuration included 100 noise agents, 10 value agents, 10 momentum agents, 1 adaptive POV agent, and 1 exchange agent. The noise agents place orders in the OB  in a random direction of fixed size, while the value agents have access to fundamental time series and operate based on variations of the mid-price related to their mid-price forecast. The momentum agents operate based on when 50 and 20-step moving averages are crossed. The adaptive POV agent provides certain liquidity to the market by placing orders at fixed intervals. Finally, the exchange agent orchestrates the interaction and integration of all the agents and the OB.

In addition to these agents, and similar to previous sections, 50 investor agents interact solely with the market makers. These investor agents place random buy or sell orders with a fixed size, always choosing the MM with the narrowest spread available in the OB, hence the cheapest one. If there were many MMs with the same cheapest bid or ask, the investor would randomly select one of them.

All the simulations begin with an opening price of \$1,000.00 and are based on a two-hour market session from 9:30 am to 11:30 am. One DQN MM agent, one Random MM (with random prices), and one Persistent MM (also with random prices but keeping the same price stable the whole simulation) are included in each simulation, as competitors. There were conducted a total of 150 independent simulations across five experiments per setup.

\subsection{Experimental results }\label{morl_sec-expsetup}

In this section, the results of the M$^3$ORL agent are presented, and they are compared with other reward-engineered approaches. The objectives are three-fold: (1) to demonstrate the effectiveness of the MORL approach in handling the market making problem managing the two objectives, (2) to show that the M$^3$ORL approach outperforms standard reward-engineered approaches in terms of metrics such as hypervolume, sparsity, and the number of undominated solutions, and (3) to enhance risk management through the use of the weight parameter while increasing the interpretability of the solution.

In the experiment, 10 different weights, $w \in \{0 , 0.1, 0.2, ... 1\}$, are trained and tested using Algorithm~\ref{alg:alg-flow_morl}. In the training phase, different configurations are started based on the presented $w$. Every weight represents a utility function, hence, a prioritization or balance between both objectives. This $w$ can also be used as an inventory risk aversion coefficient. The higher the $w$, the less averse to inventory risk. A total of five independent experiments are conducted for each $w$. Once all the agents are trained, one trained agent per weight is randomly picked, and five different tests are carried out.

\begin{figure}[H]
 \centering
  \includegraphics[width=1\linewidth]{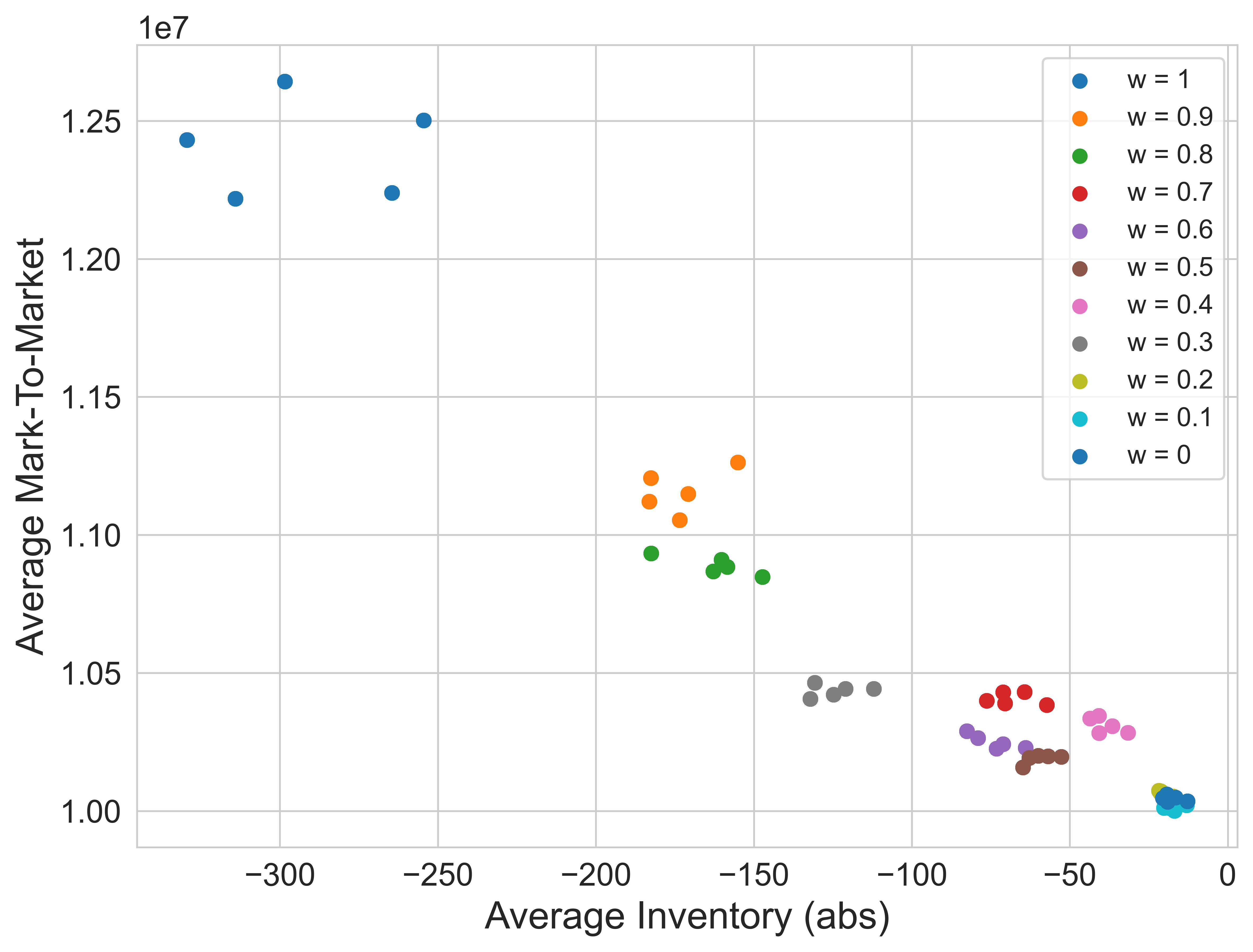}
  \caption[Test results - Pareto]{Test results: Each data point on the plot represents the average value of a simulation conducted with a specific weight. Five simulations were performed for each weight, as indicated in the plot. Notably, the plot demonstrates the relationship between controlling the inventory (lower value of 'w') and achieving a lower level of sparsity in the solution, and vice versa.}\label{fig:_morl_results_2}
\end{figure}

Examining the first results, the average outcomes for each testing weight are analyzed in \autoref{fig:_morl_results_2}, which showcases the trade-off between MtM and inventory control, denoted by $R_{1}(s,a)$ and $R_{2}(s,a)$ respectively. Each data point represents the average value of MtM and inventory corresponding to a specific $w$, obtained from multiple (5) experiments. Notably, there is a clear trend where increasing the $w$ prioritizes the MtM objective, leading to higher levels of both MtM and inventory. Note that, for the sake of explanation, the inventory level is presented always in negative values.

\begin{figure}[H]
  \centering
  \includegraphics[width=1\linewidth]{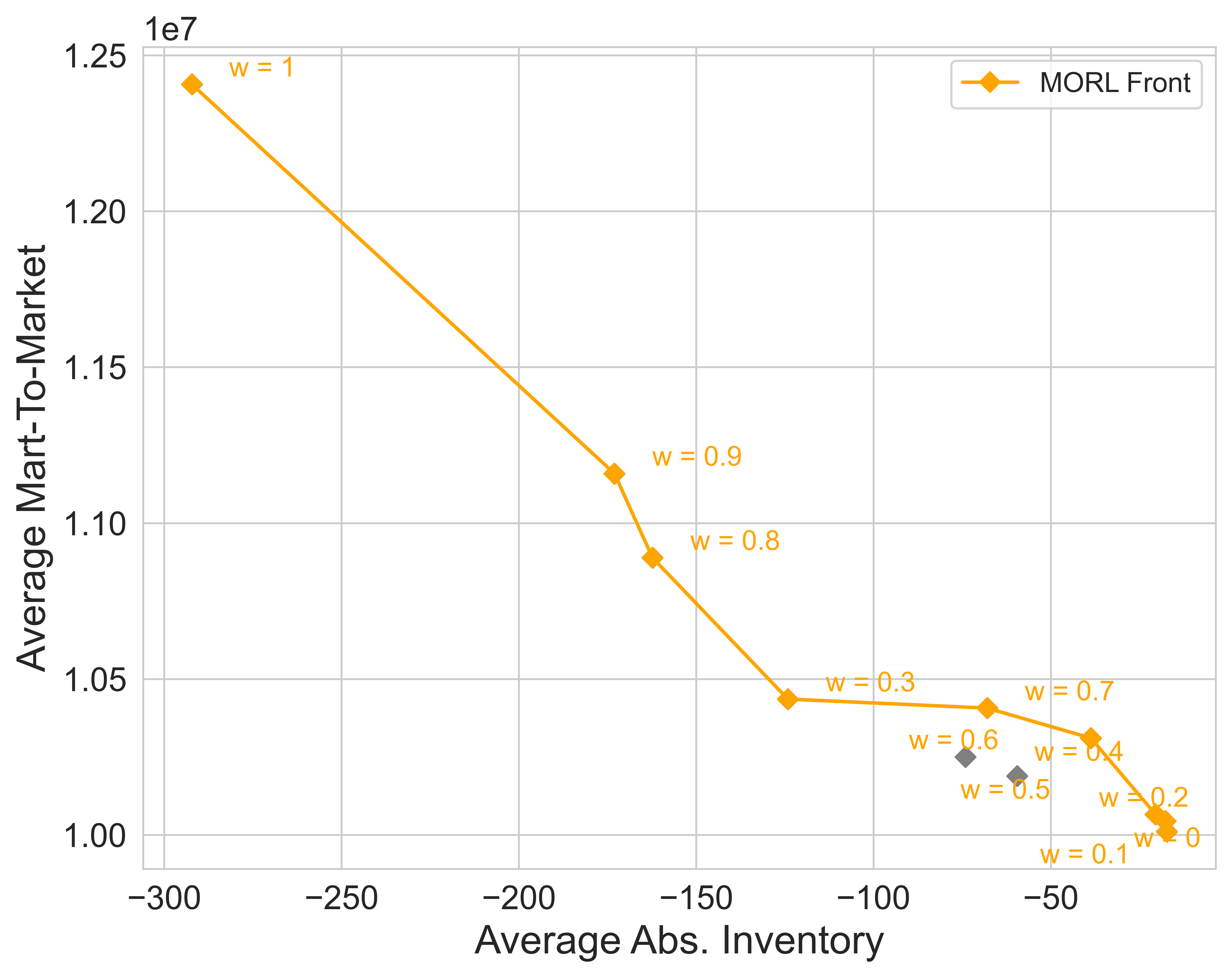}
\caption[MORL Undominated front.]{ MORL Undominated front. The plot displays all the solutions generated by the MORL algorithm during the testing stage, distinguishing between dominated solutions (grey) and non-dominated solutions (orange). Each data point represents the average of five runs per weight.  }
\label{fig-morlresults}
\end{figure}

To further evaluate the performance of different $w$, the PF for each weight is presented in \autoref{fig-morlresults}. Among the 11 weights tested, 9 yielded undominated solutions (orange points), indicating favorable outcomes. However, 2 points remained dominated by the rest of the solutions (grey points). It is worth mentioning that as the $w$ approaches 1, the standard deviation of the average solutions also increases (details in Table \ref{table-values}), signifying greater instability when focusing solely on profitability. Conversely, when incentivizing inventory control, the sparsity of average points in \autoref{fig:_morl_results_2} is reduced.

\begin{table}[H]
\begin{center}
\begin{minipage}{1.1\textwidth}
\caption[MORL and RE-W market making test results.]{MORL and RE-W market making Test Results. This table provides detailed data from the experiments, showcasing the performance in terms of average MtM and Inventory. It is evident that the weight parameter plays a significant role in modulating both goals, as indicated in the figures. It is also remarkable how the MORL approach manages the inventory more efficiently than RE-W in higher and lower weights.}
\label{table-values}
\begin{tabular*}{\textwidth}{@{\extracolsep{\fill}}lcccccccccc@{\extracolsep{\fill}}}
& \multicolumn{3}{c}{MORL MtM}
& \multicolumn{2}{c}{MORL Inv}
& \multicolumn{3}{c}{RE-W MtM}
& \multicolumn{2}{c}{RE-W Inv}
\\\cmidrule(lr){2-4}\cmidrule(lr){5-6}\cmidrule(lr){7-9}\cmidrule(lr){10-11} %
Weight &  Avg[\footnote{For the sake of legibility the MtM has been adapted to start at 1 instead of $1^7$. Same with standard deviation.}] & $\sigma$  & Var\%  &  Avg & $\sigma$ & Avg[\footnote{For the sake of legibility the MtM has been adapted to start at 1 instead of $1^7$. Same with standard deviation.}] & $\sigma$  & Var\%  &  Avg & $\sigma$  \\
\midrule
$W=1$ &$1,23$&$35$&$23,95\%$&$-106$&$23$ & $1,23$ & $275$ & $27,62\%$ & $1860$ & $177$
\\
$W=0.9$ &$1,11$&$32$&$11,82\%$ &$-55$&$10$ & $1,37$ & $265$ & $37,89\%$ & $-3986$ & $565$
\\
$W=0.8$ &$1,09$&$25$&$8,78\%$&$146$&$10$ & $1,10$ & $31$ & $10,72\%$ & $135$ & $8$
\\
$W=0.7$ &$1,04$&$12$&$4,10\%$&$24$&$5$ & $1,04$ & $29$ & $4,09\%$ & $-17$ & $9$
\\
$W=0.6$ &$1,02$&$13$&$2,46\%$&$-50$&$7$ & $1,05$ & $25$ & $6,24\%$ & $-66$ & $10$
\\
$W=0.5$ &$1,01$&$10$&$2,01\%$&$-34$&$2$ & $1,04$ & $11$ & $4,42\%$ & $-9$ & $7$
\\
$W=0.4$ &$1,03$&$16$&$3,10\%$&$-9$&$4$ & $0,99$ & $10$ & $-0,44\%$ & $22$ & $4$
\\
$W=0.3$ &$1,04$&$21$&$4,60\%$&$-90$&$5$ & $1,0$ & $5$ & $1,08\%$ & $14$ & $2$
\\
$W=0.2$ &$1,006$&$11$&$0,64\%$&$4$&$3$ & $1,0$ & $2$ & $0,47\%$ & $-3$ & $1$
\\
$W=0.1$ &$1,001$&$2$&$0,13\%$&$-1$&$2$ & $1,01$ & $3$ & $1,01\%$ & $-10$ & $2$
\\
 $W=0$  & $1,004$&$4$&$0,50\%$ &$1$&$1$ & $1,02$ & $3$ & $0,38\%$ & $12$ & $1$
 \\ 
\end{tabular*}
\end{minipage}
\end{center}
\end{table}

Overall, our approach demonstrates its ability to address the market making challenge, providing a range of undominated solutions to choose from based on the desired utility function. The PF analysis showcases the effectiveness of the MORL approach in achieving a balance between profitability and inventory control, with a high ratio of undominated solutions.

\subsubsection{Reward-engineered vs MORL analysis}\label{morl_sec-rewardcomparison}

To evaluate the robustness of the MORL approach and compare its performance against classical RL reward engineering alternatives, a comparison with two reward-engineered versions was conducted:

\begin{enumerate}
\item \textbf{RE-AIIF}. The first one is called \textit{Reward Engineering Alpha Inventory Impact Factor} (RE-AIIF). It defines the reward function introduced in \autoref{amm_reward-function} of the previous work that scalarizes both goals into a single-objective, including a dynamic penalty term ($Pny$) for managing inventory risk, as shown in Equation \ref{eq:reward_alternative}.

\begin{equation}
\label{eq:reward_alternative}
    R(s,a)_i = E_i + PnL_i - HgC_i - Pny_i
\end{equation}

In that equation, $E_i$ represents the reward from the buy and sell profits at $t_i$, $PnL_i$ is the value earned or lost due to inventory value change, and $HgC_i$ is the hedging cost. The penalty term can manage the amount of inventory held dynamically according to an impact factor (AIIF), as detailed in \autoref{morl_eq-pny}:

\begin{equation}
\label{morl_eq-pny}
Pny_i =  AIIF \cdot min\{ | R_i |, |  R_i \cdot \frac{\overline{inv_i}}{\overline{thr_i}} |  \}
\end{equation}

The AIIF coefficient balances between both objectives, where a lower value indicates less inventory control and higher MtM as a result, and vice versa. The following AIIFs were tested: $AIIF = {0, 0.2, 0.5, 0.8, 1, 1.5, 2, 5, 10, 100}$. $R_i$ represents the MtM reward, defined as $R_i = E_i + PnL_i - HgC_i$. Additionally, $\overline{inv_i}$ represents the average inventory value held in the last $n$ time steps, while $\overline{thr_i}$ is a dynamic threshold computed at every time step based on the cash-to-inventory value ratio. 

This reward function was compared against other reward functions from different works, yielding favorable results. The experimental data obtained from the original work can be found in \autoref{table-ratios}. 

\item \textbf{RE-W}. For the second comparison, a reward function called \textit{Reward Engineering} (Weighted) (RE-W) has been developed, which is described by \autoref{eq:reward_eng_weighted}.

\begin{equation}
\label{eq:reward_eng_weighted}
     R(s,a) =  w \cdot (E_i + PnL_i - HgC_i) + (1-w) \cdot (- \alpha \cdot | I_{i} |  - HgC_i)
\end{equation}

This function aims to provide a straightforward approach by scalarizing both goals using weighted terms. It incorporates the same terms as the MORL approach, but with the inclusion of weights ($w$) to combine the subterms into a single reward value based on a specific utility function. It is important to note that in the MORL approach, similar weights are used, but they are utilized to guide exploration and balance between subgoals rather than aggregating their rewards. The motivation behind designing this weighted reward function is to compare the MORL approach with a pure weighted reward-engineered approach, ensuring that at least the same reward terms are considered as inputs for the comparison. 
As stated, both reward subterms are similar to Equations \ref{eq:reward_mtm_morl} and \ref{eq:reward_inventory_morl}.
\end{enumerate}

Apart from reward functions, and to ensure a fair comparison among methods, most of the experimental parameters were kept consistent, including the ABIDES environment configuration, the number and type of agents, the training episode lengths (150), the number of training episodes per approach (5), the DQN NN architectures, and the state and action spaces.

\begin{figure}[H]
  \centering  \includegraphics[width=1\linewidth]{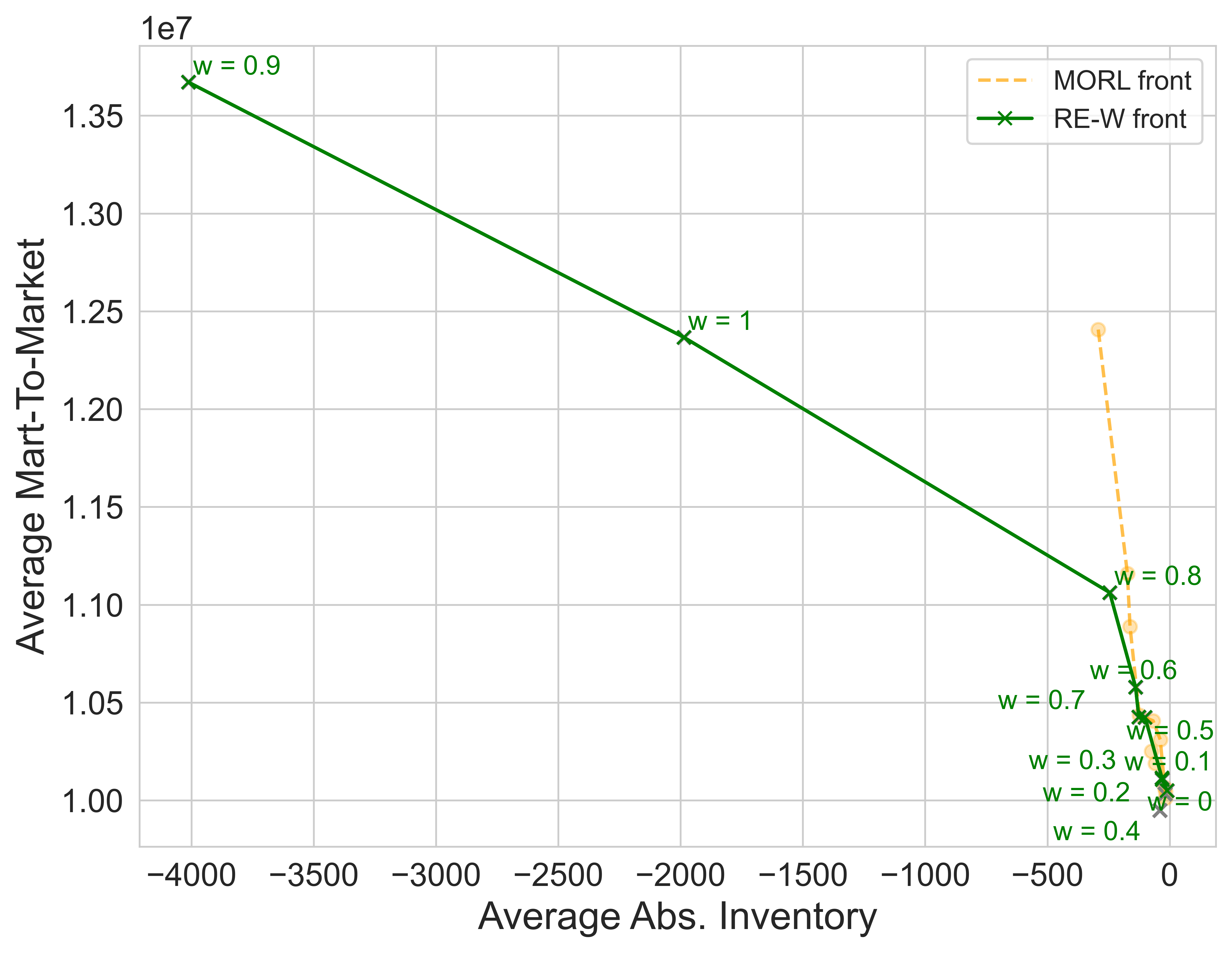}    
    \caption[Reward-engineered undominated fronts. More details]{Reward-engineered undominated fronts. The plots display all the solutions generated by the reward-engineered algorithm RE-W during the testing stage, distinguishing between dominated solutions (grey) and non-dominated solutions (orange). Each data point represents the average of five runs per weight.}
    \label{fig:test_right}
\end{figure}

In the initial results, it is observed that MORL and RE-AIIF solutions have similar undominated fronts, as depicted in \autoref{fig:test_left}. However, the RE-W alternative returns different undominated solutions, diverging from the MORL set (\autoref{fig:test_right}). Using the proposed metrics to compare the three alternatives, the following observations were made:

\begin{figure}[H]
\centering
  \centering  \includegraphics[width=1\linewidth]{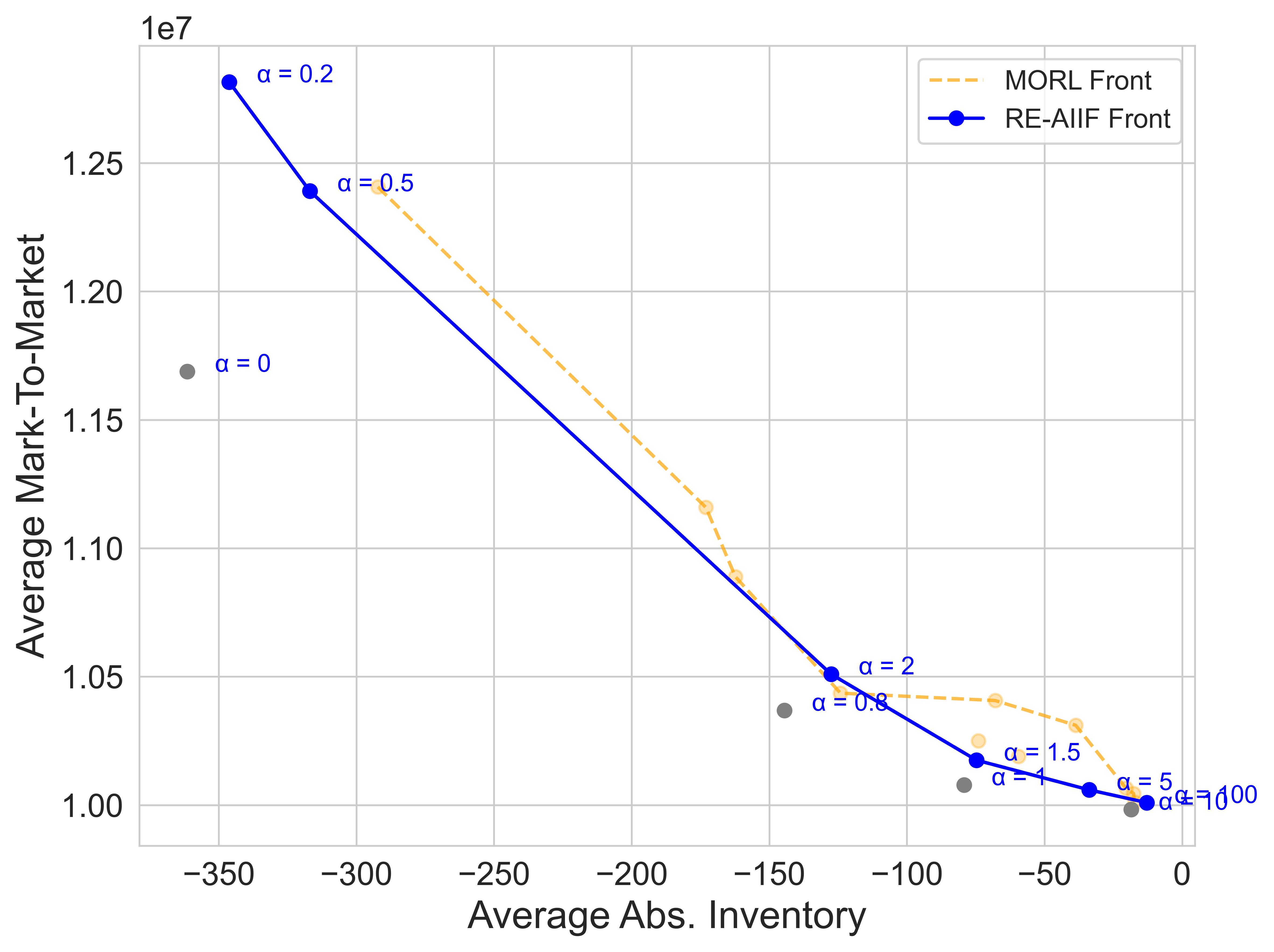}
    \caption[Reward-engineered undominated fronts.]{Reward-engineered undominated fronts. The plots display all the solutions generated by the reward-engineered algorithm RE-AIIF during the testing stage, distinguishing between dominated solutions (grey) and non-dominated solutions (orange). Each data point represents the average of five runs per weight.}\label{fig:test_left}
\end{figure}

\begin{figure}[H]
  \centering
  \includegraphics[width=\linewidth]{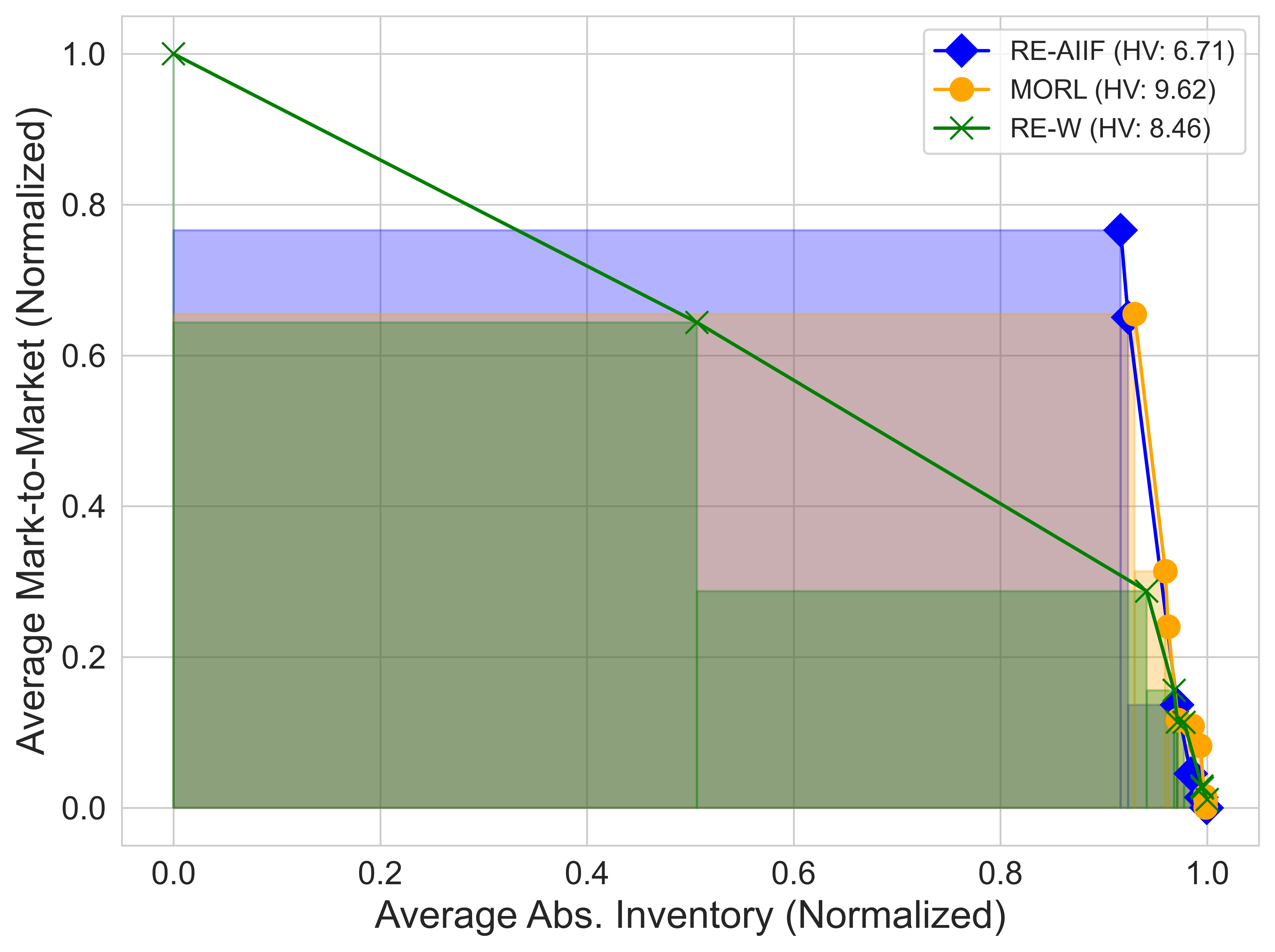}
  \caption[Comparison of normalized hypervolumes among algorithms]{Comparison of normalized hypervolumes among algorithms:  The extent of the objective space covered by each algorithm determines its performance, with a larger coverage indicating a better algorithm. In this sense, MORL has the larger hypervolume (9.62), followed by RE-W (8.46) and RE-AIIF (6.71).}\label{fig:paretocomb_plus_hypervol_normalized}
  \end{figure}

\textbf{1. Hypervolumes:}   \autoref{fig:paretocomb_plus_hypervol_normalized} and Table \ref{table-metrics} illustrate that the MORL approach covers a larger area compared to the classical alternatives: 9.71 vs 9.56 (RE-W) and 6.8 (RE-AIIF). To calculate the hypervolumes, all the solutions are first normalized to the range [0, 1], with a little extra margin. This is achieved by applying min-max normalization to every point, independently for each axis. For this exercise, the Nadir point is chosen as the reference point (\autoref{fig:pareto-explain}, \cite{Deb2014}).

\begin{figure}[H]
  \centering
  \includegraphics[width=\linewidth]{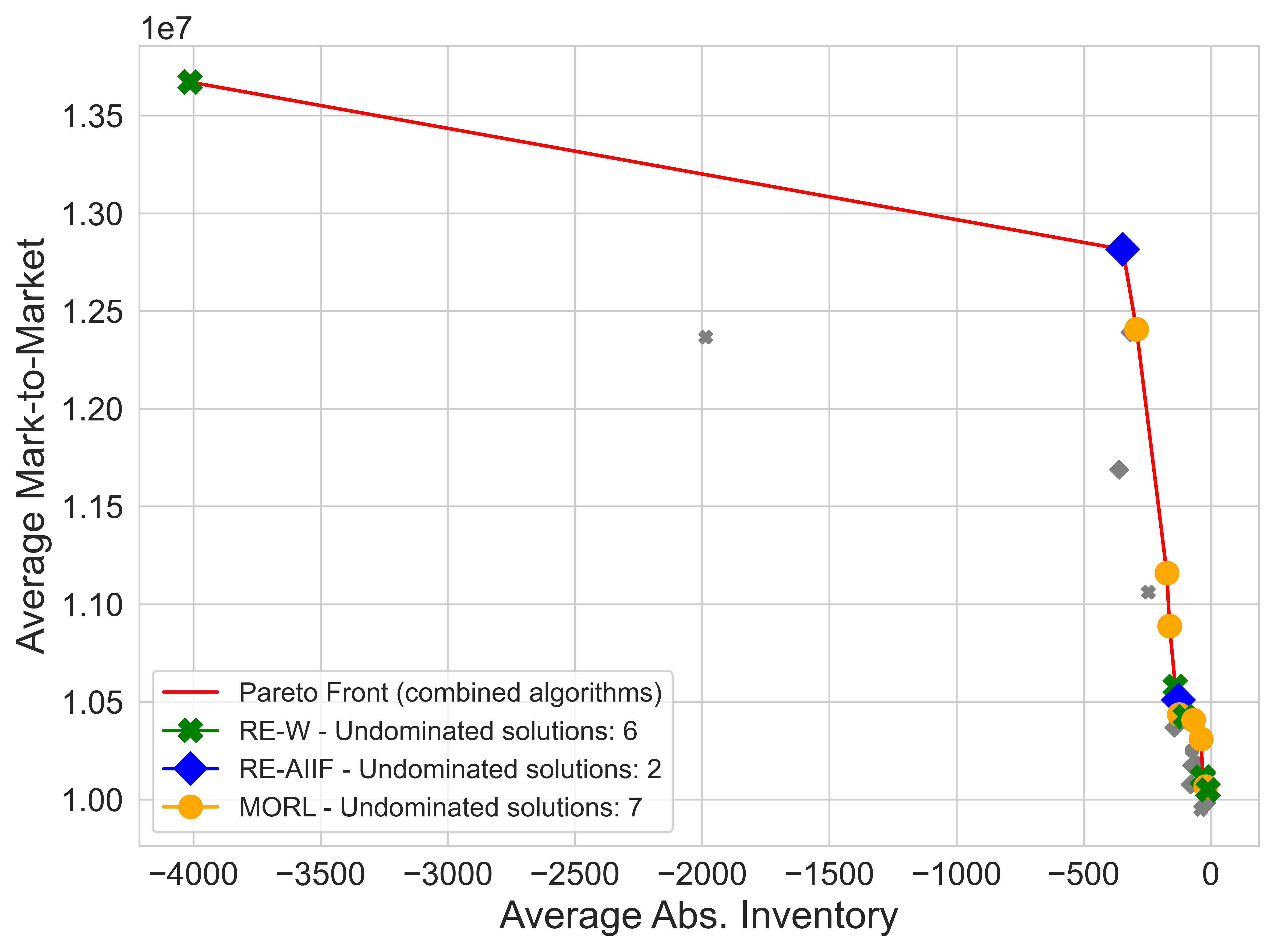}
  \caption[Combined Pareto front from undominated solutions]{Combined Pareto front from undominated solutions. The plot showcases the undominated solutions obtained from the three algorithms. In our case, MORL contributes the most solutions (7 out of 15), followed by the RE-W (6 out of 15), and the pure reward-engineered solution RE-AIIF (2 out of 15).  }
\label{fig:paretocomb_plus_hypervol}
\end{figure}

\textbf{2. Undominated solutions:} If all the solutions are aggregated into one set, as shown in \autoref{fig:paretocomb_plus_hypervol}, the MORL approach exhibits a higher number of undominated solutions (7 out of 15) compared to the classical approaches: 6 (RE-W) and 2 (RE-AIIF) out of 15. A higher number of undominated solutions indicates a higher quality front, as the algorithm provides more non-dominated solutions.

\textbf{3. Sparsity:} Finally, the average Euclidean distance between the undominated solutions is calculated to assess their proximity. Once again, the MORL approach demonstrates a lower sparsity between undominated solutions compared to the classical approaches: 299k vs 453k (RE-W) and  561k (RE-AIIF).

All the multi-objective metrics are detailed in Table \ref{table-metrics}. In addition to these metrics, the MtM and Inventory values from the MORL and RE-W experiments can be reviewed in Table \ref{table-values}. As for the RE-AIIF experiments, all the values can be reviewed in Table \ref{morl_table-ratios}).

\begin{table}[H]
\begin{center}
\begin{minipage}{\textwidth}
\caption[Comparison of multi-objective metrics: hypervolume, sparsity, and undominated solutions.]{Comparison of multi-objective metrics: hypervolume, sparsity, and undominated solutions. The performance of the MORL, RE-AIIF, and RE-W experiments is evaluated using the three key metrics. The results demonstrate that the MORL approach outperforms the other two reward-engineered algorithms across all three metrics, indicating its superior performance and effectiveness in handling the multi-objective nature of the problem.}\label{morl_table-ratios}
\label{table-metrics}
\begin{tabular*}{\textwidth}{@{\extracolsep{\fill}}lcccccc@{\extracolsep{\fill}}}
& \multicolumn{1}{@{}c@{}}{Hypervolume\footnotemark[1]}
& \multicolumn{2}{@{}c@{}}{Sparsity\footnotemark[2]}
& \multicolumn{2}{@{}c@{}}{Undominated} 
\\\cmidrule{2-2}\cmidrule{3-4} \cmidrule{5-6}%
Experiment &  Surface & Avg  &  $\sigma$ & Amount  \\
\midrule
\textbf{MORL}  &$ \textbf{9.62}$ &     $\approx \textbf{299k} $ & $ \approx\textbf{385k} $ & $ \textbf{7 (46.6\%)} $ 
 \\ 
RE-AIIF  & $6.71$ &     $\approx561k  $ & $ \approx674k $ & $ 2 (13.3\%)  $
\\ 
RE-W   & $8.46$ &     $\approx453k  $ & $ \approx515k $ & $ 6 (40\%)  $
\end{tabular*}
\footnotetext{Note 1: Normalized hypervolume.}
\footnotetext{Note 2: High values due to MtM figures.}

\end{minipage}
\end{center}
\end{table}

\subsection{Analysis: The relevance of trend information in the proposed method}\label{morl:Trends}

One relevant aspect of financial markets is related to trends, where prices typically continue moving upwards or downwards for unknown periods. In this regard, many algorithms and strategies take advantage of this behavior \cite{SZAKMARY2010409,rohrbach2017momentum,5331484}. Classical autoregressive models, such as ARIMA, are commonly used as inputs for various strategies.

In this dissertation, this aspect has not been directly considered. Although some instant variation information, such as the mid-price variation between the last timestep and the present, is included, there is no explicit consideration of trends in the current MM strategies. To evaluate whether this omission could represent a potential weakness in the proposed methods, an additional experiment has been conducted using the MORL agents. To capture short-term price trends, Exponential Moving Average (EMA) information has been added to the initial state space of the RL MM agents. EMA is preferred over the standard moving average because it is more sensitive to recent price changes, which is beneficial for high-frequency strategies. This momentum metric is commonly used in intraday strategies.

For this additional experiment, two different types of agents have been compared:

\begin{itemize}
    \item The MORL Agent without any additional information, as used in \autoref{morl_sec-morl}, where the standard state space of 8 variables is applied.
    \item Three MORL Agents that include 3 additional variables in their state space: EMA-L (long), EMA-S (short), and EMA-L Slope, resulting in a total of 11 state variables. The purpose of using two different EMA metrics is to allow the agent to detect when they cross each other, signaling different phases of the trend. 
\end{itemize}

The only difference between EMA-L and EMA-S is the number of data points considered in each option, with EMA-L exhibiting a slower behavior compared to EMA-S. EMA-L and EMA-S are computed as shown in \autoref{eq:ema}:

\begin{equation}
    \text{EMA}_t = \alpha \cdot x_t + (1 - \alpha) \cdot \text{EMA}_{t-1}
\label{eq:ema}
\end{equation}

\noindent where $\text{EMA}_t$ is the Exponential Moving Average at time $t$, $\alpha$ is the smoothing factor, typically calculated as $\alpha = \frac{2}{1+N}$, with $N$ being the number of periods. $x_t$ is the value at time $t$, and $\text{EMA}_{t-1}$ is the Exponential Moving Average from the previous time step.

EMA-L Slope, on the other hand, is the difference in EMA-L between two points, as represented in \autoref{eq:ema_slope}. This aims to capture the smooth trend in the market.

\begin{equation}
    \text{EMA-L Slope}_t = \text{EMA-L}_{t} - \text{EMA-L}_{t-n}
\label{eq:ema_slope}
\end{equation}

\begin{figure}[H]
    \centering
    \begin{minipage}[b]{0.48\textwidth}
        \centering
        \includegraphics[width=\textwidth]{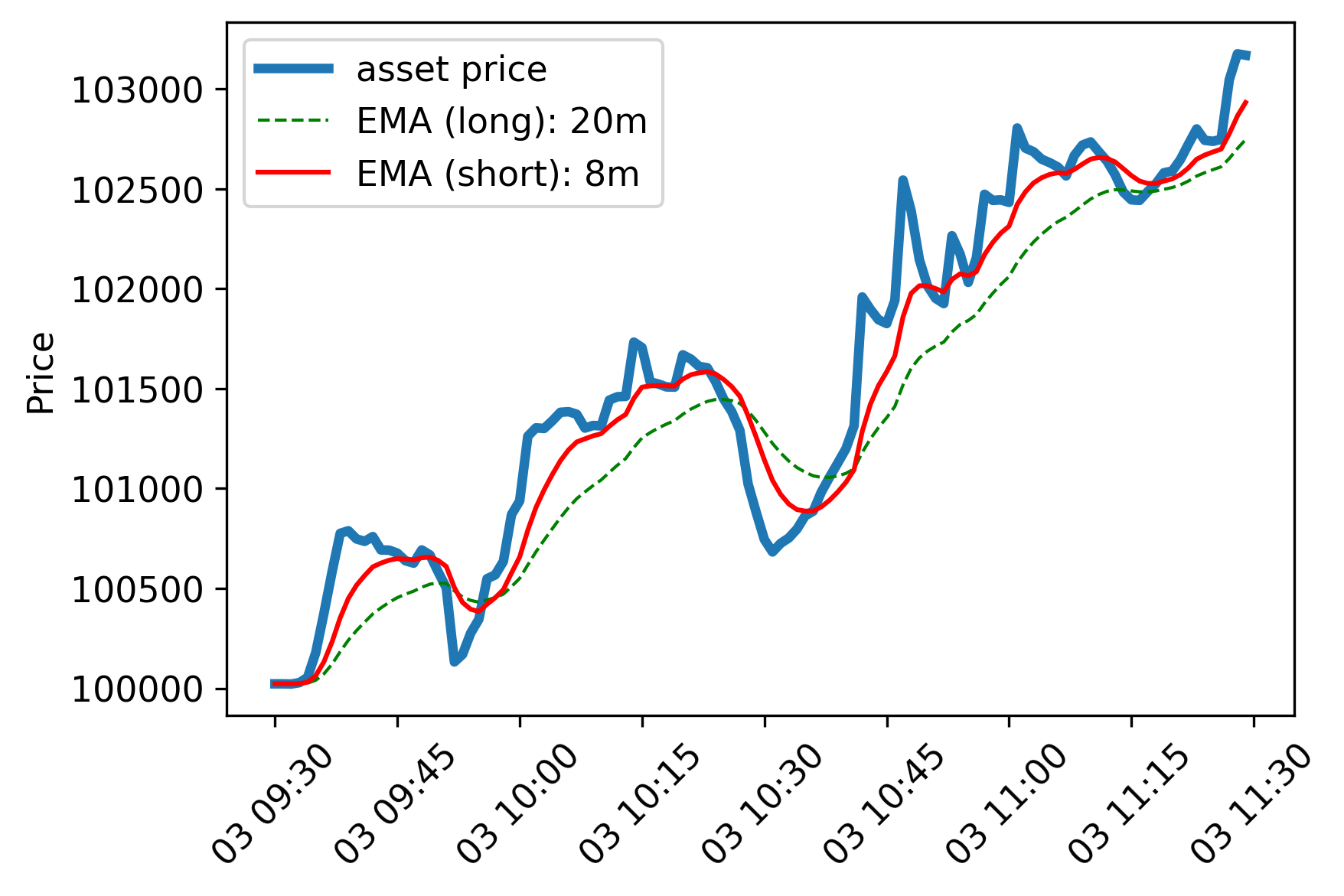}
    \end{minipage}
    \hfill
    \begin{minipage}[b]{0.48\textwidth}
        \centering
        \includegraphics[width=\textwidth]{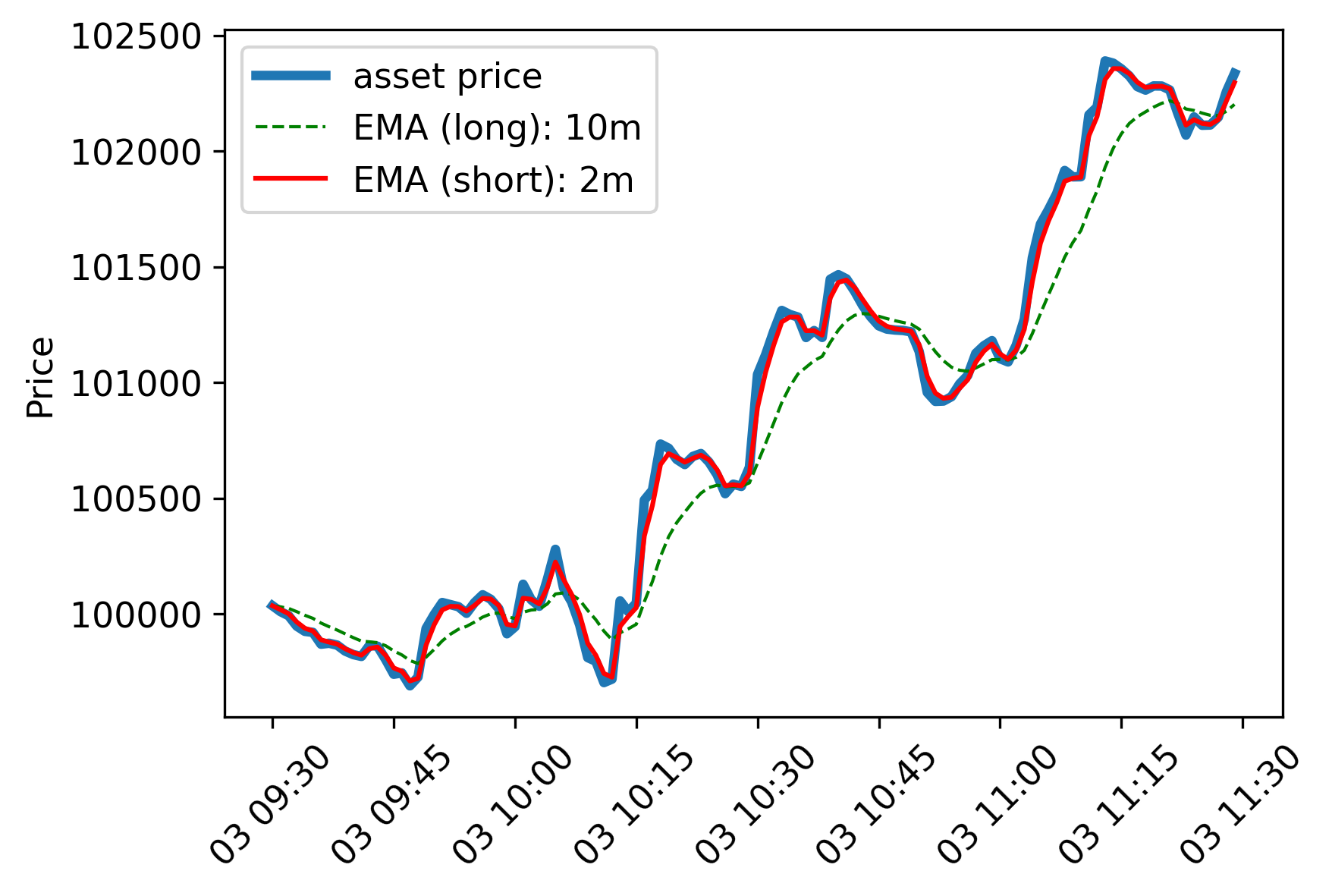}
    \end{minipage}     
     \caption{EMAs example of different trading sessions: EMA-L 20 minutes + EMA-S 8 minutes (left) and EMA-L 10 minutes + EMA-S 2 minutes (right).}
     \label{fig:emas_prices_example}
\end{figure}

In the experiment, three different EMA-L/EMA-S configurations have been tested: (20,8), (15,4), and (10,2), with EMA-L Slopes of 20, 15, and 10 minutes respectively. For instance, in the first case (20, 8) indicates the use of an EMA-L of 20 minutes and an EMA-S of 8 minutes.
\autoref{fig:emas_prices_example} shows two ABIDES trading sessions with different EMAs calculated based on the setup. 150 simulations per experiment (5 seeds) have been run, keeping the same parameters and hyperparameters the same as in the MORL \autoref{morl_sec-morl}.

\begin{figure}[H]
    \centering
    \includegraphics[width=1\textwidth]{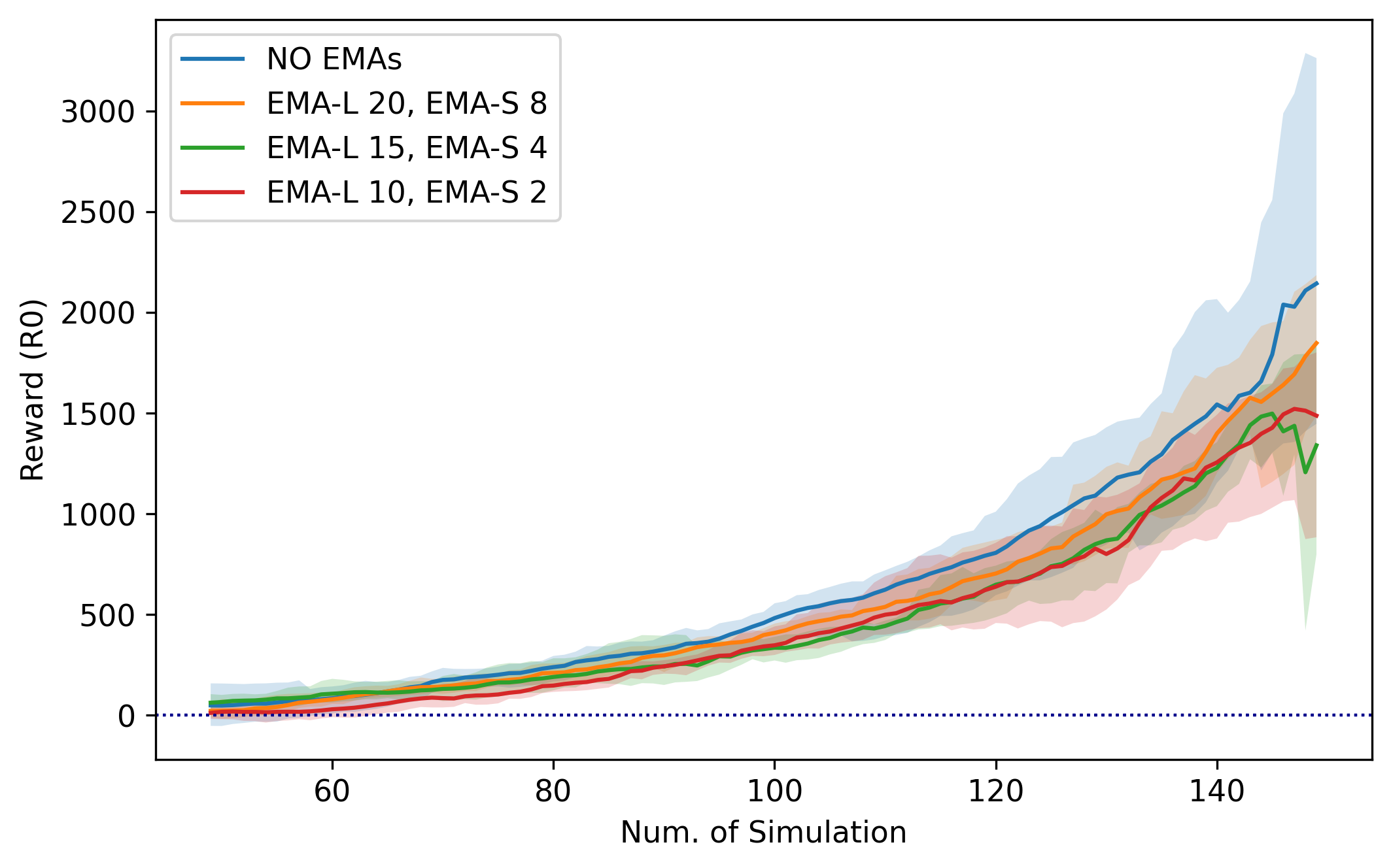}
    \caption{Training results of agents with and without EMA variables in the state space.}
    \label{fig:training_ema}
\end{figure}

\textbf{Results}: Initial results from the training stage indicate that agents with additional EMA variables in their state space perform worse, as shown in \autoref{fig:training_ema}.

\begin{figure}[H]
        \centering
        \includegraphics[width=1\textwidth]{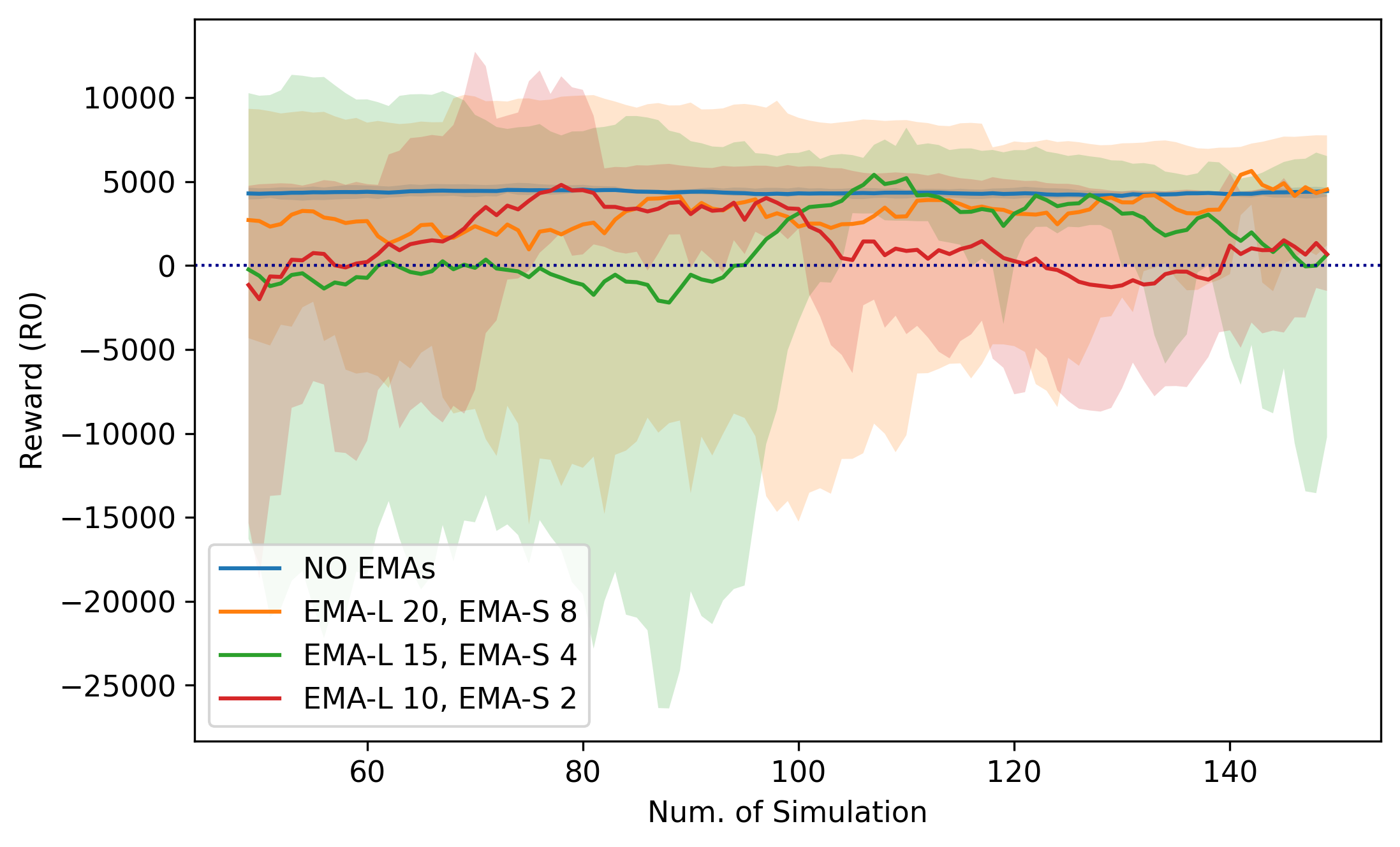}
        \caption{Testing results of agents with and without EMA variables}
        \label{fig:testing_EMA}
\end{figure}

Testing the final neural network from the training stage over 150 sessions confirms that including trend information does not improve performance in our MM strategy. In fact, it results in noisier behavior compared to the standard MORL agent, as depicted in \autoref{fig:testing_EMA}.

\begin{figure}[H]
        \centering
        \includegraphics[width=1\textwidth]{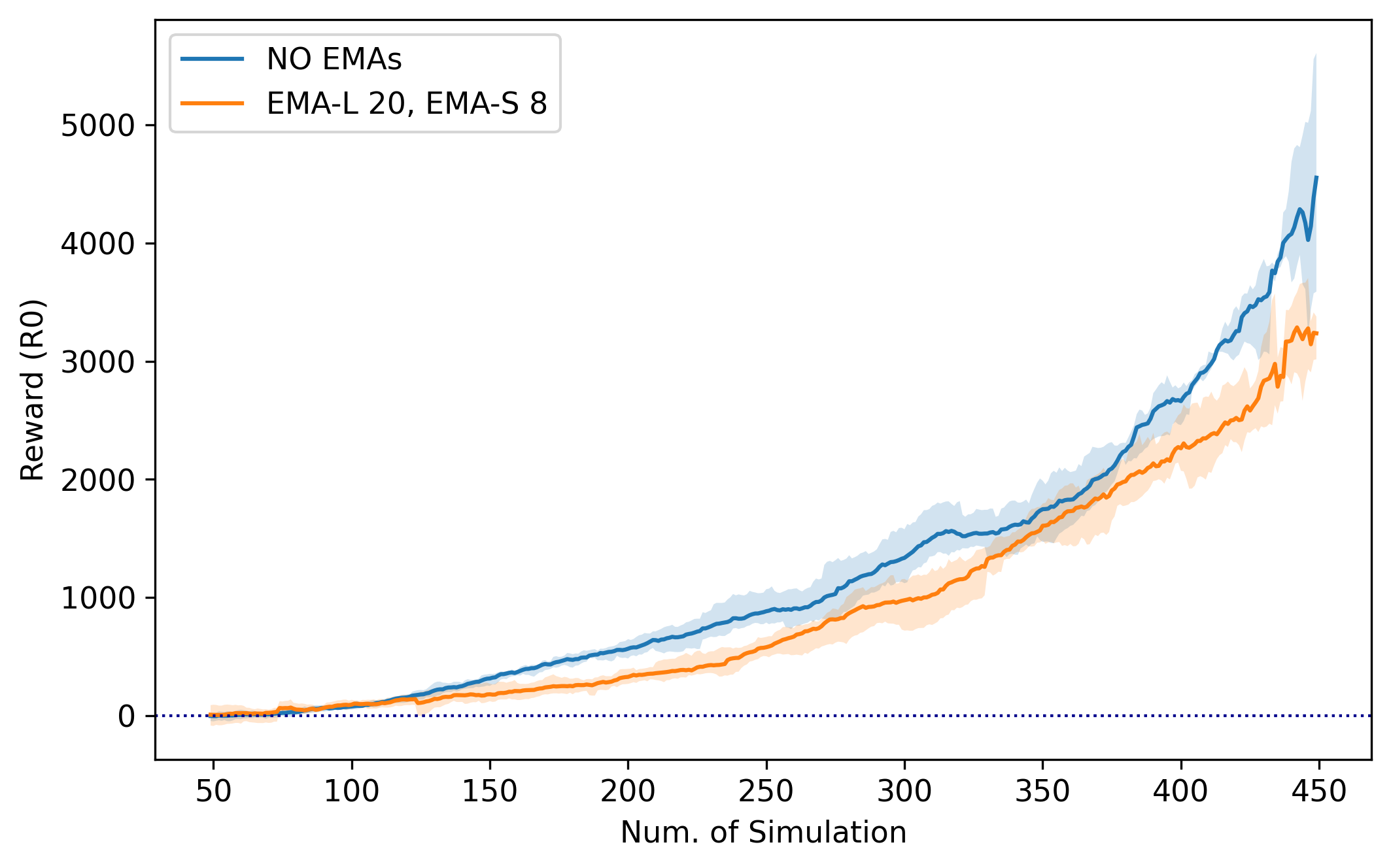}
        \caption{Training results of extended simulations with EMA and standard agents.}
        \label{fig:ema_longtrain}
\end{figure}

There are several possible reasons for this degradation. One possibility is the increased complexity of the state space, which expands from 8 to 11 variables. To rule out this factor, an additional experiment was conducted by increasing the number of simulations from 150 to 450 to see if performance improved compared to the standard agent. Due to computational constraints, only one EMA agent (the best-performing one from the previous training) was tested in this validation. As seen in \autoref{fig:ema_longtrain}, this increase proportionally affects both agents without any significant improvement in the EMA agent.

Other factors contributing to this lack of improvement could also be considered. For instance, the nature of the strategy, which is based primarily on instantaneous spreads and inventory management, and characterized by a high number of operations within short timeframes (HFT style), may be less reliant on long-term or short-term price trends. High-frequency trading strategies typically prioritize rapid execution and immediate market conditions, making them less sensitive to the momentum and trend-following signals that EMAs are designed to capture. It is possible that other types of strategies, particularly those with longer holding periods or a focus on trend-following, would benefit more from incorporating EMA variables. 

Therefore, while the introduction of EMA variables did not improve the performance of the current strategy, it does not rule out their potential utility in other trading approaches that depend more heavily on market trends and price momentum.

\subsection{Discussion } \label{morl_sec-conclusion}

The results of the experiments demonstrate that the MORL approach consistently outperforms single-objective reward-engineered methods when controlling both goals simultaneously (\autoref{table-values}). Specifically, it surpasses the classical approach in multi-objective key metrics such as hypervolume, sparsity, and the number of undominated solutions, indicating significant improvement (\autoref{table-metrics}). The separate treatment of sub-objectives within the MORL approach enables better specialization and stability in the solutions. Additionally, compared to a fully reward-engineered approach (w/o weights), the MORL approach offers several distinct advantages in the context of market making problems: (1) By treating profitability and inventory management as separate objectives, the MORL approach eliminates the need for time-consuming reward engineering and removes any subjective influence on reward definition; (2) The separation of the two objectives within the framework allows the agent to specialize and comprehend the nuances of both objectives, resulting in improved overall performance; (3) The separation of subobjectives also enhances the explainability of the subsequent selection process, as it highlights the impact of MtM improvement versus inventory management and vice versa; (4) Furthermore, the MORL approach is highly adaptable to changes in the final utility function, as it relies on a single factor (weight, denoted as $w$) that can be easily adjusted over time. This allows for dynamic risk aversion adjustments in the market making agent during trading sessions. 

While similar adaptability can be achieved in weighted reward-engineered functions such as in RE-W, the MORL approach with its separate NNs for each objective offers more flexibility in modifying the architecture, Q-Network, or state space for each objective. This flexibility is particularly valuable when dealing with assets exhibiting complex behaviors, such as highly volatile stocks or assets with low liquidity. These advantages, including the ability to manage the complexity of one objective without sacrificing performance in the other, are not easily attainable with reward-engineered alternatives.

Overall, the MORL approach offers superior performance, increased adaptability, better objective specialization, and enhanced explainability compared to single-objective reward-engineered methods. It provides a robust framework for addressing multi-objective market making problems and offers significant advantages over traditional approaches.

From a financial standpoint, these tools offer a more robust approach to navigating the markets. The Pareto fronts (PFs) allow financial business users to assess how well the agent will perform by considering the balance between maximizing profits and managing inventory. By selecting the proper `w' parameter based on the operator's specific a priori utility function, strategies can be adapted to different market conditions or levels of aggressiveness. This means they can effectively tailor their approach to meet specific goals, whether it is increasing profits or minimizing risks. Ultimately, these tools provide valuable insights that help them make informed decisions and optimize their trading strategies. It also provides them with an understanding of the algorithm's limitations before deploying it in the market.
  
Nevertheless, this technique has its limitations. A weakness of this method is the necessity of having one agent per trained weight. This implies training one agent per $w$, which increases the complexity of the final solution, and the training/adjusting time, and reduces the generalization capabilities of the method. Other alternatives exist to address this issue, such as Universal Value Function Approximators (UVFAs) \cite{pmlr-v37-schaul15}, but they have shown worse performance compared to specific MORL algorithms \cite{Hayes2022}. Another weakness compared to reward-engineered versions is the increased computational time. Using one (or two) neural networks per goal, instead of just one as in the reward-engineered alternative, can increase both computational time and resource usage. Parallelization techniques may be necessary to reduce the required computing time.

\subsection{Contributions } \label{morl_contribs}

This work makes significant contributions to the optimization of market making strategies through a multi-objective RL  framework that deals perfectly with the profitability/inventory control challenge. The specific contributions are:

\begin{enumerate}
    \item \textbf{Development of a multi-objective RL framework}: Introduction of a multi-objective RL framework that outperforms previous reward engineering approaches in market making. The advantages of this framework include:
    \begin{itemize}
        \item Elimination of the need for time-consuming reward engineering, streamlining the development process.
        \item Reduction in computational time due to more efficient learning algorithms.
        \item Enhanced specialization in subgoals, allowing for more targeted strategy development.
        \item Improved capability to balance between conflicting subgoals, optimizing overall strategy effectiveness.

    \end{itemize}
    
    \item \textbf{In-depth analysis of multi-objective metrics}: Extensive evaluation of the proposed multi-objective RL framework using several key metrics, which are crucial for validating the effectiveness and robustness of the approach:
    \begin{itemize}
        \item \textbf{Hypervolume}: Utilization of hypervolume measurements to assess how well the proposed solutions cover the range of possible objective values.
        \item \textbf{Sparsity}: Analysis of sparsity to determine the distribution of solutions within the objective space, ensuring diverse solution sets.
        \item \textbf{Number of undominated solutions}: Examination of the count of undominated solutions to evaluate the competitiveness of solutions against each other.
    \end{itemize}
    These metrics provide a comprehensive framework for validating the effectiveness of the multi-objective RL approaches and demonstrating their superiority over traditional single-objective optimization techniques.
\end{enumerate}

Additionally, the introduction of trend information has been tested by incorporating EMA variables into the state space to assess their impact on the proposed method. The results indicate that, in this case, the inclusion of this information does not provide any advantage. Although this was tested on MORL agents within this section, these conclusions can be extended to the rest of the agents presented.

This work establishes a market making agent capable of robustly managing both profitability and inventory control, paving the way for further enhancements. The agent's dual focus not only strengthens its operational efficiency in real-time market conditions but also sets a foundation for future advancements in market making strategies. By integrating multi-objective RL, the agent effectively balances the complex trade-offs between achieving high returns and maintaining optimal inventory levels. This balance is crucial for sustaining long-term viability and competitiveness in the dynamic landscape of financial trading.

\chapter{Automation under non-stationarity}\label{chap_nonstationarity}

Having developed an RL market making agent capable of maintaining profitability while managing inventory, the next step is to ensure the agent's autonomous adaptation to environmental changes. Given the dynamic nature of financial markets, having a strategy that can seamlessly adjust to these fluctuations is essential. As mentioned earlier, financial markets are inherently non-stationary, making it crucial to address this property. In the upcoming section, a modification of the previously established M$^3$ORL agent will be introduced. This enhancement aims to transform the agent into an autonomous entity capable of navigating the evolving landscapes of financial markets robustly.

\section{Dealing with non-stationarity: Continual learning}\label{back-cl}

As stated in \autoref{sota_nonstation}, one usual technique to adapt to changes in the environment is letting the agent, or the ML model, be fine-tuned with new incoming experiences or data without losing previous knowledge. This is also known as continual learning (CL). In this regard, there are ML techniques that deal with this issue. One of the goals of the work presented in this chapter is to examine and compare three different strategies that address the challenges of CL: freezing layers (FL), data rehearsal (DR), and elastic weight consolidation (EWC). The purpose is to understand the strengths and weaknesses of each approach and to establish a benchmark for evaluating the algorithm proposed in this dissertation. This section provides an overview of each of these strategies.

\noindent \textbf{Freezing layers}: This method strategically fixes the weights of certain layers within the network, typically the early layers responsible for capturing basic and general features, while allowing the weights of later, more task-specific layers to adjust and learn from new tasks \cite{Sorrenti2023}. The rationale behind freezing the initial layer/s is to preserve the fundamental knowledge they have, which is assumed to be broadly applicable across various tasks. On the other hand, the adaptable deeper layers facilitate the learning of new, task-specific nuances. Let  \( \Theta \) represent the entirety of parameters within the NN, \( \Theta_f \) denote the subset of parameters located within the frozen layers, and \( \Theta_t \) signifies the subset of parameters residing within the trainable layers. During training on a new task, the update rules for the parameters are:

\begin{enumerate}
    \item For frozen layer parameters \( \theta_f \in \Theta_f \):
    \[
    \theta_f^{(\text{new})} = \theta_f^{(\text{old})}
    \]
    i.e., the parameters in frozen layers remain constant.

    \item For trainable layer parameters \( \theta_t \in \Theta_t \):
    \[
    \theta_t^{(\text{new})} = \theta_t^{(\text{old})} - \alpha \frac{\partial L}{\partial \theta_t}
    \]
    i.e., the parameters in trainable layers are updated via gradient descent, where \( \alpha \) is the learning rate and \( \frac{\partial L}{\partial \theta_t} \) is the gradient of the loss function \( L \).
\end{enumerate}

This technique is not only applied on CL tasks, it is also indeed very common in fine-tuning tasks \cite{ShenLiuQinSavvidesCheng2021, Goutam2020}, moreover when big pre-trained models are involved.

\noindent \textbf{Elastic weight consolidation}: another alternative, which is ``parameter's selective'' instead of ``layer's selective'', is EWC \cite{doi:10.1073/pnas.1611835114}. This approach mirrors, in some way, the brain's method of synaptic consolidation, allowing for continued learning in artificial NNs by ensuring key parameters remain close to their previously established values. To achieve this, the Fisher information matrix is calculated once the agent is trained, using the training dataset. This way, a relative importance weight factor is obtained, which is used as a penalty in further training. This approach prevents the deterioration of performance on earlier tasks while new tasks are being learned. Due to the inherent over-parameterization in deep NNs, where numerous parameter configurations can lead to similar outcomes, EWC strategically restricts updates to a specific zone that maintains accuracy for earlier tasks. This restriction is realized through a quadratic penalty that acts akin to a variable-strength spring, anchoring the parameters based on their relevance to the task just completed. EWC meticulously modulates the learning trajectory to prevent significant deviations from the solutions of all preceding tasks, thereby harmonizing the retention of previous knowledge with the acquisition of new information, according to the loss function of \autoref{losswec}.

\begin{equation}\label{losswec}
    L(\theta) = L_B(\theta) + \sum_{i} \frac{\lambda}{2} Fisher_i (\theta_i - \theta^*_{A,i})^2
\end{equation}

Where $L_B(\theta)$ is the loss for task B only (the new one), $\lambda$ sets the importance of the old task relative to the new one, $Fisher_i$ is the Fisher information matrix concerning the previous task's parameters, $\theta^{*}_{A,i}$ represents the parameters of the previous task $A$, and $i$ indexes each parameter. 

\noindent \textbf{Data rehearsal}: The last scrutinized technique, compatible with the aforementioned methods, leverages the reuse of previous experiences to mitigate the forgetfulness of older information upon the introduction of new data \cite{ROBINS1995,9412614, Atkinson2021}. Particularly relevant in the context of off-policy RL algorithms, this approach involves the utilization of old replay buffers in conjunction with new transitions during the training phase. In essence, multiple distinct replay buffers exist, each accumulating experiences based on varied criteria. This work explores the integration of two complementary replay buffers — one containing the initial agent's training data and the other housing new experiences acquired during the testing phase — in varying proportions. Specifically, being a dataset of $t$ experiences \(D = \{(s,a,s',r)_0, \ldots, (s,a,s',r)_t\}\), two of them are considered: $D_{\text{old}}$ with old experiences and $D_{\text{new}}$, containing newly gathered experiences during the testing phase. The combined replay buffer $D_{\text{cl}}$, employed for CL, is formulated as in \autoref{eq:datarehearsal}.

\begin{equation}
    D_{cl} = \gamma \cdot D_{old} + (1 - \gamma) \cdot D_{new}
    \label{eq:datarehearsal}
\end{equation}

Here, $\gamma \in [0, 1]$ represents the proportion in which the old experiences are incorporated relative to the new experiences, effectively balancing the influence of historical and recent data in the CL process. This approach allows the RL algorithm to balance between both data distributions, forcing it to learn a more general policy.

\section{Discounted Thompson sampling}
\label{back_dts}

As part of the proposed MM agent, it is necessary to introduce the discounted Thompson sampling (dTS) technique. Classical TS \cite{dc35850b-2ca1-314f-9e0d-470713436b17} is a multi-armed Bayesian bandit strategy employed in sequential decision-making scenarios, designed to strike a balance between leveraging existing knowledge for immediate gains (exploitation) and gathering new data to enhance future outcomes (exploration). This algorithm efficiently tackles a wide spectrum of challenges, earning it widespread adoption due to its computational effectiveness and versatility. In the TS algorithm with Bernoulli arms, a Bernoulli sampling model coupled with a Beta($\alpha$, $\beta$) prior distribution can be selected. Consequently, as depicted by \autoref{thomsomeq}, the posterior distribution for each round will follow a beta distribution: 
\begin{equation}\label{thomsomeq}
I_i \sim \text{Beta}(\alpha + S_{i}, \beta + F_{i}), \quad i = 1,2,3, \ldots, k; 
\end{equation}
\newline
 $S_i$  denotes the cumulative number of successes and  $F_i$  denotes the cumulative number of failures. As the number of rounds progresses, the beta distribution increasingly narrows in on the empirical mean, reflecting a more precise estimation of the underlying probability with each additional round. This default setup is good where a stationary environment is present, where coefficients grow according to the evolution of the trials. The problem arises when the underlying dynamics change, making it difficult to adapt these coefficients to the new probability distributions, as it has to forget some way the last inputs. To mitigate this, there is one modification of the TS algorithm called \textit{discounted Thompson sampling} \cite{8842548, Asyuraa2021}. In this alternative, two coefficients modulate the $\alpha$ and $\beta$ parameters in every update, according to the Equations \ref{discthomsom_a} and \ref{discthomsom_b}:

\begin{equation}\label{discthomsom_a}
\begin{aligned}
S_i = S_i + r_s,  & \text{ if } c_i = a_i,  \\
F_i = F_i + r_f, & \text{ if } c_i \neq  a_i 
\end{aligned}
\end{equation}

\begin{equation}\label{discthomsom_b}
S_i = \gamma S_i, F_i = \gamma F_i t\text{,    }  \forall \gamma \in [0, 1]
\end{equation}

At every round $i$, the dTS algorithm selects an arm $c_i$ based on the current $\alpha$ and $\beta$ coefficients. If the chosen arm is the winning arm $a_i$, the success counter $S_i$ is incremented by $r_s$; otherwise, the failure counter $F_i$ is incremented by $r_f$. Additionally, both coefficients are discounted by $\gamma$ as described in Equation \ref{discthomsom_b}. Therefore, the $\gamma$ coefficient allows us to adjust the significance of recent experiences relative to older ones, depending on the characteristics of the specific environment. 

\section{Methodology: Multi-objective policy selection in dynamic markets}\label{sec-methods}

As described, the market making challenge involves balancing two main objectives: (i) maximizing MM profitability and (ii) minimizing inventory levels to reduce devaluation risks. To tackle these objectives, a MORL methodology (M$^3$ORL MM) was introduced in \autoref{morl_sec-morl}. This approach allows for the autonomous learning of these objectives within a specialized multi-objective framework. After learning a set of multi-objective policies across various contexts, it is proposed to assemble a library with these policies. The POW-dTS algorithm, outlined in \autoref{subsec:powdts}, is then tasked with selecting the appropriate combination of these pre-trained policies from the library based on the evolving market conditions. This technique will be an essential component of the autonomous MM agent.

\subsection{Market Maker as a MOMDP}
\label{subsec:mmmomdp}

In our RL framework, the RL MM agent is conceptualized as the M$^3$ORL MM agent presented in \autoref{morl_sec-morl}, and in previous sections as well. In this regard, our MM is a DQN-based agent equipped with two pairs of NNs (train and target), conformed by 3 hidden layers with 32 neurons per NN, each managing a specific goal separately. The state and action spaces, as well as the multi-objective reward vector, remain identical to those of the MORL agent.

\subsection{Policy Weighting through discounted Thompson sampling (POW-dTS) Algorithm}
\label{subsec:powdts}

To achieve good performance with M$^3$ORL MMs in non-stationary markets, the following POW-dTS algorithm is proposed. In essence, it is a dynamic agent weighting mechanism that utilizes dTS to alternate among different policies sequentially. The primary goal is to manage effectively a set of M$^3$ORL MM pre-trained agents (or policies) by periodically recalibrating their weights based on real performance, thereby optimizing the decision-making process over time along an evolving market. The intuition behind this is that each pre-trained agent is allocated a duration of time according to the weight that dTS assigns to it. It is important to remark that, instead of alternating agents every time step, they are allowed to perform actions during their corresponding time steps, and weights are calculated accordingly. This approach is deemed more appropriate as RL is designed to act based not only on the immediate reward but also on subsequent ones. Alternating the agents in blocks in this manner ensures they better exploit their respective policies.

The algorithm is divided into two functions. Algorithm \ref{algo_1} provides the general flow, while Algorithm \ref{alg_2} focuses on the evaluation and recalibration part of the coefficients and weights of the dTS. Initially, as outlined in Algorithm \ref{algo_1}, the algorithm sets up initial parameters and structures. Some parameters must be passed to the algorithm: the library of agents/policies to be used, $A$, the modifiers of $\alpha$ and $\beta$ coefficients, $\alpha_{inc}$ and $\beta_{inc}$, the gamma discount parameter of the dTS algorithm, $\gamma$, how often the algorithm should be recalibrated, $rounds_{exp}$, the number of testing time steps per recalibration round, $rounds_{recal}$, and how many total rounds the agents will perform their actions in the exploitation stage (according to their weights), $exp_{ts}$. Some variables also need to be initialized: the recalibration flag $recal$, which indicates whether the algorithm is performing an update (line 1 in Algorithm \ref{algo_1}); the initial alpha and beta coefficients for all the agents (line 2); the default initial weights per agent (line 3); the actuation range per agent, denoted by the array $secs$ (line 4); and a starting agent $agt$ (line 5). The sections included in the variable $secs$ will be populated during the recalibration stage, according to this structure: $[(x_i, x_e)_0, \ldots, (x_i,x_e)_n]$, where both $x_i$ and $x_e$ represent the initial and ending time steps for each agent, respectively. Lastly, the counter $ts_{test}$ is initialized to 0 (line 6).

\begin{algorithm}[H]
\caption{POW-dTS(A, $\alpha_{inc}$, $\beta_{inc}$, $\gamma$, $rounds_{exp}$, $rounds_{recal}$, $exp_{ts}$ )}
\label{algo_1}
\begin{algorithmic}[1]
\State Set recalibration flag: $\text{recal} \gets \text{False}$, 
\State Initialize alpha and beta coefficients per agent: $\text{coefs} \gets [(1,1)_0, \ldots, (1,1)_n]$
\State Initialize agent's weights: $\text{weights} \gets \left[\frac{1}{n}, \ldots, \frac{1}{n}\right]$
\State Define sections array: $\text{secs} \gets [\text{ }]$ 
\State Select an initial agent: $\text{agt} \gets A[0]$

\State Init $ts_{test} \gets 0$
\State 
\For{$t = 0$ to $T$}
    \State rew $\gets$ step(agt) \Comment{Execute action by current agent $\textit{agt}$ and collect reward}
    \If{$t \mod rounds_{exp} = 0$} \Comment{Recalibration triggered}
        \State Set recalibration flag: $\text{recal} \gets \text{True}$ 
        \State Reset current time steps:  $ts_{recal} \gets 0$, $ts_{test} \gets 0$
        \State Set current agent index: $\text{$curr_{agt}$} \gets 0$ 
        \For{$f$ in range(len($A$))}
            \State Init results dictionary: $\text{$results$}[f] \gets [\text{ }]$
        \EndFor
    \EndIf
    \If{$\text{recal}$} \Comment{The agents' performance is tested}
        
        \State $results[curr_{agt}]$.append($rew$)  \Comment{Append the reward}
        \State  $\text{$curr_{agt}$} \gets \text{round}(ts_{recal} / rounds_{recal})$ \Comment{Update current agent index}
        
        \If{$\text{$curr_{agt}$} < \text{len}(A)$}
            \State  $\text{$agt$} \gets \text{A}[curr_{agt}]$ \Comment{Select the agent}
        \Else             
            \State $\text{recal} \gets \text{False}$ \Comment{End recalibration}
            \State $agent_{win} \gets$ max reward agent index at $results$
            \State secs $\gets$ Recalibration(A, coefs, $\alpha_{inc}$, $\beta_{inc}$ $agent_{win}$, $exp_{ts}$).\Comment{Alg. \ref{alg_2}} 
        \EndIf
        \State $ts_{recal}$ += 1        
    \Else
        \State $agt \gets$ Select the agent for the current time step $ts_{test}$ based on sections $\textit{secs}$
        \State  $ts_{test}$ += 1
    \EndIf
\EndFor
\end{algorithmic}
\end{algorithm}

Once everything has been initialized, the algorithm enters a predefined loop for a set number of iterations $T$ (line 8), with each iteration representing a single time step. During each time step, the selected and active agent, denoted as $agt$, executes a greedy action, and the outcome, or reward $rew$, is recorded (line 9). Periodically, the algorithm initiates a recalibration of the agents' weights to adapt to environmental changes (line 10). This recalibration process begins after initializing all required variables (lines 11-16), where each agent is allotted an equal number of time steps to act in a sequential manner (line 20). Throughout these recalibration periods, agents interact with the environment, earning rewards that reflect their performance (line 19). Upon evaluating all the agents or policies in $A$ (lines 21-23), the algorithm determines the agent that achieved the highest returns during the testing phase (line 25). Utilizing this data, it proceeds to an update phase, adjusting the weights to current performance levels through the \textit{Recalibration} function (line 26). Once the recalibration phase has concluded and all the new sections have been determined, the agent is selected according to the counter $ts_{test}$ (line 30). For instance, if there are 4 policies or agents with the following sections $secs = [(0,10)_{ag1}, (11,25)_{ag2}, (26,30)_{ag3}, (31,38)_{ag4}]$, and the current testing time step is 14, then agent number 2 will be active. This agent continues until the $ts{test}$ counter reaches 26, at which point agent 3 will be selected.

\begin{figure}[H]
        \centering
        \includegraphics[width=1\textwidth]{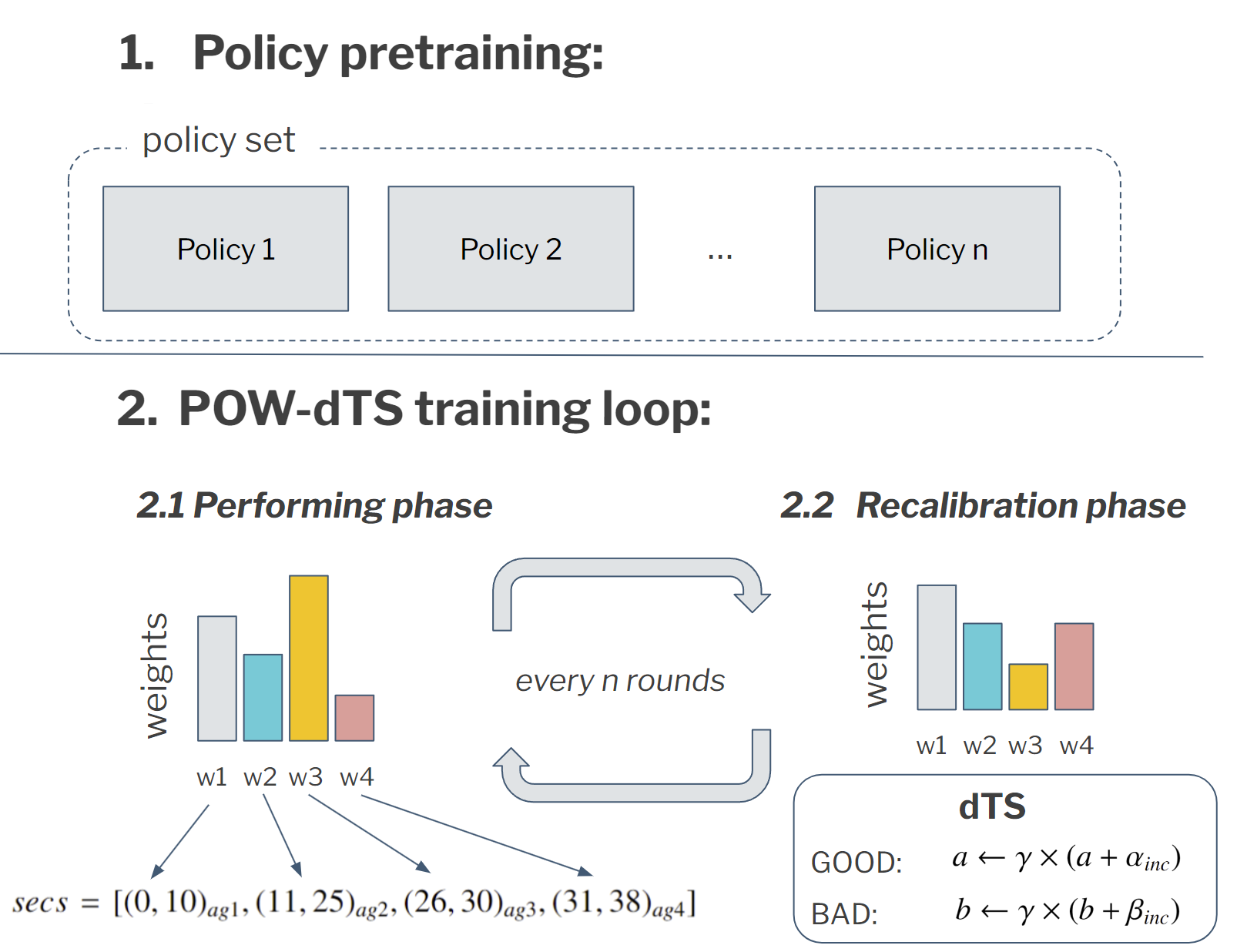}
        \caption[POW-dTS flow]{This image depicts the POW-dTS flow. In the first stage of the algorithm, a set of policies is trained to capture the different dynamics in the environment. Once the policy set is defined, the algorithm moves to the Performing phase, where it operates in the market by assigning sequential ranges of time (secs) to the policies based on calculated weights. The weights are recalculated every \(n\) steps during the recalibration phase, using discounted Thompson sampling. 

}
        \label{powdts_sample}
        \end{figure}

Exploring the recalibration function further (Algorithm \ref{alg_2}), which calculates weights and sections, it starts by processing the initial set of reference policies/agents, $A$, the current $\alpha$ and $\beta$ coefficients for every agent, $coefs$, the agent that achieved the highest reward in the previous testing phase, $agent_{win}$, and the total number of time steps during which all agents will perform their policies in the exploitation stage, $exp_{ts}$, as determined by the recently computed weights. The initial step involves selecting a winning agent based on the current $\alpha$ and $\beta$ coefficients (line 1 in Algorithm \ref{alg_2}). These coefficients influence the selection probabilities. If the POW-dTS's predicted winner aligns with the actual testing stage winner (line 8), an increase to that agent's $\alpha$ coefficient is applied (line 9) to promote its future selection; conversely, if there's a mismatch, the $\beta$ coefficient is incremented instead (line 11). This adjustment is designed to shift selection probabilities in favor of agents demonstrating superior performance. Additionally, a discount factor, $\gamma$, is applied to the coefficients of all the agents, including the rest (line 14). After recalibrating the $\alpha$ and $\beta$ coefficients, the agents' weights are recalculated (lines 20 - 24), altering the probability distribution in line with the adjusted $\alpha$ and $\beta$ coefficients and ensuring the total weights sum to 1. These revised weights then inform the determination of operational intervals for each agent (lines 25-27), proportionate to the recalculated weights. This function finally returns the sections for all the agents that are used in Algorithm \ref{algo_1} (line 30). \autoref{powdts_sample} depicts the algorithm flow.

\begin{algorithm}[H]
\caption{Recalibration(A, coefs, $\alpha_{inc}$, $\beta_{inc}$ $agent_{win}$, $exp_{ts}$)}
\label{alg_2}
\begin{algorithmic}[1]
\State $\text{$best_{thp}$} \gets \text{argmax}([\text{random.beta}(a,b) \text{ for } (a,b) \text{ in coefs}])$
\State Initialize: $init_{pos} \gets 0$
\State
\State\textit{\# $\alpha$ and $\beta$ coefficients update}
            \For{$ag$ in range(len(A))}
                \State $a, b \gets \text{coefs}[ag]$ \Comment{Current $\alpha$ and $\beta$ coefs of this agent are gathered}
                \If{$ag == \text{$agent_{win}$}$} 
                    \If{$ag == \text{$best_{thp}$} $}
                        \State $a \gets \gamma \cdot (a + \alpha_{inc}$)\Comment{Increase $\alpha$ (success)}
                    \Else
                        \State $b \gets \gamma \cdot (b + \beta_{inc}$) \Comment{Increase $\beta$ (fail)}
                    \EndIf
                \Else \Comment{For the rest a and b are discounted by gamma}
                    \State $a \gets \gamma \cdot a, b \gets \gamma \cdot b$
                \EndIf
                \State $\text{coefs}[ag] \gets (a, b)$ \Comment{Update with the new coefs}
            \EndFor
            
            \State
            \State\textit{\#Weight's adaptation}
            \State $\text{$weight_{sum}$} \gets \text{sum}([a / (a + b) \text{ for } (a, b) \text{ in } \text{coefs}])$
            \State $\text{weights} \leftarrow []$
            \For{$c$ in $\text{coefs}$} \Comment{Weights' recalibration according to \textit{a} and \textit{b}}
                \State $\text{weights}[c] \gets (\text{coefs}[c][0] / (\text{coefs}[c][0] + \text{coefs}[c][1])) / \text{$weight_{sum}$}$
            \EndFor
            \For{$s$ in $\text{secs}$} \Comment{Sections' adjustment according to weights}
                \State $\text{secs}[s] \gets (\text{$init_{pos}$}, \text{round}(\text{$init_{pos}$} + \text{$exp_{ts}$} \cdot \text{weights}[s]))$
                \State $\text{$init_{pos}$} \gets \text{$init_{pos}$} + \text{round}(\text{$exp_{ts}$} \cdot \text{weights}[s])$
            \EndFor
    \State
    \State 
    \Return secs
        
\end{algorithmic}
\end{algorithm}

It is crucial to highlight that, during the policy exploitation phase, agents are deployed in segments as per the pre-defined intervals, operating continuously until the onset of the next recalibration phase. This operational mode promotes the calculation of time intervals $secs$. For example, and as stated previously, if $x$ time steps are set aside for an agent's turn (let's call this a section), that agent will keep going until those $x$ time steps are up. Once that time is done, the next agent in line takes over. One additional and relevant point is the algorithm's adaptability to unseen scenarios, as it seeks the most profitable combination of known policies in terms of rewards.

\section{Experimental evaluation}\label{sec-expeva}

In this section, extensive experiments are performed to validate the effectiveness of the proposed POW-dTS algorithm. Specifically, \autoref{sec-expsetup} introduces the trading environment and the baseline models used for comparison. \autoref{sec-results} presents the results and subsequent discussion. Finally, in \autoref{sec_ablation}, several ablation studies are conducted to analyze the contribution of each proposed module to the trading performance.

\subsection{Experimental setup}\label{sec-expsetup}

\begin{figure}[H]
    \centering
    \begin{subfigure}[b]{\textwidth}
        \includegraphics[width=0.95\textwidth, height=0.8\textheight, keepaspectratio]{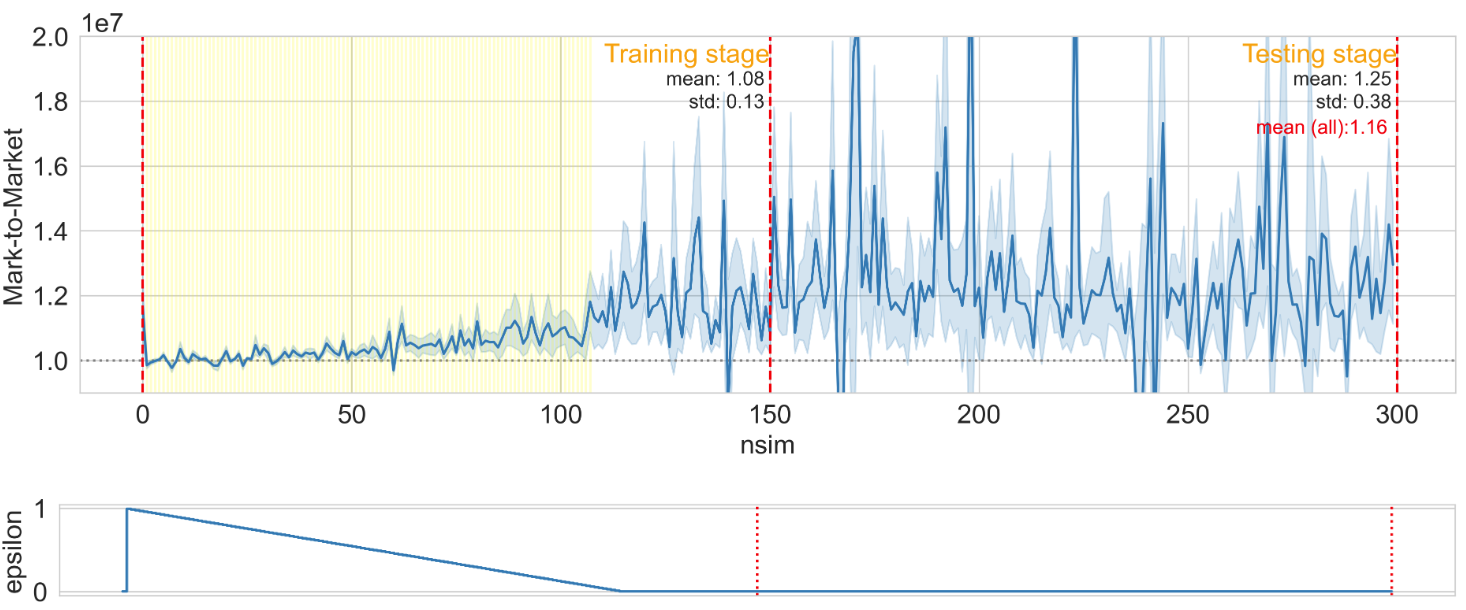}
        \caption{MtM of M$^3$ORL MM training with 0 MM competitors}
    \end{subfigure}
    
    \begin{subfigure}[b]{\textwidth}
        \includegraphics[width=0.95\textwidth, height=0.8\textheight, keepaspectratio]{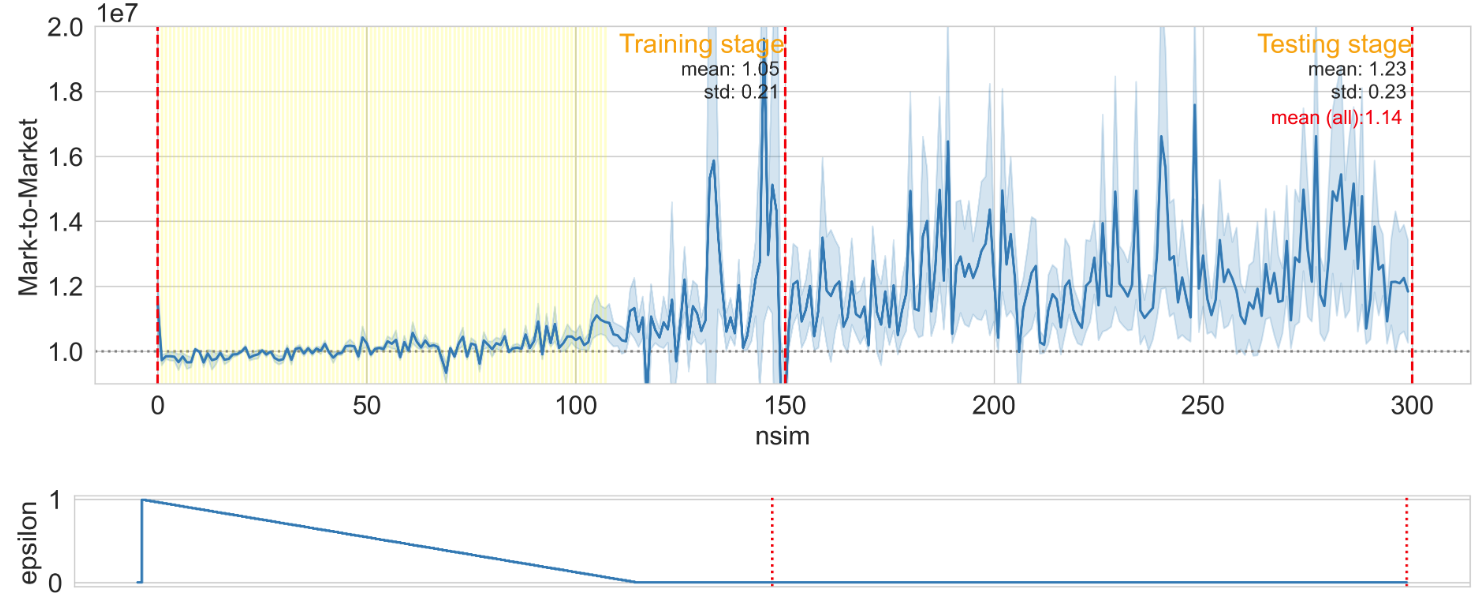}
        \caption{MtM of M$^3$ORL MM training with 1 MM competitors}
    \end{subfigure}
    \caption{MtM of M$^3$ORL MM training with 0 and with 1 MM competitors.}
\label{fig_alltrainings}

\end{figure}

\begin{figure}[H]
    
    \begin{subfigure}[b]{\textwidth}
        \includegraphics[width=0.95\textwidth, height=0.8\textheight, keepaspectratio]{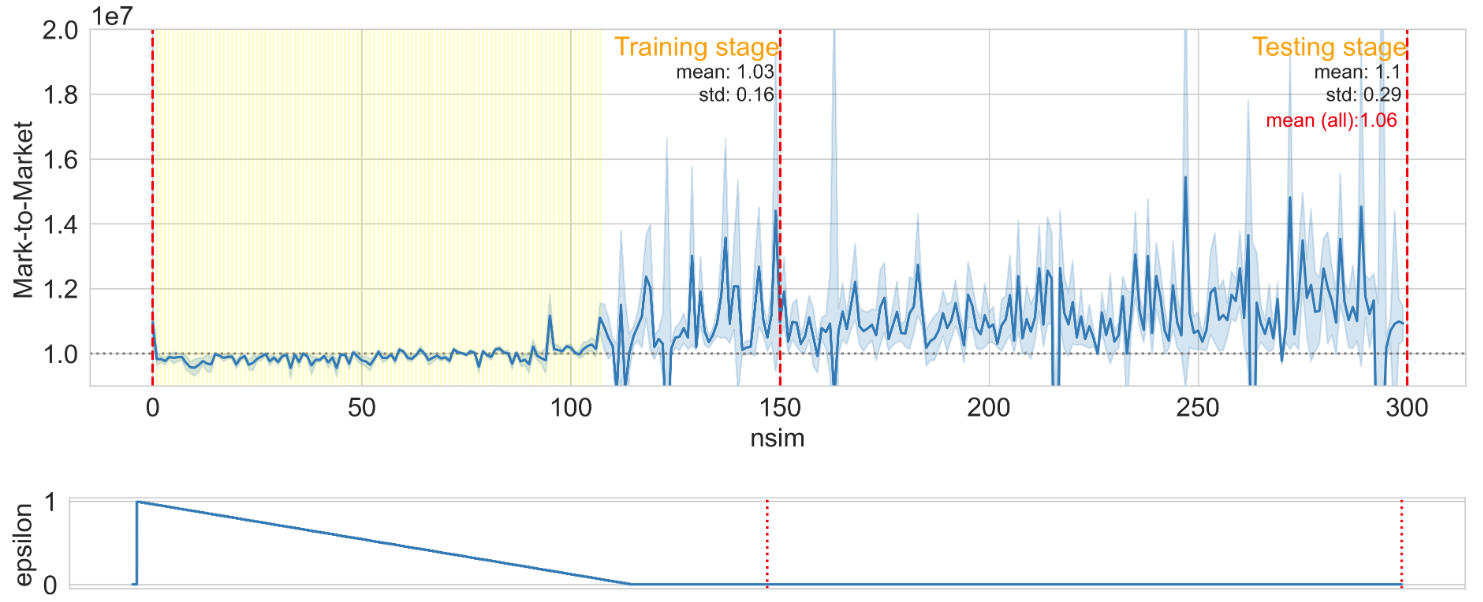}
        \caption{MtM of M$^3$ORL MM training with 5 MM competitors.}
    \end{subfigure}
    
    \begin{subfigure}[b]{\textwidth}
        \includegraphics[width=0.95\textwidth, height=0.8\textheight, keepaspectratio]{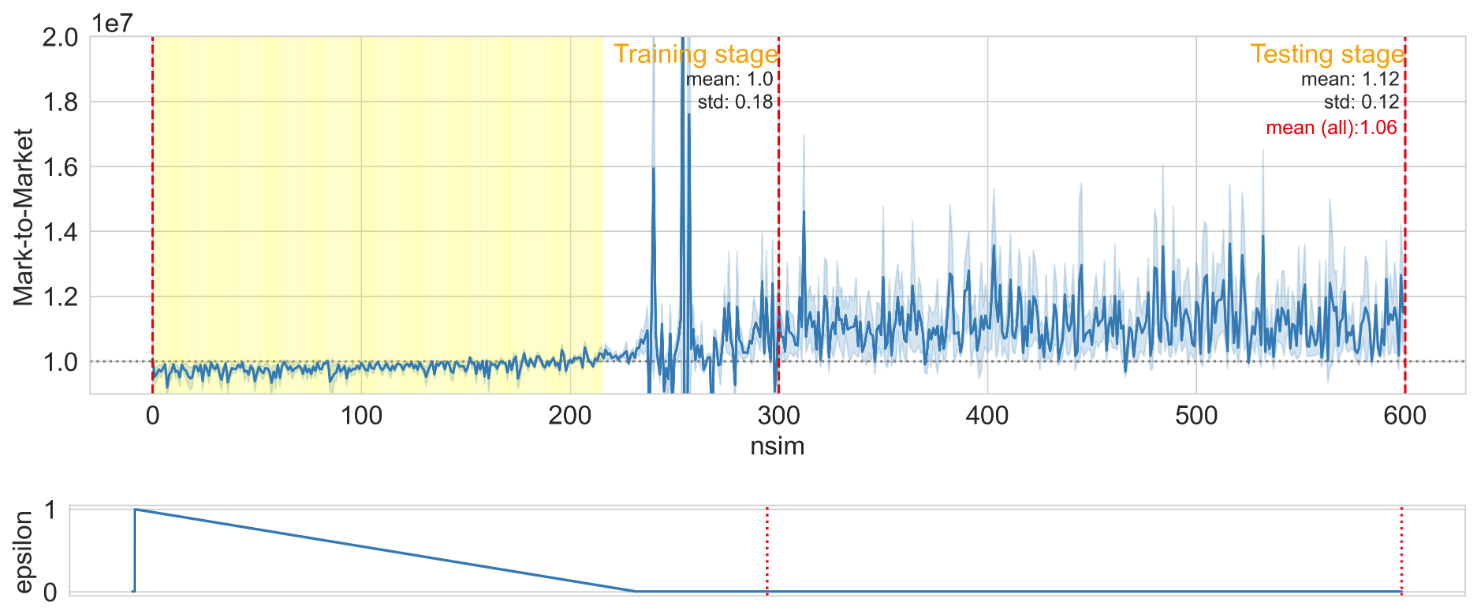}
        \caption{MtM of M$^3$ORL MM training with 7 MM competitors.}
    \end{subfigure}
    
    \caption[Optimal agents trained against 0, 1, 5, and 7 M$^3$ORL MM competitors. ]{Optimal agents trained against 0, 1, 5, and 7 M$^3$ORL MM competitors. The plot distinctly illustrates two phases: an initial training phase against $n$ additional competitors, followed by a testing phase where the agent acts greedily.}
    \label{fig_alltrainings2}
\end{figure}

 The experimental setup for the market simulation in ABIDES consisted of the following agents: 100 noise agents, 10 value agents, 10 momentum agents, 1 adaptive POV agent, and 1 exchange agent. Regarding MMs, the simulation incorporated 1 random MM, and 1 persistent MM agent, in addition to specific M$^3$ORL MM competitors, to gauge performance across different independent runs. Both perform random actions, but persistent MM keeps one random price all over a simulation. The experiments involve varying the number of M$^3$ORL MM competitors to introduce non-stationarity: 0, 1, 5, and 7, creating different contexts. Additionally, to define a set of pre-trained policies in various contexts and thus construct the library $A$ utilized as input by Algorithm \ref{algo_1}, there were trained four M$^3$ORL MM agents within each of these contextual settings: one in the absence of any RL competitors, another in the presence of a single competitor, a third in the presence of five competitors, and a fourth in the presence of seven competitors. These agents underwent training over 150 trading sessions each in the cases of 0, 1, and 5 competitors, and over 300 sessions in the case of 7 competitors due to increased complexity. All the agents were trained using an $\epsilon$-greedy strategy with epsilon decay. More details about this training stage can be found at \autoref{fig_alltrainings} and \autoref{fig_alltrainings2}.

\begin{figure}[H]
    \centering

        \includegraphics[width=1\textwidth]{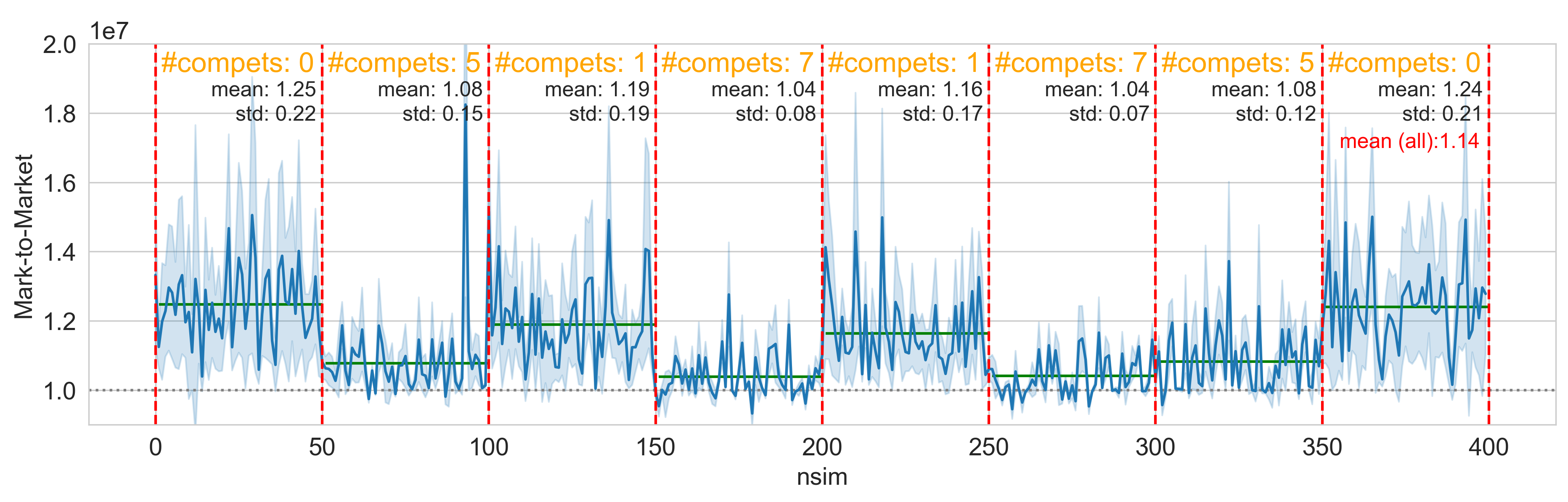}
        \caption{MtM of M$^3$ORL MM agent pre-trained with 0 MM competitors.}
        \label{fig_mtmbaseline}
\end{figure}

\begin{figure}[H]
        \includegraphics[width=1\textwidth]{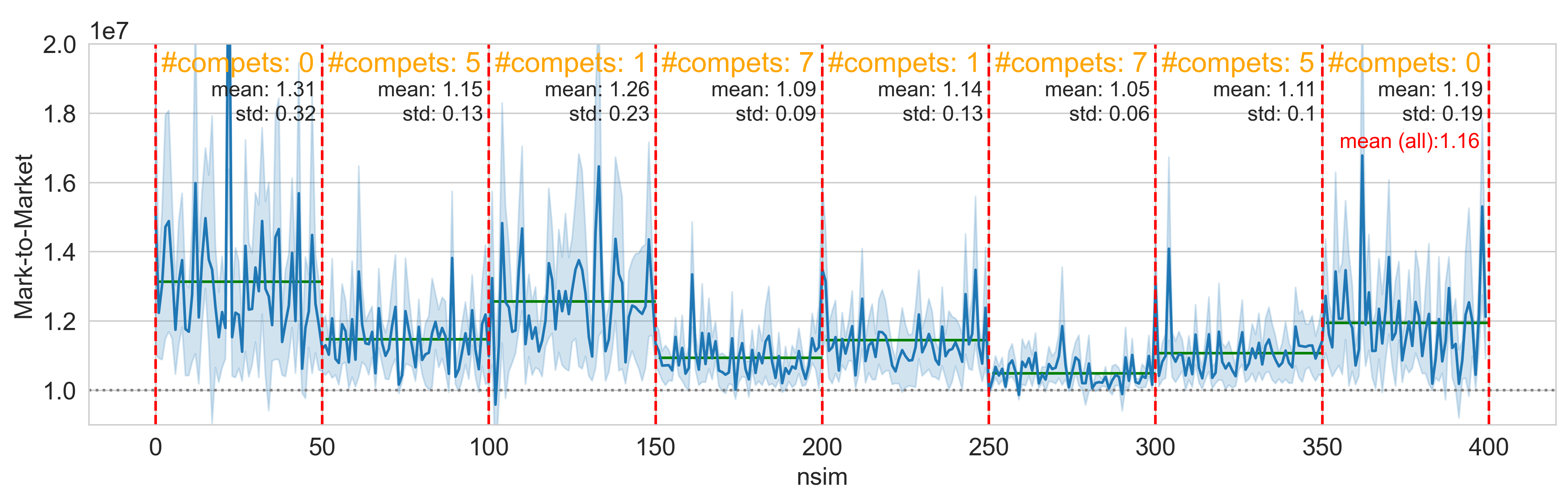}
        \caption{MtM of M$^3$ORL MM agent pre-trained with 1 MM competitor.}
        \label{fig_mtmbaseline1}
\end{figure}

\begin{figure}[H]
        \includegraphics[width=1\textwidth]{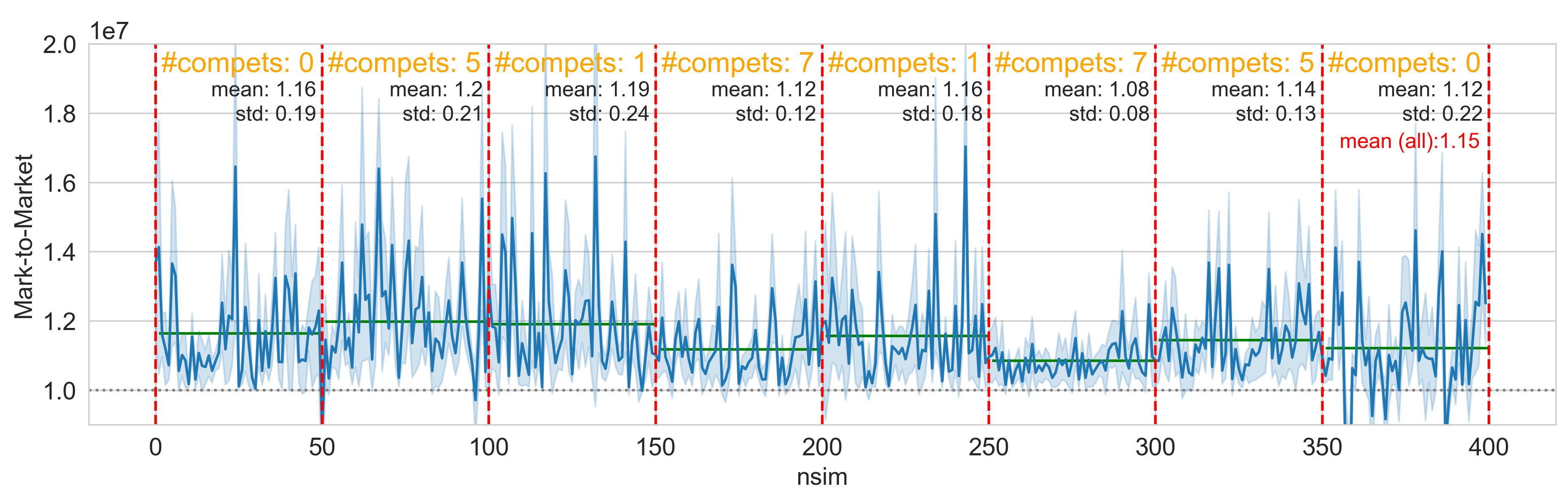}
        \caption{MtM of M$^3$ORL MM agent pre-trained with 5 MM competitors.}
        \label{fig_mtmbaseline5}
\end{figure}

\begin{figure}[H]
        \includegraphics[width=1\textwidth]{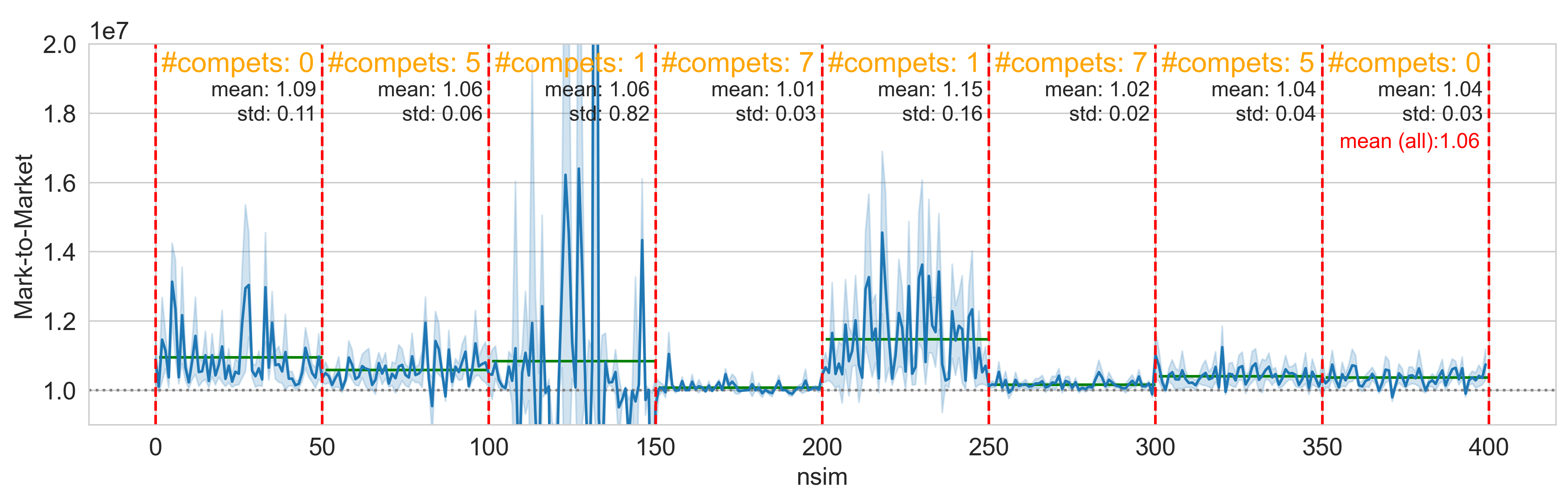}
        \caption{MtM of M$^3$ORL MM agent pre-trained with 7 MM competitors.}
        \label{fig_mtmbaseline7}
\end{figure}

Each complete experiment follows the sequence $seq$ of context changes with: 
\begin{align*}
seq = 0\rightarrow5\rightarrow1\rightarrow7\rightarrow1\rightarrow7\rightarrow5\rightarrow0, 
\end{align*}
where a different number of MM competitors are considered. 
Figures \ref{fig_mtmbaseline}-\ref{fig_mtmbaseline7} depict the evolution of each agent in library $A$ across each of the contexts in this sequence $seq$. Each context is evaluated over 50 or 80 trading sessions of 2 hours trading session each, based on the approach being tested. All the simulations started with an opening cash of \$$100.000,00$, spanning a two-hour session. The multi-goal reward is defined as 90\% of $R_1$ (MtM) with 10\% of $R_2$ (inv. control). 

The M$^3$ORL MMs trained without competitors and with one competitor yield similar results across the contexts as shown in Figures \ref{fig_mtmbaseline} and \ref{fig_mtmbaseline1}. Conversely, in Figure \ref{fig_mtmbaseline5} the M$^3$ORL MM trained against five competitors produces slightly better results than the others in competitive contexts. Finally, in Figure \ref{fig_mtmbaseline7} the M$^3$ORL MM agent trained against seven competitors faced an exceedingly hostile environment during learning, making it challenging to learn an appropriate behavioral policy, and thus exhibits the poorest evolution across contexts.

The experiments can be categorized into two distinct types: single-policy and multiple-policy methodologies. In single-policy approaches, an initial behavior policy is established, aiming to refine its effectiveness as the market dynamics evolve through contextual changes. In particular, the same M$^3$ORL MM agent is used, pre-trained with no competitors, and tries to adapt it to each new context. Each context runs for 50 rounds, or trading sessions, totaling 400 rounds. The comparison includes the following single-policy techniques:

\begin{itemize}
\item \textbf{Single-policy}: The agent is not retrained during the experiments and acts greedily in all contexts, applying only the pre-trained policy. The agent trained with no competitors is used as the \textbf{baseline} for the rest of the experiments. There are 3 additional single-policy versions according to the number of competitors the agent was trained with 1, 5, and 7. Figures \ref{fig_mtmbaseline}-\ref{fig_mtmbaseline7} represents one run of every different mentioned policy. The baseline agent corresponds to Figure \ref{fig_mtmbaseline}.
\item \textbf{CL-SingleP}: The agent mentioned as the baseline (single-policy with 0 competitors) is continuously retrained across the different contexts using only incoming experiences. This setup was tested with two different learning rates: $lr \in \{0.1, 0.05\}$.
\item \textbf{CL-FreezingL}: The technique of FL is used in this setup. The baseline agent has all the NN layers frozen (the parameters) except for the last one, which is updated with incoming experiences.
\item \textbf{CL-Rehearsal}: DR is applied here. The baseline agent is retrained across different contexts using both incoming and old experience replay buffers, maintaining a 50\% (old) / 50\% (new) replay-buffer ratio.
\item \textbf{CL-EWC}: EWC is applied in this setup with the baseline agent. The baseline agent's NNs' parameter updates are penalized based on the Fisher information matrix computed during the training stage. Three different $\lambda$ coefficients were tested: $\lambda \in \{0.5, 1, 10\}$. A higher value of $\lambda$ makes the network's parameters more resistant to change, thereby prioritizing the retention of what the network has previously learned, and vice-versa.
\end{itemize}

These single-policy techniques use only experiences gathered during greedy behavior, without any additional exploration. For comparison, adding 30 specific exploration rounds after every context change was also tested. This applies to CL-SingleP, CL-FreezingL, CL-Rehearsal, and CL-EWC setups. The motivation behind this was to evaluate if exploring other states could improve the agent's policy by applying the mentioned techniques. These experiments are denoted with an \textit{``exp''} suffix in figures and tables. In these alternative approaches, the total rounds per context are 80, comprising 30 rounds for additional exploration plus the 50 default exploitation rounds. Essentially, these exploration rounds consist of increasing the DQN epsilon to 1 for these 30 rounds, allowing the agent to explore different states and actions. It is significant to mention that these exploratory alternatives require some form of context change detection, which is not implemented in this work. Instead, for comparison, an `automatic context change detection is applied at the start of each new context.

Conversely, the multiple policy method involves the availability of a library containing diverse policies at the outset. Here, the objective shifts to the dynamic selection of the most suitable policy at any given moment, tailored to the prevailing market conditions. The following multi-policy techniques are included:

\begin{itemize}
\item \textbf{Multi-policy (Benchmark)}: This setup serves as a comparison benchmark and consists of including the best agent for each context from the beginning. This represents the best possible configuration, as all the pre-trained agents have the optimal policy for their context. It assumes the presence of a perfect change point detector and a precise agent's designator for each context.

\item \textbf{POW-dTS (ours)}:  The algorithm was tested with various $\gamma$ and $\alpha$/$\beta$ modifiers: $\gamma \in {0.4, 0.85}$ and $\alpha_{inc}, \beta_{inc} \in {1,5}$. Lower $\gamma$ values reduce the influence of older experiences, increasing forgiveness, whereas higher alpha and beta modifiers facilitate faster adaptation by the algorithm. Depending on the environment, it is important to select an appropriate $\gamma$ value. As previously noted, the algorithm employs four distinct pre-trained policies: for 0, 1, 5, and 7 M$^3$ORL MM competitors. POW-dTS was initialized with $rounds_{exp}$ = 3, indicating to launch a recalibration every 3 rounds, $rounds_{recal}$ = 150 setting the time steps per recalibration round, and $exp_{ts}$ = 750 that established the max time steps per exploitation round.
\item \textbf{Random agents}: More detailed in the ablation study \autoref{sec_ablation}, and similar to the POW-dTS setup, two alternative configurations were explored: \textit{Random Agent (Blocks)}, which assigns random weights to each block for the four agents; and \textit{Random Agent (Timesteps)}, where the concept of blocks is discarded in favor of random agent selection at every time step. Neither alternative incorporates dTS, only random sampling.
\end{itemize}

\subsection{Results}\label{sec-results}

\begin{figure}[H]
\centering
\includegraphics[width=1\textwidth]{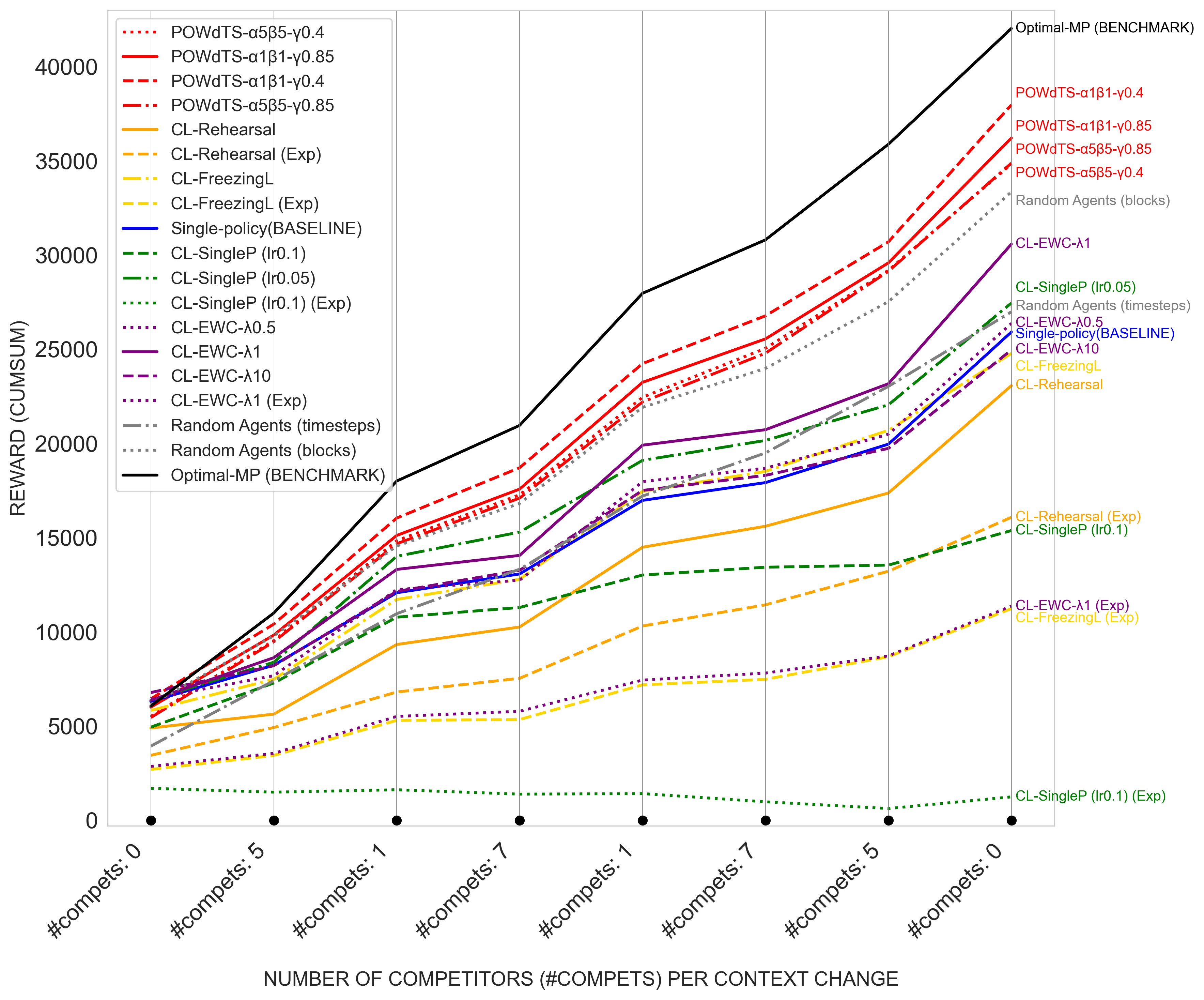}
\caption[Comparison of methods' performance in terms of accumulated mean rewards.]{Comparison of methods' performance in terms of accumulated mean rewards (5 seeds), along all the consecutive different contexts. The POW-dTS algorithm demonstrates relevant performance across each of the four configurations: $\alpha_{inc}$ and $\beta_{inc}$ coefficients, and $\gamma$.}\label{fig_comparisonplot}
\end{figure}

Table \ref{tab:results} presents the outcomes of all conducted experiments. The column labeled as \textit{Str.} indicates whether the algorithm uses only the baseline policy, which evolves according to different single-policy techniques (SP), or if the technique is based on utilizing all the pre-trained policies (MP). Two additional columns indicate whether the techniques require a change point detection mechanism (column labeled as \textit{CHDP}) and if the algorithm performs some kind of additional exploration when running (\textit{Expl.}). The table includes all the techniques used (\textit{Methods}), along with specific variants in learning rates, $\gamma$ values, or other parameters, showcasing the results. The results represent the mean values from five seeds per experiment, in terms of mean reward (\textit{Reward}), mean MtM (\textit{MtM}), and mean absolute inventory value held (\textit{\textbar Inv\textbar}). Conversely, Figure \ref{fig_comparisonplot} depicts the evolution of all the different methods throughout the experiments in terms of CUMSUM mean reward for five different seeds. This figure also illustrates how each technique performs when a new context is introduced. The best algorithms are less sensitive to very different contexts than the initial one (such as 5 and 7), while those with the worst performance suffer when transitioning to these situations.

\begin{table}[H]
\footnotesize
\begin{tabular}{@{}lccp{3.5cm}ccc@{}}
\toprule
\textbf{Str.} & \textbf{CHDP $^a$} & \textbf{Expl. $^b$} & \textbf{Methods} & \textbf{Reward$^c$} & \textbf{MtM$^c$} & \textbf{\textbar Inv\textbar$^c$} \tabularnewline
\midrule
\multirow{13}{*}{\textbf{SP}} & \multirow{3}{*}{} & & \textbf{Single-Policy} \tabularnewline
                                    & & & \quad 0 compets (Baseline)                             & 3240 $\pm$ 114 & 1.14 $\pm$ 0.01 & 323 $\pm$ 6  \tabularnewline
                                    
                                    & & & \quad 1 compets                             & 3132 $\pm$ 831 & 1.14 $\pm$ 0.02 & 502 $\pm$ 337  \tabularnewline
                                    & & & \quad 5 compets                             & 3267 $\pm$ 335 & 1.14 $\pm$ 0.02 & 531 $\pm$ 148  \tabularnewline
                                    & & & \quad 7 compets                             & 1531 $\pm$ 583 & 1.07 $\pm$ 0.02 & 884 $\pm$ 641  \tabularnewline
                               & & & \textbf{CL-SingleP} & & & \tabularnewline
                               & & & \quad lr = 0.1                 & 1923 $\pm$ 617 & 1.09 $\pm$ 0.03 & 466 $\pm$ 86 \tabularnewline
                               & & & \quad lr = 0.05               & 3432 $\pm$ 1007 & 1.15 $\pm$ 0.04 & 428 $\pm$ 109 \tabularnewline
                               & & & \textbf{CL-FreezingL}               & 3098 $\pm$ 714 & 1.13 $\pm$ 0.03 & 346 $\pm$ 63 \tabularnewline
                               & & & \textbf{CL-Rehearsal}               & 2884 $\pm$ 156 & 1.15 $\pm$ 0.01 & 723 $\pm$ 80 \tabularnewline
                               & & & \textbf{CL-EWC} & & & \tabularnewline
                               & & & \quad $\lambda =$ 0.5               & 3297 $\pm$ 266 & 1.14 $\pm$ 0.01 & 381 $\pm$ 63 \tabularnewline
                               & & & \quad $\lambda =$ 1                & 3823 $\pm$ 292 & 1.17 $\pm$ 0.01 & 414 $\pm$ 24 \tabularnewline
                               & & & \quad $\lambda =$ 10               & 3120 $\pm$ 260 & 1.14 $\pm$ 0.01 & 371 $\pm$ 42 \tabularnewline
                               & \multirow{4}{*}{} \checkmark & \checkmark & \textbf{CL-SingleP (Exp)}           & 156 $\pm$ 229 & 1.01 $\pm$ 0.01 & 268 $\pm$ 274 \tabularnewline
                               & \checkmark & \checkmark & \textbf{CL-FreezingL (Exp)}         & 1405 $\pm$ 381 & 1.06 $\pm$ 0.02 & 240 $\pm$ 37 \tabularnewline
                               & \checkmark & \checkmark & \textbf{CL-Rehearsal (Exp)}         & 2011 $\pm$ 282 & 1.08 $\pm$ 0.01 & 313 $\pm$ 67 \tabularnewline
                               & \checkmark & \checkmark & \textbf{CL-EWC (Exp)}               & 1421 $\pm$ 240 & 1.06 $\pm$ 0.01 & 222 $\pm$ 33 \tabularnewline
\midrule
\multirow{8}{*}{\textbf{MP}}  &  & & \textbf{POW-dTS (ours)} & & & \tabularnewline
                               &  &   & \quad $\alpha_{inc}$=5  $\beta_{inc}$=5  $\gamma$=0.4  & 4354 $\pm$ 391 & 1.19 $\pm$ 0.02 & 483 $\pm$ 62 \tabularnewline
                               &  &   & \quad $\alpha_{inc}$=1  $\beta_{inc}$=1  $\gamma$=0.85   & 4527 $\pm$ 135 & 1.2 $\pm$ 0.01 & 428 $\pm$ 19 \tabularnewline
                               &  &   & \quad \textbf{\boldmath{$\alpha_{inc}$=1  $\beta_{inc}$=1  $\gamma$=0.4}}    & \bfseries 4747 $\pm$ \bfseries 441 & \bfseries 1.21 $\pm$ 0.02 & \bfseries 448 $\pm$ 55 \tabularnewline
                               & &   & \quad $\alpha_{inc}$=5  $\beta_{inc}$=5  $\gamma$=0.85   & 4360 $\pm$ 305 & 1.19 $\pm$ 0.02 & 479 $\pm$ 29 \tabularnewline
                               &  & & \textbf{Rand. Agents (ablation)} & & & \tabularnewline
                               &  &   & \quad blocks   & 4170 $\pm$ 166 & 1.18 $\pm$ 0.01 & 453 $\pm$ 9 \tabularnewline
                               &  &   & \quad time steps   & 3374 $\pm$ 41 & 1.15 $\pm$ 0.0 & 459 $\pm$ 8 \tabularnewline
                               & \checkmark &  & \textbf{\textit{Benchmark: Optimal-MP}}      & \textit{5254 $\pm$ 100} &  \textit{1.22 $\pm$ 0.01} &  \textit{450 $\pm$ 8 }\tabularnewline
\bottomrule

\end{tabular}
\footnotesize{$^a$ Change point detection needed, $^b$ It perform some exploration rounds. $^c$ Average values.}
\caption[Results of non-stationary M$^3$ORL MM strategies in terms of rewards, MtM, and absolute inventory value. ]{Results of non-stationary M$^3$ORL MM strategies in terms of rewards, MtM, and absolute inventory value. POW-dTS obtain the best results in reward and MtM (bold) after optimal benchmark (italic).}
\label{tab:results}
\end{table}

Based on the results presented in Table \ref{tab:results} and Figure \ref{fig_comparisonplot}, it is evident that POW-dTS outperforms the other options by a significant margin. It nearly matches the ideal benchmark, which represents the best possible outcome aimed to achieve. This observation is crucial because the benchmark alternative relies on the following assumptions: firstly, a flawless change point detector that identifies context changes perfectly from the beginning; secondly, an impeccable policy mapping algorithm that assigns the best policy to each new context starting from the first time step. Our POW-dTS experiments involved four different setups, varying parameters such as $\gamma$ in the range $\{0.4, 0.85\}$ and $\alpha_{inc}, \beta_{inc}$ within $\{1,5\}$. These setups exhibited robust behavior, as indicated by the results. As mentioned earlier, lower values of $\gamma$ result in a more forgiving approach to older data, while higher values of $\alpha_{inc}$ and $\beta_{inc}$ amplify the reinforcement of successes and failures, respectively. The most successful outcomes were associated with $\alpha_{inc}, \beta_{inc} = 1$, suggesting that smoother adjustments of the beta distribution parameters are preferable in the current environment. 

\begin{figure}[H]
\centering
\includegraphics[width=1\textwidth]{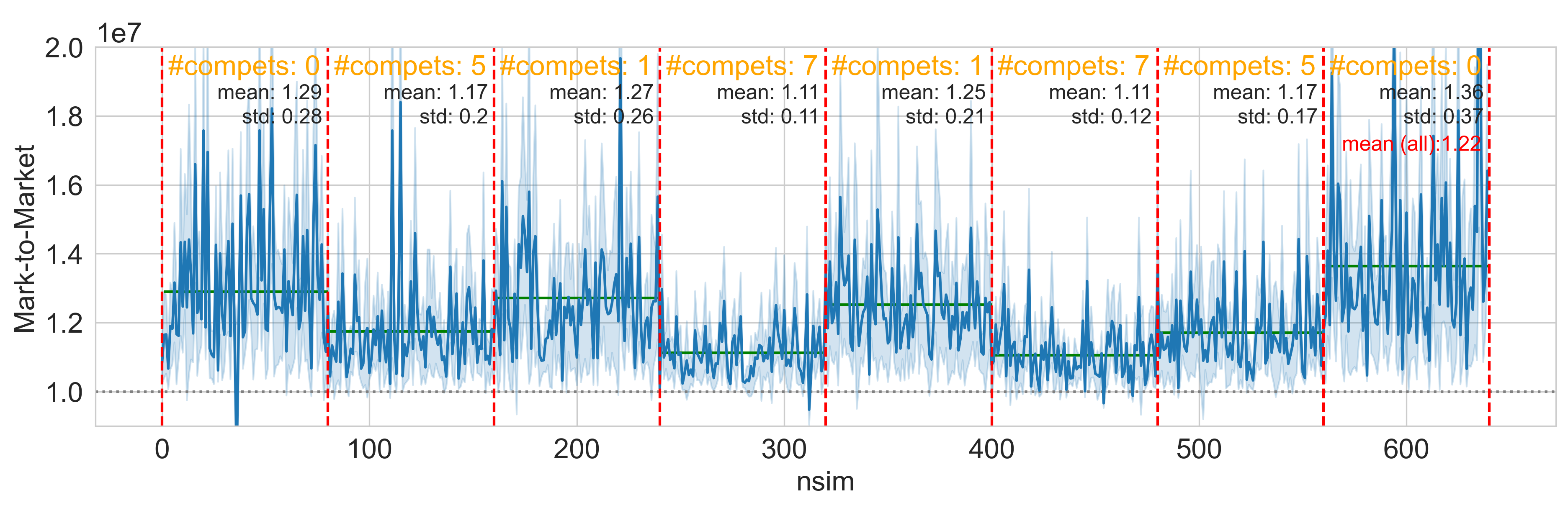}
\caption[Performance image of one experiment using POW-dTS with $\alpha_{inc} = 1$, $\beta_{inc} = 1$, and $\gamma = 0.4$ in terms of MtM.]{The picture illustrates the strong performance of the POW-dTS algorithm in terms of MtM across various scenarios. The average value (mean (all): 1.22) is close to the average obtained by optimal-MP. Parameters used: $\alpha_{inc} = 1$, $\beta_{inc} = 1$, and $\gamma = 0.4$.}\label{fig_powdts_round}
\end{figure}

\autoref{fig_powdts_round} depicts one round performed by POW-dTS in terms of MtM, using $\alpha_{inc}$, $\beta_{inc} = 1$, and $\gamma = 0.4$. It is worth noting that while these parameters worked well in our case, they might need adaptation for situations with different levels of volatility.

\begin{figure}[H]
\centering
\includegraphics[width=1\textwidth]{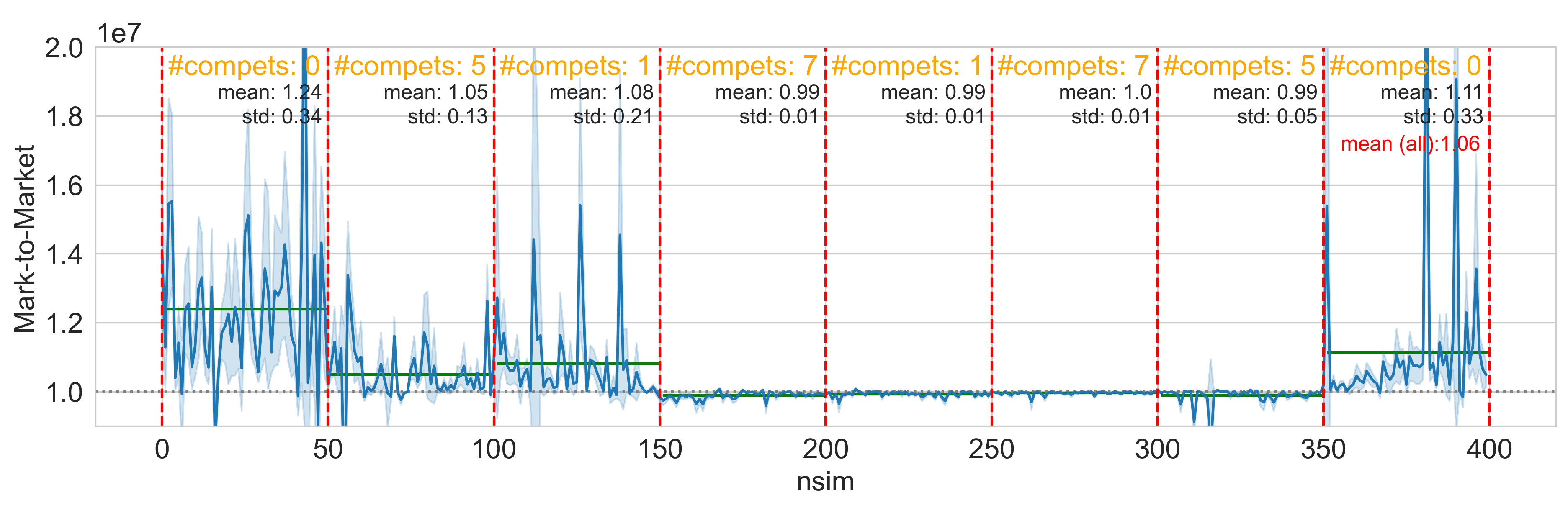}
\caption[Impact of a vanilla CL strategy of the baseline agent]{The picture clearly illustrates the impact of a vanilla CL strategy of the baseline agent with a learning rate of lr=0.1, in terms of MtM (CL-SingleP experiment). This effect, clearly noticeable in the sequence of contexts 7-1-7-5, is commonly referred to as catastrophic forgetting. This is due to the loss of previously acquired information in the NN.}\label{fig_mtm_cf}
\end{figure}

Regarding the CL-SingleP strategy, it is observed that using the default learning rate of lr = 0.1 leads to inferior results, primarily due to significant catastrophic forgetting, as depicted in Figure \ref{fig_mtm_cf}. The effect of new data on the agent's NNs, significantly degrading its performance, is evident. However, employing a lower learning rate, as in CL-SingleP (lr=0.05), mitigates this adverse effect, allowing the policy to surpass the baseline. This strategy, previously discussed in \autoref{back-cl}, presents an interesting approach to circumvent this issue.

Another well-performing technique, according to the results, is EWC, moreover when applying a $\lambda$ = 1. However, this method's effectiveness is sensitive to the precise selection of this lambda parameter, as shown by the lower performance under other values.

\begin{figure}[H]
\centering
\includegraphics[width=1\textwidth]{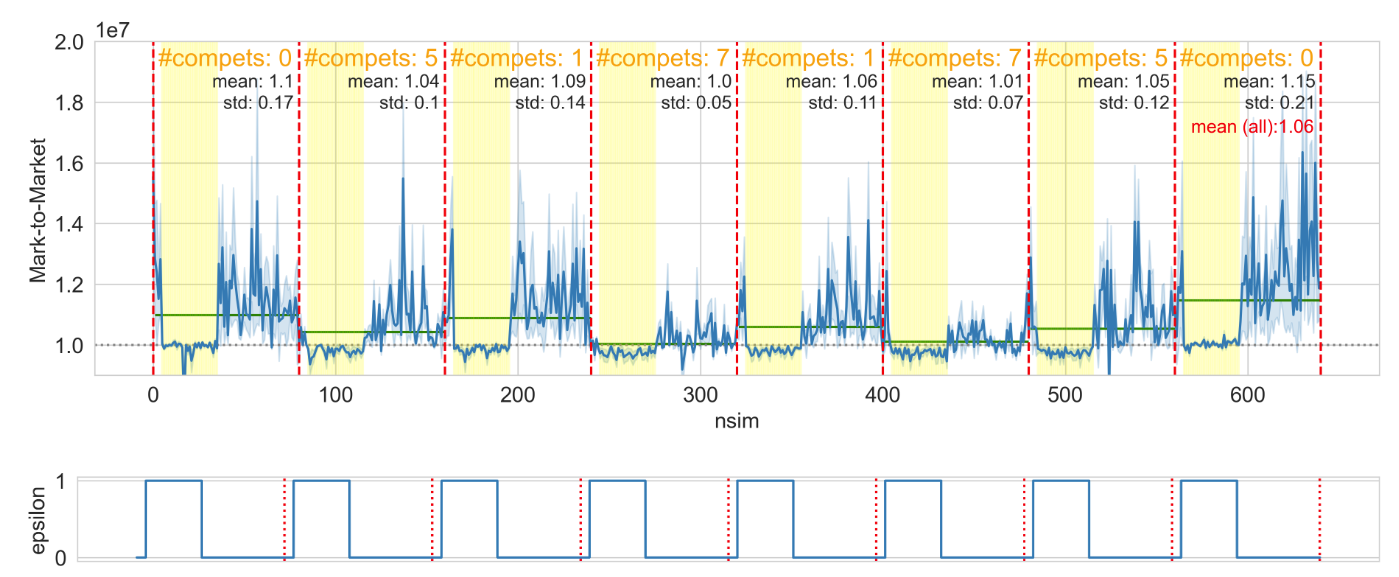}
\caption[CL-EWC-$\lambda$1 (exp) experiment.]{The CL-EWC-$\lambda$1 (exp) experiment's figure shows the progression of MtM when employing an EWC strategy with $\lambda = 1$. The exploration rounds per context, marked in yellow, indicate a decline in returns due to this exploration phase. No substantial improvement is observed after these rounds. Additionally, the plot below displays the $\epsilon$ value of the epsilon-greedy strategy executed by the MM.}\label{fig_mtm_exploration}
\end{figure}

It is also crucial to highlight the noticeable impact of exploration rounds on the overall reward for methods that include this action, attributed partially to the decreased returns during these exploration phases. This effect is depicted in Figure \ref{fig_mtm_exploration}, demonstrating the anticipated impact of exploration rounds on performance. These additional rounds did not substantially improve the policy to warrant their inclusion.

Next, the \textit{pros and cons} observed with the POW-dTS algorithm are outlined:
\textbf{Among its notable strengths}, the algorithm demonstrates robust adaptability to unseen contexts by leveraging knowledge from familiar ones, securing significantly high and optimal returns when compared to other methods. Moreover, it stands out as a computationally affordable method, offering quick adjustments through TS. Importantly, it operates independently of change points or known context transition probabilities/patterns, which positively affects computational resources while minimizing additional errors. \textbf{Conversely, on the downside}, it necessitates the periodic evaluation of each agent's performance, which could affect overall returns if executed too frequently. Finding a balance between exploitation and updating phases is critical, varying by the specific environment. Additionally, precise calibration of $\alpha_{inc}$ and $\beta_{inc}$ values, along with the $\gamma$ factor, is essential, as incorrect parameter settings may negatively impact returns. Another critical point is POW-dTS's reliance on pre-trained policies, necessitating prior access to these contexts for effective agent pre-training. This last aspect makes the method less suitable for scenarios encompassing a wide array of contexts.

\subsection{Ablation study}\label{sec_ablation}

To analyze the impact of the different components within POW-dTS, 
two additional experiments were conducted: one retaining the blocks but omitting the dTS part, termed \textit{Rand. Agents (blocks)}, and another performing random selection of agents at every time step, named \textit{Rand. Agents (timesteps)}. It is immediately evident that both variants outperform the baseline, suggesting that policy weighting alone is a potent approach in such types of environments. Moreover, the exceptionally good performance of random agents operating in blocks is particularly noteworthy. Alternating between different policies while allowing them to work sequentially significantly enhances performance. This is coherent with the intrinsic characteristics of RL algorithms, which consider not only immediate rewards but also future ones. This configuration allows them to exploit certain trajectories, an advantage that the time-step-based alternative lacks. Nonetheless, dTS algorithm shines when it comes to adjusting the weights according to the agent's performance, providing a substantial boost to the solution. The four distinct parameter configurations tested, with $\alpha_{inc}$ and $\beta_{inc}$ each set to either 1 or 5, and $\gamma$ set to either 0.4 or 0.85, have yielded significant results. 

\section{Discussion} \label{sec-conclusion}

Dealing with non-stationarity is not a trivial task. Financial markets are complex environments where stationarity is not usually present. In this work, a novel approach to address this challenge, specifically within the multi-objective market making problem, has been proposed. Based on experimental results, POW-dTS has proven to be very robust in these types of tasks. Starting with a set of policies trained differently, POW-dTS can assign weights to them to create composite policies that approximate the optimal scenario. A key advantage of this method is its independence from identifying change points or mapping transition probabilities. This is particularly beneficial as it avoids potential errors introduced by these techniques and proves advantageous when environmental shifts are subtle—scenarios where change point detectors may fail. Not to mention the savings in computational resources typically required by these methods. Additionally, since POW-dTS relies on a composite policy, it adapts to unknown contexts by selecting the most effective combination of known policies. This aspect is crucial because known and stable environments that remain unchanged over time are seldom encountered. It is more realistic to expect a dynamic environment that evolves. This constant change challenges the common assumption of context stability required by many algorithms that depend on change point detection or transition probabilities.

For future work, it is confident that this method can be applied in other domains where non-stationarity is a significant factor. Notably, heating, ventilation, and air conditioning systems (HVAC), which require dynamic adjustment to environmental changes and occupancy patterns, present a promising area for applying our approach. Similarly, energy trading strategies, where market conditions fluctuate rapidly, could greatly benefit from the adaptability our method offers. Beyond these examples, sectors such as autonomous vehicles, where conditions change moment-to-moment and adaptive web systems, which must adjust to user behavior in real time, could also see substantial improvements in efficiency and effectiveness. Investigating these applications further could not only validate the versatility of this method but also contribute to advancements in these fields.

\section{Contribution}\label{contrib_stationarity}

The main contribution of this chapter is addressing the non-stationarity problem in financial markets faced by MMs through the development of a new algorithm, Policy Weighting through discounted Thompson sampling (POW-dTS). The key aspects of this contribution are enumerated as follows:

\begin{enumerate}
    \item \textbf{Innovative algorithm design:} Development of POW-dTS, which combines a dynamic form of classical TS with policy weighting. This adaptation allows the algorithm to adjust dynamically to the evolving conditions of the markets without requiring artifacts like change point detectors or prior knowledge of the environment’s dynamics, such as transition probabilities or reward functions.
    \item \textbf{Comparative analysis with classical techniques:} exploration and comparison of classical training techniques for non-stationary environments. This involves assessing the performance of methods such as \text{EWC}, \text{FL}, and \text{DR}, and comparing these with the proposed method.
    \item \textbf{Ablation study:} Inclusion of an ablation study to evaluate the crucial components of the proposed algorithm, helping to highlight the contributions of individual elements to the overall effectiveness of the approach.
    \item \textbf{Evaluation of catastrophic forgetting:} Examination of the impact of CF within these environments, providing insights into how significantly this phenomenon affects the stability and reliability of learning algorithms in non-stationary settings.
    \item \textbf{Broader applicability:} While the primary objective is to present a solution for multi-objective market making within non-stationary markets, the developed methodology also has potential applicability to a broader spectrum of tasks requiring flexible and adaptive strategies in environments where non-stationarity is prevalent.
\end{enumerate}

These points collectively underscore the novel approach taken to solve the non-stationarity problem in financial markets. While the primary objective is to present a solution for multi-objective market making within non-stationary markets, the methodology developed has potential applicability to a broader spectrum of tasks that need flexible and adaptive strategies, where non-stationarity is present.

\chapter{Conclusion}\label{chap_conclusion}

The dissertation comprehensively addresses the multifaceted challenges of market making through the innovative application of RL, exploring the evolution from a single-agent model focused on profitability to a fully autonomous multi-policy system capable of adapting dynamically to complex and non-stationary financial environments.

\textbf{Starting with establishing RL in market making:}
The journey begins with the foundational exploration of applying deep Q-learning to simulate market making scenarios. This initial approach establishes the viability of RL to inject liquidity and ensure profitability in simulated stock markets. The focus here is to demonstrate the capacity of RL agents to learn and adapt strategies, even with additional competitors, without human intervention, setting a precedent for the subsequent, more complex investigations. And not only this but also to understand the learned strategies and how these agents perform when competing against other RL and non-RL MMs.

\textbf{Advancing to sophisticated inventory management:}
With the foundation established, the journey progresses to address one of the most significant challenges in market making: inventory management. At this stage, the dissertation introduces two different approaches. The first strategy, based on reward engineering, represents a major leap from traditional static models, offering enhanced flexibility and risk mitigation. This first development implements two dynamic factors in the reward function: the \textit{Alpha Inventory Impact Factor} (AIIF) and the \textit{Dynamic Inventory Threshold Factor} (DITF). These factors allow the MM to adapt to the evolving situation of its assets, enabling it to manage inventory risk more dynamically by adjusting the ratio based on the cash-to-inventory value ratio. This approach is compared to other published alternatives, showcasing better performance in terms of inventory control, profitability, and risk management.

The second and more advanced approach to dealing with inventory control was based on the use of a MORL framework. Instead of aggregating both goals (profitability and inventory control) into one complex reward function, an RL MM with two pairs of NNs is designed. In this stage, the use of PFs comes into play, offering a sophisticated tool to navigate these trade-offs. This stage of the journey not only highlights the robustness of multi-objective RL frameworks but also enhances the decision-making process by providing clear, quantifiable multi-objective metrics for evaluating different strategies. It is demonstrated how this strategy is more robust and stable, due mainly to the specialization of the agent on both tasks simultaneously. This phase strengthens the agent's ability to operate in a complex decision environment, preparing it for full autonomy.

\textbf{Culminating in autonomous adaptation:}
The final leg of the journey brings the most ambitious and innovative contribution: developing an algorithm capable of autonomously adapting to non-stationary market conditions. This stage introduces the POW-dTS algorithm, which integrates the TS algorithm with policy weighting. This advancement allows the market making agent to continuously adjust to changing market dynamics with only a set of different pre-trained policies, demonstrating exceptional adaptability and learning.

Regarding the objectives presented in \autoref{sec-goals}, it can be asserted that they have been successfully accomplished. Throughout this dissertation, it has been demonstrated that RL is a suitable tool for creating market making agents capable of operating in competitive scenarios. Market making agents have been trained not only to evaluate performance in non-competitive scenarios but also to explore the impact of competitors in terms of MtM and inventory control. Even when applying transfer learning with pre-trained policies in different environments, their effectiveness has been validated and relevant insights on policy performance have been extracted. This clearly validates the \textbf{first objective}, \textit{evaluate RL for profitable market making in competitive scenarios}.

For the \textbf{second objective}, which involves managing inventory while remaining profitable, two different techniques were proposed. First, a reward-engineered agent capable of dynamically managing inventory based on the cash/inventory value ratio. This approach was compared to other reward functions from relevant works, achieving better results in both profitability and inventory management. The dynamic nature of the reward function, which adapts to the ongoing situation of the MM, allows it to be more aggressive at certain moments without increasing risk. The second approach turned the agent into a M$^3$ORL MM by separating the reward function into a vector of sub-goals instead of a single scalar. This approach enabled the agent to learn faster while avoiding the issues associated with reward engineering. Furthermore, it outperformed the reward-engineered option in various multi-objective metrics, demonstrating the power of this alternative solution.

Finally, regarding the \textbf{third objective} of dealing with non-stationarity, a novel algorithm called POW-dTS is proposed. This algorithm enables the previous MORL agent to handle changes in market patterns. By pre-training a set of policies for different situations, the algorithm modulates each one to best fit the current market. This algorithm was also compared to different CL techniques, achieving better performance than those. Additionally, this algorithm can be applied to other use cases where non-stationarity is present.

To sum up, each stage of the dissertation not only addresses increasingly complex aspects of market making but also contributes to a holistic understanding of how RL can be effectively applied in financial markets. The journey—from establishing basic RL applications in market making to developing autonomous agents capable of adapting to dynamic environments—illustrates a significant evolution in computational finance. This academic research not only paves the way for future studies but also promises substantial implications for practical applications in finance.

\section{Limitations of the presented research}

In this work, a method for addressing the market making problem from a reinforcement learning perspective has been presented. However, it is crucial to highlight the limitations of this approach when considering its application to real markets. Several challenges must be addressed before these techniques can be effectively implemented in a real-world trading environment.

All the research has been conducted using ABIDES, a financial market simulator based on agents' interaction. It offers numerous advantages, such as controlled environments and the ability to test multiple scenarios without financial risk. However, the policies trained in simulators often face significant challenges when transferred to the real world. Firstly, before applying an agent to a specific financial asset, it is necessary to develop a simulator that closely mirrors the real asset in terms of spreads, volatility, returns, and other technical indicators. The closer the simulator is to reality, the more accurate the policies learned by the agent will be. However, despite efforts to emulate these dynamics, the stochastic nature of financial markets makes them difficult to replicate with high fidelity. In other fields, the dynamics can often be derived from known effects, such as physical equations or other well-defined processes, leading to highly accurate simulations. In contrast, in finance, we face the dual challenge of replicating a stochastic environment and validating this simulation against real market data. Moreover, markets evolve, with their dynamics changing in response to various factors, including economic conditions, geopolitical events, and investor sentiment. This necessitates continuous monitoring of the simulator’s performance and regular fine-tuning to adapt to these changes, adding another layer of complexity to the process.

Therefore, the use of simulators to learn trading strategies should be complemented with other techniques, such as combining real and simulated experiences, using domain randomization techniques \cite{tobin2017domain} to encourage generalization, or conducting online training directly in the market. Even when a simulator is validated against a specific asset, there will always be some discrepancy between simulation and reality. This is where the concept of Sim2Real \cite{10.1109/ICRA.2018.8460528,hofer2021sim2real}, short for ``simulation to reality'' becomes relevant. Sim2Real focuses on the transfer of skills and models trained in simulation to the real world. It addresses the challenge of bridging the gap between simulated environments and real-world conditions with different techniques, which is an essential step for deploying effective trading strategies.

An alternative approach to overcoming the limitations of simulation-based training is to train RL agents directly in the market using online learning. While this method circumvents some of the issues associated with simulators, it introduces new challenges. Online training has two main disadvantages compared to simulation: \textbf{the speed of training and the associated costs}. In simulators, training times can be significantly reduced, as they are limited only by computational power and the efficiency of the simulation environment. In contrast, real-world training relies on the availability of data in real-time, which imposes a clear constraint on training speed. Regarding costs, training agents directly in the market incurs various expenses, such as transaction fees, slippage, and the cost of operational errors. Depending on the strategy being developed, these costs can range from manageable to prohibitive. For high-frequency or complex strategies, the financial impact of errors or suboptimal decisions during training can be substantial.

To summarize, while RL-based methods hold promise for market making and other financial applications, their deployment in real markets requires careful consideration of the limitations of simulation, the need for robust validation techniques, and the practical challenges of online learning. A combination of simulation, real-world testing, and advanced adaptation strategies is essential for developing effective and resilient trading systems.

\section{Future work}

This section highlights several key areas that are worth further exploration as part of future work. The first \autoref{fut:challenges} delves into strategies for addressing the limitations of the current method, offering potential alternatives and improvements to enhance its performance. Next, \autoref{fut:llms} explores the role of large language models (LLMs) in solving the MM problem,  showcasing their potential not only in this specific context but also in broader applications within trading.

\subsection{Future directions in simulated and real-world market making}\label{fut:challenges}

The foundation of a robust approach to market making through the use of RL has been established. Although all research thus far has utilized a market simulator, significant work remains before these methods can be applied in the real world. If the goal is to pre-train a MM agent in a simulated stock market before transferring the learning to the real world, it is crucial to replicate a similar simulated environment. Therefore, developing methods that assist in the \textbf{creation and validation} of these simulators is a critical line of research. Ensuring that the simulator closely resembles the asset intended for operation is essential. Designing agents and similarity metrics that align with real price movements is a complex task, but it is necessary for effective MM training in a simulated environment. Additionally, \textbf{bridging the gap between simulated and real environments} (Sim2Real) is a key requirement. Exploring different techniques to address this challenge is an important step before deploying the agent in the real world \cite{10.1145/3533271.3561755, 10.1145/3383455.3422561}.

Conversely, if the option is to bypass simulators and directly \textbf{learn policies in the real environment}, this approach typically incurs higher costs, in addition to increased training times. For online training, a more cautious learning policy must be designed to minimize potential financial losses while still enabling effective learning. Therefore, safe RL learning techniques \cite{JMLR:v16:garcia15a} should be explored or proposed to allow the agent to learn directly in real markets in a safe manner. This could undoubtedly be an additional line of research.

Regarding other algorithms, other lines of work can also be pursued. In the area of dealing with non-stationarity, some interesting algorithms can be tested and compared to our solution. One similar approach presented in this thesis is called \textit{Probabilistic Policy Reuse in a RL Agent}~\cite{10.1145/1160633.1160762}, where the authors select the current policy of the agent according to a probabilistic method. This method shares some common points with our approach, and it would be interesting to compare it with POW-dTS in terms of performance. Additionally, there are other methods not tested, which rely on changepoint detectors, such as \cite{Padakandla2019}. Changepoint detectors, such as ODCP\footnote{Online Parametric Dirichlet Changepoint} or ECP\footnote{E-Divisive Change Point detection}, which have to be correctly tuned, introduce uncertainty into the algorithm, as they need to be robustly trained to avoid false positives and false negatives, especially when changes in the environment are not abrupt. However, this could be an interesting line of work, not only related to testing these types of algorithms but also to designing robust changepoint detectors, which is a matter of research itself.

\subsection{LLMs and their application in trading strategies}\label{fut:llms}

Large Language Models (LLMs)~\cite{NIPS2017_3f5ee243} have undoubtedly opened a new universe of possibilities in machine learning. Their ability to process vast amounts of information and extract relationships from data has enabled new capabilities across various domains. Designing trading and, therefore, market making strategies that benefit from these capabilities is an important area of research and deserves a brief analysis. In this regard, two main groups of applications can be explored: LLMs integrated with RL and LLMs apart from RL. Regarding LLMs integrated with RL, the combination of RL and LLMs can enhance agents in many ways. For instance, LLMs can improve the state space by incorporating new information directly from their outputs. The order book can be tokenized as a sequence of text, allowing the LLM to extract relevant information, or other types of unstructured data—such as news, events, tweets, and sentiment—can be incorporated. Furthermore, multimodal LLMs~\cite{10386743} (models that accept not only text but also images or other types of inputs) can be used to include visual elements like order book snapshots, candlestick patterns, or other technical information, enriching the state information gathered from the environment. Another potential application of LLMs in RL is using transformer layers (the main component of LLMs) in value or policy neural networks to infer complex patterns. By integrating transformer architectures, the RL agent may capture long-range dependencies and complex temporal patterns in market data, thereby enhancing its decision-making capabilities. In addition, incorporating risk assessments generated by LLMs into the RL framework can lead to more robust trading strategies. For example, LLMs can analyze textual data to estimate market volatility or detect potential \textit{black swan} events\cite{taleb2010black}, which the RL agent can use to adjust its policy accordingly. 

Additionally, LLMs can also be used as standalone agents, apart from RL. These agents could operate based on various inputs, including the unstructured data mentioned earlier, and can be tested or fine-tuned according to specific requirements. They can act as `intelligent' competitors in simulations, increasing the complexity of simulated markets and resulting in more robust policies. Finally, LLMs can also be used more broadly, such as to predict market trends or generate trading signals based on textual data. For instance, LLMs can process earnings call transcripts to predict stock movements or potential impacts, which can then inform trading decisions or be used as input for other models. Additionally, LLMs can be used to generate synthetic financial data to augment training datasets, improving the robustness of trading algorithms, especially when historical data is limited. They could even generate plausible market scenarios or price movements.

All of these are examples of potential applications where LLMs can positively impact trading and market making strategies, making them worth exploring.


\clearpage
\addcontentsline{toc}{chapter}{Bibliography}

\printbibliography

@article{Ganesh2019,
	title        = {Reinforcement Learning for Market Making in a Multi-agent Dealer Market},
	author       = {Sumitra Ganesh and Nelson Vadori and Mengda Xu and Hua Zheng and Prashant Reddy and Manuela Veloso},
	year         = 2019,
	month        = 11,
	url          = {http://arxiv.org/abs/1911.05892}
}

@report{Chan2001,
	title        = {An Electronic Market-Maker},
	author       = {Nicholas Tung Chan and Christian Shelton},
	year         = 2001
}

@article{Spooner2018,
	title        = {Market Making via Reinforcement Learning},
	author       = {Thomas Spooner and John Fearnley and Rahul Savani and Andreas Koukorinis},
	year         = 2018,
	month        = 4,
	booktitle    = {Proceedings of the 17th International Conference on Autonomous Agents and MultiAgent Systems},
	location     = {Stockholm, Sweden},
	publisher    = {International Foundation for Autonomous Agents and Multiagent Systems},
	address      = {Richland, SC},
	series       = {AAMAS '18},
	pages        = {434–442},
	url          = {http://arxiv.org/abs/1804.04216},
	numpages     = 9
}

@article{Byrd2019,
	title        = {ABIDES: Towards High-Fidelity Market Simulation for AI Research},
	author       = {David Byrd and Maria Hybinette and Tucker Hybinette Balch},
	year         = 2019,
	month        = 4,
	booktitle    = {Proceedings of the 2020 ACM SIGSIM Conference on Principles of Advanced Discrete Simulation},
	location     = {Miami, FL, Spain},
	publisher    = {Association for Computing Machinery},
	address      = {New York, NY, USA},
	series       = {SIGSIM-PADS '20},
	pages        = {11–22},
	doi          = {10.1145/3384441.3395986},
	isbn         = 9781450375924,
	url          = {http://arxiv.org/abs/1904.12066},
	numpages     = 12
}

@article{Avellaneda2008,
	title        = {High-frequency trading in a limit order book},
	author       = {Marco Avellaneda and Sasha Stoikov},
	year         = 2008,
	month        = 4,
	journal      = {Quantitative Finance},
	volume       = 8,
	pages        = {217--224},
	doi          = {10.1080/14697680701381228},
	issn         = 14697688,
	issue        = 3
}

@article{Patel2018,
	title        = {Optimizing Market Making using Multi-Agent Reinforcement Learning},
	author       = {Yagna Patel},
	year         = 2018,
	month        = 12,
	url          = {http://arxiv.org/abs/1812.10252}
}

@phdthesis{watkins1989,
	title        = {Learning from Delayed Rewards},
	author       = {Watkins, Christopher},
	year         = 1989,
	address      = {Cambridge, UK},
	school       = {King's College}
}

@inproceedings{Chakraborty2011,
	title        = {Market making and mean reversion},
	author       = {Tanmoy Chakraborty and Michael Kearns},
	year         = 2011,
	journal      = {Proceedings of the ACM Conference on Electronic Commerce},
	pages        = {307--313},
	doi          = {10.1145/1993574.1993622},
	isbn         = 9781450302616
}

@article{Lu2018,
	title        = {Order-book modelling and market making strategies},
	author       = {Xiaofei Lu and Frédéric Abergel},
	year         = 2018,
	month        = 6,
	url          = {http://arxiv.org/abs/1806.05101}
}

@inproceedings{Wang2011,
	title        = {The agent-based simulation of inventory-based model's impact on market maker trading mechanism},
	author       = {Yi Wang and Hongtao Zhou and Wei Zeng},
	year         = 2011,
	journal      = {IWACI 2011},
	pages        = {601--604},
	doi          = {10.1109/IWACI.2011.6160079},
	isbn         = 9781612843735
}

@misc{Shap,
	title        = {SHAP (SHapley Additive exPlanations)},
	author       = {Scott Lundberg},
	year         = 2018,
	howpublished = {https://shap.readthedocs.io/en/latest/index.html}
}

@misc{Scikit-Learn,
	title        = {Scikit-Learn (Machine learning in Python)},
	author       = {David Cournapeau},
	year         = 2007,
	howpublished = {https://scikit-learn.org/stable/}
}

@article{Mnih2015,
	title        = {Human-level control through deep reinforcement learning},
	author       = {Volodymyr Mnih and Koray Kavukcuoglu and David Silver and Andrei A. Rusu and Joel Veness and Marc G. Bellemare and Alex Graves and Martin Riedmiller and Andreas K. Fidjeland and Georg Ostrovski and Stig Petersen and Charles Beattie and Amir Sadik and Ioannis Antonoglou and Helen King and Dharshan Kumaran and Daan Wierstra and Shane Legg and Demis Hassabis},
	year         = 2015,
	month        = 2,
	journal      = {Nature},
	publisher    = {Nature Publishing Group},
	volume       = 518,
	pages        = {529--533},
	doi          = {10.1038/nature14236},
	issn         = 14764687,
	issue        = 7540,
	pmid         = 25719670
}

@article{hendershott2003electronic,
	title        = {Electronic trading in financial markets},
	author       = {Hendershott, Terrence},
	year         = 2003,
	journal      = {IT Professional Magazine},
	publisher    = {IEEE Computer Society},
	volume       = 5,
	number       = 4,
	pages        = 10
}

@incollection{HUBERMAN2005,
	title        = {A Simple Approach to Arbitrage Pricing Theory},
	author       = {Gur Huberman},
	year         = 2005,
	booktitle    = {Theory of Valuation},
	publisher    = {World Scientific},
	pages        = {289--308},
	doi          = {10.1142/9789812701022_0009}
}

@article{Chordia2013,
	title        = {High-Frequency Trading},
	author       = {Tarun Chordia and Amit Goyal and Bruce N. Lehmann and Gideon Saar},
	year         = 2013,
	journal      = {{SSRN} Electronic Journal},
	publisher    = {Elsevier {BV}},
	doi          = {10.2139/ssrn.2278347}
}

@article{https://doi.org/10.1111/j.1540-6261.1983.tb03834.x,
	title        = {Information Effects on the Bid-Ask Spread},
	author       = {Copeland, Thomas E. and Galai, Dan},
	year         = 1983,
	journal      = {The Journal of Finance},
	volume       = 38,
	number       = 5,
	pages        = {1457--1469},
	doi          = {10.1111/j.1540-6261.1983.tb03834.x}
}

@article{GARCIA2020104021,
	title        = {Learning adversarial attack policies through multi-objective reinforcement learning},
	author       = {Javier García and Rubén Majadas and Fernando Fernández},
	year         = 2020,
	journal      = {Engineering Applications of Artificial Intelligence},
	volume       = 96,
	pages        = 104021,
	doi          = {10.1016/j.engappai.2020.104021},
	issn         = {0952-1976},
	url          = {https://www.sciencedirect.com/science/article/pii/S0952197620303043}
}

@article{roijers2013survey,
  title={A survey of multi-objective sequential decision-making},
  author={Roijers, Diederik M and Vamplew, Peter and Whiteson, Shimon and Dazeley, Richard},
  journal={Journal of Artificial Intelligence Research},
  volume={48},
  pages={67--113},
  year={2013}
}

@book{jin2006multi,
  title={Multi-objective machine learning},
  author={Jin, Yaochu},
  volume={16},
  year={2006},
  publisher={Springer Science \& Business Media}
}

@incollection{deb2016multi,
  title={Multi-objective optimization},
  author={Deb, Kalyanmoy and Sindhya, Karthik and Hakanen, Jussi},
  booktitle={Decision sciences},
  pages={161--200},
  year={2016},
  publisher={CRC Press}
}

@article{van2014multi,
  title={Multi-objective reinforcement learning using sets of pareto dominating policies},
  author={Van Moffaert, Kristof and Now{\'e}, Ann},
  journal={The Journal of Machine Learning Research},
  volume={15},
  number={1},
  pages={3483--3512},
  year={2014},
  publisher={JMLR. org}
}

@article{nguyen2020multi,
  title={A multi-objective deep reinforcement learning framework},
  author={Nguyen, Thanh Thi and Nguyen, Ngoc Duy and Vamplew, Peter and Nahavandi, Saeid and Dazeley, Richard and Lim, Chee Peng},
  journal={Engineering Applications of Artificial Intelligence},
  volume={96},
  pages={103915},
  year={2020},
  publisher={Elsevier}
}

@inproceedings{ijcai2020p615,
	title        = {Data-Driven Market-Making via Model-Free Learning},
	author       = {Zhong, Yueyang and Bergstrom, YeeMan and Ward, Amy},
	year         = 2020,
	month        = 7,
	booktitle    = {Proceedings of the Twenty-Ninth International Joint Conference on Artificial Intelligence, {IJCAI-20}},
	publisher    = {International Joint Conferences on Artificial Intelligence Organization},
	pages        = {4461--4468},
	doi          = {10.24963/ijcai.2020/615},
	note         = {Special Track on AI in FinTech},
	editor       = {Christian Bessiere}
}

@inproceedings{Mani2019,
	title        = {Applications of Reinforcement Learning in Automated Market-Making},
	author       = {Mohammad Ali Mani and Steve Phelps},
	year         = 2019,
	booktitle    = {GAIW, May 2019, Montreal, Canada},
	url          = {https://nms.kcl.ac.uk/simon.parsons/publications/conferences/gaiw19.pdf}
}

@article{Selser_2021,
	title        = {Optimal Market Making by Reinforcement Learning},
	author       = {Matias Selser and Javier Kreiner and Manuel Maurette},
	year         = 2021,
	journal      = {{SSRN} Electronic Journal},
	publisher    = {Elsevier {BV}},
	doi          = {10.2139/ssrn.3829984}
}

@inproceedings{Lokhacheva2020/01,
	title        = {Reinforcement Learning Approach for Market-Maker Problem Solution},
	author       = {K.A. Lokhacheva and D.I. Parfenov and I.P. Bolodurina},
	year         = {2020},
	booktitle    = {Proceedings of the International Session on Factors of Regional Extensive Development (FRED 2019)},
	publisher    = {Atlantis Press},
	pages        = {256--260},
	doi          = {10.2991/fred-19.2020.52},
	isbn         = {978-94-6252-882-6},
	issn         = {2352-5428}
}

@article{GUEANT,
	title        = {Dealing with the Inventory Risk. A solution to the market making problem},
	author       = {Olivier Guéant and Charles-Albert Lehalle and Joaquin Fernandez Tapia},
	year         = 2011,
	month        = 5,
	journal      = {Mathematics and Financial Economics},
	doi          = {10.1007/s11579-012-0087-0}
}

@book{symposium,
	title        = {Reinforcement learning for high-frequency market making},
	author       = {Ye-Sheen Lim and Denise Gorse},
	year         = 2018,
	publisher    = {26th European Symposium on Artificial Neural Networks, Computational Intelligence and Machine Learning ESANN 2018 : Bruges, Belgium, April 25, 26, 27, 2018 : proceedings},
	isbn         = 9782875870476,
	url          = {https://www.esann.org/sites/default/files/proceedings/legacy/es2018-50.pdf}
}

@article{Zhang_2020,
	title        = {Reinforcement Learning for Optimal Market Making with the Presence of Rebate},
	author       = {Ge Zhang and Ying Chen},
	year         = 2020,
	journal      = {{SSRN} Electronic Journal},
	publisher    = {Elsevier {BV}},
	doi          = {10.2139/ssrn.3646753}
}

@article{Gasperov2021,
	title        = {Market Making with Signals through Deep Reinforcement Learning},
	author       = {Bruno Gasperov and Zvonko Kostanjcar},
	year         = 2021,
	journal      = {IEEE Access},
	publisher    = {Institute of Electrical and Electronics Engineers Inc.},
	volume       = 9,
	pages        = {61611--61622},
	doi          = {10.1109/ACCESS.2021.3074782},
	issn         = 21693536
}

@article{HO198147,
	title        = {Optimal dealer pricing under transactions and return uncertainty},
	author       = {Thomas Ho and Hans R. Stoll},
	year         = 1981,
	journal      = {Journal of Financial Economics},
	volume       = 9,
	number       = 1,
	pages        = {47--73},
	doi          = {10.1016/0304-405X(81)90020-9},
	issn         = {0304-405X}
}

@inproceedings{8329917,
	title        = {MoVEMo: A Structured Approach for Engineering Reward Functions},
	author       = {Mallozzi, Piergiuseppe and Pardo, Raul and Duplessis, Vincent and Pelliccione, Patrizio and Schneider, Gerardo},
	year         = 2018,
	booktitle    = {2018 Second IEEE International Conference on Robotic Computing (IRC)},
	volume       = {},
	number       = {},
	pages        = {250--257},
	doi          = {10.1109/IRC.2018.00053}
}

@inproceedings{7965896,
	title        = {Deep reward shaping from demonstrations},
	author       = {Hussein, Ahmed and Elyan, Eyad and Gaber, Mohamed Medhat and Jayne, Chrisina},
	year         = 2017,
	booktitle    = {2017 International Joint Conference on Neural Networks (IJCNN)},
	volume       = {},
	number       = {},
	pages        = {510--517},
	doi          = {10.1109/IJCNN.2017.7965896}
}

@article{Adams2022,
	title        = {A survey of inverse reinforcement learning},
	author       = {Stephen Adams and Tyler Cody and Peter A. Beling},
	year         = 2022,
	month        = 8,
	journal      = {Artificial Intelligence Review},
	publisher    = {Springer Science and Business Media B.V.},
	doi          = {10.1007/s10462-021-10108-x},
	issn         = 15737462
}

@article{Behboudian2022,
	title        = {Policy invariant explicit shaping: an efficient alternative to reward shaping},
	author       = {Behboudian, Paniz and Satsangi, Yash and Taylor, Matthew E. and Harutyunyan, Anna and Bowling, Michael},
	year         = 2022,
	month        = 2,
	day          = {01},
	journal      = {Neural Computing and Applications},
	volume       = 34,
	number       = 3,
	pages        = {1673--1686},
	doi          = {10.1007/s00521-021-06259-1},
	issn         = {1433-3058}
}

@article{K.J.2022,
author={K. J., Prabuchandran
and Singh, Nitin
and Dayama, Pankaj
and Agarwal, Ashutosh
and Pandit, Vinayaka},
title={Change point detection for compositional multivariate data},
journal={Applied Intelligence},
year={2022},
month={01},
day={01},
volume={52},
number={2},
pages={1930-1955},
issn={1573-7497},
doi={10.1007/s10489-021-02321-6},
url={https://doi.org/10.1007/s10489-021-02321-6}
}

@inproceedings{10.1145/3490354.3494433,
	title        = {ABIDES-Gym: Gym Environments for Multi-Agent Discrete Event Simulation and Application to Financial Markets},
	author       = {Amrouni, Selim and Moulin, Aymeric and Vann, Jared and Vyetrenko, Svitlana and Balch, Tucker and Veloso, Manuela},
	year         = 2022,
	booktitle    = {Proceedings of the Second ACM International Conference on AI in Finance},
	location     = {Virtual Event},
	publisher    = {Association for Computing Machinery},
	address      = {New York, NY, USA},
	series       = {ICAIF '21},
	doi          = {10.1145/3490354.3494433},
	isbn         = 9781450391481,
	articleno    = 30,
	numpages     = 9
}

@inproceedings{10.1145/3383455.3422554,
	title        = {Generating Synthetic Data in Finance: Opportunities, Challenges and Pitfalls},
	author       = {Assefa, Samuel A. and Dervovic, Danial and Mahfouz, Mahmoud and Tillman, Robert E. and Reddy, Prashant and Veloso, Manuela},
	year         = 2021,
	booktitle    = {Generating Synthetic Data in Finance: Opportunities, Challenges and Pitfalls},
	location     = {New York, New York},
	publisher    = {Association for Computing Machinery},
	address      = {New York, NY, USA},
	series       = {ICAIF '20},
	doi          = {10.1145/3383455.3422554},
	isbn         = 9781450375849,
	articleno    = 44,
	numpages     = 8
}

@article{9383217,
	title        = {Improving Generalization in Reinforcement Learning–Based Trading by Using a Generative Adversarial Market Model},
	author       = {Kuo, Chia-Hsuan and Chen, Chiao-Ting and Lin, Sin-Jing and Huang, Szu-Hao},
	year         = 2021,
	journal      = {IEEE Access},
	volume       = 9,
	number       = {},
	pages        = {50738--50754},
	doi          = {10.1109/ACCESS.2021.3068269}
}

@article{doi:10.1080/1350486X.2021.1967767,
	title        = {Deep Learning for Market by Order Data},
	author       = {Zihao Zhang and Bryan Lim and Stefan Zohren},
	year         = 2021,
	journal      = {Applied Mathematical Finance},
	publisher    = {Routledge},
	volume       = 28,
	number       = 1,
	pages        = {79--95},
	doi          = {10.1080/1350486X.2021.1967767}
}

@article{gueant:hal-02862554,
	title        = {{Optimal market making}},
	author       = {Gu{\'e}ant, Olivier},
	year         = 2017,
	month        = 8,
	journal      = {{Applied Mathematical Finance}},
	publisher    = {{Taylor \& Francis (Routledge): SSH Titles}},
	volume       = 24,
	number       = 2,
	pages        = {112--154},
	doi          = {10.1080/1350486X.2017.1342552},
	url          = {https://hal.archives-ouvertes.fr/hal-02862554},
	hal_id       = {hal-02862554},
	hal_version  = {v1}
}

@article{Ait_Sahalia_2017,
	title        = {High Frequency Market Making: Implications for Liquidity},
	author       = {Yacine Ait-Sahalia and Mehmet Saalam},
	year         = 2017,
	journal      = {{SSRN} Electronic Journal},
	publisher    = {Elsevier {BV}},
	doi          = {10.2139/ssrn.2908438}
}

@article{Dixon_2017,
	title        = {High Frequency Market Making with Machine Learning},
	author       = {Matthew Francis Dixon},
	year         = 2017,
	journal      = {{SSRN} Electronic Journal},
	publisher    = {Elsevier {BV}},
	doi          = {10.2139/ssrn.2868473}
}

@article{10.1007/s11704-014-3312-6,
	title        = {An Intelligent Market Making Strategy in Algorithmic Trading},
	author       = {Li, Xiaodong and Deng, Xiaotie and Zhu, Shanfeng and Wang, Feng and Xie, Haoran},
	year         = 2014,
	month        = 8,
	journal      = {Front. Comput. Sci.},
	publisher    = {Springer-Verlag},
	address      = {Berlin, Heidelberg},
	volume       = 8,
	number       = 4,
	pages        = {596–608},
	doi          = {10.1007/s11704-014-3312-6},
	issn         = {2095-2228},
	issue_date   = {August    2014},
	numpages     = 13
}

@article{ganesh2019reinforcement,
	title        = {Reinforcement Learning for Market Making in a Multi-agent Dealer Market},
	author       = {Sumitra Ganesh and Nelson Vadori and Mengda Xu and Huabao Zheng and Prashant P. Reddy and Manuela M. Veloso},
	year         = 2019,
	journal      = {ArXiv},
	volume       = {abs/1911.05892}
}

@article{8756240,
	title        = {A New Model for the Multi-Objective Multiple Allocation Hub Network Design and Routing Problem},
	author       = {Demir,Ibrahim and Ergin, Fatma Corut and Kiraz, Berna},
	year         = 2019,
	journal      = {IEEE Access},
	volume       = 7,
	number       = {},
	pages        = {90678--90689},
	doi          = {10.1109/ACCESS.2019.2927418}
}

@inbook{Deb2014,
	title        = {Multi-objective Optimization},
	author       = {Deb, Kalyanmoy and Deb, Kalyanmoy},
	year         = 2014,
	booktitle    = {Search Methodologies: Introductory Tutorials in Optimization and Decision Support Techniques},
	publisher    = {Springer US},
	address      = {Boston, MA},
	pages        = {403--449},
	doi          = {10.1007/978-1-4614-6940-7_15},
	isbn         = {978-1-4614-6940-7},
	url          = {https://doi.org/10.1007/978-1-4614-6940-7_15}
}

@inproceedings{10.1007/978-3-642-15381-5_10,
	title        = {An Evolutionary Multi-objective Optimization of Market Structures Using PBIL},
	author       = {Li, Xinyang and Krause, Andreas},
	year         = 2010,
	booktitle    = {Intelligent Data Engineering and Automated Learning -- IDEAL 2010},
	publisher    = {Springer Berlin Heidelberg},
	address      = {Berlin, Heidelberg},
	pages        = {78--85},
	isbn         = {978-3-642-15381-5},
	editor       = {Fyfe, Colin and Tino, Peter and Charles, Darryl and Garcia-Osorio, Cesar and Yin, Hujun}
}

@inbook{Eschmann2021,
	title        = {Reward Function Design in Reinforcement Learning},
	author       = {Eschmann, Jonas},
	year         = 2021,
	booktitle    = {Reinforcement Learning Algorithms: Analysis and Applications},
	publisher    = {Springer International Publishing},
	address      = {Cham},
	pages        = {25--33},
	doi          = {10.1007/978-3-030-41188-6_3},
	isbn         = {978-3-030-41188-6},
	url          = {https://doi.org/10.1007/978-3-030-41188-6_3}
}

@article{Hayes2022,
	title        = {A practical guide to multi-objective reinforcement learning and planning},
	author       = {Hayes, Conor F. and R{\u{a}}dulescu, Roxana and Bargiacchi, Eugenio and K{\"a}llstr{\"o}m, Johan and Macfarlane, Matthew and Reymond, Mathieu and Verstraeten, Timothy and Zintgraf, Luisa M. and Dazeley, Richard and Heintz, Fredrik and Howley, Enda and Irissappane, Athirai A. and Mannion, Patrick and Now{\'e}, Ann and Ramos, Gabriel and Restelli, Marcello and Vamplew, Peter and Roijers, Diederik M.},
	year         = 2022,
	month        = 4,
	day          = 13,
	journal      = {Autonomous Agents and Multi-Agent Systems},
	volume       = 36,
	number       = 1,
	pages        = 26,
	doi          = {10.1007/s10458-022-09552-y},
	issn         = {1573-7454},
	url          = {https://doi.org/10.1007/s10458-022-09552-y}
}

@inproceedings{10.1007/978-3-540-89378-3_37,
	title        = {On the Limitations of Scalarisation for Multi-objective Reinforcement Learning of Pareto Fronts},
	author       = {Vamplew, Peter and Yearwood, John and Dazeley, Richard and Berry, Adam},
	year         = 2008,
	booktitle    = {AI 2008: Advances in Artificial Intelligence},
	publisher    = {Springer Berlin Heidelberg},
	address      = {Berlin, Heidelberg},
	pages        = {372--378},
	doi          = {https://doi.org/10.1007/978-3-540-89378-3},
	isbn         = {978-3-540-89378-3},
	editor       = {Wobcke, Wayne and Zhang, Mengjie}
}

@inproceedings{1599245,
	title        = {Pareto Multi Objective Optimization},
	author       = {Ngatchou, P. and Zarei, A. and El-Sharkawi, A.},
	year         = 2005,
	booktitle    = {Proceedings of the 13th International Conference on, Intelligent Systems Application to Power Systems},
	volume       = {},
	number       = {},
	pages        = {84--91},
	doi          = {10.1109/ISAP.2005.1599245}
}

@inproceedings{8477730,
	title        = {Sampling Reference Points on the Pareto Fronts of Benchmark Multi-Objective Optimization Problems},
	author       = {Tian, Ye and Xiang, Xiaoshu and Zhang, Xingyi and Cheng, Ran and Jin, Yaochu},
	year         = 2018,
	booktitle    = {2018 IEEE Congress on Evolutionary Computation (CEC)},
	volume       = {},
	number       = {},
	pages        = {1--6},
	doi          = {10.1109/CEC.2018.8477730}
}

@article{TANABE2020106078,
	title        = {An easy-to-use real-world multi-objective optimization problem suite},
	author       = {Ryoji Tanabe and Hisao Ishibuchi},
	year         = 2020,
	journal      = {Applied Soft Computing},
	volume       = 89,
	pages        = 106078,
	doi          = {https://doi.org/10.1016/j.asoc.2020.106078},
	issn         = {1568-4946},
	url          = {https://www.sciencedirect.com/science/article/pii/S1568494620300181}
}

@article{doi:10.1142/S0219622019500093,
	title        = {Directed Exploration in Black-Box Optimization for Multi-Objective Reinforcement Learning},
	author       = {Garc\'{\i}a, Javier and Iglesias, Roberto and Rodr\'{\i}guez, Miguel A. and Regueiro, Carlos V.},
	year         = 2019,
	journal      = {International Journal of Information Technology \& Decision Making},
	volume       = 18,
	number       = {03},
	pages        = {1045--1082},
	doi          = {10.1142/S0219622019500093},
	url          = {https://doi.org/10.1142/S0219622019500093},
	eprint       = {https://doi.org/10.1142/S0219622019500093}
}

@inproceedings{pmlr-v37-schaul15,
	title        = {Universal Value Function Approximators},
	author       = {Schaul, Tom and Horgan, Daniel and Gregor, Karol and Silver, David},
	year         = 2015,
	month        = 7,
	booktitle    = {Proceedings of the 32nd International Conference on Machine Learning},
	publisher    = {PMLR},
	address      = {Lille, France},
	series       = {Proceedings of Machine Learning Research},
	volume       = 37,
	pages        = {1312--1320},
	url          = {https://proceedings.mlr.press/v37/schaul15.html},
	editor       = {Bach, Francis and Blei, David}
}

@inproceedings{ngatchou2005pareto,
	title        = {Pareto multi objective optimization},
	author       = {Ngatchou, Patrick and Zarei, Anahita and El-Sharkawi, A},
	year         = 2005,
	booktitle    = {Proceedings of the 13th international conference on, intelligent systems application to power systems},
	pages        = {84--91},
	doi          = {10.1109/ISAP.2005.1599231},
	organization = {IEEE}
}

@article{Vamplew2011,
	title        = {Empirical evaluation methods for multiobjective reinforcement learning algorithms},
	author       = {Vamplew, Peter and Dazeley, Richard and Berry, Adam and Issabekov, Rustam and Dekker, Evan},
	year         = 2011,
	month        = 7,
	day          = {01},
	journal      = {Machine Learning},
	volume       = 84,
	number       = 1,
	pages        = {51--80},
	doi          = {10.1007/s10994-010-5232-5},
	issn         = {1573-0565},
	url          = {https://doi.org/10.1007/s10994-010-5232-5}
}

@inproceedings{7360024,
	title        = {Performance metrics in multi-objective optimization},
	author       = {Riquelme, Nery and Von Lücken, Christian and Baran, Benjamin},
	year         = 2015,
	booktitle    = {2015 Latin American Computing Conference (CLEI)},
	volume       = {},
	number       = {},
	pages        = {1--11},
	doi          = {10.1109/CLEI.2015.7360024}
}

@inproceedings{dewey2014reinforcement,
	title        = {Reinforcement learning and the reward engineering principle},
	author       = {Dewey, Daniel},
	year         = 2014,
	booktitle    = {2014 AAAI Spring Symposium Series},
	url          = {https://aaai.org/papers/07704-7704-reinforcement-learning-and-the-reward-engineering-principle/}
}

@article{garcia2017incremental,
	title        = {Incremental reinforcement learning for multi-objective robotic tasks},
	author       = {Garc{\'\i}a, Javier and Iglesias, Roberto and Rodr{\'\i}guez, Miguel A and Regueiro, Carlos V},
	year         = 2017,
	journal      = {Knowledge and Information Systems},
	publisher    = {Springer},
	volume       = 51,
	pages        = {911--940}
}

@article{vamplew2022scalar,
	title        = {Scalar reward is not enough: A response to Silver, Singh, Precup and Sutton (2021)},
	author       = {Vamplew, Peter and Smith, Benjamin J and K{\"a}llstr{\"o}m, Johan and Ramos, Gabriel and R{\u{a}}dulescu, Roxana and Roijers, Diederik M and Hayes, Conor F and Heintz, Fredrik and Mannion, Patrick and Libin, Pieter JK and others},
	year         = 2022,
	journal      = {Autonomous Agents and Multi-Agent Systems},
	publisher    = {Springer},
	volume       = 36,
	number       = 2,
	pages        = 41
}

@article{Vicente2023,
	title        = {Automated market maker inventory management with deep reinforcement learning},
	author       = {Vicente, {\'O}scar Fern{\'a}ndez and Fern{\'a}ndez, Fernando and Garc{\'i}a, Javier},
	year         = 2023,
	day          = 24,
	journal      = {Applied Intelligence},
	doi          = {10.1007/s10489-023-04647-9},
	issn         = {1573-7497},
	url          = {https://doi.org/10.1007/s10489-023-04647-9}
}

@article{YOO2021487,
	title        = {A Dynamic Penalty Function Approach for Constraint-Handling in Reinforcement Learning},
	author       = {Haeun Yoo and Victor M. Zavala and Jay H. Lee},
	year         = 2021,
	journal      = {IFAC-PapersOnLine},
	volume       = 54,
	number       = 3,
	pages        = {487--491},
	doi          = {https://doi.org/10.1016/j.ifacol.2021.08.289},
	issn         = {2405-8963},
	url          = {https://www.sciencedirect.com/science/article/pii/S2405896321010624},
	note         = {16th IFAC Symposium on Advanced Control of Chemical Processes ADCHEM 2021}
}

@article{Zakaria_2021,
	title        = {A study of multiple reward function performances for vehicle collision avoidance systems applying the DQN algorithm in reinforcement learning},
	author       = {N J Zakaria and M I Shapiai and N Wahid},
	year         = 2021,
	month        = 8,
	journal      = {IOP Conference Series: Materials Science and Engineering},
	publisher    = {IOP Publishing},
	volume       = 1176,
	number       = 1,
	pages        = {012033},
	doi          = {10.1088/1757-899X/1176/1/012033},
	url          = {https://dx.doi.org/10.1088/1757-899X/1176/1/012033}
}

@article{OBERLECHNER2004407,
	title        = {Information sources, news, and rumors in financial markets: Insights into the foreign exchange market},
	author       = {Thomas Oberlechner and Sam Hocking},
	year         = 2004,
	journal      = {Journal of Economic Psychology},
	volume       = 25,
	number       = 3,
	pages        = {407--424},
	doi          = {https://doi.org/10.1016/S0167-4870(02)00189-7},
	issn         = {0167-4870},
	url          = {https://www.sciencedirect.com/science/article/pii/S0167487002001897}
}

@article{MCINISH2002287,
	title        = {After-hours trading of NYSE stocks on the regional stock exchanges},
	author       = {Thomas H McInish and Bonnie F {Van Ness} and Robert A {Van Ness}},
	year         = {2002},
	journal      = {Review of Financial Economics},
	volume       = {11},
	number       = {4},
	pages        = {287--297},
	doi          = {https://doi.org/10.1016/S1058-3300(02)00060-5},
	issn         = {1058-3300},
	url          = {https://www.sciencedirect.com/science/article/pii/S1058330002000605},
	
}

@article{MANUCA1996134,
	title        = {Stationarity and nonstationarity in time series analysis},
	author       = {Radu Manuca and Robert Savit},
	year         = {1996},
	journal      = {Physica D: Nonlinear Phenomena},
	volume       = {99},
	number       = {2},
	pages        = {134--161},
	doi          = {https://doi.org/10.1016/S0167-2789(96)00139-X},
	issn         = {0167-2789},
	url          = {https://www.sciencedirect.com/science/article/pii/S016727899600139X}
}

@article{doi:10.1177/0093650217705528,
	title        = {Intraday News Trading: The Reciprocal Relationships Between the Stock Market and Economic News},
	author       = {Nadine Strauß and Rens Vliegenthart and Piet Verhoeven},
	year         = 2018,
	journal      = {Communication Research},
	volume       = 45,
	number       = 7,
	pages        = {1054--1077},
	doi          = {10.1177/0093650217705528},
	url          = {https://doi.org/10.1177/0093650217705528},
	note         = {PMID: 30443092},
	eprint       = {https://doi.org/10.1177/0093650217705528}
}

@book{Sutton1998,
	title        = {Reinforcement Learning: An Introduction},
	author       = {Sutton, Richard S. and Barto, Andrew G.},
	year         = 2018,
	publisher    = {The MIT Press},
	address      = {Cambridge},
	edition      = {Second}
}

@article{Padakandla2019,
	title        = {Reinforcement Learning in Non-Stationary Environments},
	author       = {Sindhu Padakandla and Prabuchandran K. J and Shalabh Bhatnagar},
	year         = 2019,
	month        = 5,
	doi          = {10.1007/s10489-020-01758-5},
	url          = {http://arxiv.org/abs/1905.03970 http://dx.doi.org/10.1007/s10489-020-01758-5}
}

@article{Asyuraa2021,
	title        = {Empirical evaluation on discounted Thompson sampling for multi-armed bandit problem with piecewise-stationary Bernoulli arms},
	author       = {F C Asyuraa and S Abdullah and T E Sutanto},
	year         = 2021,
	month        = 1,
	journal      = {Journal of Physics: Conference Series},
	publisher    = {IOP Publishing},
	volume       = 1722,
	number       = 1,
	pages        = {012096},
	doi          = {10.1088/1742-6596/1722/1/012096},
	url          = {https://dx.doi.org/10.1088/1742-6596/1722/1/012096}
}

@article{8842548,
	title        = {Spectrum Sensing Across Multiple Service Providers: A Discounted Thompson Sampling Method},
	author       = {Zhou, Min and Wang, Tianyu and Wang, Shaowei},
	year         = 2019,
	journal      = {IEEE Communications Letters},
	volume       = 23,
	number       = 12,
	pages        = {2402--2406},
	doi          = {10.1109/LCOMM.2019.2941717}
}

@article{10.1145/3459991,
	title        = {A Survey of Reinforcement Learning Algorithms for Dynamically Varying Environments},
	author       = {Padakandla, Sindhu},
	year         = 2021,
	month        = 7,
	journal      = {ACM Comput. Surv.},
	publisher    = {Association for Computing Machinery},
	address      = {New York, NY, USA},
	volume       = 54,
	number       = 6,
	doi          = {10.1145/3459991},
	issn         = {0360-0300},
	url          = {https://doi.org/10.1145/3459991},
	issue_date   = {July 2022},
	articleno    = 127,
	numpages     = 25
}

@inproceedings{hadoux:hal-01200817,
	title        = {Sequential Decision-Making under Non-stationary Environments via Sequential Change-point Detection},
	author       = {Hadoux, Emmanuel and Beynier, Aur{\'e}lie and Weng, Paul},
	year         = 2014,
	month        = Sep,
	booktitle    = {Learning over Multiple Contexts (LMCE)},
	address      = {Nancy, France},
	url          = {https://hal.science/hal-01200817},
	hal_id       = {hal-01200817},
	hal_version  = {v1}
}

@article{JMLR:v17:14-037,
	title        = {Addressing Environment Non-Stationarity by Repeating Q-learning Updates},
	author       = {Sherief Abdallah and Michael Kaisers},
	year         = 2016,
	journal      = {Journal of Machine Learning Research},
	volume       = 17,
	number       = 46,
	pages        = {1--31},
	url          = {http://jmlr.org/papers/v17/14-037.html}
}

@inbook{Choi2001,
	title        = {Hidden-Mode Markov Decision Processes for Nonstationary Sequential Decision Making},
	author       = {Choi, Samuel P. M. and Yeung, Dit-Yan and Zhang, Nevin L.},
	year         = 2001,
	booktitle    = {Sequence Learning: Paradigms, Algorithms, and Applications},
	publisher    = {Springer Berlin Heidelberg},
	address      = {Berlin, Heidelberg},
	pages        = {264--287},
	doi          = {10.1007/3-540-44565-X\_12},
	isbn         = {978-3-540-44565-4},
	editor       = {Sun, Ron and Giles, C. Lee}
}

@article{ROBINS1995,
	title        = {Catastrophic Forgetting, Rehearsal and Pseudorehearsal},
	author       = {Anthony Robins},
	year         = 1995,
	journal      = {Connection Science},
	publisher    = {Taylor \& Francis},
	volume       = 7,
	pages        = {123--146},
	doi          = {10.1080/09540099550039318},
	url          = {https://doi.org/10.1080/09540099550039318},
	issue        = 2
}

@inbook{Chen2018,
	title        = {Continual Learning and Catastrophic Forgetting},
	author       = {Chen, Zhiyuan and Liu, Bing},
	year         = 2018,
	booktitle    = {Lifelong Machine Learning},
	publisher    = {Springer International Publishing},
	address      = {Cham},
	pages        = {55--75},
	doi          = {10.1007/978-3-031-01581-6\_4},
	isbn         = {978-3-031-01581-6}
}

@inproceedings{9412614,
	title        = {Rethinking Experience Replay: a Bag of Tricks for Continual Learning},
	author       = {P. Buzzega and M. Boschini and A. Porrello and S. Calderara},
	year         = 2021,
	month        = 1,
	booktitle    = {2020 25th International Conference on Pattern Recognition (ICPR)},
	publisher    = {IEEE Computer Society},
	address      = {Los Alamitos, CA, USA},
	volume       = {},
	pages        = {2180--2187},
	doi          = {10.1109/ICPR48806.2021.9412614},
	issn         = {1051-4651}
}

@article{Atkinson2021,
	title        = {Pseudo-rehearsal: Achieving deep reinforcement learning without catastrophic forgetting},
	author       = {Craig Atkinson and Brendan McCane and Lech Szymanski and Anthony Robins},
	year         = 2021,
	journal      = {Neurocomputing},
	volume       = 428,
	pages        = {291--307},
	doi          = {https://doi.org/10.1016/j.neucom.2020.11.050},
	issn         = {0925-2312},
	url          = {https://www.sciencedirect.com/science/article/pii/S0925231220318439}
}

@article{Lange2022,
	title        = {A Continual Learning Survey: Defying Forgetting in Classification Tasks},
	author       = {M De Lange and R Aljundi and M Masana and S Parisot and X Jia and A Leonardis and G Slabaugh and T Tuytelaars},
	year         = 2022,
	month        = 7,
	journal      = {IEEE Transactions on Pattern Analysis; Machine Intelligence},
	publisher    = {IEEE Computer Society},
	volume       = 44,
	pages        = {3366--3385},
	doi          = {10.1109/TPAMI.2021.3057446},
	issn         = {1939-3539},
	city         = {Los Alamitos, CA, USA},
	issue        = {07}
}

@inproceedings{Sorrenti2023,
	title        = {Selective Freezing for Efficient Continual Learning},
	author       = {A Sorrenti and G Bellitto and F Salanitri and M Pennisi and C Spampinato and S Palazzo},
	year         = 2023,
	month        = 10,
	journal      = {2023 IEEE/CVF International Conference on Computer Vision Workshops (ICCVW)},
	publisher    = {IEEE Computer Society},
	pages        = {3542--3551},
	doi          = {10.1109/ICCVW60793.2023.00381},
	url          = {https://doi.ieeecomputersociety.org/10.1109/ICCVW60793.2023.00381},
	city         = {Los Alamitos, CA, USA}
}

@inproceedings{10.1145/1160633.1160779,
	title        = {Improving reinforcement learning with context detection},
	author       = {da Silva, Bruno C. and Basso, Eduardo W. and Perotto, Filipo S. and C. Bazzan, Ana L. and Engel, Paulo M.},
	year         = 2006,
	booktitle    = {Proceedings of the Fifth International Joint Conference on Autonomous Agents and Multiagent Systems},
	location     = {Hakodate, Japan},
	publisher    = {Association for Computing Machinery},
	address      = {New York, NY, USA},
	series       = {AAMAS '06},
	pages        = {810–812},
	doi          = {10.1145/1160633.1160779},
	isbn         = 1595933034,
	numpages     = 3
}

@article{ShenLiuQinSavvidesCheng2021,
	title        = {Partial Is Better Than All: Revisiting Fine-tuning Strategy for Few-shot Learning},
	author       = {Shen, Zhiqiang and Liu, Zechun and Qin, Jie and Savvides, Marios and Cheng, Kwang-Ting},
	year         = 2021,
	month        = 5,
	journal      = {Proceedings of the AAAI Conference on Artificial Intelligence},
	volume       = 35,
	number       = 11,
	pages        = {9594--9602},
	doi          = {10.1609/aaai.v35i11.17155},
	url          = {https://ojs.aaai.org/index.php/AAAI/article/view/17155},
	abstractnote = {The goal of few-shot learning is to learn a classifier that can recognize unseen classes from limited support data with labels. A common practice for this task is to train a model on the base set first and then transfer to novel classes through fine-tuning or meta-learning. However, as the base classes have no overlap to the novel set, simply transferring whole knowledge from base data is not an optimal solution since some knowledge in the base model may be biased or even harmful to the novel class. In this paper, we propose to transfer partial knowledge by freezing or fine-tuning particular layer(s) in the base model. Specifically, layers will be imposed different learning rates if they are chosen to be fine-tuned, to control the extent of preserved transferability. To determine which layers to be recast and what values of learning rates for them, we introduce an evolutionary search based method that is efficient to simultaneously locate the target layers and determine their individual learning rates. We conduct extensive experiments on CUB and mini-ImageNet to demonstrate the effectiveness of our proposed method. It achieves the state-of-the-art performance on both meta-learning and non-meta based frameworks. Furthermore, we extend our method to the conventional pre-training + fine-tuning paradigm and obtain consistent improvement.}
}

@article{Goutam2020,
	title        = {LayerOut: Freezing Layers in Deep Neural Networks},
	author       = {Goutam, Kelam and Balasubramanian, S. and Gera, Darshan and Sarma, R. Raghunatha},
	year         = 2020,
	month        = 9,
	day          = {08},
	journal      = {SN Computer Science},
	volume       = 1,
	number       = 5,
	pages        = 295,
	doi          = {10.1007/s42979-020-00312-x},
	issn         = {2661-8907},
	url          = {https://doi.org/10.1007/s42979-020-00312-x}
}

@article{doi:10.1073/pnas.1611835114,
	title        = {Overcoming catastrophic forgetting in neural networks},
	author       = {James Kirkpatrick  and Razvan Pascanu  and Neil Rabinowitz  and Joel Veness  and Guillaume Desjardins  and Andrei A. Rusu  and Kieran Milan  and John Quan  and Tiago Ramalho  and Agnieszka Grabska-Barwinska  and Demis Hassabis  and Claudia Clopath  and Dharshan Kumaran  and Raia Hadsell},
	year         = 2017,
	journal      = {Proceedings of the National Academy of Sciences},
	volume       = 114,
	number       = 13,
	pages        = {3521--3526},
	doi          = {10.1073/pnas.1611835114},
	url          = {https://www.pnas.org/doi/abs/10.1073/pnas.1611835114},
	eprint       = {https://www.pnas.org/doi/pdf/10.1073/pnas.1611835114}
}

@article{dc35850b-2ca1-314f-9e0d-470713436b17,
	title        = {On the Likelihood that One Unknown Probability Exceeds Another in View of the Evidence of Two Samples},
	author       = {William R. Thompson},
	year         = 1933,
	journal      = {Biometrika},
	publisher    = {[Oxford University Press, Biometrika Trust]},
	volume       = 25,
	number       = {3/4},
	pages        = {285--294},
	doi          = {10.1093/biomet/25.3-4.285},
	issn         = {00063444},
	url          = {http://www.jstor.org/stable/2332286}
}

@article{french1999catastrophic,
	title        = {Catastrophic forgetting in connectionist networks},
	author       = {Robert M. French},
	year         = 1999,
	journal      = {Trends in Cognitive Sciences},
	volume       = 3,
	number       = 4,
	pages        = {128--135},
	doi          = {https://doi.org/10.1016/S1364-6613(99)01294-2},
	issn         = {1364-6613},
	url          = {https://www.sciencedirect.com/science/article/pii/S1364661399012942}
}

@article{WU2020142,
	title        = {Adaptive stock trading strategies with deep reinforcement learning methods},
	author       = {Xing Wu and Haolei Chen and Jianjia Wang and Luigi Troiano and Vincenzo Loia and Hamido Fujita},
	year         = 2020,
	journal      = {Information Sciences},
	volume       = 538,
	pages        = {142--158},
	doi          = {https://doi.org/10.1016/j.ins.2020.05.066},
	issn         = {0020-0255},
	url          = {https://www.sciencedirect.com/science/article/pii/S0020025520304692}
}

@inproceedings{noda2010recursive,
	title        = {Recursive Adaptation of Stepsize Parameter for Non-stationary Environments},
	author       = {Noda, Itsuki},
	year         = 2009,
	booktitle    = {Principles of Practice in Multi-Agent Systems},
	publisher    = {Springer Berlin Heidelberg},
	address      = {Berlin, Heidelberg},
	pages        = {525--533},
	doi          = {https://doi.org/10.1007/978-3-642-11814-2_5},
	isbn         = {978-3-642-11161-7},
	editor       = {Yang, Jung-Jin and Yokoo, Makoto and Ito, Takayuki and Jin, Zhi and Scerri, Paul}
}

@article{george2006adaptive,
	title        = {Adaptive stepsizes for recursive estimation with applications in approximate dynamic programming},
	author       = {George, Abraham P and Powell, Warren B},
	year         = 2006,
	journal      = {Machine learning},
	publisher    = {Springer},
	volume       = 65,
	pages        = {167--198},
	doi          = {https://doi.org/10.1007/s10994-006-8365-9}
}

@inproceedings{10.1145/3600211.3604730,
author = {Hussain, Mubarak},
title = {Can AlphaGo be apt subjects for Praise/Blame for "Move 37"?},
year = {2023},
publisher = {Association for Computing Machinery},
address = {New York, NY, USA},
url = {https://doi.org/10.1145/3600211.3604730},
doi = {10.1145/3600211.3604730},
booktitle = {Proceedings of the 2023 AAAI/ACM Conference on AI, Ethics, and Society},
pages = {977–979},
numpages = {3},
location = {Montreal, Canada},
series = {AIES '23}
}

@inproceedings{haarnoja2018soft,
  title={Soft actor-critic: Off-policy maximum entropy deep reinforcement learning with a stochastic actor},
  author={Haarnoja, Tuomas and Zhou, Aurick and Abbeel, Pieter and Levine, Sergey},
  booktitle={International conference on machine learning},
  pages={1861--1870},
  year={2018},
  organization={PMLR}
}

@article{Williams92,
  author = {Williams, R. J.},
  journal = {Machine Learning},
  pages = {229--256},
  title = {Simple statistical gradient-following algorithms for connectionist reinforcement learning},
  volume = 8,
  year = 1992
}

@article{SchulmanWDRK17,
  author       = {John Schulman and
                  Filip Wolski and
                  Prafulla Dhariwal and
                  Alec Radford and
                  Oleg Klimov},
  title        = {Proximal Policy Optimization Algorithms},
  journal      = {CoRR},
  volume       = {abs/1707.06347},
  year         = {2017},
  url          = {http://arxiv.org/abs/1707.06347},
  eprinttype    = {arXiv},
  eprint       = {1707.06347},
  bibsource    = {dblp computer science bibliography, https://dblp.org}
}

@article{browne2012survey,
  title={A survey of monte carlo tree search methods},
  author={Browne, Cameron B and Powley, Edward and Whitehouse, Daniel and Lucas, Simon M and Cowling, Peter I and Rohlfshagen, Philipp and Tavener, Stephen and Perez, Diego and Samothrakis, Spyridon and Colton, Simon},
  journal={IEEE Transactions on Computational Intelligence and AI in games},
  volume={4},
  number={1},
  pages={1--43},
  year={2012},
  publisher={IEEE}
}

@inproceedings{10.1109/ICRA.2018.8460528,
author = {Peng, Xue Bin and Andrychowicz, Marcin and Zaremba, Wojciech and Abbeel, Pieter},
title = {Sim-to-Real Transfer of Robotic Control with Dynamics Randomization},
year = {2018},
publisher = {IEEE Press},
url = {https://doi.org/10.1109/ICRA.2018.8460528},
doi = {10.1109/ICRA.2018.8460528},
booktitle = {2018 IEEE International Conference on Robotics and Automation (ICRA)},
pages = {1–8},
numpages = {8},
location = {Brisbane, Australia}
}

@article{hofer2021sim2real,
  title={Sim2real in robotics and automation: Applications and challenges},
  author={Hofer, Sebastian and Bekris, Kostas and Handa, Ankur and Gamboa, Juan Camilo and Mozifian, Melissa and Golemo, Florian and Atkeson, Chris and Fox, Dieter and Goldberg, Ken and Leonard, John and others},
  journal={IEEE transactions on automation science and engineering},
  volume={18},
  number={2},
  pages={398--400},
  year={2021},
  publisher={IEEE}
}

@article{MCCAULEY2008820,
title = {Nonstationarity of efficient finance markets: FX market evolution from stability to instability},
journal = {International Review of Financial Analysis},
volume = {17},
number = {5},
pages = {820-837},
year = {2008},
issn = {1057-5219},
doi = {https://doi.org/10.1016/j.irfa.2008.02.004},
url = {https://www.sciencedirect.com/science/article/pii/S1057521908000045},
author = {Joseph L. McCauley}
}

@article{Schmitt_2013,
doi = {10.1209/0295-5075/103/58003},
url = {https://dx.doi.org/10.1209/0295-5075/103/58003},
year = {2013},
month = {9},
publisher = {EDP Sciences, IOP Publishing and Società Italiana di Fisica},
volume = {103},
number = {5},
pages = {58003},
author = {Thilo A. Schmitt and Desislava Chetalova and Rudi Schäfer and Thomas Guhr},
title = {Non-stationarity in financial time series: Generic features and tail behavior},
journal = {Europhysics Letters}
}

@phdthesis{procacci2023non,
  title={Non stationarity and market structure dynamics in financial time series},
  author={Procacci, Pier Francesco},
  year={2023},
  school={UCL (University College London)}
}

@InProceedings{pmlr-v119-fedus20a,
  title = 	 {Revisiting Fundamentals of Experience Replay},
  author =       {Fedus, William and Ramachandran, Prajit and Agarwal, Rishabh and Bengio, Yoshua and Larochelle, Hugo and Rowland, Mark and Dabney, Will},
  booktitle = 	 {Proceedings of the 37th International Conference on Machine Learning},
  pages = 	 {3061--3071},
  year = 	 {2020},
  editor = 	 {III, Hal Daumé and Singh, Aarti},
  volume = 	 {119},
  series = 	 {Proceedings of Machine Learning Research},
  publisher =    {PMLR},
  url = 	 {https://proceedings.mlr.press/v119/fedus20a.html}
}

@ARTICLE{8466590,
  author={Adadi, Amina and Berrada, Mohammed},
  journal={IEEE Access}, 
  title={Peeking Inside the Black-Box: A Survey on Explainable Artificial Intelligence (XAI)}, 
  year={2018},
  volume={6},
  number={},
  pages={52138-52160},
  doi={10.1109/ACCESS.2018.2870052}}

@inproceedings{tobin2017domain,
  title={Domain randomization for transferring deep neural networks from simulation to the real world},
  author={Tobin, Josh and Fong, Rachel and Ray, Alex and Schneider, Jonas and Zaremba, Wojciech and Abbeel, Pieter},
  booktitle={2017 IEEE/RSJ international conference on intelligent robots and systems (IROS)},
  pages={23--30},
  year={2017},
  organization={IEEE}
}

@inproceedings{10.1145/1160633.1160762,
author = {Fern\'{a}ndez, Fernando and Veloso, Manuela},
title = {Probabilistic policy reuse in a reinforcement learning agent},
year = {2006},
isbn = {1595933034},
publisher = {Association for Computing Machinery},
address = {New York, NY, USA},
url = {https://doi.org/10.1145/1160633.1160762},
doi = {10.1145/1160633.1160762},
booktitle = {Proceedings of the Fifth International Joint Conference on Autonomous Agents and Multiagent Systems},
pages = {720–727},
numpages = {8},
location = {Hakodate, Japan},
series = {AAMAS '06}
}

@misc{us_market_makers,
  author       = {{AlphaTrade}},
  title        = {List of U.S. market makers},
  howpublished = {\url{https://web.archive.org/web/20090122074413/http://www.alphatrade.com/techSupport/marketMakers.html}},
  note         = {Archived from the original on 2009-01-22. Retrieved 2008-10-31}
}

@misc{canadian_market_makers,
  author       = {{AlphaTrade}},
  title        = {List of market makers in Canada},
  howpublished = {\url{https://web.archive.org/web/20070109084732/http://www.alphatrade.com/techSupport/canadianIDs.html}},
  note         = {Archived from the original on 2007-01-09. Retrieved 2008-10-31}
}

@inproceedings{NIPS2017_3f5ee243,
 author = {Vaswani, Ashish and Shazeer, Noam and Parmar, Niki and Uszkoreit, Jakob and Jones, Llion and Gomez, Aidan N and Kaiser, \L ukasz and Polosukhin, Illia},
 booktitle = {Advances in Neural Information Processing Systems},
 editor = {I. Guyon and U. Von Luxburg and S. Bengio and H. Wallach and R. Fergus and S. Vishwanathan and R. Garnett},
 pages = {},
 publisher = {Curran Associates, Inc.},
 title = {Attention is All you Need},
 url = {https://proceedings.neurips.cc/paper_files/paper/2017/file/3f5ee243547dee91fbd053c1c4a845aa-Paper.pdf},
 volume = {30},
 year = {2017}
}

@INPROCEEDINGS{10386743,
  author={Wu, Jiayang and Gan, Wensheng and Chen, Zefeng and Wan, Shicheng and Yu, Philip S.},
  booktitle={2023 IEEE International Conference on Big Data (BigData)}, 
  title={Multimodal Large Language Models: A Survey}, 
  year={2023},
  volume={},
  number={},
  pages={2247-2256},
  doi={10.1109/BigData59044.2023.10386743}}

@article{SZAKMARY2010409,
title = {Trend-following trading strategies in commodity futures: A re-examination},
journal = {Journal of Banking \& Finance},
volume = {34},
number = {2},
pages = {409-426},
year = {2010},
issn = {0378-4266},
doi = {https://doi.org/10.1016/j.jbankfin.2009.08.004},
url = {https://www.sciencedirect.com/science/article/pii/S037842660900199X},
author = {Andrew C. Szakmary and Qian Shen and Subhash C. Sharma}
}

@article{rohrbach2017momentum,
  title={Momentum and trend following trading strategies for currencies revisited-combining academia and industry},
  author={Rohrbach, Janick and Suremann, Silvan and Osterrieder, Joerg},
  journal={Available at SSRN 2949379},
  year={2017}
}

@INPROCEEDINGS{5331484,
  author={Fong, Simon and Tai, Jackie},
  booktitle={2009 Fifth International Joint Conference on INC, IMS and IDC}, 
  title={The Application of Trend Following Strategies in Stock Market Trading}, 
  year={2009},
  volume={},
  number={},
  pages={1971-1976},
  doi={10.1109/NCM.2009.402}}

@book{taleb2010black,
  title={The Black Swan:: The Impact of the Highly Improbable: With a new section:" On Robustness and Fragility"},
  author={Taleb, Nassim Nicholas},
  volume={2},
  year={2010},
  publisher={Random house trade paperbacks}
}

@inproceedings{10.1145/3533271.3561755,
author = {Bai, Yuanlu and Lam, Henry and Balch, Tucker and Vyetrenko, Svitlana},
title = {Efficient Calibration of Multi-Agent Simulation Models from Output Series with Bayesian Optimization},
year = {2022},
isbn = {9781450393768},
publisher = {Association for Computing Machinery},
address = {New York, NY, USA},
url = {https://doi.org/10.1145/3533271.3561755},
doi = {10.1145/3533271.3561755},
booktitle = {Proceedings of the Third ACM International Conference on AI in Finance},
pages = {437–445},
numpages = {9},
location = {New York, NY, USA},
series = {ICAIF '22}
}

@inproceedings{10.1145/3383455.3422561,
author = {Vyetrenko, Svitlana and Byrd, David and Petosa, Nick and Mahfouz, Mahmoud and Dervovic, Danial and Veloso, Manuela and Balch, Tucker},
title = {Get real: realism metrics for robust limit order book market simulations},
year = {2021},
isbn = {9781450375849},
publisher = {Association for Computing Machinery},
address = {New York, NY, USA},
url = {https://doi.org/10.1145/3383455.3422561},
doi = {10.1145/3383455.3422561},
booktitle = {Proceedings of the First ACM International Conference on AI in Finance},
articleno = {2},
numpages = {8},
location = {New York, New York},
series = {ICAIF '20}
}

@article{JMLR:v16:garcia15a,
  author  = {Javier Garc{{\'i}}a and Fern and o Fern{{\'a}}ndez},
  title   = {A Comprehensive Survey on Safe Reinforcement Learning},
  journal = {Journal of Machine Learning Research},
  year    = {2015},
  volume  = {16},
  number  = {42},
  pages   = {1437--1480},
  url     = {http://jmlr.org/papers/v16/garcia15a.html}
}

\appendix
\chapter* {Appendix}
\renewcommand{\thechapter}{A}

\section{Appendix: Additional AIIF plots}\label{appendix_aiifplots}
Additional experiments were performed changing the AIIF factors. The results are shown as follows:
 \begin{figure}[ht]
        \centering        \includegraphics[width=1\textwidth]{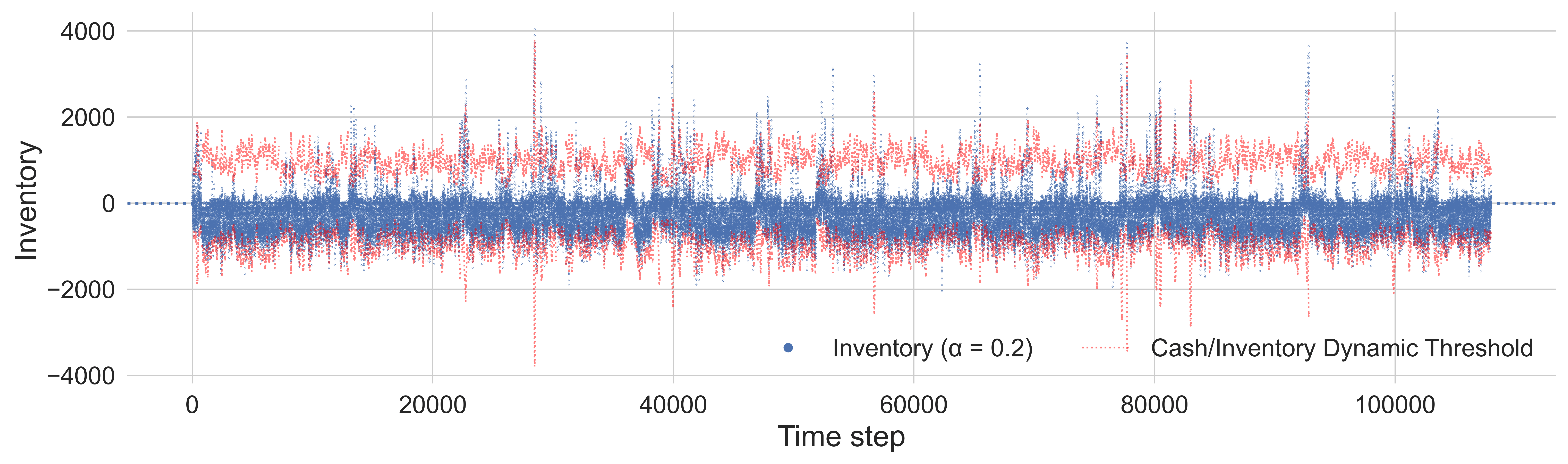}
        \caption[DQN MM instant inventories held and thresholds along the testing experiment every time step with $AIIF = 0.2$]{DQN MM instant inventories held and thresholds along the testing experiment every time step with $alpha$ factor applied $AIIF = 0.2$.}
        \label{}
        \end{figure}

 \begin{figure}[ht]
        \centering        \includegraphics[width=1\textwidth]{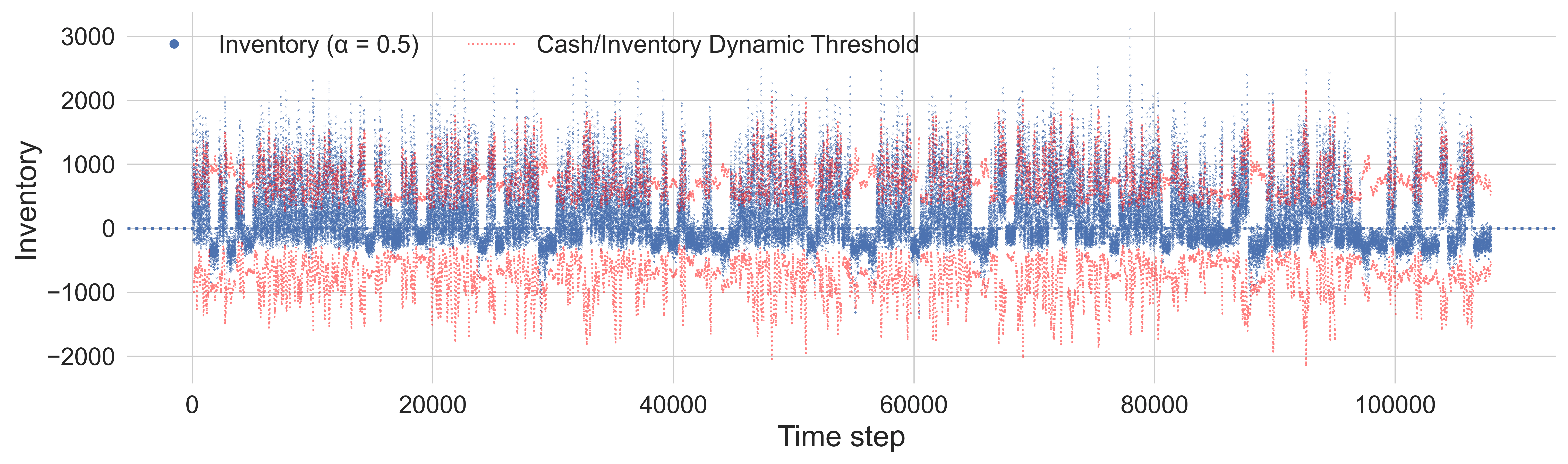}
        \caption[DQN MM instant inventories held and thresholds along the testing experiment every time step with $AIIF = 0.5$]{DQN MM instant inventories held and thresholds along the testing experiment every time step with $alpha$ factor applied $AIIF = 0.5$.}
        \label{}
        \end{figure}

 \begin{figure}[ht]
        \centering        \includegraphics[width=1\textwidth]{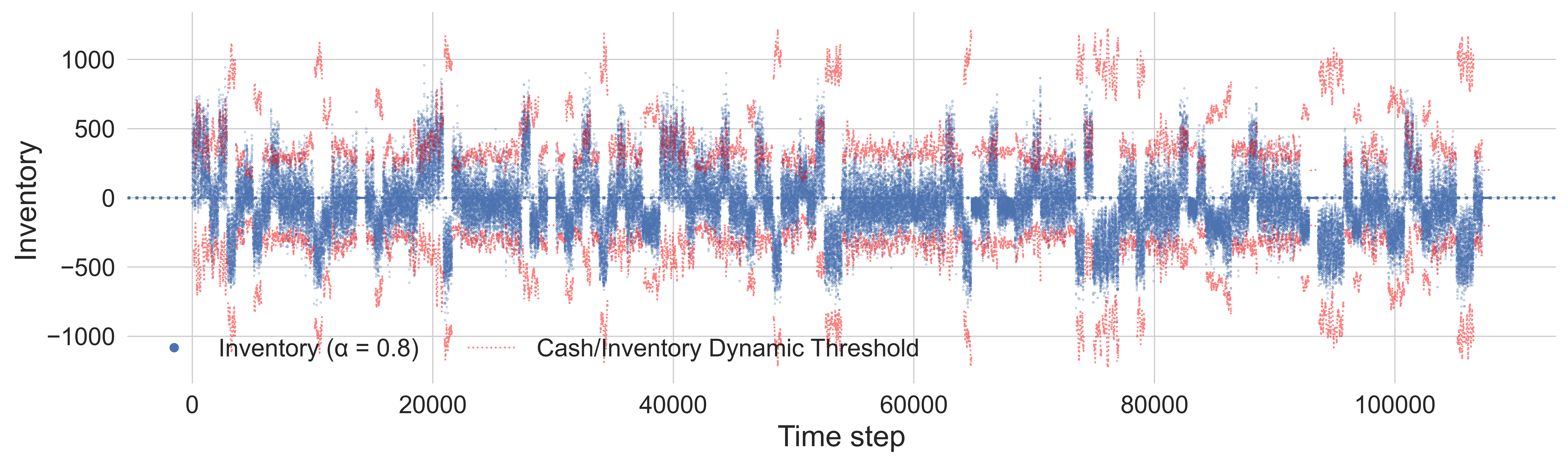}
        \caption[DQN MM instant inventories held and thresholds along the testing experiment every time step with $AIIF = 0.8$]{DQN MM instant inventories held and thresholds along the testing experiment every time step with $alpha$ factor applied $AIIF = 0.8$.}
        \label{}
        \end{figure}
        
 \begin{figure}[ht]
        \centering        \includegraphics[width=1\textwidth]{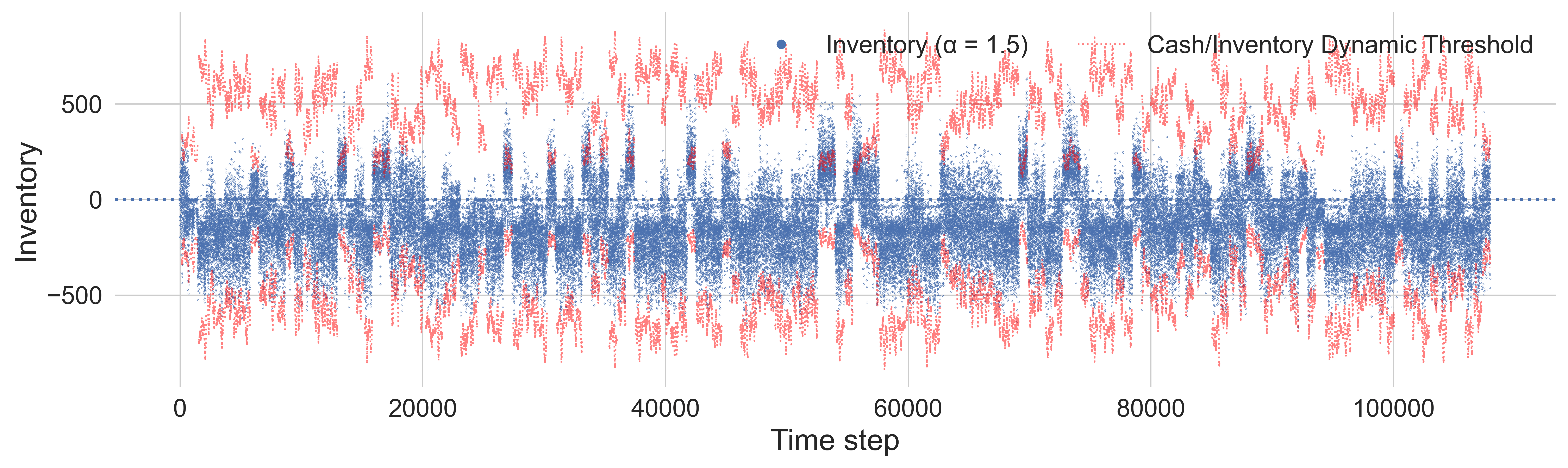}
        \caption[DQN MM instant inventories held and thresholds along the testing experiment every time step with $AIIF = 1.5$]{DQN MM instant inventories held and thresholds along the testing experiment every time step with $alpha$ factor applied $AIIF = 1.5$.}
        \label{}
        \end{figure}

 \begin{figure}[ht]
        \centering        \includegraphics[width=1\textwidth]{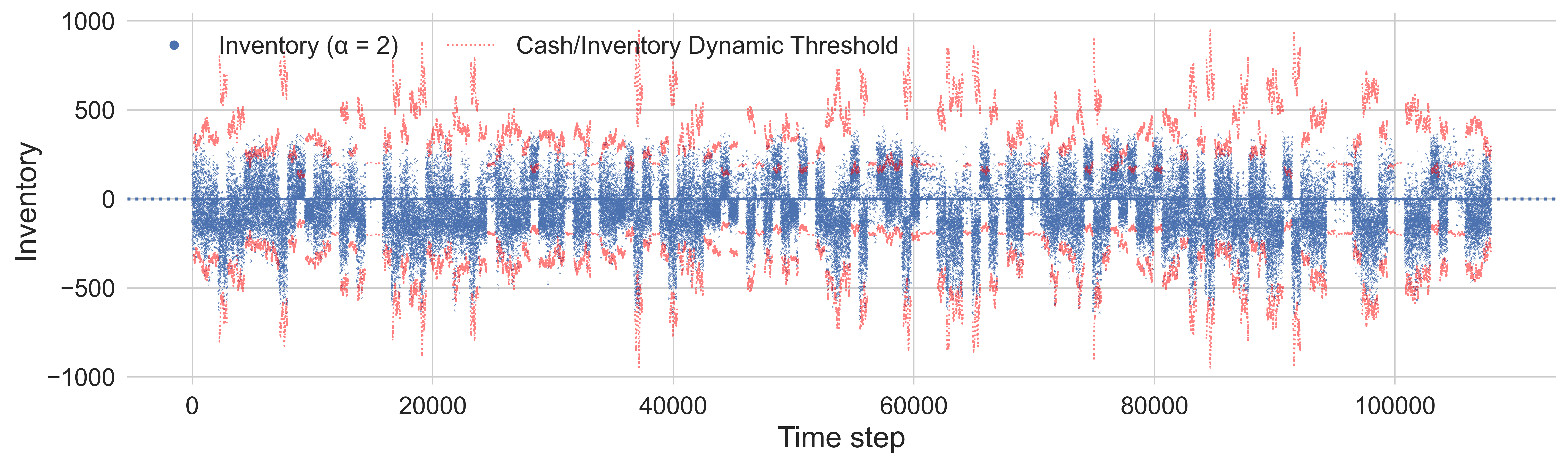}
        \caption[DQN MM instant inventories held and thresholds along the testing experiment every time step with $AIIF = 2$]{DQN MM instant inventories held and thresholds along the testing experiment every time step with $alpha$ factor applied $AIIF = 2$.}
        \label{}
        \end{figure}

 \begin{figure}[ht]
        \centering        \includegraphics[width=1\textwidth]{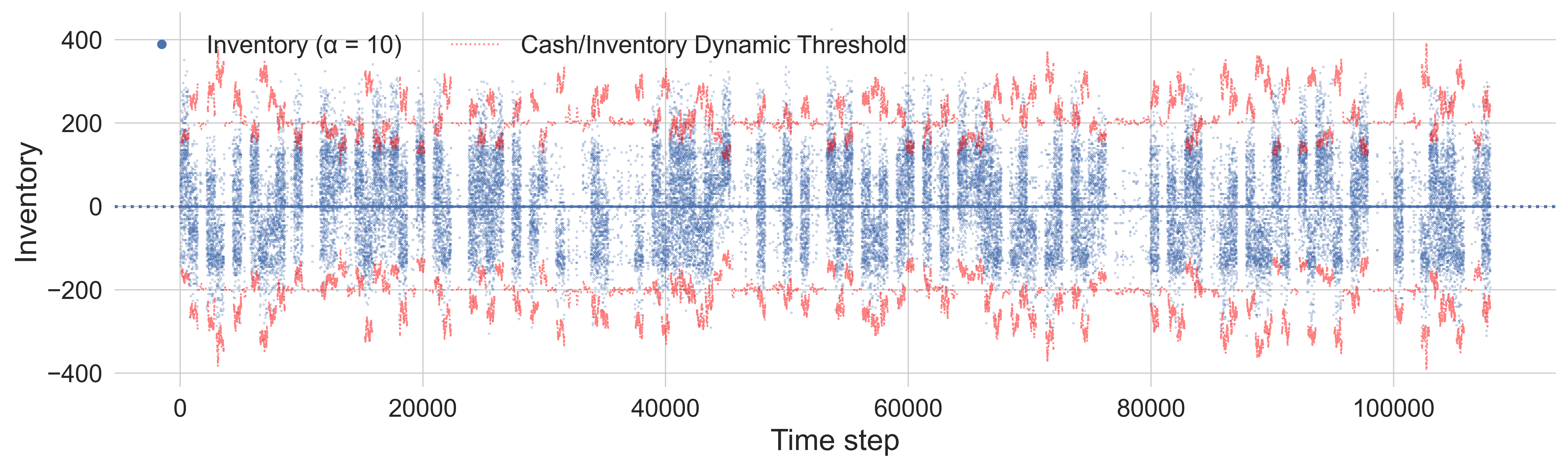}
        \caption[DQN MM instant inventories held and thresholds along the testing experiment every time step with $AIIF = 10$]{DQN MM instant inventories held and thresholds along the testing experiment every time step with $alpha$ factor applied $AIIF = 10$.}
        \label{}
        \end{figure}

 \begin{figure}[ht]
        \centering
        \includegraphics[width=1\textwidth]{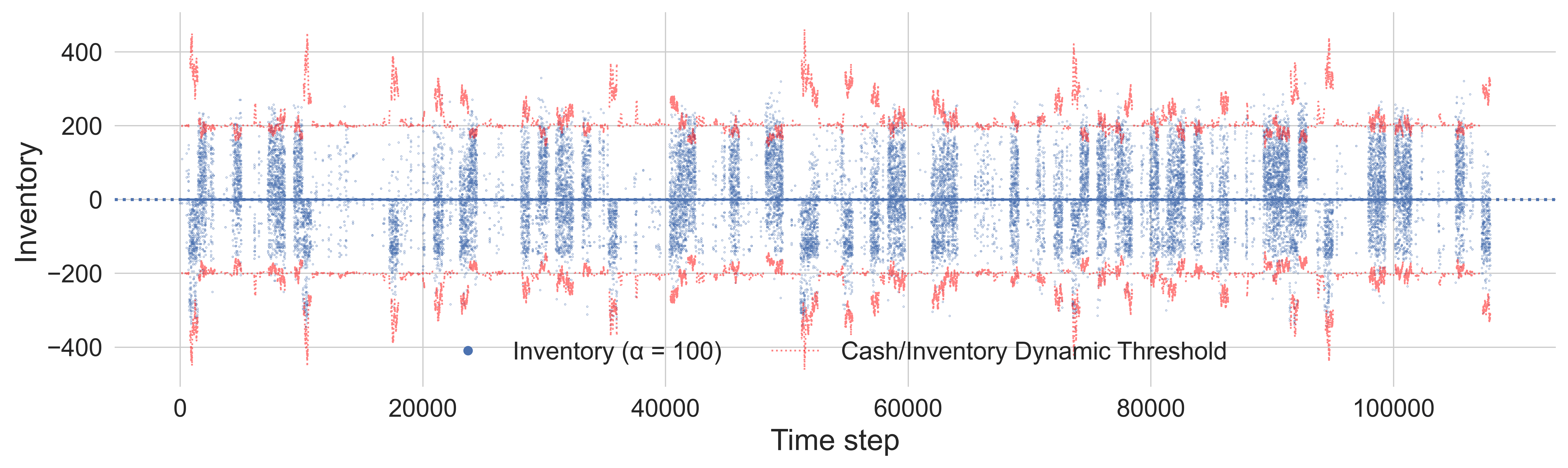}
        \caption[DQN MM instant inventories held and thresholds along the testing experiment every time step with $AIIF = 100$]{DQN MM instant inventories held and thresholds along the testing experiment every time step with $alpha$ factor applied $AIIF = 1005$.}
        \label{}
        \end{figure}

\end{document}